\documentclass[11pt, a4paper]{article}

\usepackage[top=1in, bottom=1in, left=1in, right=1in]{geometry}
\usepackage{amssymb}
\usepackage{amsmath}
 \usepackage{amsthm}

\usepackage{amsfonts}
\usepackage{hyperref}
\usepackage{enumitem}
\usepackage{titlesec}
\usepackage{secdot}
\usepackage{algorithm}
\usepackage{algpseudocode}
\sectiondot{subsection}
\sectiondot{subsubsection}
\usepackage{fancyhdr}
\usepackage{color}
\usepackage{authblk}
\usepackage{tcolorbox}
\usepackage{subcaption}
\usepackage{titlesec}
\usepackage{relsize}
\usepackage{mathtools}
\theoremstyle{plain}
\newtheorem{theorem}{Theorem}

\usepackage{pgf,tikz}
\usepackage{graphicx} % Required for inserting images

\usepackage{xcolor}

\title{Diffusion-Based Stochastic Operator Networks for Uncertainty Quantification in Stochastic Partial Differential Equations}

\author[1]{Phuoc Toan Huynh}
\author[2]{Richard Archibald}
\author[1]{Feng Bao}

\affil[1]{Department of Mathematics, Florida State University, Tallahassee, FL, USA}
\affil[2]{Computer Science and Mathematics Division, Oak Ridge National Laboratory, Oak Ridge, Tennessee, USA}

\date{}

\begin{document}

\maketitle
\begin{abstract}
We introduce a novel framework for uncertainty quantification of solution operators associated with stochastic partial differential equations (SPDEs). Although SPDEs play a central role in modeling complex physical systems under uncertainty, their practical use typically requires specifying the magnitude and structure of model uncertainties that are often unknown and difficult to infer from noisy measurements. To address this challenge, we develop a stochastic operator-learning framework that learns directly from noisy data and outputs both a mean solution field and a quantification of uncertainty. The proposed method, namely the Stochastic Operator Network (SON), is constructed by combining the structure of the Deep Operator Network (DeepONet) with Stochastic Neural Networks (SNNs) to model stochasticity and enable probabilistic prediction. The training procedure is carried out by minimizing a Hamiltonian-type loss and optimizing the resulting objective using the Stochastic Maximum Principle. Numerical experiments on benchmark SPDEs under multiple uncertainty sources demonstrate the accuracy and robustness of the proposed method in capturing solution structure and quantifying predictive uncertainty.
\end{abstract}
{}
\section{Introduction}

Partial differential equations (PDEs) are fundamental tools for modeling complex physical systems and have been widely used across scientific and engineering applications. However, many real-world problems involve intrinsic uncertainties arising from incomplete physical knowledge, imperfect observations, environmental variability, and unresolved multiscale processes. These uncertainties may appear, for example, in initial or boundary conditions, unresolved physical processes, or heterogeneous material properties, and can significantly impact predictive accuracy. To obtain reliable and uncertainty-aware predictions, such effects are often incorporated directly into the mathematical formulation through random inputs, stochastic coefficients, or stochastic perturbations, leading to stochastic partial differential equations (SPDEs).

Deriving numerical solutions to SPDEs has thus become a central focus of the uncertainty quantification (UQ) community, where extensive efforts have been devoted to developing efficient solvers that can accurately characterize and propagate uncertainty in high-dimensional, nonlinear dynamical systems (see, e.g.,~\cite{Abgrall2013, Abgrall2017, adjointFBSDE_Bao2019, bao2016first, bao2014hybrid, bao2011numerical, Barth2012, Geraci2016, KNIO2006, Petrella2019, Walton2025} and the references therein). Although traditional methods are effective for solving SPDEs and propagating uncertainty from stochastic input data to model predictions, they often require substantial computational cost, especially for time-dependent and multiscale problems~\cite{Ozen2017, Pranesh2018}.

One approach to reducing the computational cost of solving dynamical systems is to develop neural-network (NN)-based solvers that transfer much of the computational burden to offline training. Various modern machine-learning (ML)-based UQ methods have been developed to characterize uncertainties in stochastic dynamical systems and SPDE models~\cite{Solano2016, Harmon2025, Jung2024, Cho2022, Tokareva2024}. However, many such solvers are designed for a specific problem setting and generally need to be retrained when applied to systems with different parameters, inputs, or physical features.

Meanwhile, operator learning has emerged as a promising framework for approximating solution operators associated with differential equations~\cite{Bhattacharya2021, Chen1995, Guo2024, Lu2022, Rahman2022}. Once trained, a neural operator can be evaluated rapidly for new inputs drawn from the same problem class, making it an attractive surrogate for repeated-query tasks. However, in practice, the training data may contain measurement noise, numerical approximation errors, or intrinsic stochastic variability, particularly when generated from SPDEs. Since the corresponding solutions are random processes or random fields rather than deterministic functions, standard neural networks may overfit the stochastic fluctuations present in the training data, leading to poor generalization on unseen samples~\cite{Lin2023}. Therefore, it is important to integrate UQ techniques into operator-learning frameworks to maintain computational efficiency while providing reliable uncertainty estimates.

Several approaches have been proposed to enable UQ for operator learning, including ensemble-based and Bayesian training schemes such as UQDeepONet~\cite{Yang2022} and Bayesian DeepONet~\cite{Lin2023}. While these methods can provide meaningful uncertainty estimates, they often incur significant computational overhead, with costs that scale with the ensemble size or the number of posterior samples. As a result, they may become prohibitively expensive for large-scale real-world applications without access to high-performance computing resources. Another line of research builds upon ConvPDE-UQ networks~\cite{Winovich2019ConvPDEUQ} by integrating operator learning with Bayesian optimization and active learning to produce efficient predictions of both solution means and uncertainties~\cite{Winovich2026}. However, this approach mainly focuses on learning solution operators from noisy training data for a given PDE, rather than learning solution operators associated directly with SPDEs.

\vspace{0.5em}

In this work, we develop a Stochastic Operator Network (SON) approach for learning solution operators associated with SPDEs. SON is a probabilistic ML framework that combines Stochastic Neural Networks (SNNs)~\cite{Archibald2024} with the DeepONet architecture. An SNN can be viewed as a stochastic extension of neural ordinary differential equations (neural ODEs), where the evolution of hidden states in a deep neural network is interpreted as a discretized ODE system~\cite{Chen2018, Dupont2019, Gerstberger1997, Haber2018}. More specifically, a Brownian diffusion term with a trainable scaling coefficient is introduced into the hidden-state dynamics, thereby transforming the ODE system into an SDE~\cite{Archibald2024, Bao2022, Jia2019, Kong2020, Liu2019}. This stochastic diffusion component is used to characterize uncertainty arising from model randomness and noisy data.

Similar to DeepONet, SON aims to learn a solution operator whose inputs consist of physical problem data, such as forcing terms or initial conditions. However, unlike DeepONet, SON produces probabilistic solution predictions. The randomness is not prescribed as an explicit external input variable; instead, it is encoded in the stochastic diffusion components of the model and sampled during prediction. Therefore, SON is designed to perform uncertainty quantification for operator learning conditioned on deterministic problem inputs, rather than learning a finite-dimensional random-variable-to-solution map. In a recent study~\cite{Ryan2026}, a preliminary SON architecture was introduced for learning simple stochastic solution operators associated with stochastic \textit{ordinary} differential equations (SDEs). In that approach, SON was built upon the DeepONet architecture by replacing the branch network with an SNN. The SNN output was then combined with the trunk output to produce the model prediction. In this way, the randomness in the prediction was encoded through the branch network, and both the deterministic DeepONet parameters and the stochastic diffusion parameters were optimized simultaneously within a single training procedure. The SON training mechanism was formulated by adapting the SNN training framework, which is based on the Stochastic Maximum Principle (SMP), to learn the magnitude of uncertainty. While this formulation demonstrated the feasibility of stochastic operator learning for SDEs, it becomes computationally prohibitive when more expressive architectures are required for SPDE solution operators and may also suffer from reduced training stability.

To develop a SON framework capable of learning complex SPDE models, our objective is to reduce the computational cost and complexity of the training process while maintaining the predictive accuracy of SON. The main idea is to decouple the learning of the deterministic solution operator from the learning of stochastic uncertainty structure, leading to a two-phase training strategy designed to further improve training efficiency. In Phase I, a deterministic neural operator, such as DeepONet, is trained on the available noisy dataset to obtain a baseline approximation of the solution operator. In Phase II, the output of the deterministic model is passed through the SNN component, which learns the stochastic quantities and generates probabilistic solution predictions. In this way, each phase focuses on a distinct task: Phase I captures the overall structure of the solution trajectories, while Phase II models the uncertainty around the deterministic prediction. 
Moreover, this separation allows the architecture in each phase to be tailored to its specific objective: Phase I can employ a sufficiently expressive deterministic neural operator to learn the solution structure, whereas Phase II can use a simpler and more stable drift component to preserve the mean behavior, so that the computational effort can be concentrated on learning the diffusion term.
This two-phase training strategy avoids jointly training all DeepONet and SNN components and eliminates the need to modify the target loss function used in the existing SNN training framework. As a result, the proposed framework reduces the complexity of the training procedure while preserving the uncertainty quantification capability of SON.

The rest of the paper is organized as follows. Section~\ref{SON} summarizes the background material needed to construct SON. In particular, Subsection~\ref{DeepONet} reviews the theoretical foundation of DeepONet, Subsection~\ref{SNNs} introduces the methodology and training framework of SNNs, and Subsection~\ref{SONs} discusses how SNNs are integrated into DeepONet to formulate SON. We then present the proposed two-phase training strategy for SON in Subsection~\ref{TwoP_SON}. Section~\ref{NumericalExps} presents numerical experiments illustrating the effectiveness of the proposed approach. Finally, concluding remarks are given in Section~\ref{Conclusion}.

\section{Diffusion-Based Stochastic Operator Networks}\label{SON}
In this section, we introduce the \textit{diffusion-based stochastic operator network} framework. We begin with a brief overview of the operator learning paradigm and adopt Deep Operator Networks (DeepONets) as the prototype architecture used in this work. Next, we present the stochastic neural network (SNN) methodology, which incorporates an additive diffusion term into deep neural networks (DNNs) to characterize and quantify predictive uncertainty. Finally, we integrate the DeepONet architecture with the SNN formulation to develop the stochastic operator network (SON) framework, trained by an efficient two-phase procedure, for solving stochastic partial differential equations.

\subsection{Deep operator networks}\label{DeepONet}
DeepONets provide a surrogate representation for mappings between two Banach spaces~\cite{Lu2021}. Given an operator $G$ whose input is a function $u$ from a functional space $V$, DeepONet produces an approximation of $G(u)$ evaluated at a set of points $y$ in the domain of the output function. To enable the training of the approximation network, the input function $u$ is discretized by sampling its values at a finite set of $m$ locations $\{x_1, \hdots, x_m\}$, referred to as ``sensors''. 

%In our experiments, we use the same $m$ sensors for every input, but more flexible schemes where sensor locations vary by sample have also been explored (see, e.g.,~ \cite{Zhang2023}).

The theoretical foundation of operator learning is the universal operator approximation theorem (see \cite{Chen1995}). Specifically, let $K_1$ be a compact set in a Banach space $X$ and $K_2$ be a compact set in $\mathbb{R}^d$. Let $V\subset C(K_1)$ be a compact set, where $C(K_1)$ denotes the Banach space of all continuous functions on $K_1$ equipped with the norm
$\|f\|_{C(K_1)} := \max\limits_{x \in K_1}|f(x)|$.
Then the operator $G$ to be learned can be expressed as $
G: V \to C(K_2)$, $u \mapsto G(u)$,
so that $G(u)(y)$ is well-defined for $u\in V$ and $y\in K_2$. The universal operator approximation is stated as:

% \begin{theorem}\label{univapproxop}
% For any $\epsilon>0$, there exist positive integers $n,p,m$, constants
% $c_i^k,\ \xi_{ij}^k,\ \theta_i^k\in\mathbb{R}$, vectors $w_k\in\mathbb{R}^d$, scalars $\zeta_k\in\mathbb{R}$,
% points $x_j\in K_1$ (with $i=1,\ldots,n$, $k=1,\ldots,p$, $j=1,\ldots,m$), and a continuous non-polynomial function $\sigma$
% such that
% \[
% \left|G(u)(y) - \sum_{k=1}^{p}
% \underbrace{\sum_{i=1}^{n} c_i^k\, \sigma\!\left(\sum_{j=1}^{m}\xi_{ij}^k\,u(x_j)+\theta_i^k\right)}_{\text{Branch}}
% \underbrace{\sigma(w_k\cdot y + \zeta_k)}_{\text{Trunk}}
% \right| < \epsilon
% \]
% for all $u\in V$ and $y\in K_2$.
% \end{theorem}

% Theorem~\eqref{univapproxop} corresponds to a shallow (single-hidden-layer) branch and trunk construction, which can be limiting in practice. To gain expressivity, the authors in~\cite{Lu2021} extend Theorem~\eqref{univapproxop} to remove these constraints and enable arbitrary branch and trunk architectures. The generalized version of Theorem~\eqref{univapproxop} is given as follows:
\begin{theorem}
\label{univapproxop2}
\vspace{-0.5em}
For any $\epsilon >0$, there exist positive integers $m, p,$ continuous vector functions $\pmb{g}: \mathbb{R}^m \to \mathbb{R}^p$, $\pmb{f}: \mathbb{R}^d \to \mathbb{R}^p$, and $x_1, \hdots, x_m \in K_1$, such that
\begin{equation}
\label{ineq_appro2}
\begin{array}{l}
\Big|G(u)(y) - \langle \underbrace{\pmb{g}(u(x_1), u(x_2), ... u(x_m))}_{\text{Branch}}, \underbrace{\pmb{f}(y)}_{\text{Trunk}} \rangle\Big| < \epsilon, 
\end{array}
\end{equation}
holds for all $u \in V$ and $y \in K_2$, where $\langle \cdot, \cdot \rangle$ denotes the dot product in $\mathbb{R}^p$. Furthermore, the functions $\pmb{g}$ and $\pmb{f}$ can be chosen as diverse classes of neural networks, which satisfy the classical universal approximation theorem of functions, for examples, (stacked/unstacked) fully connected neural networks, residual neural networks and convolutional neural networks.
\end{theorem}

%We then define the approximation $\hat{G}(u)$ of $G(u)$ as the output of the dot-product in Eq.~\eqref{ineq_appro2}. 
Theorem \ref{univapproxop2} allows the branch and trunk networks differ in depth and width, provided they both output a $p$-dimensional vector. Although a bias term is not required in the universal approximation theorem, adding bias may increase the performance by reducing the generalization error~\cite{Lu2021}. As a result, the dot product in Eq.~\eqref{ineq_appro2} can be replaced by
\begin{equation}
\label{output_DeepONet_bias}
\tilde{G}(u)(y) = \left\langle \tilde{\pmb{g}}\left(u(x_1), \hdots, u(x_m)\right), \pmb{f}(y) \right\rangle +b_0,
\end{equation}
where $\tilde{\pmb{g}} = \pmb{g} + \pmb{b}$ with $\pmb{b} = \left[b_1, \hdots, b_p\right]^T$ and $\{\left\{b_k\right\}^{p}_{1}, b_0\}$ is the set of bias.

\vspace{0.25em}
The training process of DeepONet can become computationally expensive as the number of training samples and the output resolution increase, because the dataset is typically formed by taking the Cartesian product between input functions and output evaluation points, so that each input function is paired with many spatial or space--time locations. To mitigate this cost, we adopt Decoder-DeepONet~\cite{Chen2024}, an extension of DeepONet designed to handle high-resolution outputs more efficiently. Decoder-DeepONet has several important features: it avoids the Cartesian-product construction used in standard DeepONet, allows each input function to be evaluated on its own spatial or space--time grid, and replaces the dot-product output in Eq.~\eqref{output_DeepONet_bias} with a general decoder function $d(\cdot,\cdot)$, which can improve predictive accuracy. The following theorem provides the theoretical justification for this architecture~\cite{Chen2024}:
% \begin{theorem}
% \label{dotproduct_appro}
% For any $\epsilon>0$, there are $x_1, \hdots, x_m \in K_1$, continuous vector functions $\pmb{g}: \mathbb{R}^m \rightarrow \mathbb{R}^p$, \; $\pmb{f}: \mathbb{R}^d \rightarrow \mathbb{R}^p$, \; $\pmb{d}: \mathbb{R}^{2p} \rightarrow \mathbb{R}$, such that
% \begin{equation}
% \label{ineq_dotproduct}
% \vert \left\langle \pmb{g}\left(u(x_1), \hdots, u(x_m)\right), \pmb{f}(y)\right\rangle - \pmb{d}\left(\pmb{g}, \pmb{f}\right) \vert < \dfrac{1}{2}\epsilon,
% \end{equation}
% holds for all $u \in V$ and $y \in K_2$, where the decoder network $\pmb{d}$ needs to satisfy the universal approximation theorem of functional, and all the notations above is shared from Theorem~\eqref{univapproxop2}.

% \end{theorem}
\vspace{-0.25em}
\begin{theorem}
\label{decoder_appro}
For any $\epsilon>0$, there are $x_1, \hdots, x_m \in K_1$ and continuous vector functions $\pmb{g}: \mathbb{R}^m \rightarrow \mathbb{R}^p$, \; $\pmb{f}: \mathbb{R}^d \rightarrow \mathbb{R}^p$, \; $\pmb{d}: \mathbb{R}^{2p} \rightarrow \mathbb{R}$, such that
\begin{equation}
\label{ineq_decoder_appro}
\begin{array}{l}
\vert G(u)(y) - \pmb{d}\left(\pmb{g}\left(u(x_1), \hdots, u(x_m)\right), \pmb{f}(y)\right) \vert < \epsilon,
\end{array}
\end{equation}
holds for all $u \in V$ and $y \in K_2$.
\end{theorem}
%This modification increases the expressive power of the framework, allowing multidimensional outputs and enabling efficient learning from unaligned data~\cite{Chen2024}. 

Although Decoder-DeepONet allows for varying discretizations of $K_2$ across input functions $u$ in $K_1$, in this study we restrict attention to a fixed discretization of $K_2$ all training samples. In other words, the trunk network receives an identical set of grid points in $K_2$ for each training instance.

% One can obtain the bound~\eqref{ineq_decoder_appro} by replacing the right-hand side of~\eqref{ineq_appro2} with $\frac{1}{2}\epsilon$, then applying the triangle inequality to combine the resulting bound with~\eqref{ineq_dotproduct}. 
\vspace{0.5em}

A key limitation of DeepONet and its variants is that they are designed to produce deterministic outputs and therefore do not directly capture the stochasticity present in solutions of stochastic partial differential equations (SPDEs). To address this limitation, we develop a stochastic operator networks (SONs) framework, which integrates the DeepONet architecture with a diffusion-based stochastic neural network mechanism to enable probabilistic operator learning.

In what follows, we briefly present the mathematical formulation of stochastic neural networks (SNNs) (see \cite{Bao2022}) and summarize their training mechanism via a sample-wise backpropagation algorithm. We then describe how SNNs are integrated into the DeepONet architecture to construct SONs for SPDEs, followed by an efficient two-phase training strategy tailored to learning solutions of SPDEs.

\subsection{Stochastic neural networks}\label{SNNs}
The stochastic neural network (SNN) is designed to incorporate uncertainty into conventional deep neural network (DNN) architectures. It extends the neural ordinary differential equation (Neural ODE) framework~\cite{Chen2018, Dupont2019, Haber2018} \textit{by introducing an additive stochastic diffusion term} in each hidden layer, so that the input is propagated through a system of stochastic differential equations (SDEs). This formulation allows the model to generate multiple solution trajectories, which can be used to characterize the probability distribution underlying the training dataset~\cite{Bao2022}. Specifically, the discrete-layer SNN considered in this work is defined by
\begin{equation}
\label{SNN_1layer}
\begin{array}{l}
A_{n+1} = A_n + h f(A_n,\theta_n) + \sqrt{h}\, g(A_n,\theta_n)\omega_n, \qquad n=0,1,\ldots,N-1,
\end{array}
\end{equation}
where \(A_n \in \mathbb{R}^d\) denotes the state of the network at the \(n\)th layer, \(f\) is the drift term corresponding to a DNN architecture, \(g\) is a matrix-valued diffusion function taking values in \(\mathbb{R}^{d\times r}\) that controls and characterizes the magnitude and structure of the uncertainty in the SNN, \(\theta_n\) denotes the set of parameters at the \(n\)th layer, \(h>0\) is the step-size that can serve as a stabilizing factor for the DNN, and \(\omega_n\) is a standard \(r\)-dimensional Gaussian random vector.

In the standard supervised learning setting, training a neural network amounts to minimizing a loss function over the network parameters. For Neural ODEs, this procedure is equivalent to an optimal control problem, where the parameters are treated as control variables and the network dynamics are constrained by an ODE. In the same spirit, the training of the SNN model~\eqref{SNN_1layer} can be formulated as a stochastic optimal control problem.

To this end, let \(T>0\) be a fixed pseudo-terminal time, and consider the limit \(N\to\infty\). In this limit, the discrete dynamics in Eq.~\eqref{SNN_1layer} formally converge to the continuous-time SDE
\begin{equation}
\label{SNN_continuous}
\begin{array}{l}
A_T = A_0 + \mathlarger{\int}_0^T f(A_t,\theta_t)\,dt + \mathlarger{\int}_0^T g(A_t,\theta_t)\,dW_t,
\end{array}
\end{equation}
where \(W=\{W_t\}_{0\le t\le T}\) is a standard Brownian motion associated with the i.i.d. Gaussian sequence \(\{\omega_n\}_n\) in Eq.~\eqref{SNN_1layer}. The time-dependent parameter \(\theta_t\) in Eq.~\eqref{SNN_continuous} is then treated as a control process, and the training problem is formulated as the stochastic optimal control problem
\begin{equation}
\begin{array}{l}
\label{optimal_control}
J(\theta^*)=\inf\limits_{\theta\in\Theta} J(\theta),
\end{array}
\end{equation}
where \(\Theta\) is a convex admissible control set and the cost functional \(J\) is defined by
\begin{align*}
J(\theta)=\mathbb{E}\left[\Phi(A_T,\Gamma)+\mathlarger{\int}_0^T r(A_t,\theta_t)\,dt\right].
\end{align*}
Here, \(\Gamma\) denotes the target random variable associated with the training data, \(\Phi(A_T,\Gamma):=\|A_T-\Gamma\|_{\text{loss}}\) is the terminal loss, and \(\int_0^T r(A_t,\theta_t)\,dt\) is the running cost.

To solve the stochastic optimal control problem defined by 
Eqs.~\eqref{SNN_continuous}--\eqref{optimal_control}, we further reformulate it as 
an optimization problem characterized through the following Hamiltonian:
\begin{equation}
\label{Hamiltonian_loss}
 H(a, b, c, u) = r(a, u) + b^{\top}f(a, u)
+\mathrm{tr}\left(c^{\top}g(a, u)\right).
\end{equation}
For any admissible control $\theta\in\Theta$ satisfying 
$\mathbb{E}[|\theta|^2]<\infty$, let $A_{t,\theta}$ denote the corresponding state 
process solving Eq.~\eqref{SNN_continuous}. We then consider the associated 
adjoint processes $(B_{t,\theta},C_{t,\theta})$, which satisfy (see \cite{Bao_Control_20, liang2024convergence, Liang_2025_FoDS})
\begin{equation}
\label{adjoint_SDE}
dB_{t,\theta}
=
-\bigl(r_a(A_{t,\theta},\theta_t)
+f_a(A_{t,\theta},\theta_t)^{\top}B_{t,\theta}
+g_a(A_{t,\theta},\theta_t)^{\top}C_{t,\theta}\bigr)\,dt
-C_{t,\theta}\,dW_t,
\end{equation}
with terminal condition
\[
B_{T,\theta}=\Phi_a(A_{T,\theta},\Gamma).
\]
Here, \(r_a\), \(f_a\), \(g_a\), and $\Phi_a$ denote the corresponding derivatives 
with respect to \(a\), and \(C_{t,\theta}\) is the martingale term in the backward 
adjoint equation.

The optimal control problem~\eqref{optimal_control}, governed by the 
SDE~\eqref{SNN_continuous}, can then be converted into an optimization problem 
involving the Hamiltonian~\eqref{Hamiltonian_loss} evaluated at the state 
variable $A_{t,\theta}$ and the adjoint pair \((B_{t,\theta},C_{t,\theta})\). 
This reformulation is justified by the Stochastic Maximum Principle 
(SMP)~\cite{Andersson2009,Sun2023}, which can be stated as follows.
\begin{theorem}
\label{theorem: SMP}
Assume that \(\theta^*\) is an optimal control for the control 
problem~\eqref{optimal_control}, and let \(A_{t,\theta^*}\) be the solution to 
Eq.~\eqref{SNN_continuous} corresponding to \(\theta^*\). Then, there exist 
adjoint processes \((B_{t,\theta^*},C_{t,\theta^*})\) satisfying the backward 
SDE~\eqref{adjoint_SDE} associated with \(A_{t,\theta^*}\) and \(\theta_t^*\). 
Moreover, \(\theta_t^*\) satisfies the pointwise minimization condition:
\begin{equation}
\label{optimize_Hamiltonian}
H(A_{t,\theta^*},B_{t,\theta^*},C_{t,\theta^*},\theta_t^*)
=
\inf_{\theta\in\Theta}
H(A_{t,\theta^*},B_{t,\theta^*},C_{t,\theta^*},\theta).
\end{equation}
\end{theorem}

Theorem~\ref{theorem: SMP} provides a necessary condition for \(\theta^*\) to be an optimal control of Eq.~\eqref{optimal_control}. Alternatively, the stochastic control problem~\eqref{optimal_control} can be studied through the Dynamic Programming Principle (DPP), which formally leads to the associated Hamilton--Jacobi--Bellman (HJB) equation~\cite{FlemingRishel1975,FlemingSoner2006, Yong1999}. However, the high dimensionality of neural network parameter spaces makes the direct solution of the resulting HJB equations computationally intractable.

To clarify the relation between the optimal control problem~\eqref{optimal_control} and the pointwise Hamiltonian minimization problem~\eqref{optimize_Hamiltonian}, we follow the approach in~\cite{Ma1999,Yong1999} to derive the identity
\begin{equation}
\label{loss_relation}
\begin{array}{rl}
 \nabla_{\theta}J(\theta)\big|_t
&=\mathbb{E}\left[(f_\theta(A_{t,\theta},\theta_t))^{\top}B_{t,\theta}+(g_\theta(A_{t,\theta},\theta_t))^{\top}C_{t,\theta}+
(r_\theta(A_{t,\theta},\theta_t))^{\top}\right]
\\ &=
\mathbb{E}\left[
\nabla_{\theta}H(A_{t,\theta},B_{t,\theta},C_{t,\theta},\theta_t)
\right]. 
\end{array}
\end{equation}
Therefore, the optimal control problem~\eqref{optimal_control} can be approached by applying stochastic gradient descent (SGD) using the Hamiltonian~\eqref{Hamiltonian_loss}, rather than differentiating the cost functional \(J\) directly~\cite{Bao2020a,Bottou2018}:
\begin{equation}
\label{stochastic_gradient_descent}
\begin{array}{l}
\theta_t^{i+1}
=
\mathcal{P}_{\Theta}\!\left(
\theta_t^i
-\alpha\,\widehat{\mathbb{E}}\!\left[
\nabla_\theta H(A_t^i,B_t^i,C_t^i,\theta_t^i)
\right]
\right)
\vspace{0.1cm}\\
\hspace{0.8cm}
=
\mathcal{P}_{\Theta}\!\Big(
\theta_t^i
-\alpha \big[
r_\theta(A_t^i,\theta_t^i)
+
(f_\theta(A_t^i,\theta_t^i))^\top B_t^i
+
(g_\theta(A_t^i,\theta_t^i))^\top C_t^i
\big]
\Big),
\end{array}
\end{equation}
where \(A_t^i\) is a simulated trajectory of the forward SDE~\eqref{SNN_continuous} under the control \(\theta^i\), and \((B_t^i,C_t^i)\) solve the adjoint equation~\eqref{adjoint_SDE} corresponding to \(A_t^i\). Here, \(\widehat{\mathbb{E}}[\cdot]\) denotes a Monte Carlo estimator of the expectation in~\eqref{loss_relation}, for example, using a single trajectory or a minibatch of trajectories, \(\alpha>0\) is the learning rate, and \(\mathcal{P}_{\Theta}\) denotes the projection onto the set of admissible controls. The convergence of the iteration~\eqref{stochastic_gradient_descent} under standard smoothness assumptions is established in~\cite{Archibald2024}.

Since both the optimal control problem~\eqref{optimal_control} and the Hamiltonian minimization problem~\eqref{optimize_Hamiltonian} are formulated in continuous time, they must be discretized in order to train the SNN model~\eqref{SNN_1layer}. This requires numerical solvers for both the forward and backward SDEs~\eqref{SNN_continuous}--\eqref{adjoint_SDE}.

The forward and backward SDEs~\eqref{SNN_continuous}--\eqref{adjoint_SDE} are solved over a uniform partition of the pseudo-time interval:
\[
\Pi_N=\{t_n: 0=t_0<t_1<\cdots<t_N=1\},
\]
where \(N\) denotes the number of SNN layers and \(h=1/N\) is the step size. We adopt classical numerical schemes for the forward and backward SDEs~\cite{levyBSDE_Bao2019, Bao_DA_BSDE, Bao_2015_DCDS, Zhao2006}:
\begin{equation}
\label{Discretized_SDEs}
\begin{array}{l}
A^N_{n+1}=A^N_n+f(A^N_n,\theta_n)\,h+g(A_n^N,\theta_n)\,\sqrt{h} \omega_n, \vspace{0.1cm}\\
B^N_n=\mathbb{E}_n^A\!\left[B^N_{n+1}\right]
+h\,\mathbb{E}_n^A\!\left[(f_a(A^N_{n+1},\theta_n))^\top B^N_{n+1}
+\bigl(r_a(A^N_{n+1},\theta_{n+1})\bigr)^\top\right], \vspace{0.1cm}\\
C^N_n=\mathbb{E}_n^A\!\left[\dfrac{B^N_{n+1}(\Delta W_n)^\top}{h}\right],
\end{array}
\end{equation}
where \(A^N_{n+1}\), \(B_n^N\), and \(C_n^N\) are numerical approximations of \(A_{t_{n+1},\theta}\), \(B_{t_n,\theta}\), and \(C_{t_n,\theta}\), respectively, and
\[
\mathbb{E}_n^A[\cdot]:=\mathbb{E}[\cdot\mid A_{n}]
\]
denotes the conditional expectation given \(A_{n}\) \cite{Bao2022_Kernel, Bao2017meshfree}. Evaluating the conditional expectation \(\mathbb{E}_n^A[\cdot]\) typically involves high-dimensional integration, which is computationally challenging due to the curse of dimensionality. To overcome this difficulty, the authors in~\cite{Archibald2024,Bao2022} proposed a sample-wise backpropagation approach, in which the expectation \(\mathbb{E}_n^A[\cdot]\) is approximated using only a single simulated state process.

More precisely, at each layer \(n\), we draw one sample \(\epsilon_n\sim\mathcal{N}(0,\pmb{I}_r)\) that represents $\omega_n$ in Eq~\eqref{Discretized_SDEs} and use it to simulate the discretized forward and backward equations. This yields the following sample-wise approximations \(A_n\) of \(A_{t_n,\theta}\) and \((B_n,C_n)\) of \((B_{t_n,\theta},C_{t_n,\theta})\):
\begin{equation}
\label{Discretized_SDEs_SampleWise}
\begin{array}{l}
A_{n+1}=A_n+f(A_n,\theta_n)\,h+g(A_n,\theta_n)\,\sqrt{h}\epsilon_n, \vspace{0.1cm}\\
B_n=B_{n+1}+h(f_a(A_{n+1},\theta_n))^\top B_{n+1}
+\bigl(r_a(A_{n+1},\theta_{n+1})\bigr)^\top, \vspace{0.1cm}\\
C_n=\dfrac{B_{n+1}\epsilon_n^\top}{\sqrt{h}}.
\end{array}
\end{equation}
These updates form the core of the sample-wise backward propagation procedure for training the SNN. We then obtain the following discretization of the Hamiltonian loss in Eq.~\eqref{Hamiltonian_loss}
\begin{equation}
\small
\label{discretized_Ham}
\tilde{H}(\pmb{A},\pmb{B},\pmb{C},\pmb{\theta})
=
\frac{1}{m_B\times N}
\sum_{n=0}^{N-1}
\Bigl(
B^{\top}_n f(A_n,\theta_n)
+
\operatorname{tr}\;\bigl(g(A_n,\theta_n)^{\top}C_n\bigr)
\Bigr),
\end{equation}
where $m_B$ is a pre-chosen size of a mini-batch from the training dataset, $\pmb{A}$, $\pmb{B}$, $\pmb{C}$, and $\pmb{\theta}$ denote the collections of $\{A_n\}_{n=0}^{N-1}$, $\{B_n\}_{n=0}^{N-1}$, $\{C_n\}_{n=0}^{N-1}$, and $\{\theta_n\}_{n=0}^{N-1}$, respectively. The control $\pmb{\theta} = \{\theta_n\}_{n=0}^{N-1}$ is then updated through the discretization scheme of Eq.~\eqref{stochastic_gradient_descent}:
\begin{equation}
\label{discretized_SGD}
\begin{array}{l}
\theta_{n}^{i+1}=\mathcal{P}_{\Theta_N}\!\left(
\theta_{n}^i-\alpha\,
\nabla_{\theta_n} \tilde{H}(\pmb{A},\pmb{B},\pmb{C},\pmb{\theta})\right),
\end{array}
\end{equation}
where $\Theta_N = \Theta \cap \mathrm{C}_N$ with a piece-wise constant approximation set $\mathrm{C}_N:= \left\{u\vert u = \sum\limits^{N-1}_{n=0}a_n \pmb{1}_{[t_n, t_{n+1})}\right\}$.
% % Finally, we integrate the SNN training procedure into the DeepONet architecture to obtain a unified stochastic operator‐learning framework, which we name the Stochastic Operator Network.}

% \section{Operator Learning Framework}

\subsection{Construction of stochastic operator networks}\label{SONs}
In this section, we discuss how the DeepONet architecture can be used to extend SNN to a stochastic operator learning framework. This extension leads to the Stochastic Operator Network (SON), which will enable probabilistic prediction for operator learning tasks. In this work, we adopt the decoder-based architecture of Decoder-DeepONet rather than the standard dot-product output of DeepONet~\cite{Ryan2026}. The goal of SON is to learn a stochastic solution operator $G$ that maps an input function $u \in V$ to the SPDE solution field, namely $G(u)(y,\omega)$, where $y \in K_2$ denotes the evaluation location and $\omega$ represents the uncertainty modeled in Eq.~\eqref{SNN_1layer}. To achieve this, SON formulates the branch network as an SNN. Its output, now also including a diffusion term to capture stochasticity, is then combined with the trunk output through a decoder architecture, following the idea of Decoder-DeepONet.

Let \(A_0=\pmb{u}=[u(x_1),\hdots,u(x_m)]^{\top}\) be the discretization of the input function \(u\) at the sensor locations, which serves as the input to the branch network. We define the following forward SDE on the pseudo-time interval \([0,1]\), discretized into \(N\) pseudo-time steps:
\begin{equation}
\label{SDE_A}
A_{n+1}=A_n+f(A_n,\theta_n)h+g(A_n,\theta_n)\sqrt{h}\,\epsilon_n,
\qquad n=0,\ldots,N-1,
\end{equation}
with initial condition \(A_0\), where \(h=\frac{1}{N}\) and \(\epsilon_n\sim\mathcal{N}(0,\pmb{I}_r)\), as introduced in Eq.~\eqref{Discretized_SDEs_SampleWise}. Here, the drift term $f(\cdot,\theta_n)$ is represented by a neural network with trainable parameters $\theta_n$ at pseudo-time step $n$. Similarly, the diffusion term $g(\cdot,\theta_n)$ may be implemented either as a collection of trainable parameters or as a separate neural network. We emphasize that each SNN layer is allowed to have its own drift and diffusion components, rather than sharing the same ones across all layers. The branch network is then defined by the final state of the SDE~\eqref{SDE_A} after $N$ pseudo-time steps:
\[
\beta(u;\theta_\beta,\epsilon):=A_N(u,\epsilon;\theta_\beta),
\]
where $\theta_\beta=\{\theta_n\}_{n=0}^{N-1}$ collects all trainable parameters associated with the drift and diffusion terms.

For the trunk network, since the entire discretization of the domain is used as input, we may employ either a standard feed-forward neural network or a convolutional block network. Its output is designed to have the same dimension as $\beta(u;\theta_\beta,\epsilon)$ and is denoted by
\[
\tau(\pmb{y};\theta_\tau),
\]
where $\theta_\tau$ collects all trainable parameters of the trunk network. The SON then combines the outputs of the branch and trunk networks through an operation $\pmb{d}$:
\begin{equation}
\label{StoDeepONet}
\hat{G}_{\text{SON}}(u)(y,\epsilon)
=
\pmb{d}\bigl(\beta(u;\theta_\beta,\epsilon),\tau(\pmb{y};\theta_\tau)\bigr)+b_0,
\end{equation}
where $\pmb{d}(\cdot,\cdot)$ represents a decoder network as in Decoder-DeepONet, and $b_0$ is a trainable bias parameter. If one uses the architecture of the standard DeepONet in~\cite{Lu2021}, the decoder $\pmb{d}(\cdot, \cdot)$ in Eq.~\eqref{StoDeepONet} is replaced by an inner-product $\left\langle \cdot, \cdot \right\rangle$ as in~\cite{Ryan2026}.

We next construct the target loss function for training SON based on the Hamiltonian loss in Eq.~\eqref{Hamiltonian_loss}. However, the original Hamiltonian loss in Eq.~\eqref{Hamiltonian_loss} does not include the parameters associated with the trunk network and the decoder, since it only involves the drift and diffusion terms in the branch network. Therefore, a modification of the Hamiltonian loss in Eq.~\eqref{Hamiltonian_loss} is required. The main difficulty, however, is that the standard SNN framework assumes that the state process preserves the same dimension across all layers, whereas this assumption may fail after the branch and trunk outputs are combined through the decoder $\pmb{d}$. To address this issue, we follow~\cite{Sun2023} and adopt a modification of SNNs that allows the output dimension to change across layers (see the last paragraph of Section~6.1.1 on page~96 of~\cite{Sun2023}).

More precisely, let
\[
A_{N+1}:=\hat{G}_{\text{SON}}(u)(\pmb{y},\epsilon),
\qquad
B_{N+1}:=\Phi_a(A_{N+1},\Gamma),
\]
where $\Gamma$ denotes the reference solution in the training dataset. If $A_{N+1}$ and $B_{N+1}$ do not have the same dimension as $A_N$, then instead of using
\[
B_N
=
B_{N+1}
+
\nabla_a\!\left(
B_{N+1}^{\top}\pmb{d}\bigl(A_N,\tau(\pmb{y};\theta_\tau)\bigr)
\right)h,
\]
we define
\[
B_N
=
\nabla_a\!\left(B^{\top}_{N+1}\pmb{d}\bigl(\beta(u;\theta_\beta,\epsilon),\tau(\pmb{y};\theta_\tau)\bigr)\right)
=
\nabla_a\!\left(
B^{\top}_{N+1}\pmb{d}\bigl(A_N,\tau(\pmb{y};\theta_\tau)\bigr)\right).
\]
We then use this $B_N$ as the initial value for the sample-wise backward SDE scheme in~\eqref{Discretized_SDEs_SampleWise}, in place of $B_{N+1}$. The resulting modified discretized Hamiltonian loss is given by
\begin{equation}
\fontsize{10.5pt}{10.5pt}\selectfont
\label{modified_Ham}
H(\pmb{A},\pmb{B},\pmb{C},\pmb{\theta})
=
\frac{1}{m_B \times (N+1)}
\left[
\sum_{n=0}^{N-1}
\Bigl(B^{\top}_n f(A_n,\theta_n)+
\operatorname{tr}\;\bigl(g(A_n,\theta_n)^{\top}C_n\bigr)
\Bigr)+B_{N+1}^{\top}\pmb{d}\bigl(A_N,\tau(\pmb{y};\theta_\tau)\bigr)
\right],
\end{equation}
where $\pmb{A}$, $\pmb{B}$, $\pmb{C}$, and $\pmb{\theta}$ denote the collections of $\{A_n\}_n$, $\{B_n\}_n$, $\{C_n\}_n$, and $\{\theta_n\}_{n=0}^{N-1} \cup \{\theta_\tau\} \cup \{\theta_d\}$, respectively, and $\theta_d$ contains the parameters introduced in the decoder. The training procedure for SON is summarized in Algorithm~\ref{SONTraining}.

\begin{algorithm}[h!]
\caption{SON Training Procedure}
\label{SONTraining}
\textbf{Input:} Training dataset, learning rate $\alpha$, the number of SNN layers $N$, number of epochs $I$, mini-batch size $m_B$, the sensor points $\{x_l\}^m_{l=0}$ from the domain $K_1$, the discretization $\pmb{y}$ of $K_2$, the size of Gaussian random variable $r$.

\textbf{Initialize:} Drift networks
$f(\cdot;\theta_{\beta,n})$, diffusion networks
$g(\cdot;\theta_{\beta,n})$, $n=0,\dots,N-1$, trunk network
$\tau(\cdot;\theta_\tau)$, and decoder
$\pmb d(\cdot,\cdot;\theta_{\pmb d})$.

For each epoch $i$ in $\{0, \hdots, I-1\}$:
\begin{enumerate}
\item Randomly draw a mini-batch of pair $\{(u_j,\pmb{y}, \Gamma_j)\}^{m_B}_{j=1}$ from the dataset, where each input function \(u_j\) is discretized at grid points \(\{x_0,\dots,x_m\}\) into the vector
$\pmb{u}_j =\;[u_j(x_0),\dots,u_j(x_m)]^{\top}$, \; $j=1, \hdots, m_B$.
\item Draw $N$ samples of noise $\pmb{\epsilon}^i_n \sim N(0, \pmb{I}_r), \; n=0, \hdots, N-1$.
\item Let $A^{i}_{0} = \pmb{U} = [\pmb{u}_1, \hdots, \pmb{u}_{m_B}]^{\top}$. For $n=0, \hdots, N-1$,
\begin{enumerate}
\item {Compute $f(A^{i}_{n}; \theta^{i}_{\beta,n})$ and $g(A^{i}_{n}; \theta^{i}_{\beta, n})$ using neural networks. The diffusion term $g(A^{i}_{n}, \theta^{i}_{\beta, n})$ can be replaced by a single trainable scalar $g^{i}_{\beta}$.}
\item $A^{i}_{{n+1}} = A^{i}_{n} + f(A^{i}_{n}; \theta^{i}_{\beta, n})h + g(A^{i}_{n}; \theta^{i}_{\beta, n}) \sqrt{h} \pmb{\epsilon}^i_n,$ with $h = 1/N$.
\end{enumerate}
The state variable at the final SNN layer is the output of the branch network: $\beta(u; \theta^i_{\beta}) := A^i_{N}$.
\item  Compute the trunk output $\tau(\pmb{y}; \theta^i_{\tau}).$
\item Compute the network output: $\hat{G}_{\text{SON}}(u)(\pmb{y}, \epsilon):= A^i_{{N+1}} = \pmb{d}\left(\beta\left(u; \theta^i_{\beta}\right), \tau(\pmb{y}; \theta^i_{\tau}) \right).$
\item Initialize the backward SDE in~\eqref{Discretized_SDEs_SampleWise}. First compute $B^{i}_{N+1}= \nabla_{a} \Phi(A^i_{N+1}, \Gamma_j).$
If $A^{i}_{N}$ and $A^{i}_{N+1}$ have the same dimension, then $B^i_{N+1}$ is used directly as the terminal condition for the backward SDE. Otherwise, the transition from $B^i_{N+1}$ to $B^i_N$ is computed through the decoder by
$B^{i}_{N}
=
\nabla_a\!\left[
(B^{i}_{N+1})^{\top}
\pmb{d}\bigl(a,\tau(\pmb{y};\theta^i_\tau)\bigr)
\right]_{a=A^i_N},$
and $B^i_N$ is then used as the terminal condition for the remaining backward SDE recursion.
\item Solve the backward SDE~\eqref{adjoint_SDE} for $\left\{\left(B^i_{n}, C^i_{n}\right)\right\}^{N-1}_1$ using sample-wise backpropagation with terminal condition obtained from Step $6$.
\item Compute the modified Hamiltonian loss $H\left(\pmb{A}^i, \pmb{B}^i, \pmb{C}^i, \pmb{\theta}^i\right)$ in~\eqref{modified_Ham}.
\item Update all trainable parameters by $\pmb\theta^{i+1}
=
\pmb\theta^i
-
\alpha
\nabla_{\pmb\theta^i}
H\left(\pmb A^i,\pmb B^i,\pmb C^i,\pmb\theta^i\right), $
where $\pmb\theta^i
=
\{\theta^i_{\beta,n}\}_{n=0}^{N-1}
\cup
\{\theta^i_\tau,\theta^i_{\pmb d}\}.$
\end{enumerate}
\end{algorithm}

\begin{algorithm}[h!]
\caption{Two-Phase-SON Training Procedure}\label{2P_SONTraining}
\textbf{Input:} Training dataset, learning rate $\alpha$, the number of SNN layers $N$, number of epochs $I$, mini-batch size $m_B$, the sensor points $\{x_l\}^m_{l=0}$ from the domain $K_1$, the discretization $\pmb{y}$ of $K_2$,.

\textbf{Phase I training:} Train the deterministic model using Decoder-DeepONet over the provided training dataset. Denote the trained model as $S_{\text{DO}}$.

\textbf{Phase II training:} For each epoch $i$ in $\{0, \hdots, I-1\}$:
\begin{enumerate}
\item Randomly draw a mini-batch of pair $\{(u_j,\pmb{y}, \Gamma_j)\}^{m_B}_{j=1}$ from the dataset, where each input function \(u_j\) is discretized at grid points \(\{x_0,\dots,x_m\}\) into the vector
\[
\pmb{u}_j =\;(u_j(x_0),\dots,u_j(x_m)), \; \; j=1, \hdots, {m_B}.
\]
\item Compute $\pmb{u}_{\text{DO}, j} = S_{\text{DO}}(\pmb{u}_j, \pmb{y})$, \; $j=1, \hdots, m_B$. Denote $\pmb{U}_{\text{DO}} = \left[\pmb{u}_{\text{DO}, 1}, \hdots, \pmb{u}_{\text{DO}, m_B}\right]$.
\item Repeat Steps 2, 3 in Algorithm~\ref{SONTraining} with initial condition $A^{i}_{0} = \pmb{U}_{\text{DO}}$. The final output $A^{i}_{N}$ for the network is the result from the final layer of the forward SDE~\eqref{SDE_A}:
\begin{align*}
\hat{G}_{\text{SON}}(u)(\pmb{y}, \epsilon) := A^{i}_{N}.
\end{align*}
\item Solve the backward SDE~\eqref{adjoint_SDE} by sample-wise backpropagation with terminal condition $B^i_{N}:= \nabla_{A^i_{N}}\Phi\left(\hat{G}_{\text{SON}}(u)(y, \epsilon), \Gamma^i\right)$.
\item Compute the discretized Hamiltonian~\eqref{discretized_Ham}.
\item Update the parameters: $\pmb{\theta}^{i+1} =\pmb{\theta}^{i} - \alpha \nabla_{\theta} H(\pmb{A}^{i},\pmb{B}^{i},\pmb{C}^{i};\pmb{\theta}^{i}),$ where $\pmb{\theta}^i = \left\{\theta^i_n\right\}^{N-1}_{i=0}$. Note that $\pmb{\theta}^{i}$ only includes the parameters from the drift and the diffusion networks.
\end{enumerate}

\end{algorithm}
\subsection{A two-phase training strategy for stochastic operator network}\label{TwoP_SON}
As seen from Algorithm~\ref{SONTraining}, the training procedure of SON updates the parameters of the branch network, the diffusion term, the trunk network, and the decoder within a single training loop. This can be computationally expensive when the target system is complex and requires large branch and trunk networks. Moreover, when the uncertainty is solution-dependent, an additional network is typically introduced to model the diffusion term, which further increases the training cost.

To address this issue, we introduce a two-phase training procedure for SON, consisting of a pre-training phase (Phase I) and a fine-tuning phase (Phase II). 
In Phase I, we train a deterministic neural operator based on Decoder-DeepONet, denoted by $S_{\text{DO}}$, using the noisy training data to obtain a baseline solver. In Phase II, we apply the SNN to the output of $S_{\text{DO}}$ and propagate this output through the SDE dynamics~\eqref{SDE_A} to produce a probabilistic prediction.

The output from Phase I is inherently limited in that it is deterministic and therefore cannot represent the stochastic component of the solution. Nevertheless, the purpose of Phase I is to capture the overall shape of the solution. Then, in Phase II, only a small refinement network is needed to parameterize the drift term in Eq.~\eqref{SDE_A} and fine-tune the deterministic component, so that the second phase focuses primarily on learning the stochasticity. 
This two-phase strategy separates the cost of learning the solution structure from that of learning stochastic uncertainty. Therefore, the deterministic solver $S_{\text{DO}}$ can be made sufficiently expressive to capture the mean solution behavior, while the drift network in Phase II can remain lightweight and stable. As a result, in Phase II, more model capacity can be allocated to the diffusion term in the SNN, allowing it to capture more complex uncertainty structures. An additional advantage of this two-phase design is that, since the SNN is applied to the output of the deterministic neural operator, it removes the need to embed the trunk network within the SNN layers. Consequently, the architecture modification previously introduced to handle dimension changes is no longer required.

The target loss function for Phase I is the loss $\Phi(\cdot,\cdot)$, while the original Hamiltonian loss~\eqref{discretized_Ham} is reused in Phase II. The pseudo-code for the proposed two-phase SON training procedure is given in Algorithm~\ref{2P_SONTraining}.

\section{Numerical Experiments}\label{NumericalExps}

% To simplify our presentation, we use the following generic form to represent the numerical solver that propagates the approximate solution of the SPDE \eqref{SPDE}:
% \begin{equation}\label{SPDE:scheme}
% {u_{t_{n+1}}} = \mathbb{F}({u_{t_n}}, t_n, \tilde{\omega}, \Delta \tilde{W}_{t_n}),
% \end{equation}
% where $\Delta \tilde{W}_{t_n}: = \tilde{W}_{t_{n+1}} - \tilde{W}_{t_n}$, and $\mathbb{F}$ denotes a forward operator that incorporates the random variables $\tilde{\omega}$ and $\Delta \tilde{W}_{t_n}$ within a UQ-enabled PDE solver. This operator advances the solution $u$ in Eq.~\eqref{SPDE} from time $t_{n}$ to $t_{n+1}$, with ${u_{t_{n+1}}}$ being a numerical approximation to ${u(\cdot, {t_{n+1}})}$. 

In this section, we present four numerical examples to demonstrate the performance of our diffusion-based SON approach in learning SPDE solutions, as well as its ability to provide well-calibrated uncertainty estimates across a range of scenarios. We begin with two time-independent PDEs: a 2D reaction–diffusion equation and a 2D advection-diffusion equation. We then consider time-dependent problems, including a 2D heat equation and a 2D Burgers’ equation, which are generally more challenging for UQ-enabled AI solvers.

\subsection{Reaction-Diffusion Equation}
\label{sec4}
We first consider the following boundary value reaction-diffusion equation:
\begin{equation}
\label{2DReactDiff}
\begin{array}{l}
 -\Delta{u(x, y)}+\kappa u(x, y) = 5 \tilde{f}(x, y), \; (x, y) \in (0, 1)^2, \vspace{0.1cm} \\
 \hspace{0.5cm} u(x, 0) = 0.5x, \; u(x, 1) = 0.5(x+1), \vspace{0.1cm} \\
 \hspace{0.5cm}  u(0, y) = 0.5y, \; u(1, y) = 0.5(y+1),
\end{array}
\end{equation}
where $\kappa = 20$. In this example, we aim to demonstrate \textit{the baseline performance} of the SON approach. To this end, we assume that the solution of the above reaction–diffusion equation is perturbed by uniform additive noise and denote the resulting stochastic solution by $u_{\tilde{f}} + \alpha\omega$, where $\omega$ is the noise sampled from $N(0, 1)$, and we set $\alpha = 0.06$ in this example. Our goal is then to learn the corresponding stochastic solution operator $G(\tilde{f})(\cdot, \cdot; \omega) \approx u_{\tilde{f}}(\cdot, \cdot) + \alpha\omega$ with respect to the forcing term $\tilde{f}$. %The size of the Gaussian random variable in Eq.~\eqref{SDE_A} is chosen to be $r=1$.

The training and testing datasets are generated by numerically solving the PDE system~\eqref{2DReactDiff} using a second-order central finite-difference scheme on a uniform mesh with step size \(h = 1/N\), where \(N = 30\). The numerical solution \(u_{\tilde{f},h}\) is computed on the \((N+1)\times (N+1)\) grid and stored as a matrix of size \((N+1)\times (N+1)\). We generate \(1800\) forcing terms \(\tilde{f}\) evaluated on the same grid points and split them into \(1500\) training samples and \(300\) testing samples. Each sample consists of three tensors: the forcing values \(\tilde{f}\), the grid points, and the reference solution \(u_{\tilde{f},h}\). 

We next describe the architecture of the neural networks used in formulating the SON. Since Phase~I uses the Decoder-DeepONet, the grid points do not need to be concatenated with the input function values. Accordingly, the branch input has size \(1500 \times 31 \times 31\), while the trunk input (grid points) has size \(1500 \times 31 \times 31 \times 2\). Although the finite-difference solver only requires \(\tilde{f}\) at interior nodes when boundary conditions are prescribed, we provide \(\tilde{f}\) on the full grid to the branch network for simplicity. The Phase~I network outputs a tensor of size \(1500 \times 31 \times 31\), which is compared with the reference solution using the mean squared error (MSE) loss to train the parameters of the branch and trunk networks. Since both the input and the output are 2D fields, we use convolutional networks rather than flattening the data into 1D vectors and applying a fully connected network. Specifically, the branch network is a 10-layer CNN, where the first two layers are projection layers that downsample the input by a factor of 2 to reduce computational cost. Similarly, the trunk network consists of six convolutional blocks, with the first two blocks serving as projection (downsampling) blocks. The outputs of the branch net and the trunk net have the same shape, $1500 \times 7 \times 7$. We then concatnate them to form a tensor of size $1500 \times 2 \times 7 \times 7$, which is fed into a decoder. The decoder first upsamples the resolution back to $1500 \times 2 \times 31 \times 31$, and then applies several additional convolutional layers for refinement.

We initialize Phase~II by using the trained deterministic model from Phase~I. Following~\cite{Sun2023}, the drift term $f$ in Eq.~\eqref{SDE_A} is parameterized by
\begin{equation}
\label{drift_phase2}
f\!\left(A,\theta_{f,t_n}\right)
= \sum_{i=1}^{L_f} a_i \, \mu\!\left(\mathrm{NN}_{i}(A; \theta_{i,t_n})\right),
\qquad n=0,\ldots,N-1,
\end{equation}
where $\{a_i\}_{i=1}^{L_f}$ are trainable weights initialized uniformly on $[-\beta,\beta]$, 
$\mathrm{NN}_i$ for $i=1,\ldots,L_f$ are refinement networks (e.g., a small multi-layer convolutional block, or a column-wise MLP applied to $X$), 
and $\mu$ denotes the sigmoid activation function. $\theta_{f, t_n}$ is then the collection of all parameters $\theta_{i, t_n}$ from the network $\mathrm{NN}_i, \; i=1, \hdots, L_f$. In general, the refinement networks may be time-dependent through the parameters $\theta_{f,t_n}$. 
In this example, we set $\beta=3$, $L_f=4$, and $N=16$, and we adopt a time-independent parameterization, i.e., $\theta_{f,t_n} \equiv \theta_f$, $n=0,\ldots,N-1.$
% \begin{align*}
% \theta_{f,t_n} \equiv \theta_f \qquad \text{for all } n=0,\ldots,N-1.
% \end{align*}
Similarly, the diffusion term is chosen as $\sigma\!\left(A,\theta_{g,t_n}\right) \equiv \sigma\left(\theta_g\right) = \sum_{l=1}^{L_g} b_l \, \mu(c_l)$,
% \begin{equation}
% \label{diffusion_phase2}
% \sigma\!\left(A,\theta_{g,t_n}\right) \equiv \sigma\left(\theta_g\right)
% = \sum_{l=1}^{L_g} b_l \, \mu(c_l),
% \qquad n=0,\ldots,N-1,
% \end{equation}
where $L_g=3$, $\{b_l\}_{l=1}^{L_g}$ are initialized uniformly on $[0.035,0.055]$, 
and $\{c_l\}_{l=1}^{L_g}$ are initialized from $0.05\,\mathcal{N}(0,1)$. $\theta_g$ is then the collection of the weights $\{b_l\}^{L_g}_{l=1}$ and the input $\{c_l\}^{L_g}_{l=1}$.
\vspace{0.5em}

\begin{figure}[h!]
\begin{minipage}{0.5\textwidth}
\includegraphics[scale = 0.18]{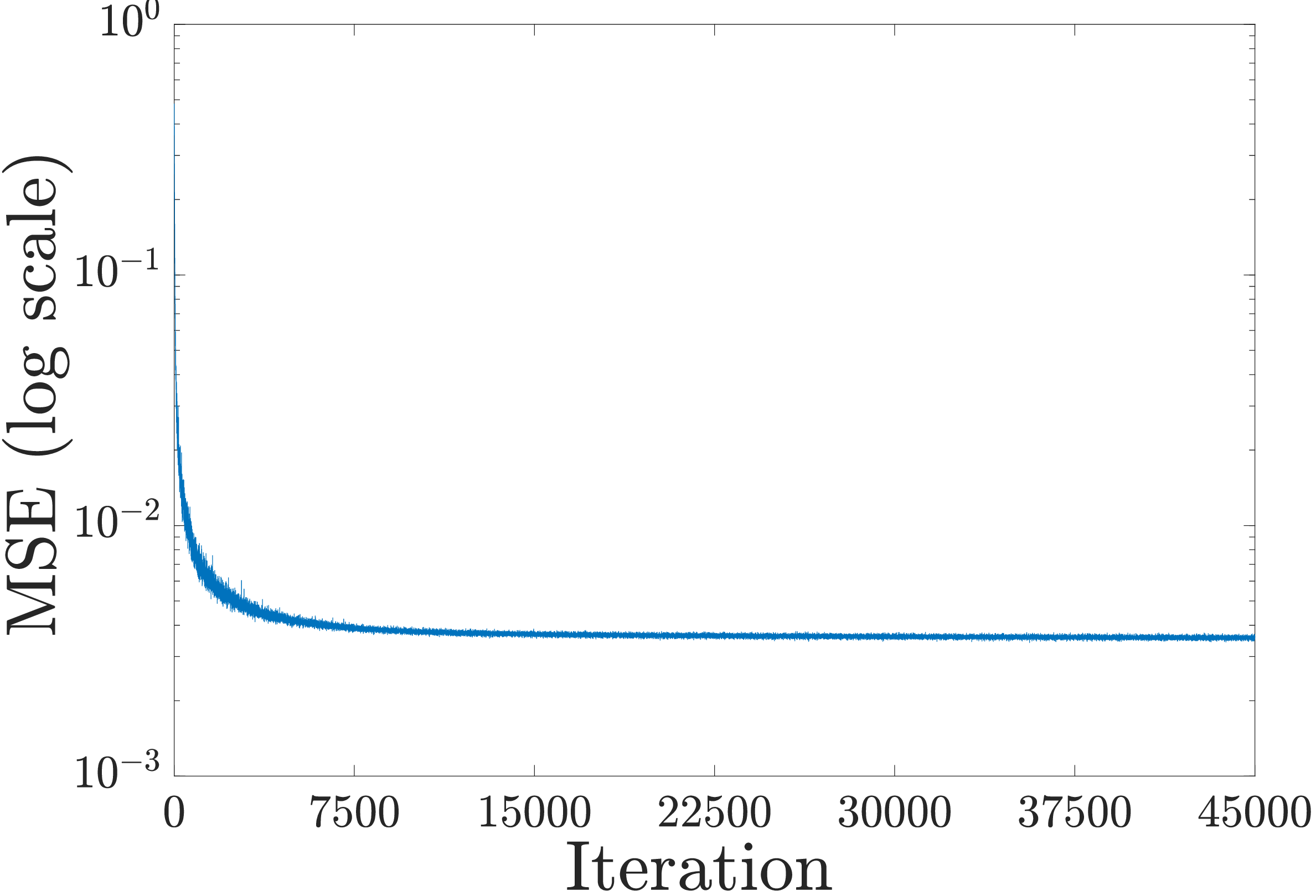}
\end{minipage}%
\begin{minipage}{0.5\textwidth}
\includegraphics[scale = 0.18]{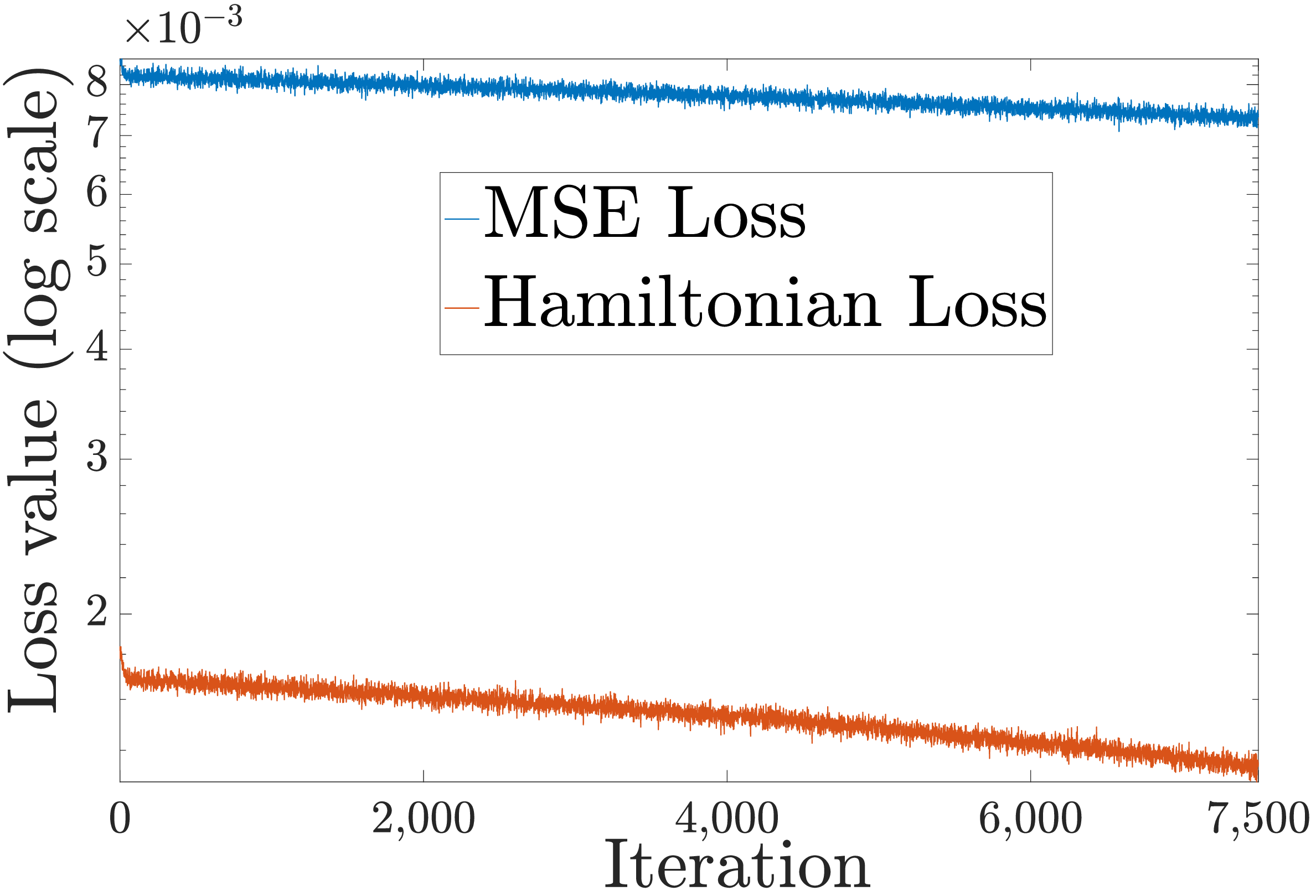}
\end{minipage}
\caption{\footnotesize [Reaction-Diffusion] (Left) MSE loss from Phase 1. (Right) MSE loss and Hamiltonian loss~\eqref{discretized_Ham} from Phase 2.}
\label{Losses_PhaseI_PhaseII}
\end{figure}

In this example, we first demonstrate the convergence behavior of the training process. We set the maximum numbers of training epochs for Phase~I to $3000$ and Phase~II to $500$, where Phase~II is designed for fine-tuning and uncertainty calibration. In both phases, each epoch consists of $100$ mini-batches. To improve computational efficiency, the training process is terminated early once the loss functions stabilize. %As a result, in this experiment Phase~I and Phase~II are trained for only $450$ and $75$ epochs, respectively.  
We employ the Adam optimizer and a cosine learning-rate scheduler in both phases, with learning rate $10^{-5}$ for Phase~I and $10^{-4}$ for Phase~II.  We present the loss curves for Phase~I and Phase~II in Figure~\ref{Losses_PhaseI_PhaseII}.  %Since the loss is recorded after each mini-batch update, we obtain $45000$ loss values for Phase~I and $7500$ for Phase~II. 
We observe that the losses in both phases decrease steadily and stabilize, indicating good convergence. Note that the MSE in Phase~II changes only slightly, as this phase primarily focuses on learning the diffusion coefficient while only mildly fine-tuning the drift term. This behavior is consistent with the design motivation of our two-phase SON framework.
% \newpage
\begin{figure}[h!]
\begin{minipage}{0.333\textwidth}
\includegraphics[scale= 0.14]{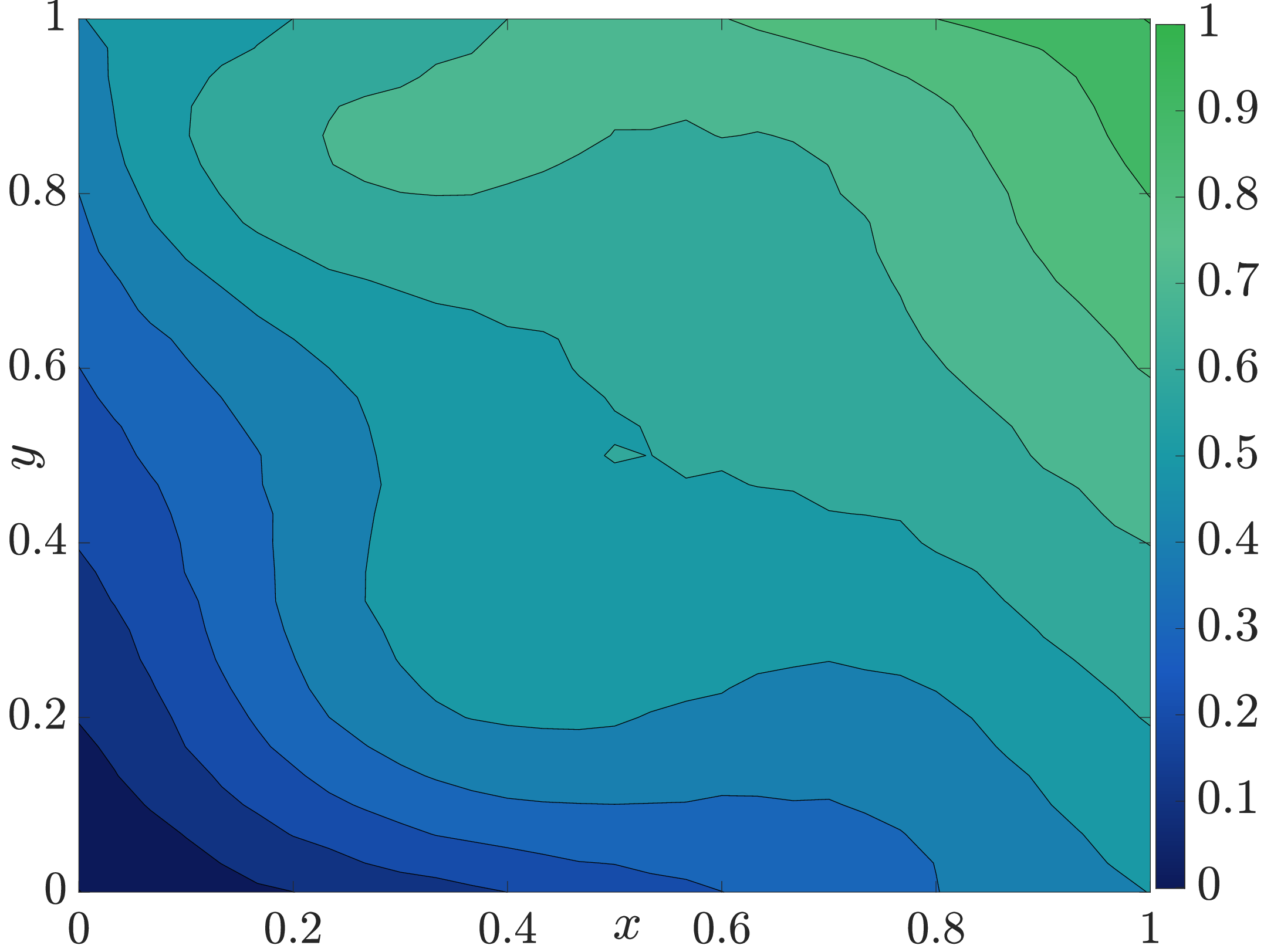}
\end{minipage}%
\begin{minipage}{0.333\textwidth}
\includegraphics[scale= 0.14]{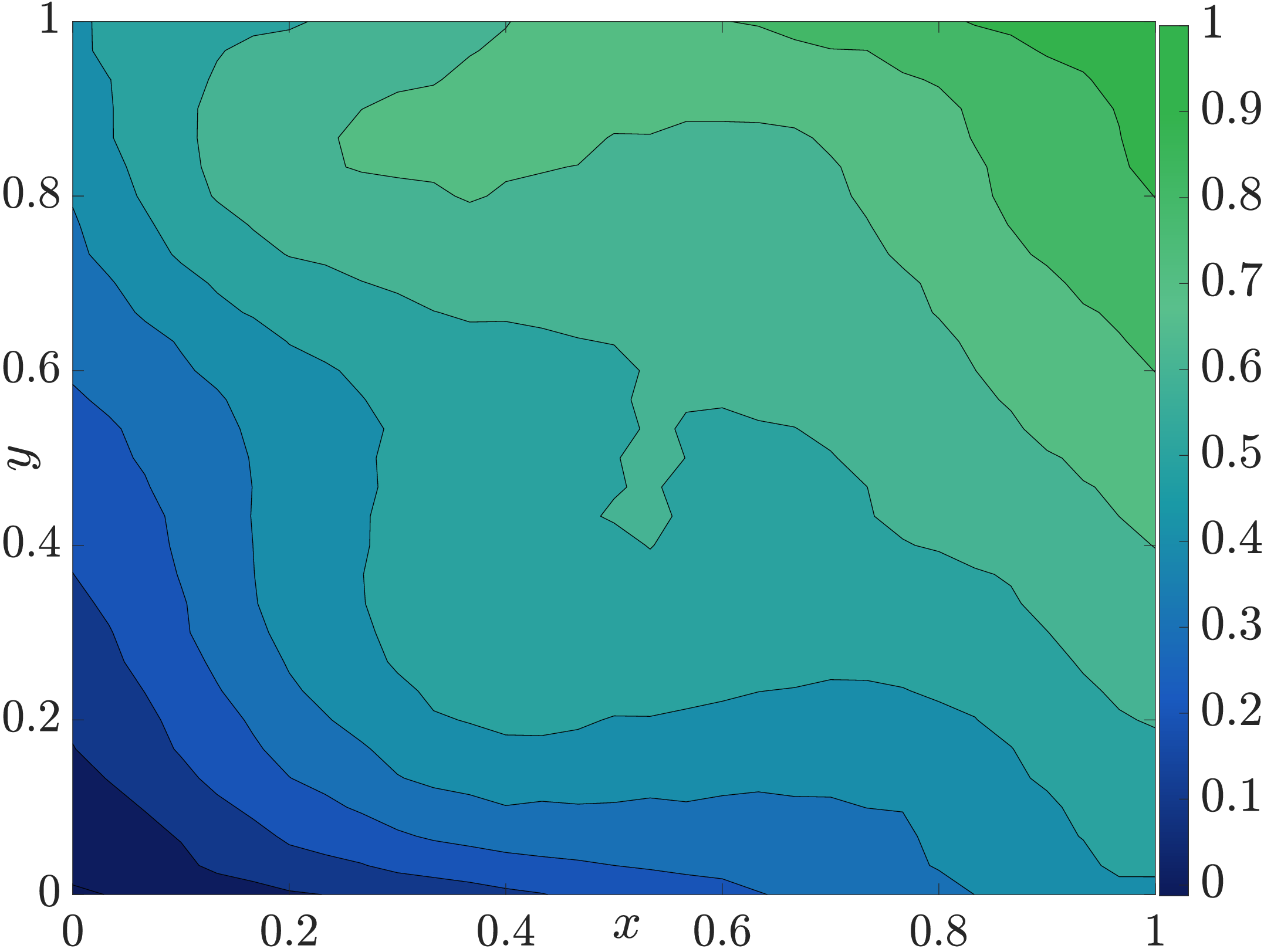}
\end{minipage}%
\begin{minipage}{0.333\textwidth}
\includegraphics[scale= 0.14]
{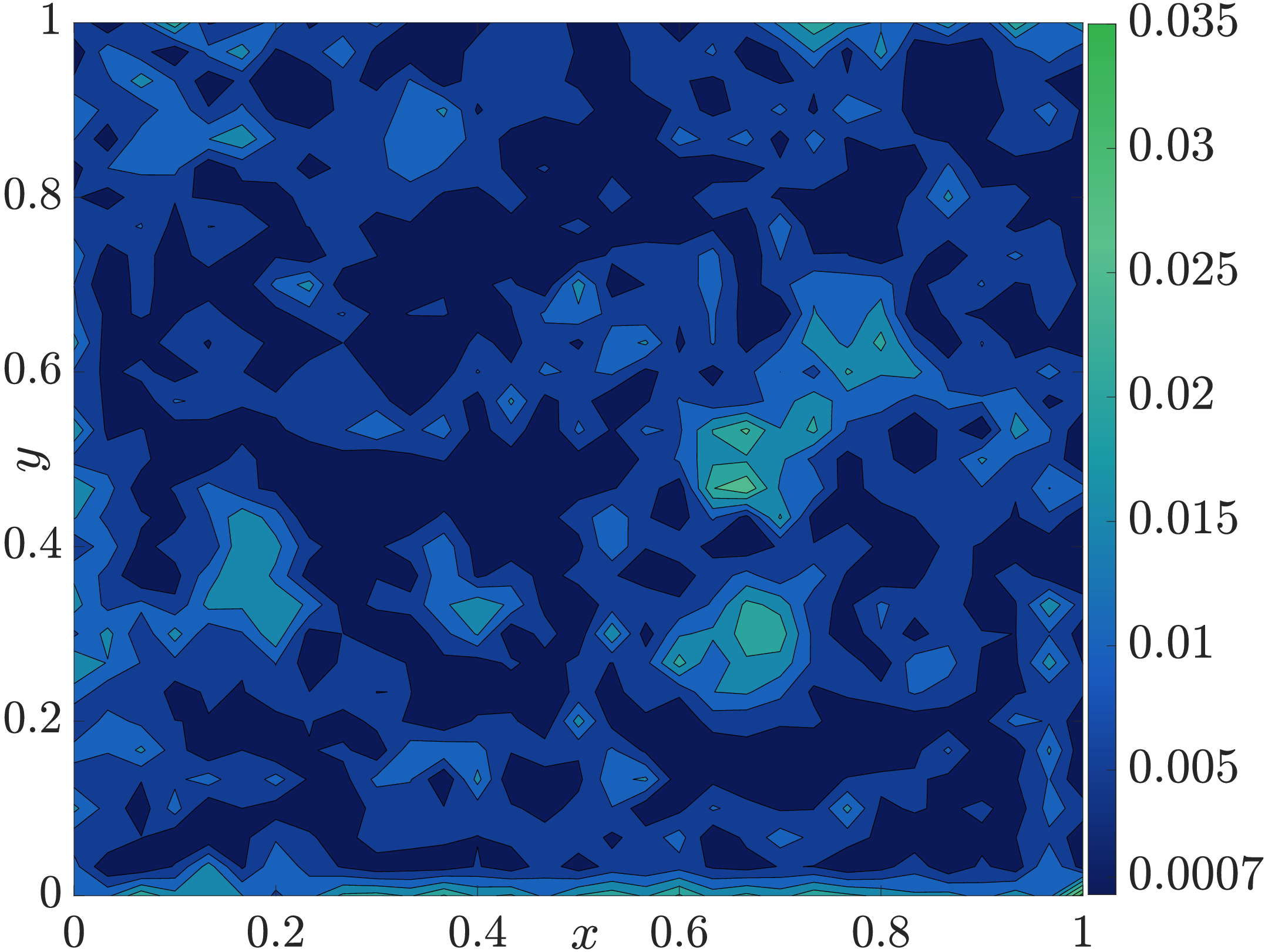}
\end{minipage}
\caption{\footnotesize [Reaction-Diffusion] Solution prediction: (Left) Reference solution mean . (Middle) Predicted solution mean. (Right) Prediction mean error.}
\label{SingleInput_2DReactDiff}
\vspace{-0.2cm}
\end{figure}
% \begin{figure}[h!]
% \begin{minipage}{0.5\textwidth}
% \hspace{0.5cm}\includegraphics[scale= 0.185]{Figures/2DReactDiff/Update/Prediction1/Mean_Input1.eps}
% \end{minipage}%
% \begin{minipage}{0.5\textwidth}
% \includegraphics[scale= 0.185]{Figures/2DReactDiff/Update/Prediction1/Mean_Input1_Redraw.eps}
% \end{minipage}%
% \caption{\footnotesize [2D Reaction-Diffusion] Single-Input Comparison. (First) Mean (Pred vs Ref.). (Right) Absolute Mean Error.}
% \label{SingleInput_2DReactDiff}
% \vspace{-0.5cm}
% \end{figure}

We next investigate the predictive performance of the SON model. %Since this is a relatively simple test case, the deterministic model from Phase I already captures the overall shape of the solution well, and Phase II only slightly refines the mean prediction. Therefore, we do not separately report the Phase I prediction in this section. 
We evaluate the performance of SON in terms of both predictive accuracy of the solution profiles and the quality of uncertainty quantification. To this end, we randomly select one input function from the testing dataset and generate $400$ predicted samples by using the trained SON. For the same input, we also generate $400$ reference solution samples from the corresponding SPDE solver. The left and middle subfigures of Figure~\ref{SingleInput_2DReactDiff} display the 2D heatmaps of the reference and predicted sample means, computed over $400$ predicted samples, while the right subfigure shows the heatmap of the prediction error between these two means. For this testing input, the reference and predicted solution surfaces nearly overlap, which is consistent with the error heatmap, where the maximum prediction error is approximately $0.035$.
% \begin{figure}[h!]
% \begin{minipage}{0.32\textwidth} 
% \includegraphics[scale= 0.235]{Figures/2DReactDiff/Update/Prediction1/Mean_Input1.eps}
% \end{minipage}%
% \begin{minipage}{0.32\textwidth}
% \includegraphics[scale= 0.235]{Figures/2DReactDiff/Update/Prediction2/Mean_Input2.eps}
% \end{minipage}%
% \begin{minipage}{0.32\textwidth} 
% \includegraphics[scale= 0.235]{Figures/2DReactDiff/Update/Prediction3/Mean_Input3.eps}
% \end{minipage}
% \caption{(First) Prediction 1. (Second) Prediction 2. (Third) Prediction 3.}
% \end{figure}

% \begin{figure}[h!]
% \begin{minipage}{0.333\textwidth} 
% \includegraphics[scale= 0.175]{Figures/2DReactDiff/Update/MeanError_Input1.eps}
% \end{minipage}%
% \begin{minipage}{0.333\textwidth} 
% \includegraphics[scale= 0.175]{Figures/2DReactDiff/Update/MeanError_Input2.eps}
% \end{minipage}%
% \begin{minipage}{0.333\textwidth} 
% \includegraphics[scale= 0.175]{Figures/2DReactDiff/Update/MeanError_Input3.eps}
% \end{minipage}
% \caption{Heatmap of Absolute Error. (First) Prediction 1. (Second) Prediction 2. (Third) Prediction 3.}
% \end{figure}

% \vspace{-0.3cm}

To demonstrate the capability of SON in quantifying uncertainty, we estimate the standard deviation (std) $\sigma$ of the predicted samples using 400 SON generated realizations. The resulting estimated std is $0.0603$, which is close to the reference value $0.06$, indicating that the SON accurately captures the magnitude of the stochastic perturbation. We repeat this experiment across eight different testing input functions and compute the average estimated std, obtaining a value of $0.0605$, which further validates the accuracy of SON’s uncertainty quantification.

\subsection{Advection-Diffusion Equation}
In the second example, we consider the following 2D advection-diffusion equation in $\Omega = [0, 1]^2$ given by
\begin{equation}
\label{2D_AdvDiff}
\left\{\begin{array}{rl}
-\varepsilon\Delta{u}(x, y) + \pmb{b}(x, y)\cdot \nabla{u(x, y)} = \tilde{f}(x,y) &  \; \text{in} \; \Omega, \vspace{0.1cm} \\
u(x, y) = w(x) & \; \text{on} \; \partial\Omega,
\end{array}\right.
\end{equation}
where $\varepsilon>0$ and the boundary condition $w(x)$ is given by
\begin{align*}
   w(x) = \left\{ \begin{array}{l}
     1, \; x \in \Gamma, \vspace{0.1cm} \\
     0, \; x \in \Omega\backslash\Gamma,
    \end{array}\right.
\end{align*}
where $\Gamma := \left\{x =0, \; y \in [0, 0.5]\right\} \cup \left\{x \in [0, 1], \; y = 0\right\}$ is a subset of $\partial\Omega.$ We set $\epsilon = 0.05$ and fix the velocity field as:
\begin{equation}
\label{VeloField}
\begin{array}{l}
\pmb{b}(x, y) = \begin{pmatrix}
\frac{1}{5}\cos(\eta(x, y)) \vspace{0.1cm} \\
\frac{1}{5}\sin(\eta(x, y))
\end{pmatrix},
\end{array}
\end{equation}
where 
\begin{align*}
\eta(x, y) = \left\{\begin{array}{ll}
0.25, \; &0 \leq x \leq  0.5, \; 0 \leq y \leq 0.5, \vspace{0.1cm}\\
0.15,\; &0.5 <x \leq  1, \; 0 \leq y \leq 0.5, \vspace{0.1cm}\\
-0.20,\; &0 \leq x \leq 0.5, \; 0.5 < y \leq 1, \vspace{0.1cm}\\
-0.35,\; &0.5 <x \leq  1, \; 0.5 < y \leq 1.
\end{array}\right.
\end{align*}
For this example, we consider two scenarios for incorporating space-dependent uncertainty: uncertainty in the forcing term $\tilde{f}(\cdot, \cdot)$ (Case 1) and uncertainty in the velocity field $\pmb{b}(\cdot, \cdot)$ (Case 2).

\subsubsection{Case 1: Space-dependent uncertainty in the right-hand side}
For Case 1, we consider a forcing term with space-dependent uncertainty, i.e., $\tilde{f}_{\xi}(x, y) = \tilde{f}(x, y) + \xi(x, y)$.
% \begin{align*}
% \begin{array}{l}
% f_{\xi}(x, y) = f(x, y) + \xi(x, y).
% \end{array}
% \end{align*}
To construct the training and testing datasets, we uniformly discretize $\Omega$ by a rectangular grid with mesh size $h=M^{-1}$, where $M=40$, and solve system~\eqref{2D_AdvDiff} using a mixed-hybrid finite element method~\cite{Brunner2014}. The forcing term and the velocity field are represented at the centers of the rectangular cells. The noise field $\xi(\cdot,\cdot)$ is discretized as a vector $\pmb{\xi}_h\in\mathbb{R}^{M^2}$, $\pmb{\xi}_h=\{\xi_j\}_{j=1}^{M^2}$,
% \begin{equation}
% \label{AdvDiff_NoiseField}
% \pmb{\xi}_h=\{\xi_j\}_{j=1}^{M^2},
% \end{equation}
where $\xi_j\sim \alpha_j\mathcal{N}(0,1)$ and each coefficient $\alpha_j$ is sampled once from the uniform distribution on $[0.0007,0.001]$, for $k=1,\ldots,M^2$. %The size of the Gaussian random variable in~\eqref{SDE_A} is chosen to be $r=1$.

In this case, we generate $2400$ forcing terms from a 2D Chebyshev polynomial space and compute the corresponding reference solutions. The resulting forcing-solution pairs are divided into $2000$ training samples and $400$ testing samples. Each sample consists of the forcing term, the grid points, and the reference solution.
% for the deterministic model in Phase I and the stochastic model in Phase II. In particular, the drift term and the diffusion term are formulated as the combination of sigmoid activation functions as in~\eqref{drift_phase2} and ~\eqref{diffusion_phase2}. 

We adopt architectures in the first example. To better capture boundary effects, we increase the complexity of the refinement network $\mathrm{NN}_i$ in Eq.~\eqref{drift_phase2} by augmenting the convolutional blocks with several column-wise MLP networks. In this experiment, we set $N=15$ and retain $L_f=4$ and $L_g=3$, as well as the distributions of the two sets of weights $\{a_i\}_{i=1}^{L_f}$ and $\{b_i\}_{i=1}^{L_g}$. However, since the noise is space-dependent, we replace the scalar coefficients $\{c_i\}_{i=1}^{L_g}$ with $M\times M$ matrices whose entries are sampled i.i.d. from $0.05\mathcal{N}(0,1)$. In each $\mathrm{NN}_i$, the MLP networks are applied to the last three columns of the feature map output by the convolutional block.

To demonstrate SON’s capability in handling uncertainty in SPDE solutions, we compare its predictive performance with that of the \textit{deterministic DeepONet}. Specifically, we randomly select one testing input and compute predictions using both models. For SON, we generate 400 samples and compute the sample mean. For the reference solution, we use the numerical solver to generate 400 samples and compute the corresponding mean.
\begin{figure}[h!]
\hspace{1cm}
% \begin{minipage}{0.333\textwidth} 
% \includegraphics[scale= 0.16]{Figures/2DAdvDiff/RHSNoise/FineTune_Phase2/Redraw/Prediction1/ErrorMap_Phase1_Input1.eps}
% \end{minipage}%
% \begin{minipage}{0.333\textwidth}
% \includegraphics[scale= 0.16]{Figures/2DAdvDiff/RHSNoise/FineTune_Phase2/Redraw/Prediction2/ErrorMap_Phase1_Input2.eps}
% \end{minipage}%
% \begin{minipage}{0.45\textwidth} 
% \includegraphics[scale= 0.18]{Figures/2DAdvDiff/RHSNoise/FineTune_Phase2/Redraw/Prediction3/ErrorMap_Phase1_Input3.eps}
% \end{minipage}
% % \begin{minipage}{0.333\textwidth}  
% % \includegraphics[scale= 0.16]{Figures/2DAdvDiff/RHSNoise/FineTune_Phase2/Redraw/Prediction1/ErrorMap_Input1.eps}
% % \end{minipage}%
% % \begin{minipage}{0.333\textwidth} 
% % \includegraphics[scale= 0.16]{Figures/2DAdvDiff/RHSNoise/FineTune_Phase2/Redraw/Prediction2/ErrorMap_Input2.eps}
% % \end{minipage}%
% \begin{minipage}{0.45\textwidth}
% \includegraphics[scale= 0.18]{Figures/2DAdvDiff/RHSNoise/FineTune_Phase2/Redraw/Prediction3/ErrorMap_Input3.eps}
% \end{minipage}
% \caption{\footnotesize [Advection-Diffusion Case 1] Heatmaps of mean errors: (left) DeepONet; (right) SON.}
% \label{AdvDiff_Compare_ErrorMap}
% \end{figure}
\begin{minipage}{0.45\textwidth} 
\includegraphics[scale= 0.18]{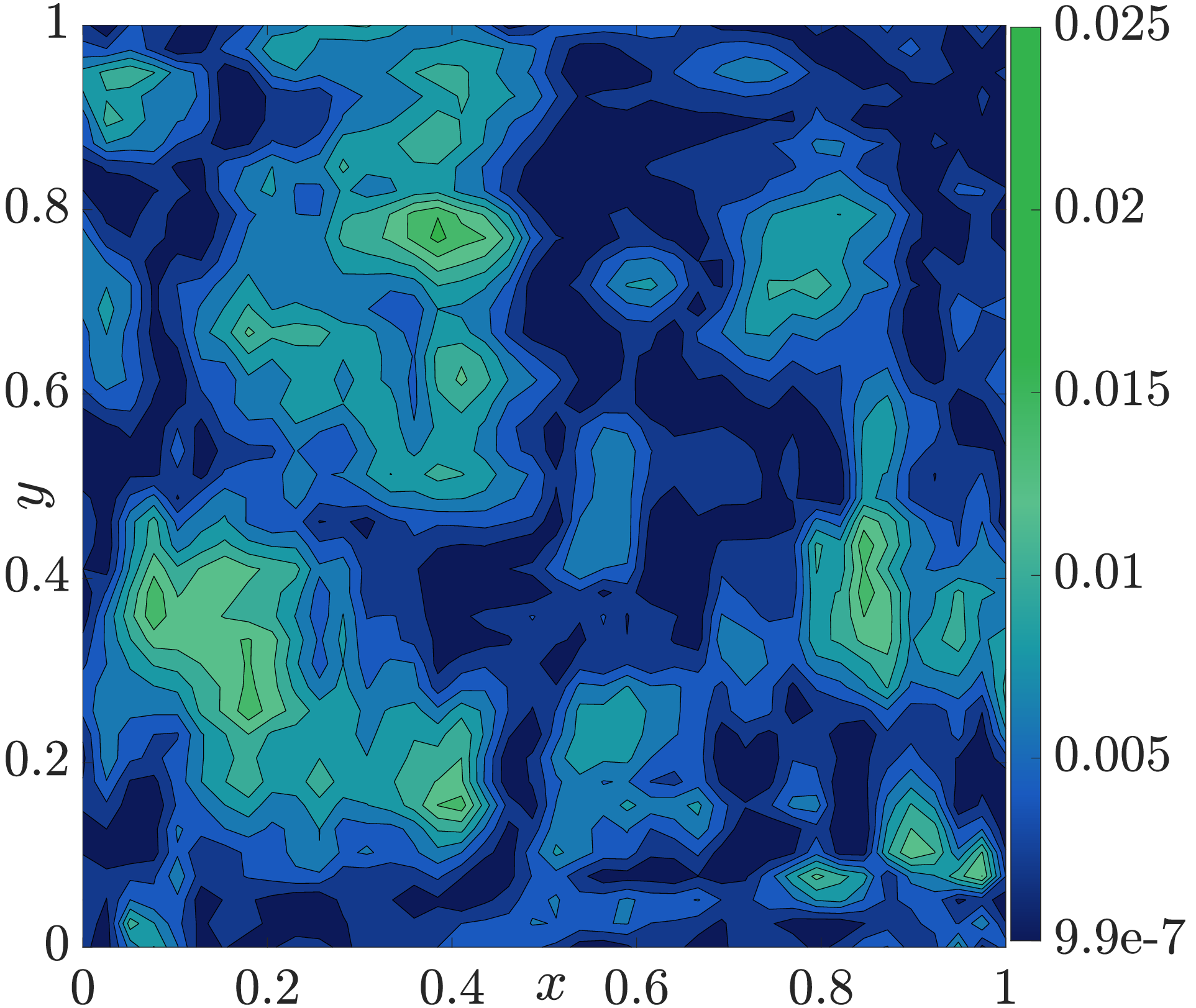}
\end{minipage}
\begin{minipage}{0.45\textwidth}
\includegraphics[scale= 0.18]{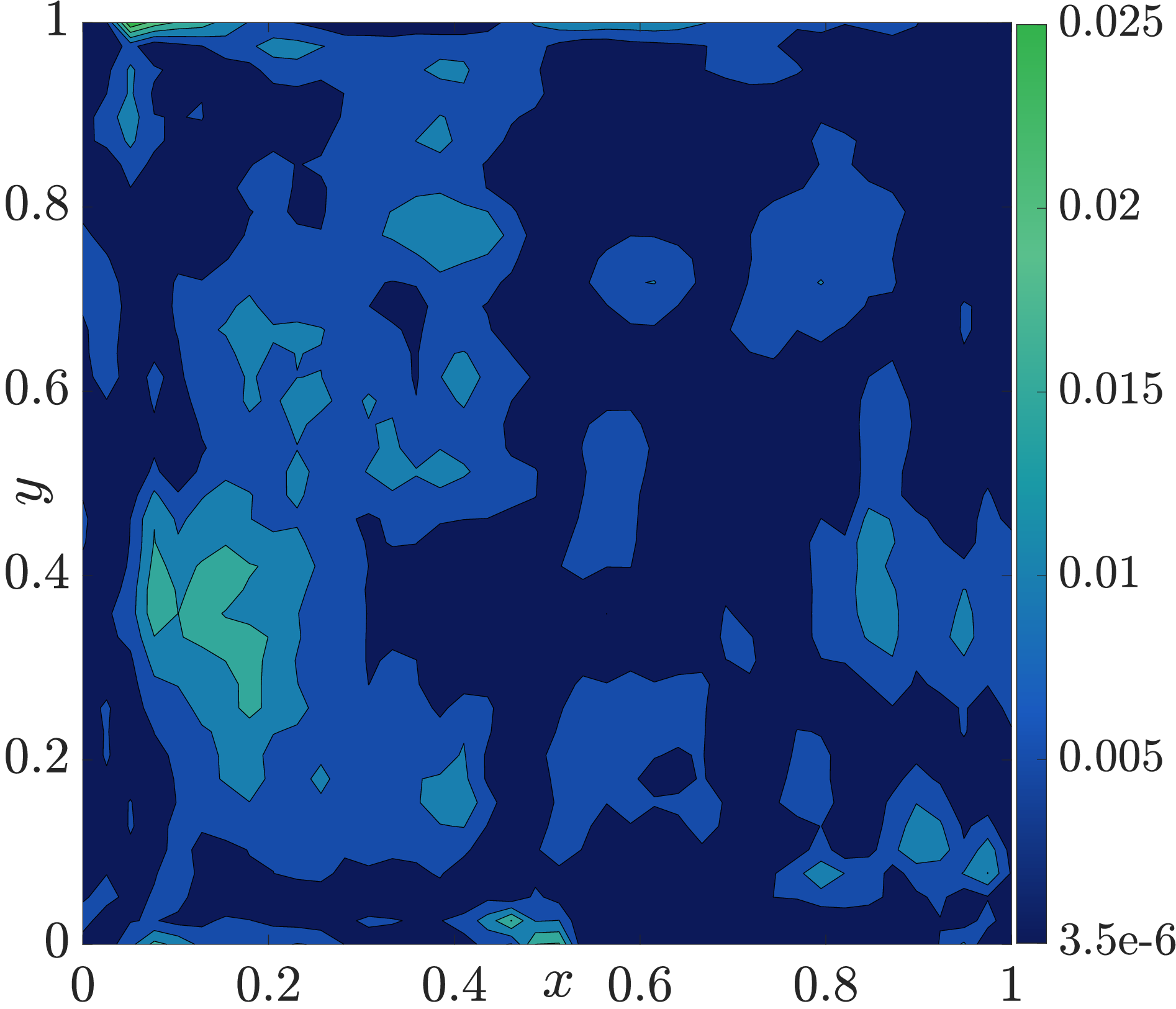}
\end{minipage}
\caption{\footnotesize [Advection-Diffusion Case 1] Heatmaps of mean errors: (Left) DeepONet; (Right) SON.}
\label{AdvDiff_Compare_ErrorMap}
\end{figure}
Figure~\ref{AdvDiff_Compare_ErrorMap} presents heatmaps of the mean error between the reference solution and the predictions from the deterministic DeepONet (left) and SON (right), respectively. This comparison shows that incorporating the diffusion term into the DeepONet architecture does not degrade predictive performance and, in fact, improves accuracy.

\begin{figure}[h!]
\begin{minipage}{0.333\textwidth} 
\includegraphics[scale= 0.15]{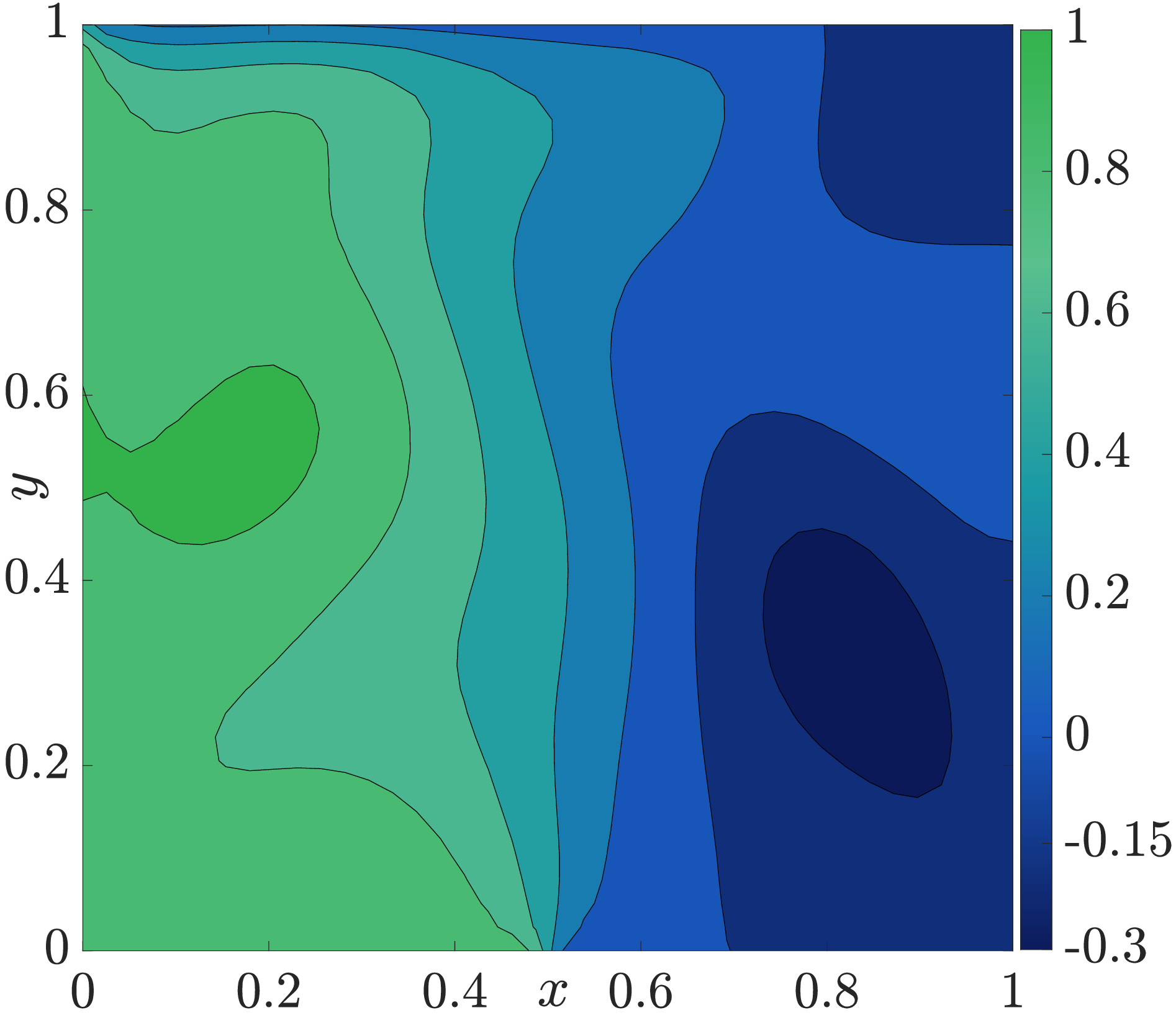}
\end{minipage}%
\begin{minipage}{0.333\textwidth}  
\includegraphics[scale= 0.15]{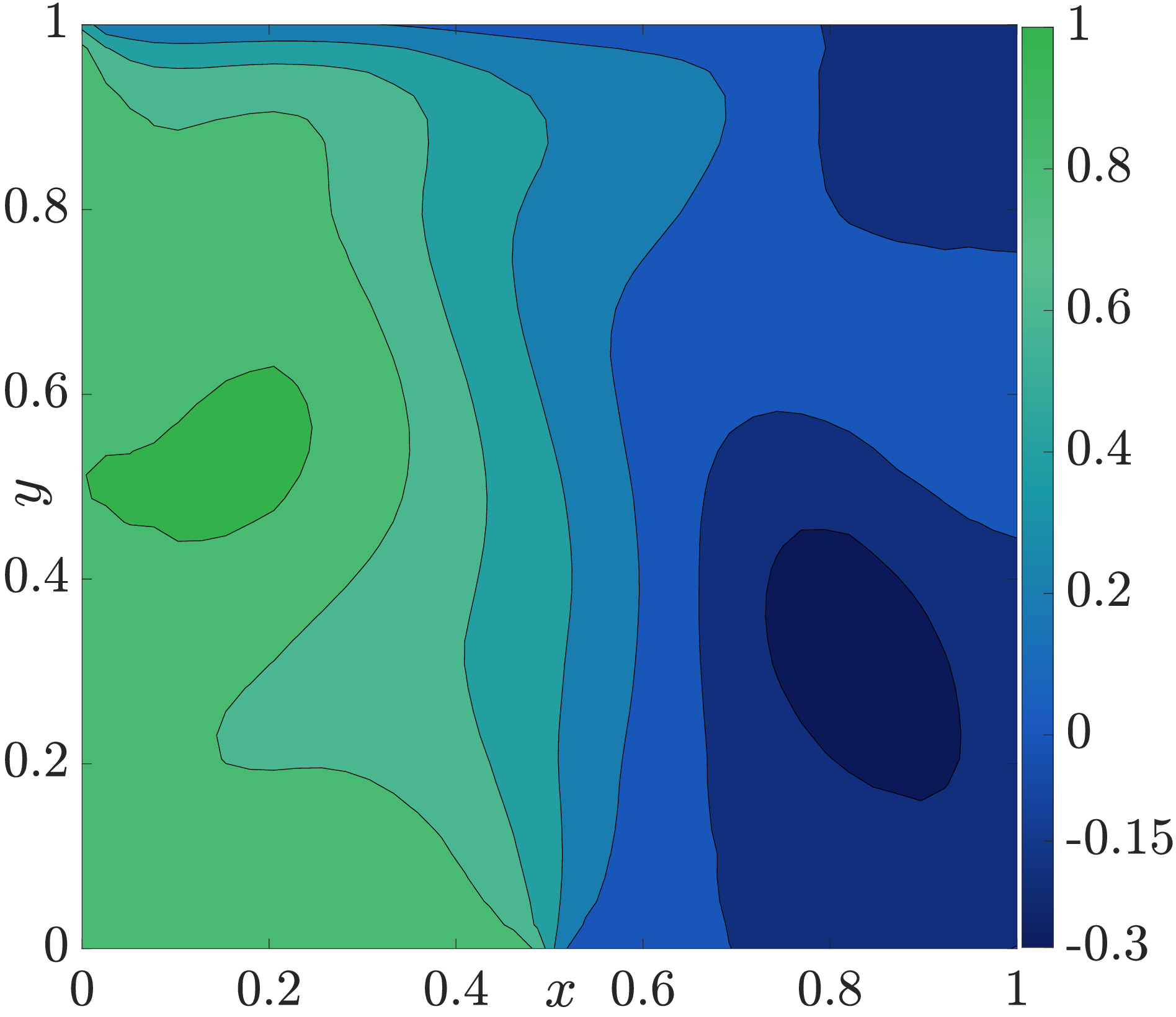}
\end{minipage}%
\begin{minipage}{0.333\textwidth} 
\includegraphics[scale= 0.15]{Figures/2DAdvDiff/RHSNoise/FineTune_Phase2/Redraw_again/ErrorMeanMap_Phase2_Input2.eps}
\end{minipage}%
\caption{\footnotesize [Advection-Diffusion Case 1] Solution prediction: (Left) Reference solution mean, (Middle) SON predicted solution mean, (Right) SON prediction mean error.}
\label{AdvDiff_RHSNoise_SingleMean}
\vspace{-0.1cm}
\end{figure}

\begin{figure}[h!]
\begin{minipage}{0.333\textwidth} 
\includegraphics[scale= 0.15]{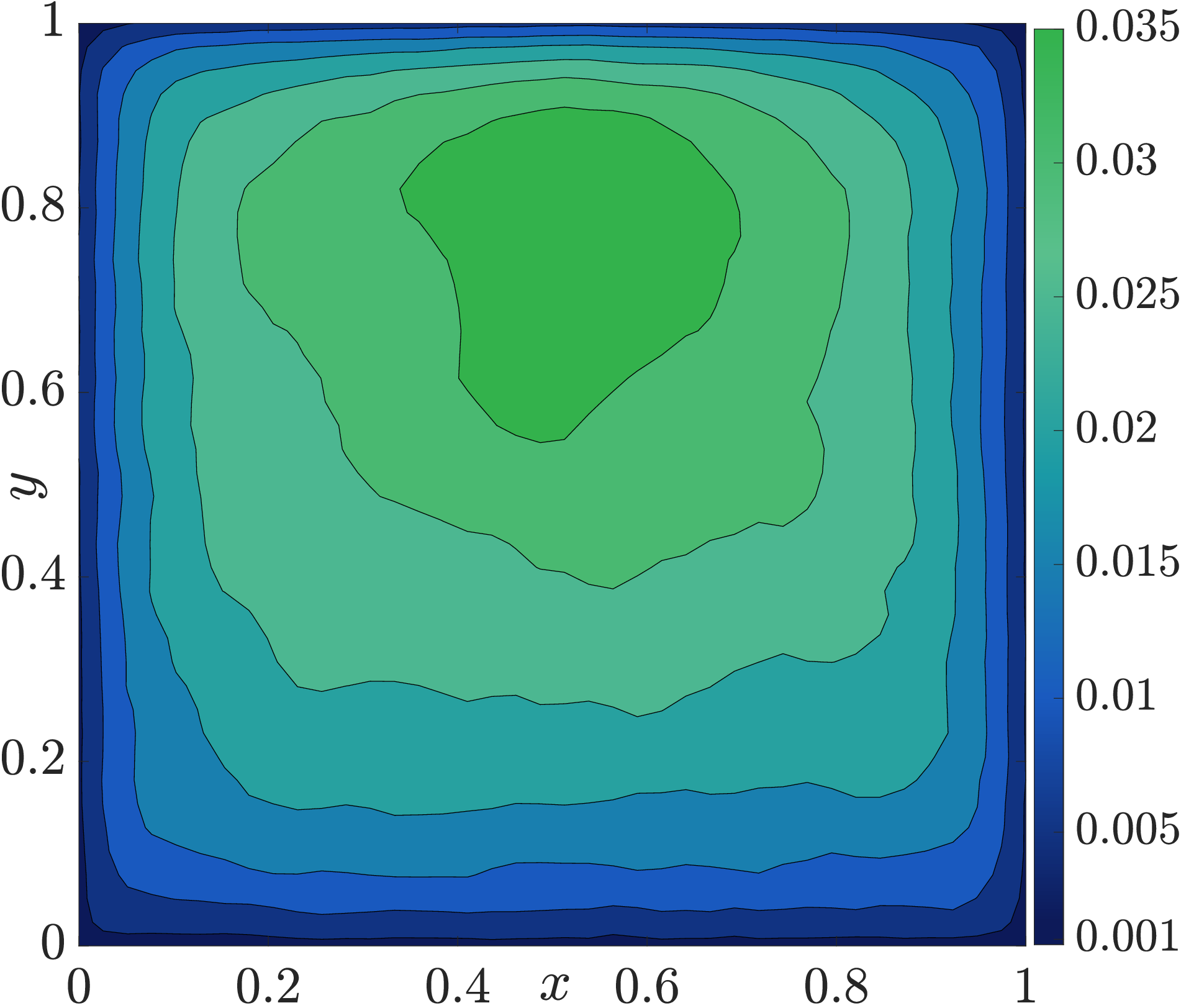}
\end{minipage}%
\begin{minipage}{0.333\textwidth}  
\includegraphics[scale= 0.15]{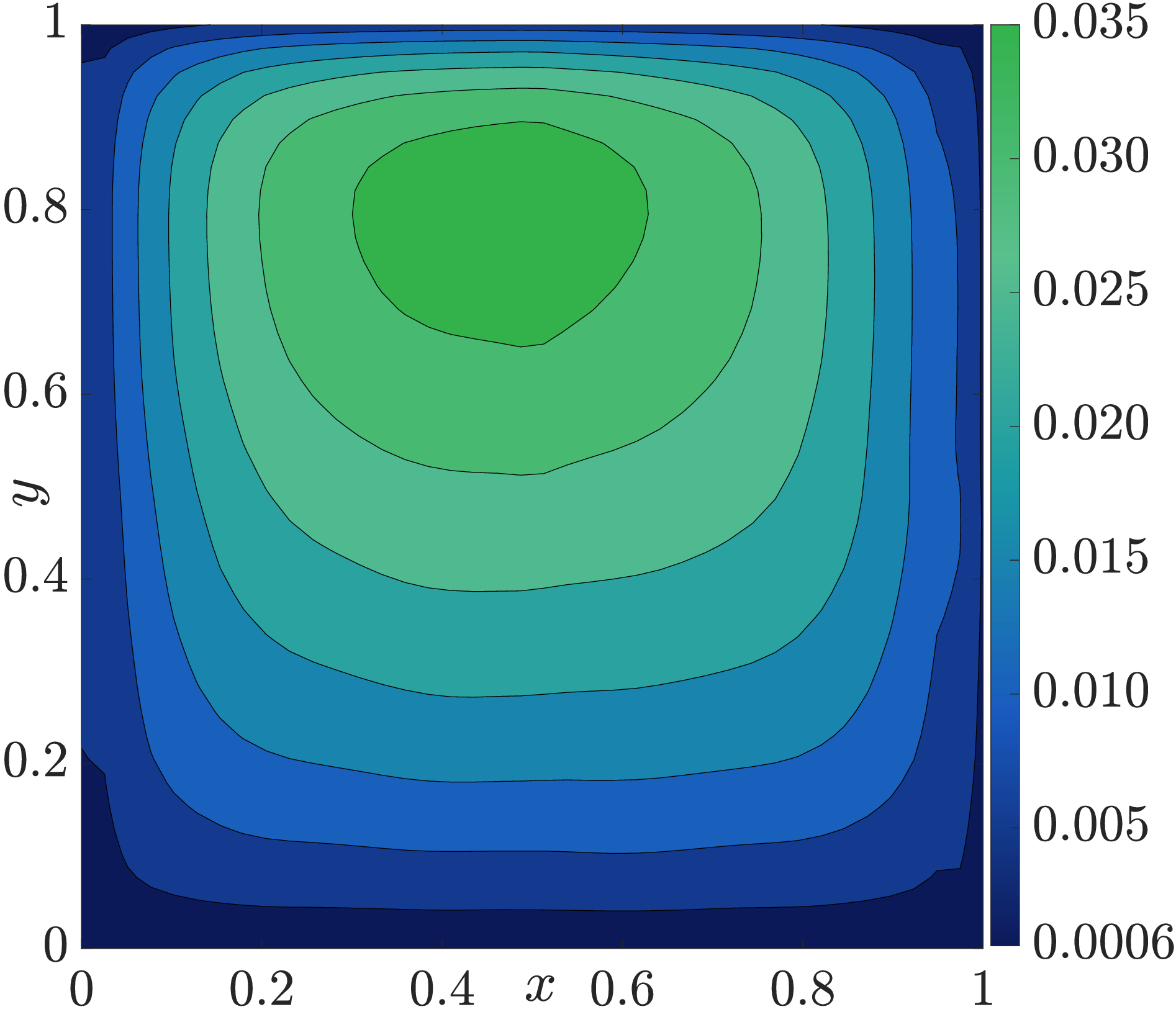}
\end{minipage}%
\begin{minipage}{0.333\textwidth} 
\includegraphics[scale= 0.15]{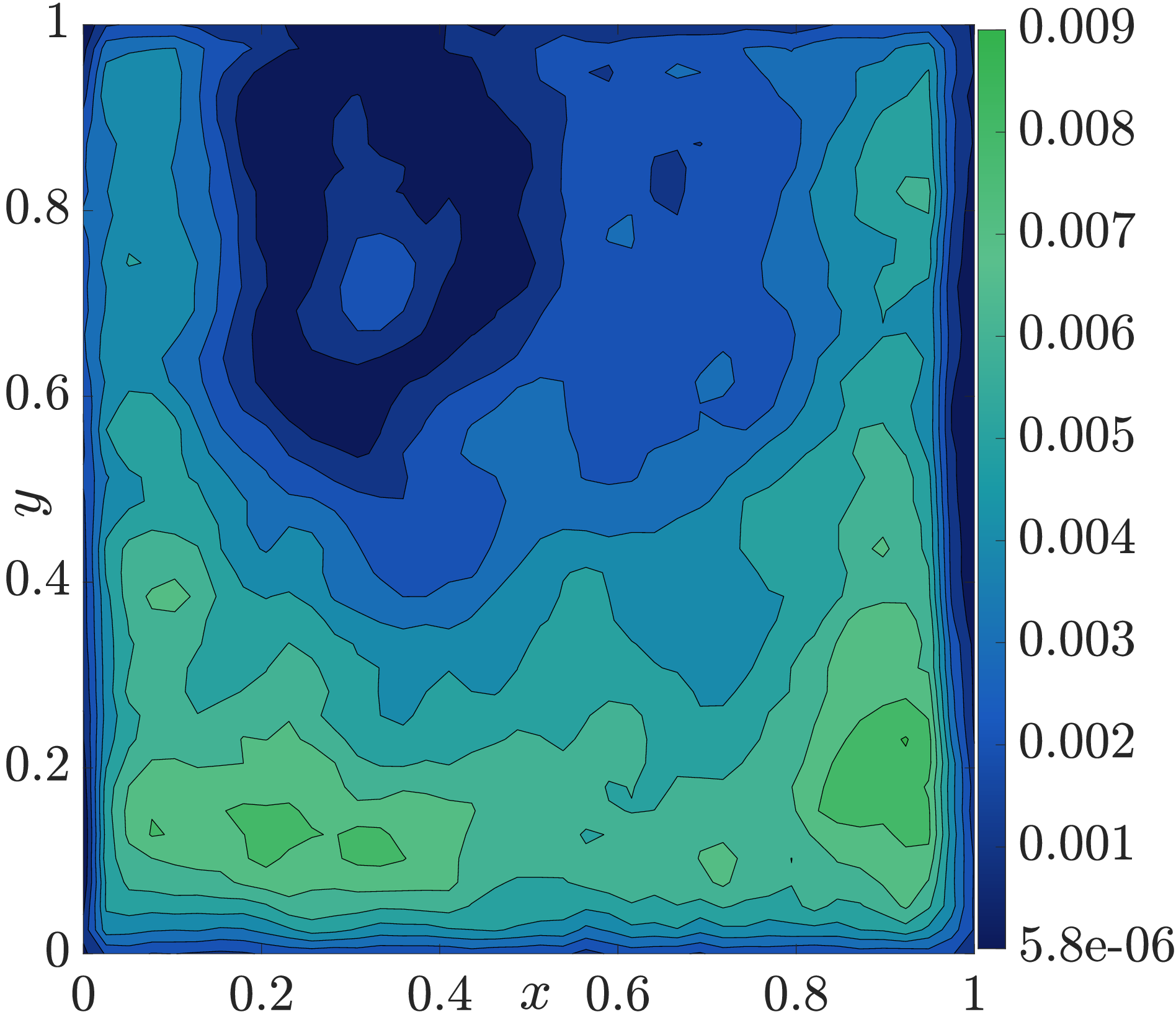}
\end{minipage}%
\caption{\footnotesize [Advection-Diffusion Case 1] Std estimation: (Left) Reference std, (Middle) SON prediction std, (Right) SON std prediction error.}
\label{AdvDiff_RHSNoise_SingleStd}
\vspace{-0.1cm}
\end{figure}

Next, we present a more detailed assessment of the SON’s performance. In Figure~\ref{AdvDiff_RHSNoise_SingleMean}, we present the sample means of the reference solution (left), the SON predicted solution (middle), along with the prediction error. The predicted mean agrees closely with the reference mean, with a maximum absolute error of approximately $0.025$.
In addition, Figure~\ref{AdvDiff_RHSNoise_SingleStd} presents the heatmaps of the standard deviations (std) of the solution samples for both the reference solution and the SON predictions, along with the corresponding error. The maximum value of the sample std error is around $0.009$ and is concentrated near the bottom and the lateral boundaries. Most of the remaining values are smaller than $0.007$. This indicates that SON accurately captures the uncertainty of the SPDE solution.

\begin{figure}[h!]
\begin{minipage}{0.333\textwidth}
\includegraphics[scale = 0.145]{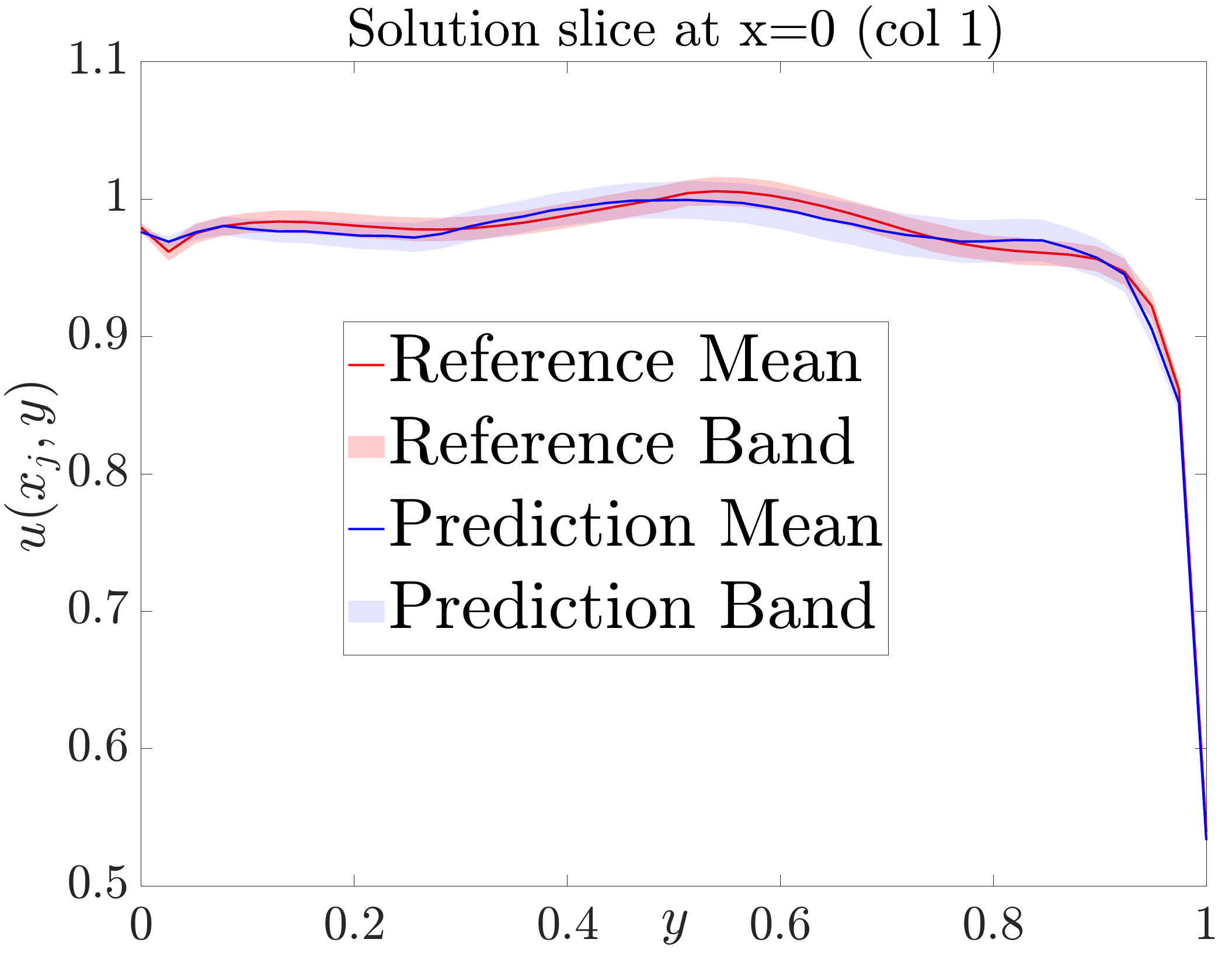}
\end{minipage}%
\begin{minipage}{0.333\textwidth}
\includegraphics[scale = 0.145]{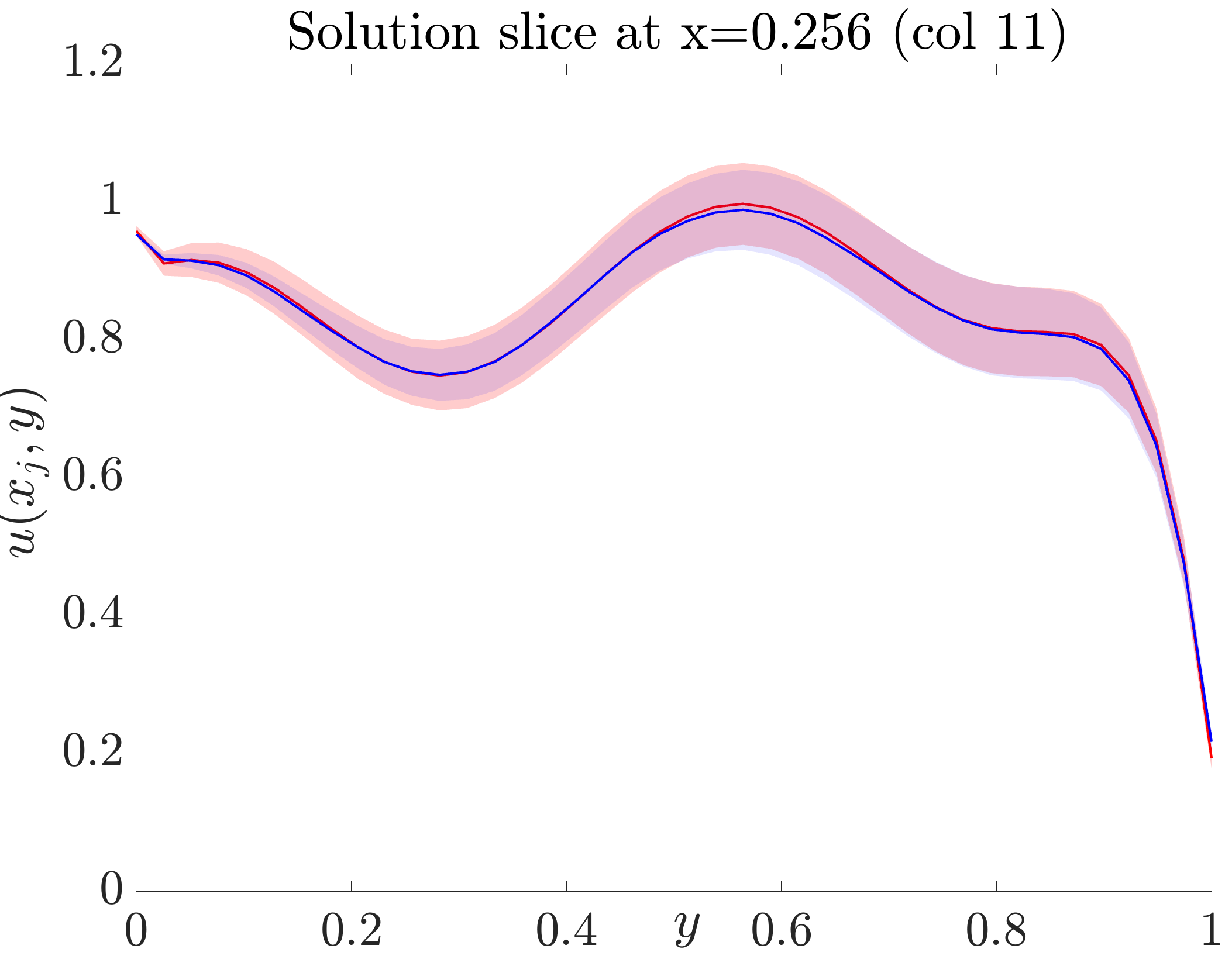}
\end{minipage}%
\begin{minipage}{0.333\textwidth}
\includegraphics[scale = 0.145]{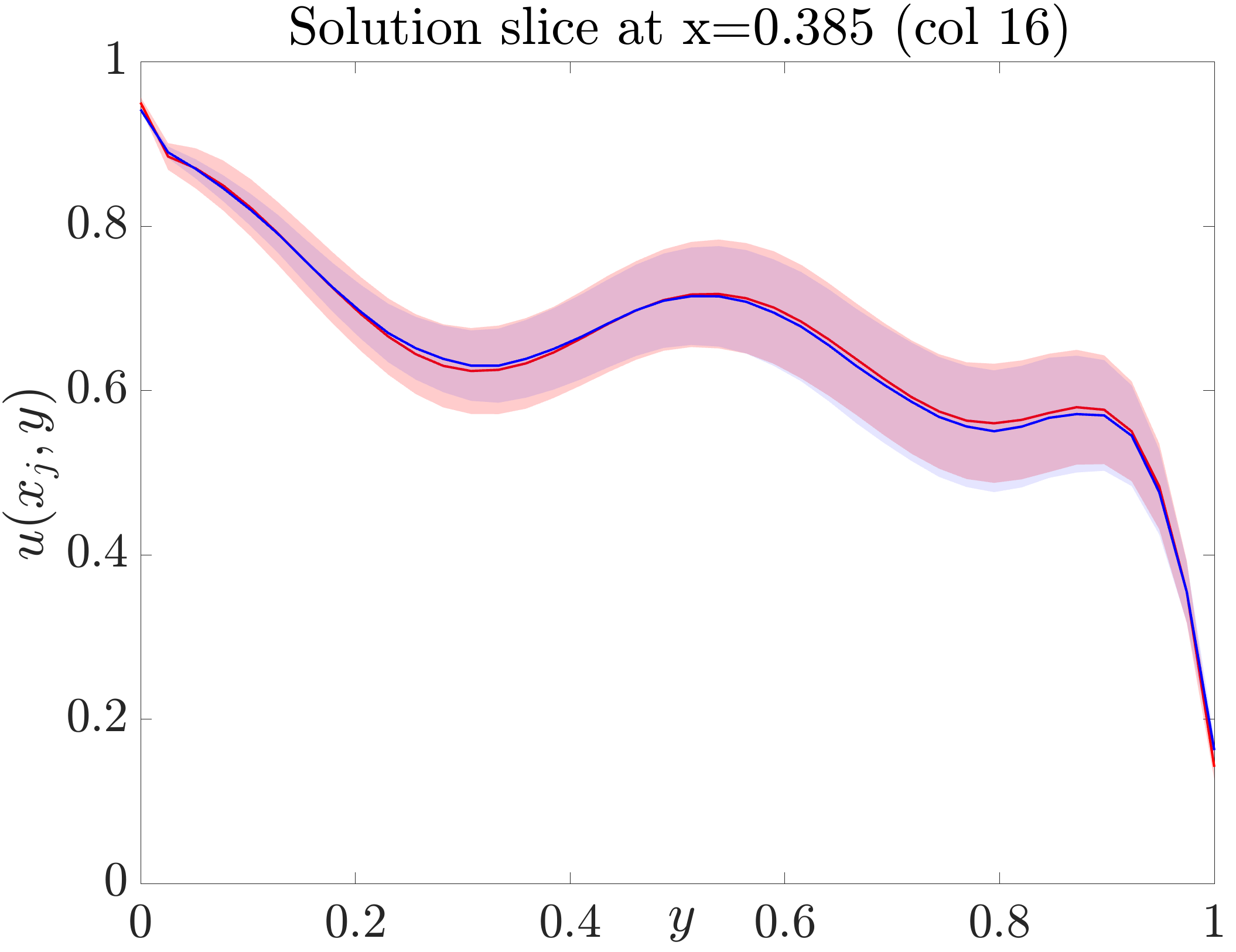}
\end{minipage}

\begin{minipage}{0.333\textwidth}
\includegraphics[scale = 0.145]{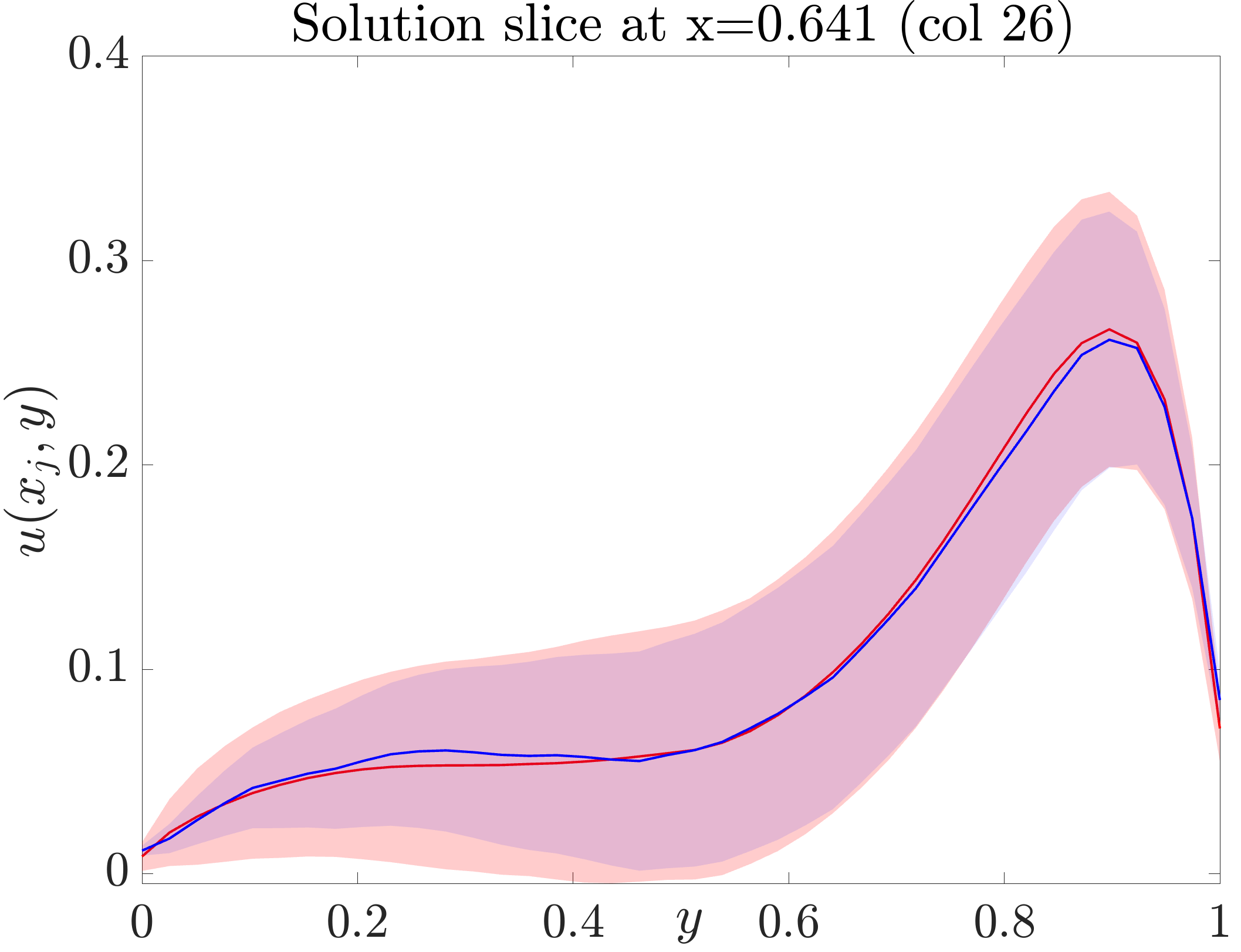}
\end{minipage}%
\begin{minipage}{0.333\textwidth}
\includegraphics[scale = 0.145]{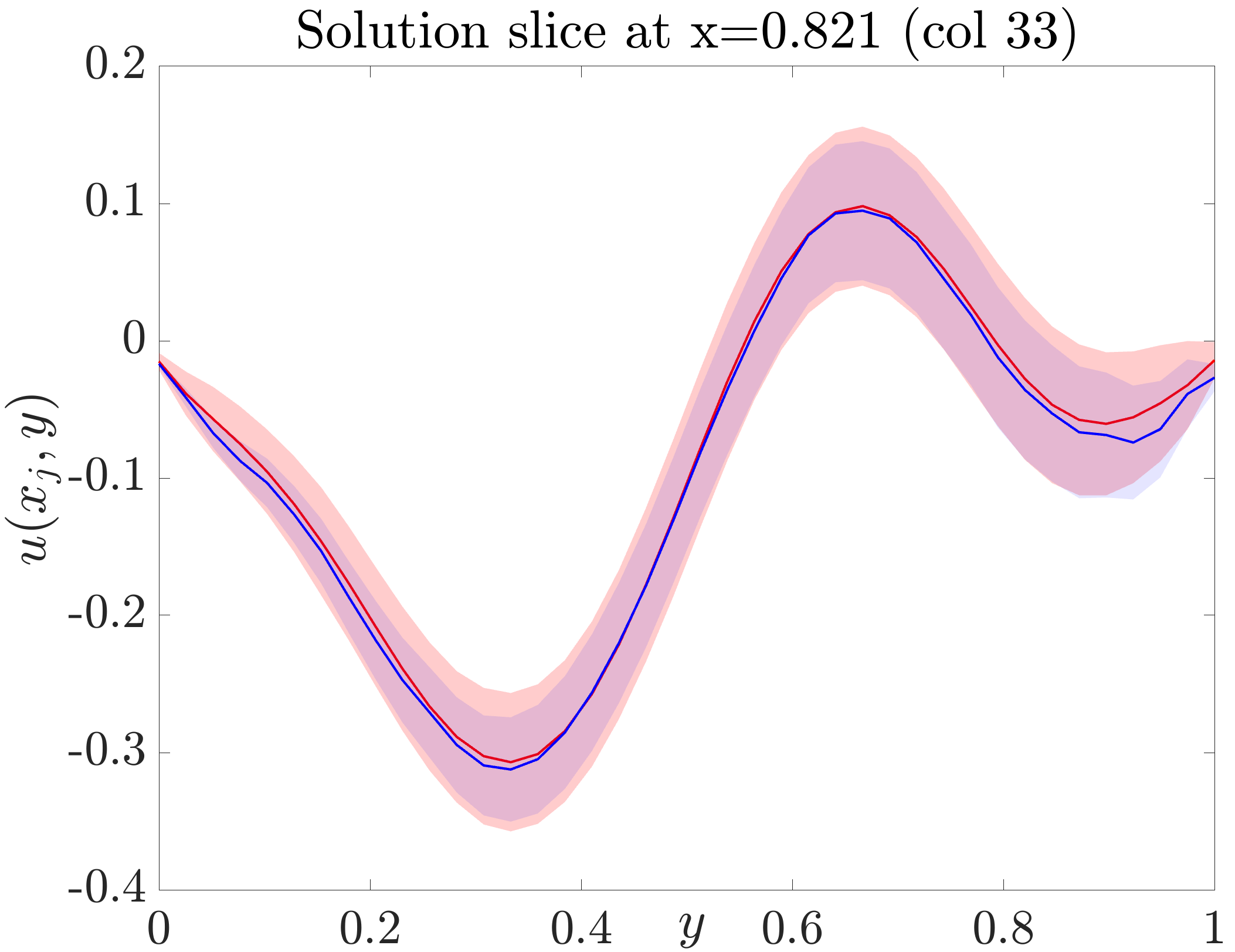}
\end{minipage}%
\begin{minipage}{0.333\textwidth}
\includegraphics[scale = 0.145]{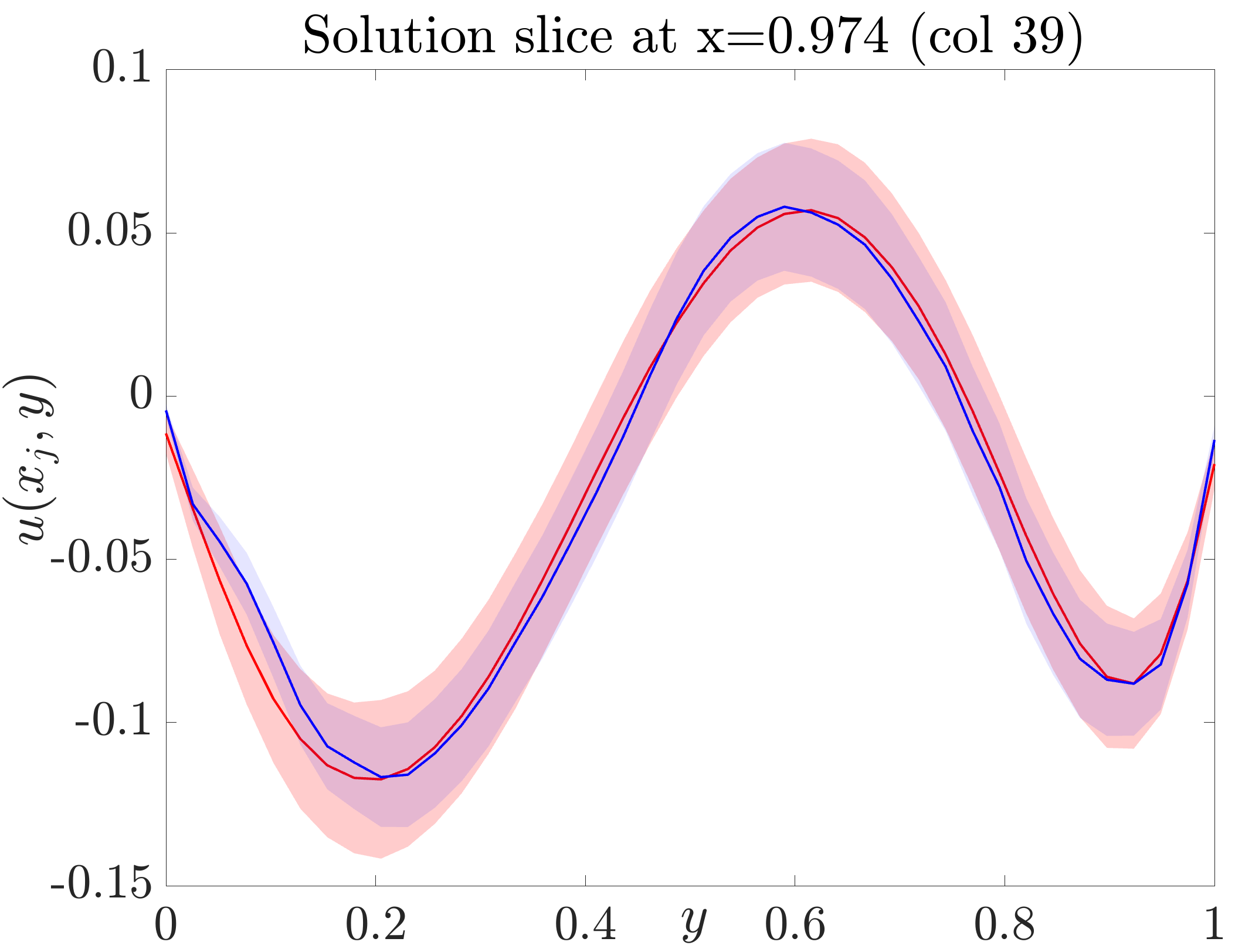}
\end{minipage}
\caption{\footnotesize [Advection-Diffusion Case 1] Cross-sections at columns 1, 11, 16, 26, 33, and 39 of the reference solution and approximation.}
\label{AdvDiff_RHSNoise_CrossSection}
\end{figure}

To further support this observation, we plot multiple cross-sections of the reference and predicted means and display their corresponding confidence bands in Figure~\ref{AdvDiff_RHSNoise_CrossSection}. The figure demonstrates a close agreement between both the means and the confidence bands of the reference and predicted solutions.

\subsubsection{Case 2: Space-dependent uncertainty in the advection field}
In Case 2, we introduce a more complex model uncertainty by incorporating space-dependent noise into the velocity field: $\pmb{b}_{\xi}(x, y) = \pmb{b}(x, y) + \xi(x, y)$.
% \begin{align*}
% \pmb{b}_{\xi}(x, y) = \pmb{b}(x, y) + \xi(x, y).
% \end{align*}
This testing case is more challenging than Case 1 because the noise is embedded in the model solver, i.e., in the stiffness matrix of the numerical solver. As a result, the induced uncertainty depends on the forcing term, and the solution components become correlated. We adopt the same numerical method as in Case 1 to discretize the system~\eqref{2D_AdvDiff} with the noisy velocity field $\pmb{b}_{\xi}$. The noise field $\xi(\cdot,\cdot)$ is formulated in the same form as Case 1, but now the scaling factors $\{\alpha_k\}_{k=1}^{M^2}$ are sampled uniformly from $[0.05, 0.2]$. The training and testing datasets are generated in the same manner as in Case 1, with the same number of samples for both datasets.

We adopt the same network architectures as in Case 1. However, because the uncertainty depends on the forcing term and has correlated entries, we choose the size of the Gaussian random variable to be $r = 16$ and reformulate the diffusion term as a layer-independent tensor of size $M\times M \times r$ where each $kth$ entry for $k=1, \hdots, r$ takes the form
\begin{equation}
\label{diffusion_phase2}
\left(\sigma\!\left(A,\theta_{n, g}\right)\right)_{k} \equiv \left(\sigma\!\left(A,\theta_{g}\right)\right)_{k} = \sum_{l=1}^{L_g} b_{l, k} \, \mu(c_{l, k}A),
\qquad n=0,\ldots,N-1,
\end{equation}
where $L_g = 4$, and for all $k=1, \hdots, r$, $\{b_{l, k}\}_{l=1}^{L_g}$ and $\{c_{l, k}\}_{l=1}^{L_g}$ are sampled uniformly from the intervals $[0.035, 0.1]$ and $[0.2, 0.5]$, respectively.

\begin{figure}[h!]
\hspace{1cm}
\begin{minipage}{0.45\textwidth} 
\includegraphics[scale= 0.18]{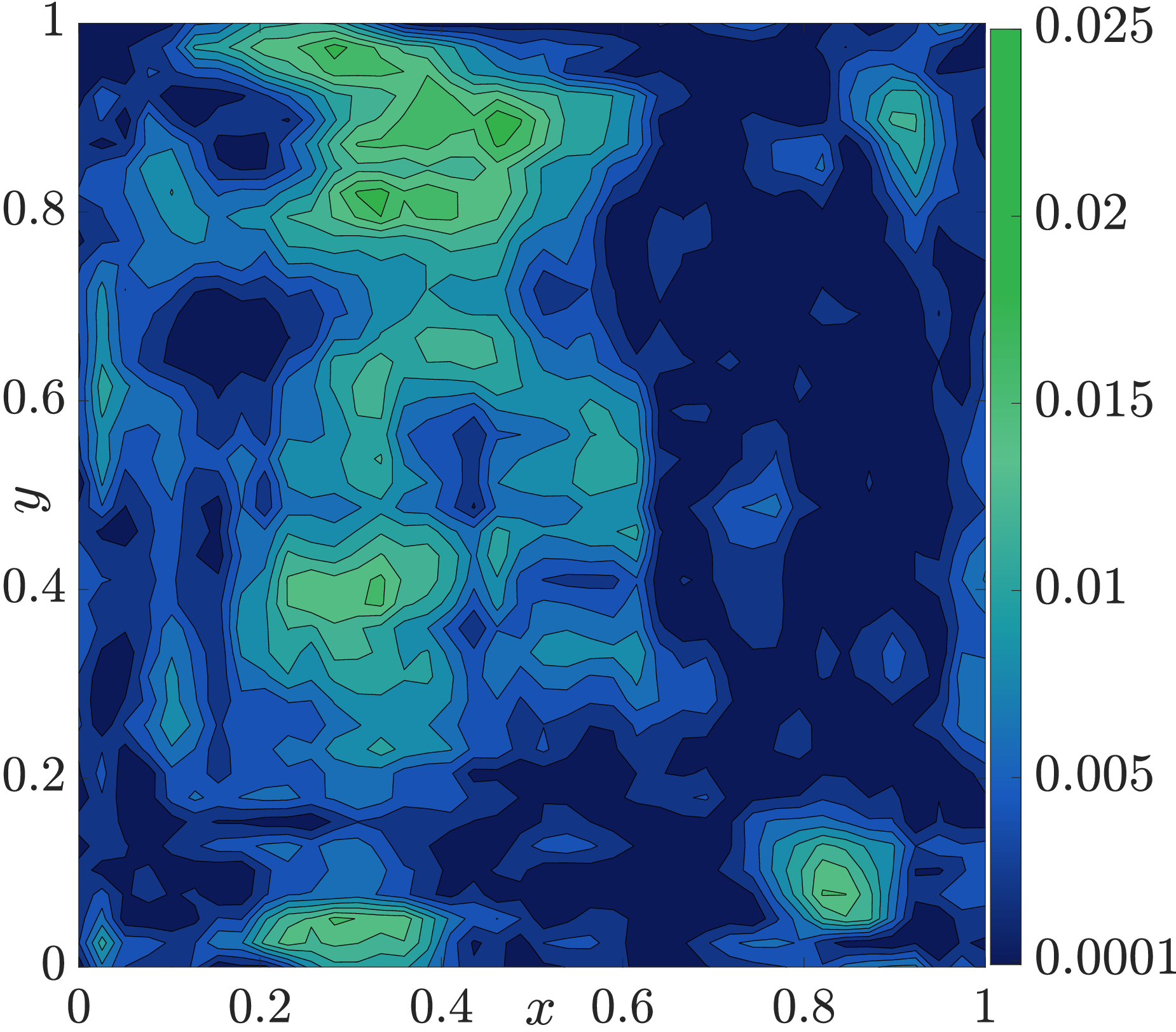}
\end{minipage}%
% \begin{minipage}{0.333\textwidth}
% \includegraphics[scale= 0.165]{Figures/2DAdvDiff/VeloNoise/Update/Redraw/Prediction2/ErrorSolMap_Phase1_Input2_Redraw.eps}
% \end{minipage}%
% \begin{minipage}{0.333\textwidth} 
% \includegraphics[scale= 0.165]{Figures/2DAdvDiff/VeloNoise/Update/Redraw/Prediction3/ErrorSolMap_Phase1_Input3_Redraw.eps}
% \end{minipage}
\begin{minipage}{0.45\textwidth} 
\includegraphics[scale= 0.18]{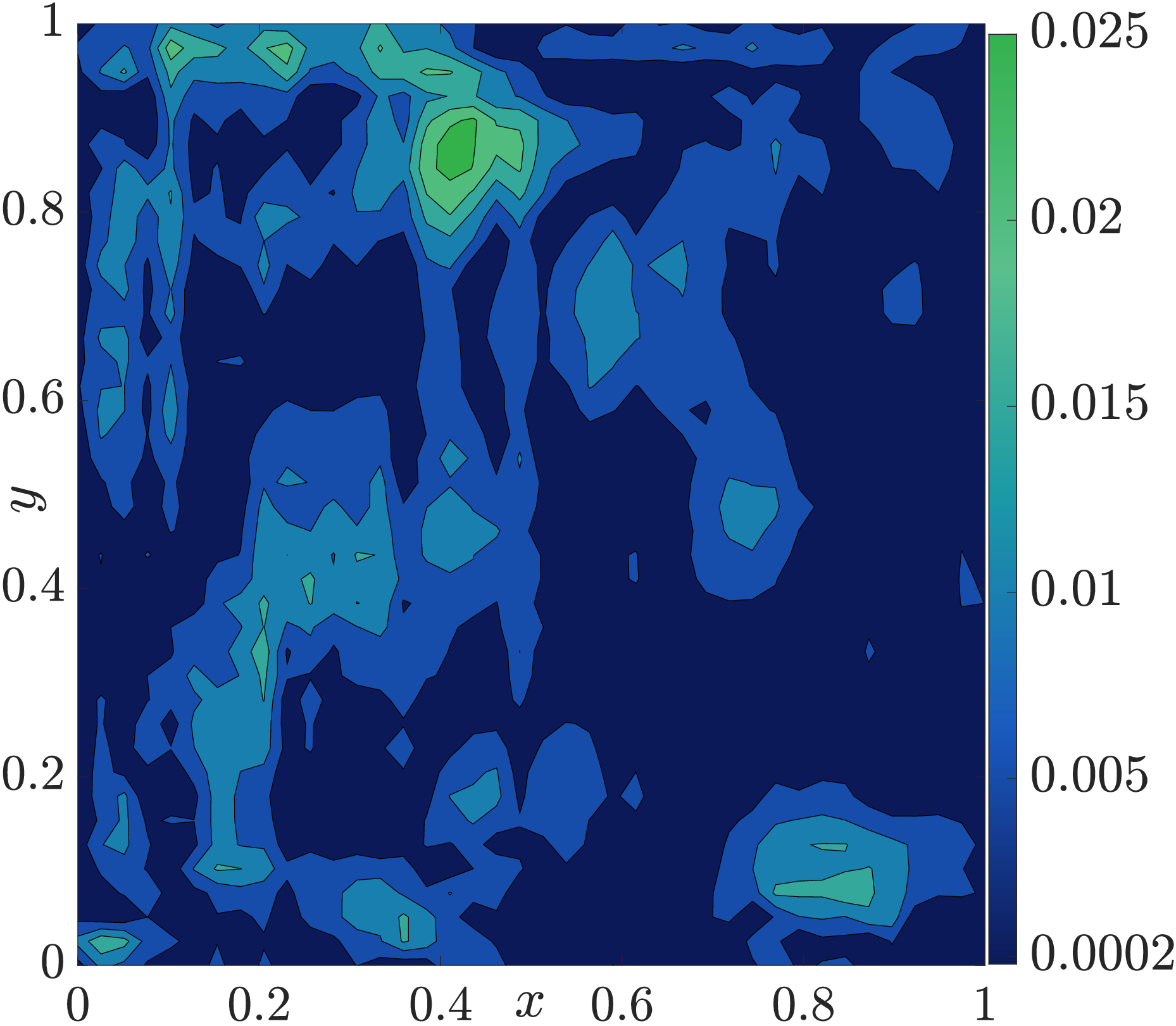}
\end{minipage}%
% \begin{minipage}{0.333\textwidth} 
% \includegraphics[scale= 0.165]{Figures/2DAdvDiff/VeloNoise/Update/Redraw/Prediction2/ErrorSolMap_Phase2_Input2_Redraw.eps}
% \end{minipage}%
% \begin{minipage}{0.333\textwidth}
% \includegraphics[scale= 0.165]{Figures/2DAdvDiff/VeloNoise/Update/Redraw/Prediction3/ErrorSolMap_Phase2_Input3_Redraw.eps}
% \end{minipage}
\caption{\footnotesize [Advection-Diffusion Case 2] Heatmaps of mean errors: (Left) DeepONet; (Right) SON.}
\label{AdvDiff_Case2_Compare_ErrorMap}
\end{figure}
\begin{figure}[h!]
\vspace{-0.1cm}
\begin{minipage}{0.333\textwidth} 
\includegraphics[scale= 0.16]{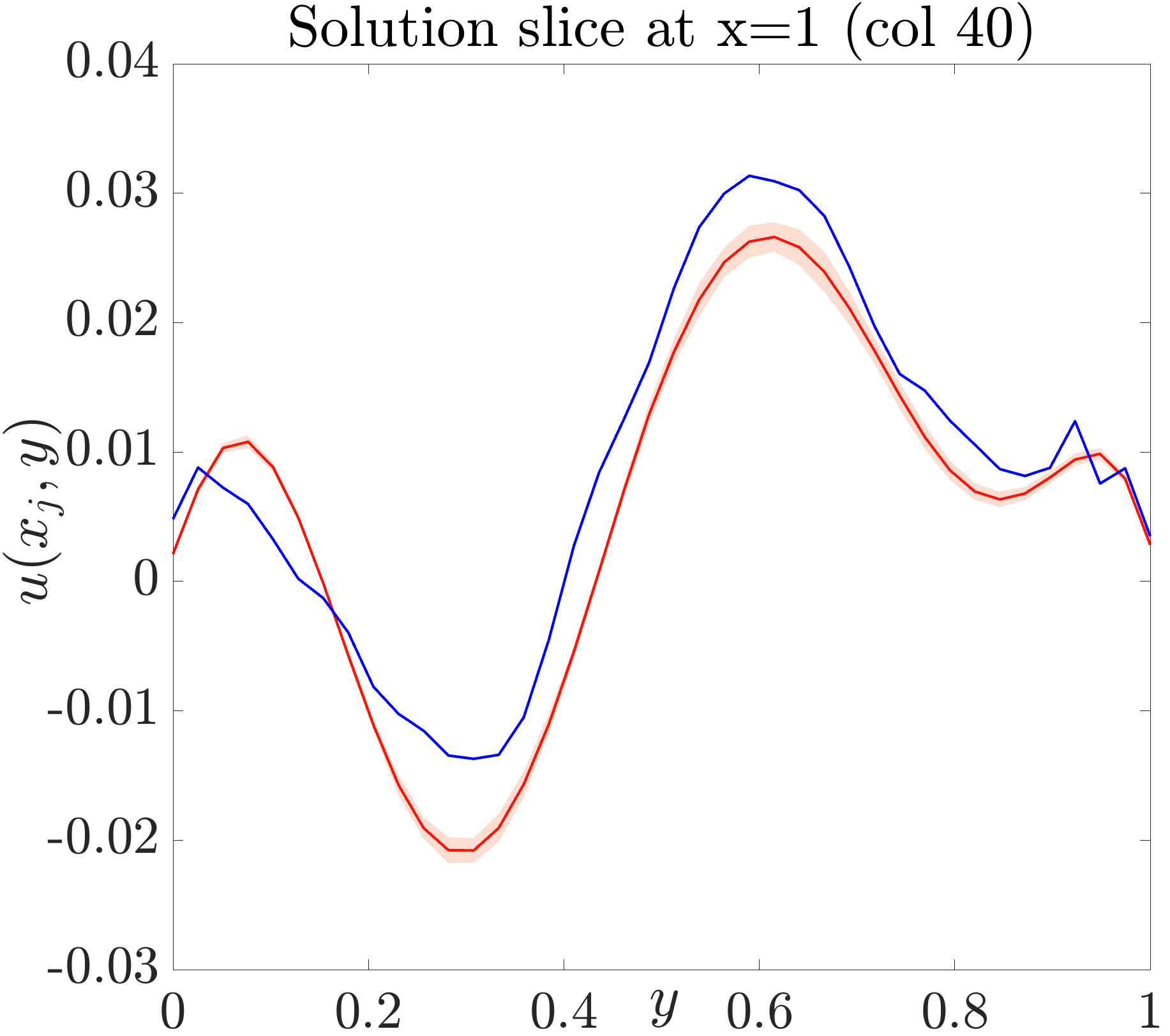}
\end{minipage}%
\begin{minipage}{0.333\textwidth}
\includegraphics[scale= 0.16]{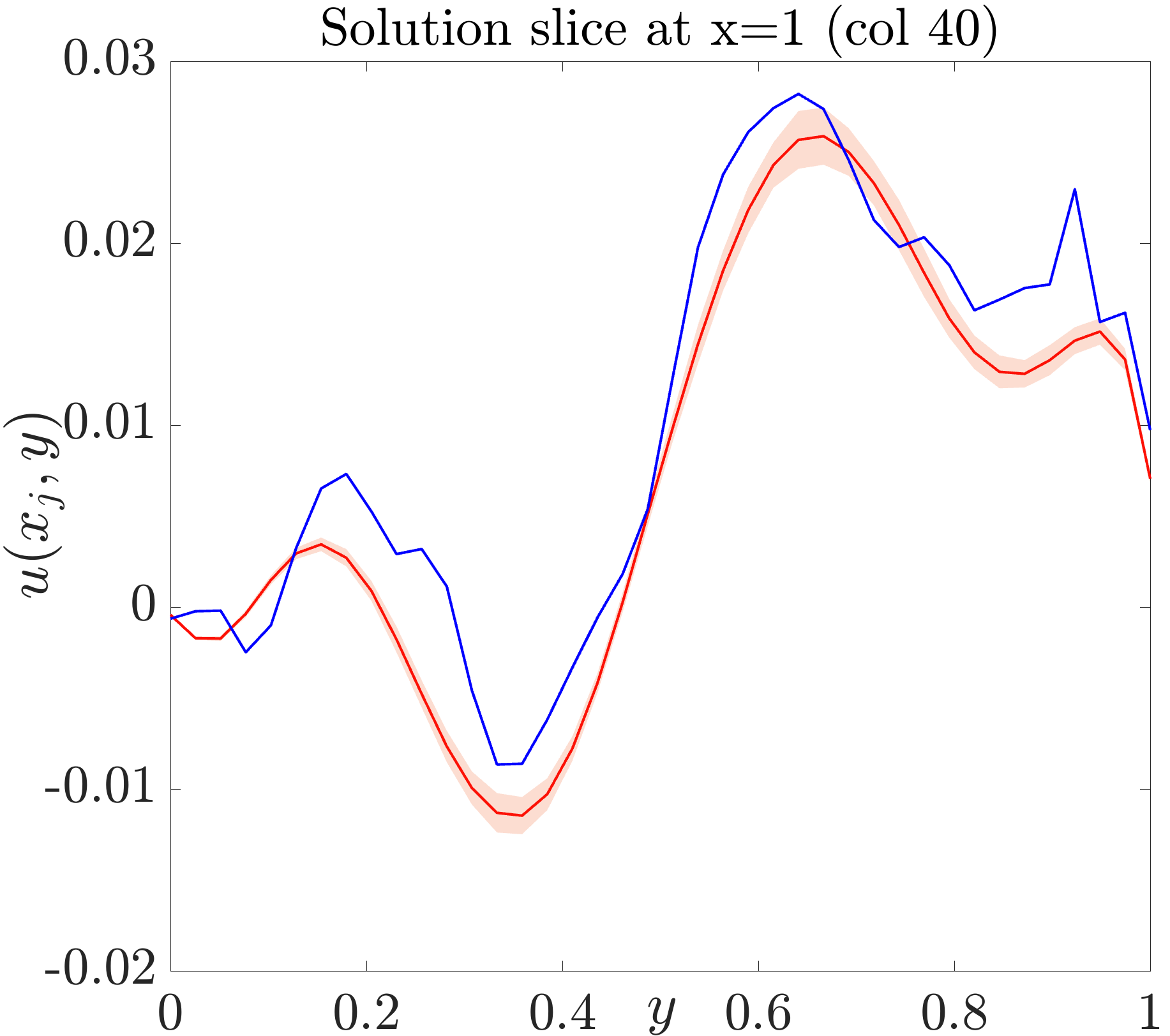}
\end{minipage}%
\begin{minipage}{0.333\textwidth} 
\includegraphics[scale= 0.16]{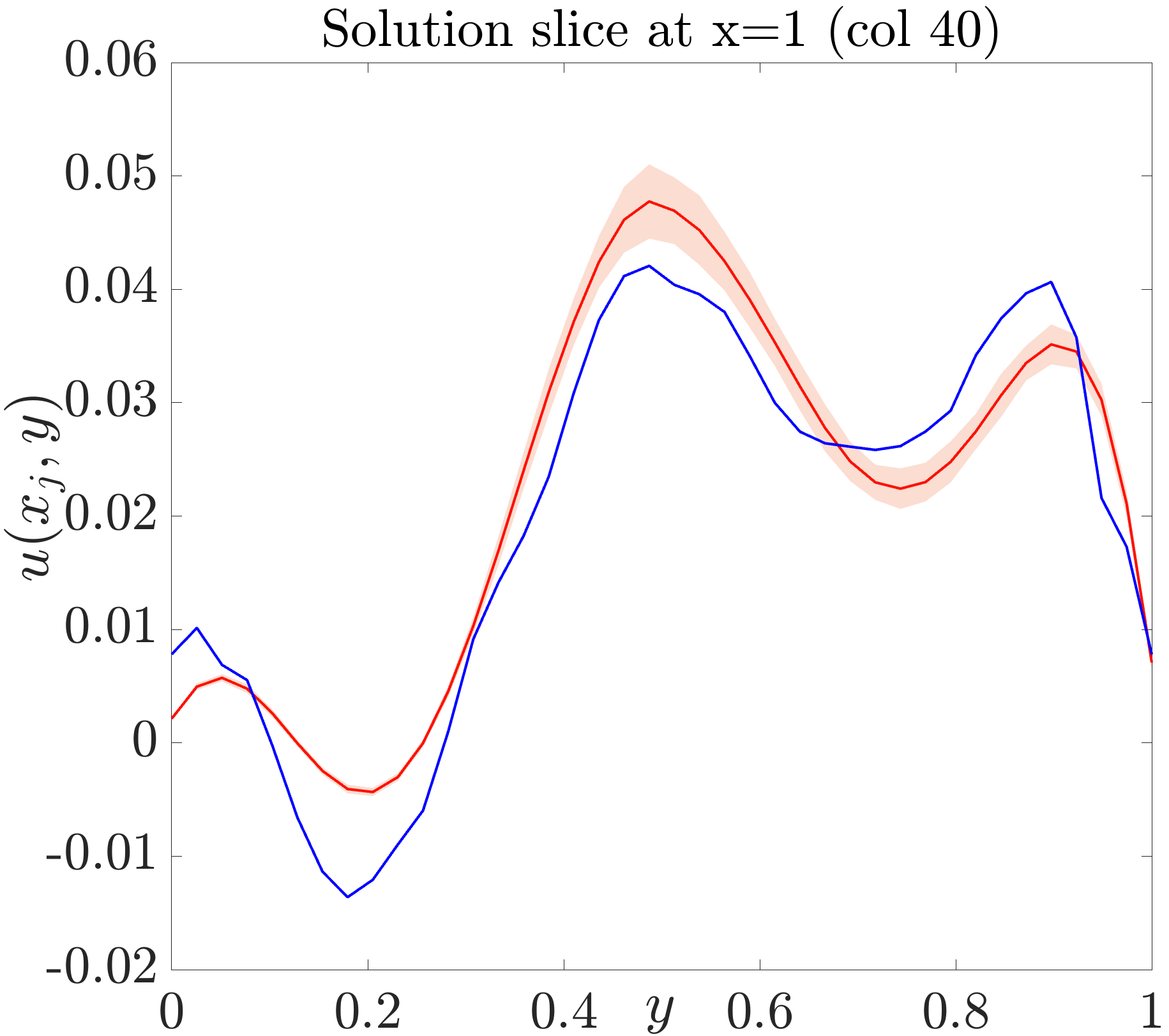}
\end{minipage}

\begin{minipage}{0.333\textwidth}  
\includegraphics[scale= 0.16]{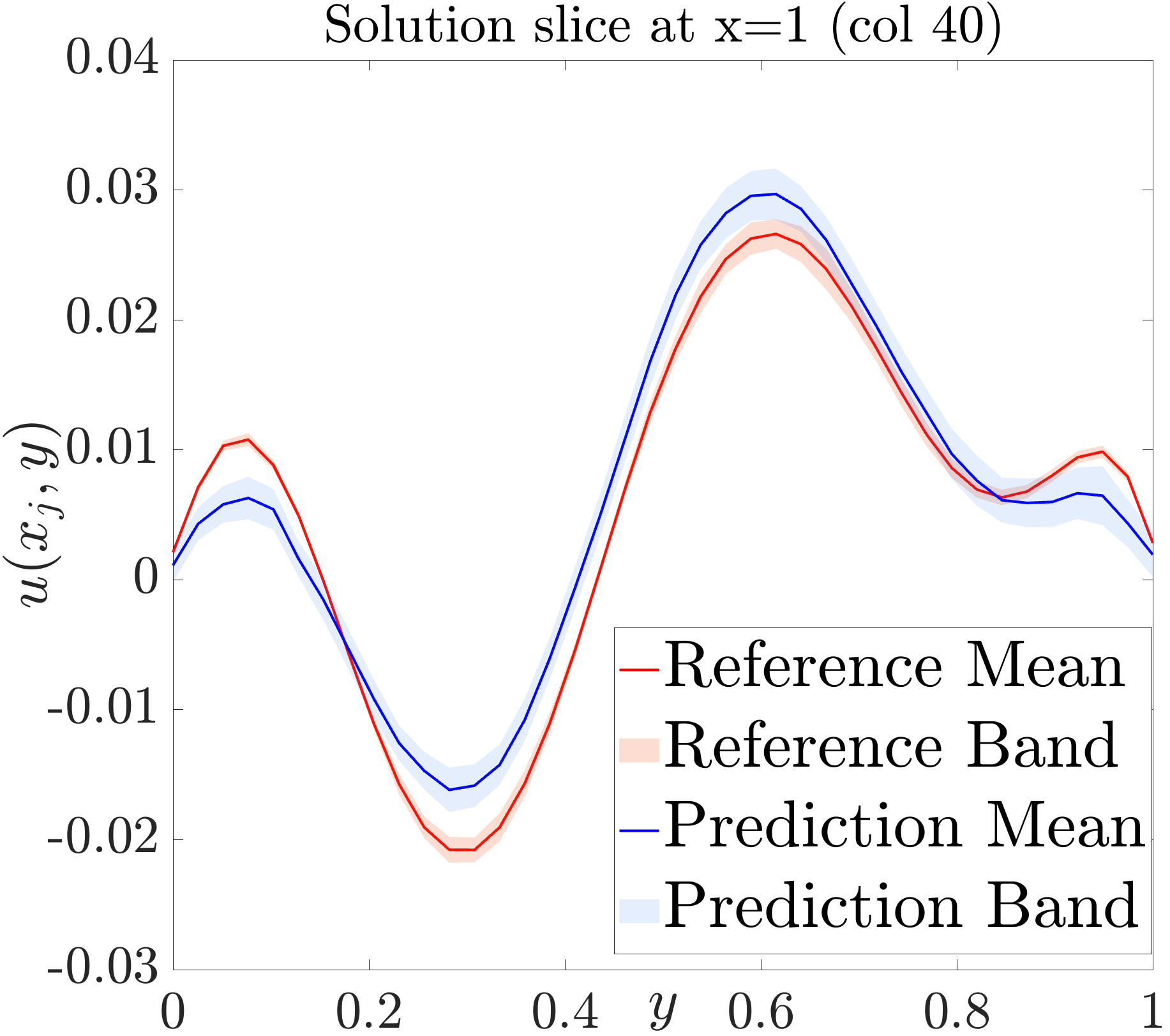}
\end{minipage}%
\begin{minipage}{0.333\textwidth} 
\includegraphics[scale= 0.16]{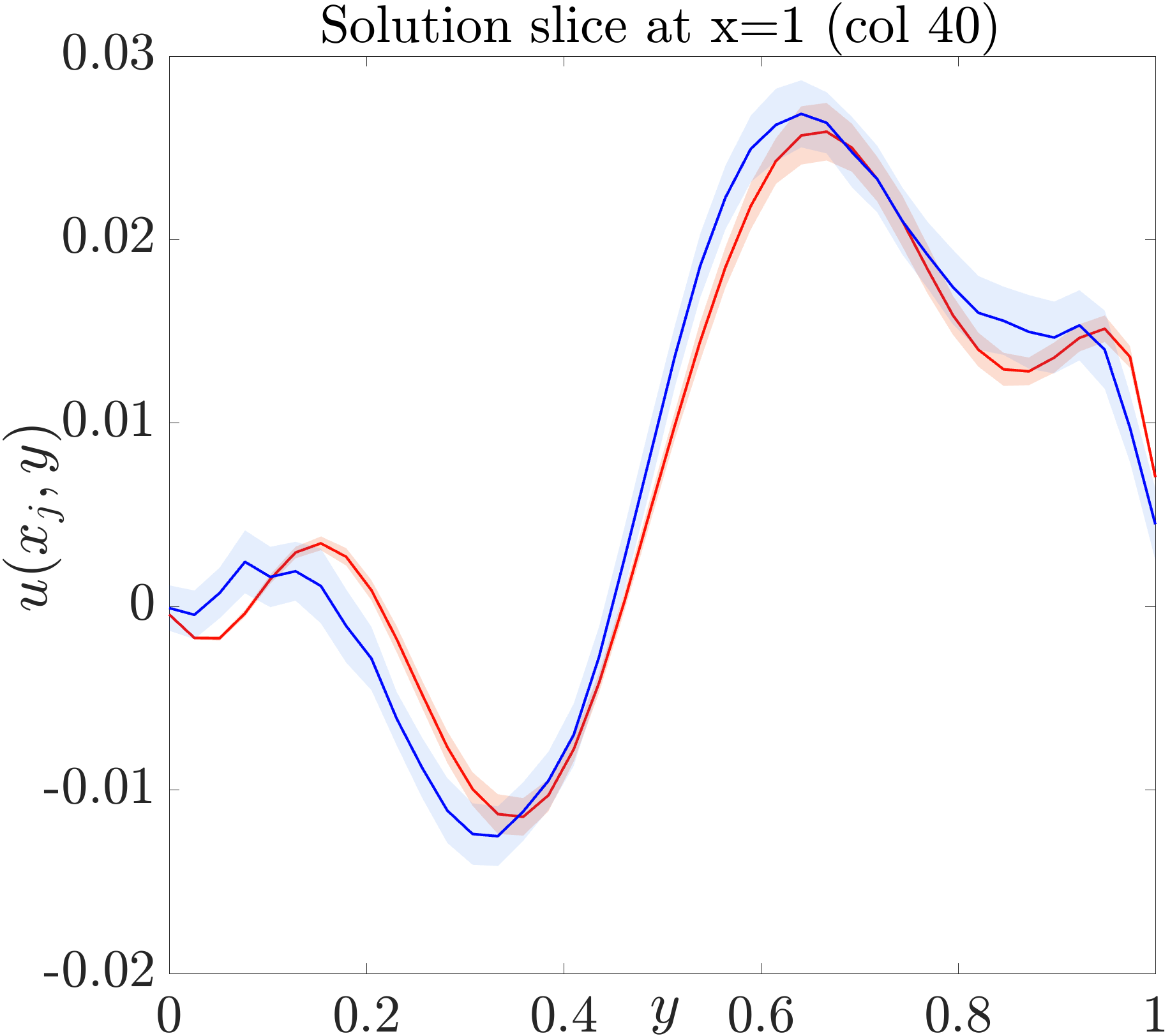}
\end{minipage}%
\begin{minipage}{0.333\textwidth}
\includegraphics[scale= 0.16]{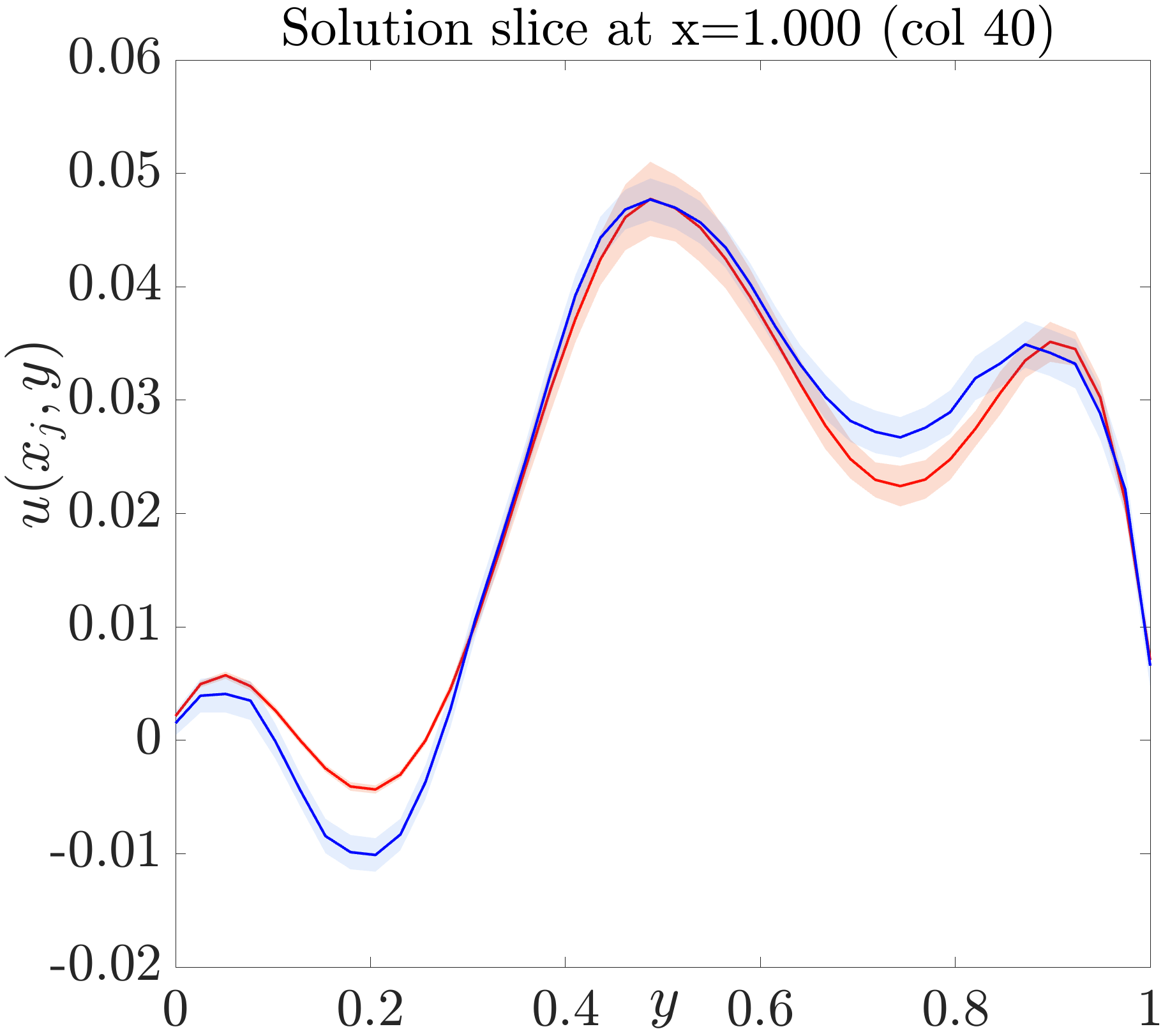}
\end{minipage}
\caption{\footnotesize [Advection-Diffusion Case 2] Cross-sections of the predicted solutions with quantitative uncertainty with three different testing inputs: (First row) DeepONet; (Second row) SON. SON achieves higher predictive accuracy than the deterministic DeepONet, and it provides reliable uncertainty quantification.}
\label{AdvDiff_Case2_Compare_CrossSections}
\end{figure}
\begin{figure}[h!]
% \begin{minipage}{0.333\textwidth}  
% \includegraphics[scale= 0.145]{Figures/2DAdvDiff/VeloNoise/Update/Prediction1/RefStd_Input1.eps}
% \end{minipage}%
% \begin{minipage}{0.333\textwidth}  
% \includegraphics[scale= 0.145]{Figures/2DAdvDiff/VeloNoise/Update/Prediction1/PredStd_Input1_v3.eps}
% \end{minipage}%
% \begin{minipage}{0.333\textwidth} 
% \includegraphics[scale= 0.145]{Figures/2DAdvDiff/VeloNoise/Update/Prediction1/ErrorStdMap_Input1_v4.eps}
% \end{minipage}
\begin{minipage}{0.333\textwidth}  
\includegraphics[scale= 0.145]{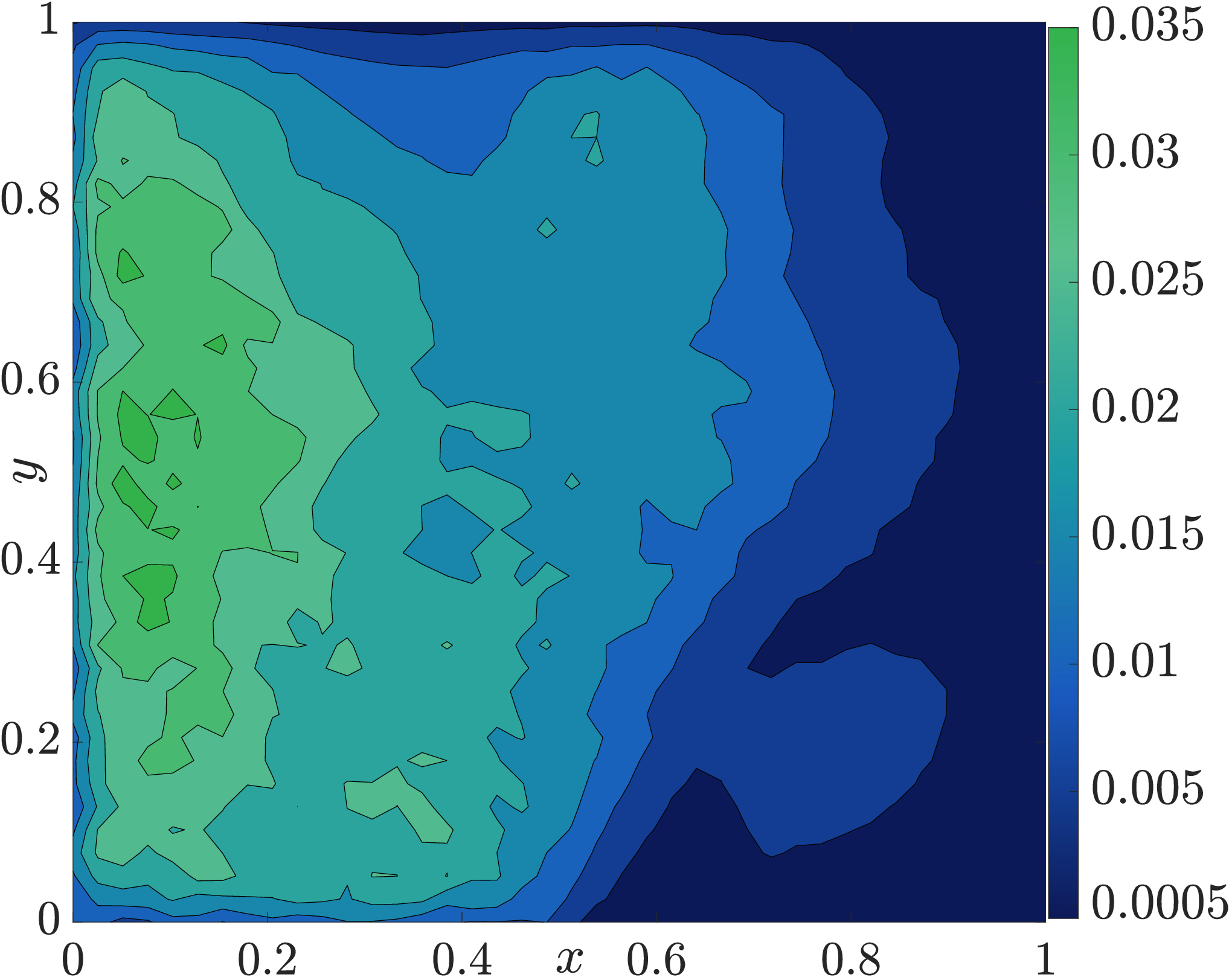}
\end{minipage}%
\begin{minipage}{0.333\textwidth}  
\includegraphics[scale= 0.145]{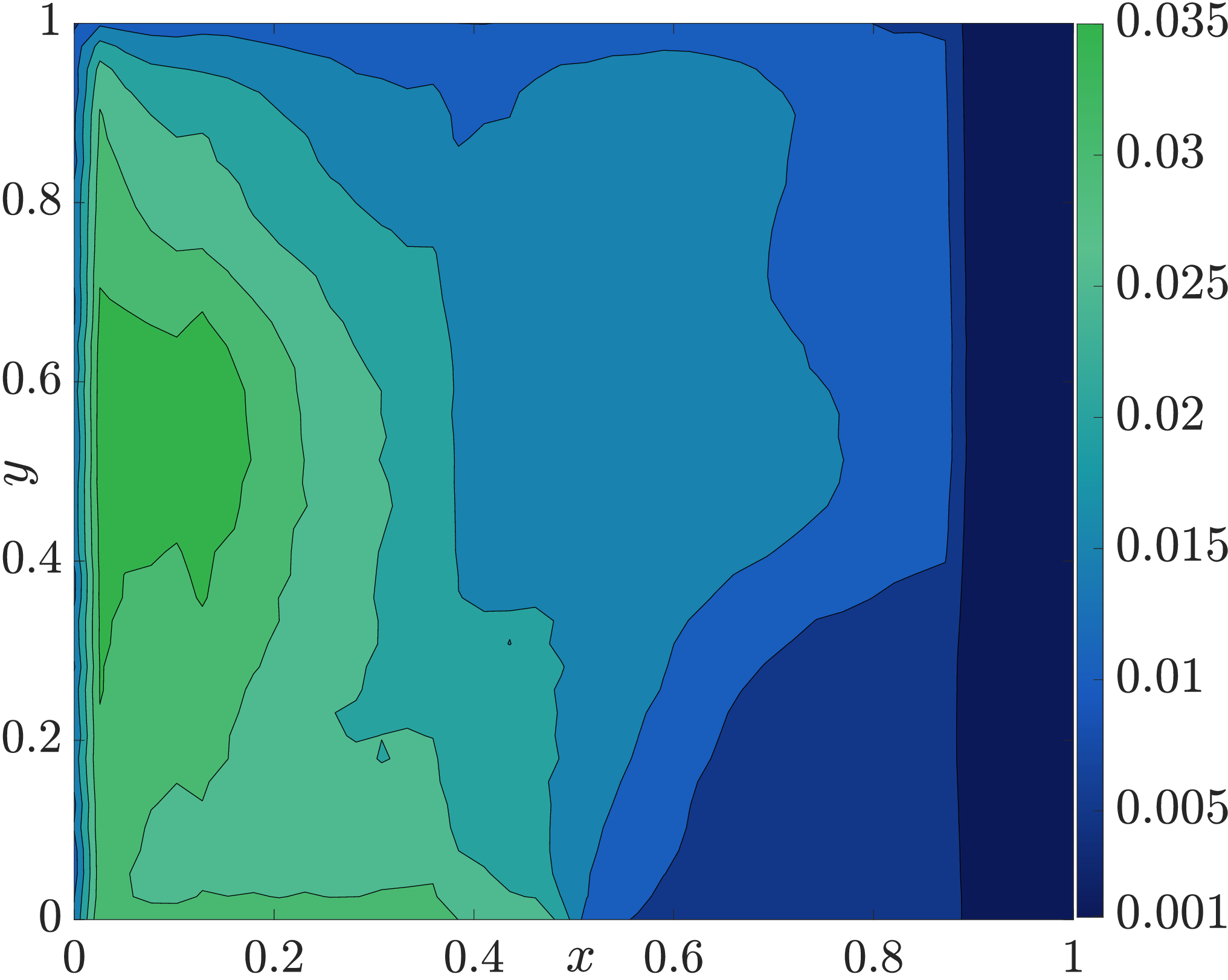}
\end{minipage}%
\begin{minipage}{0.333\textwidth} 
\includegraphics[scale= 0.145]{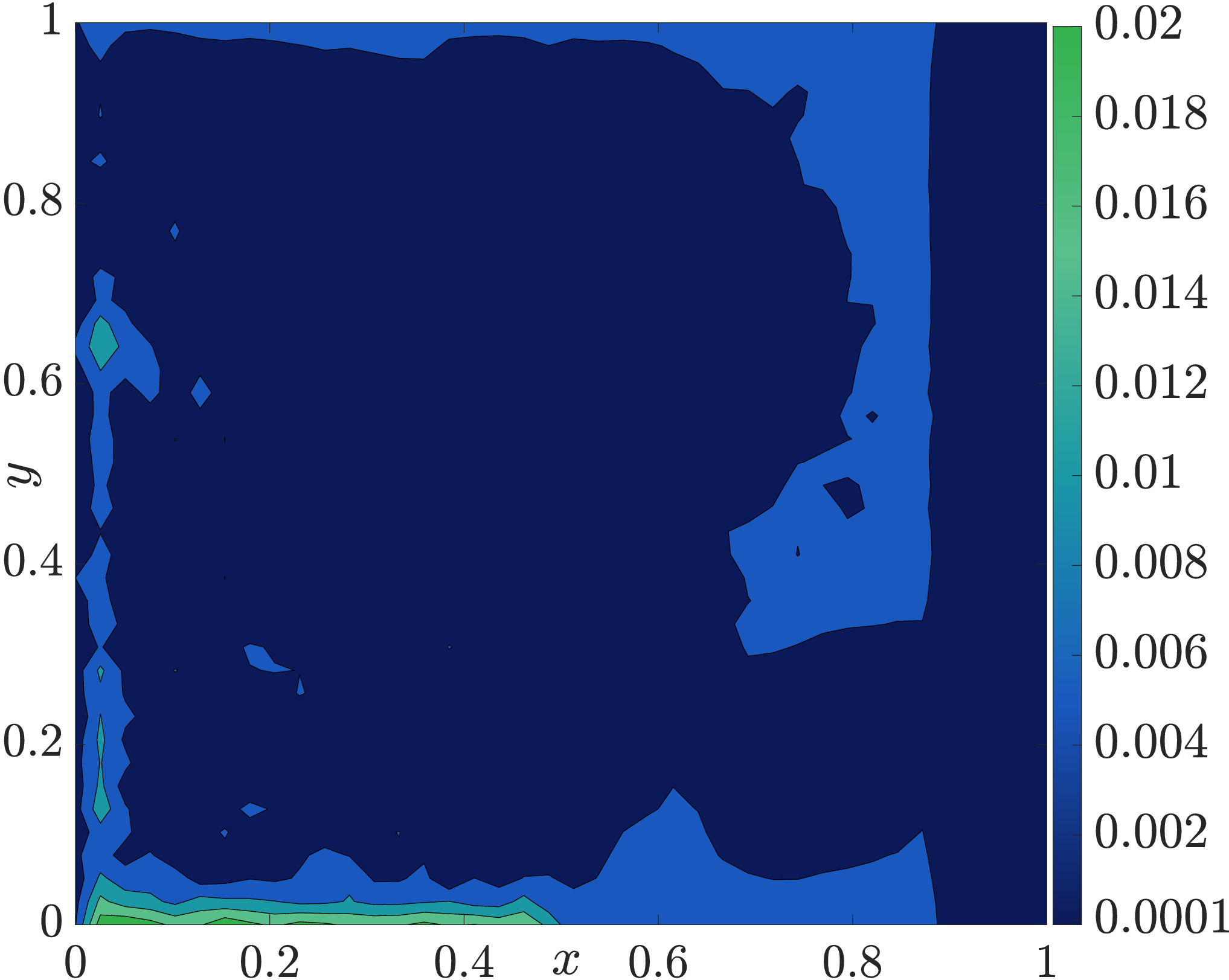}
\end{minipage}
\caption{\footnotesize [Advection-Diffusion Case 2] Std estimation: (Left) Reference std, (Middle) SON prediction std, (Right) Std prediction Error.}
\label{AdvDiff_VeloNoise_SingleStd}
\vspace{-0.1cm}
\end{figure}

We first compare the predictive accuracy between the deterministic DeepONet and the SON. In SON, Phase I and Phase II are trained for 500 and 200 epochs, respectively, for a total of 700 training epochs, while the DeepONet baseline is trained for $500$.
%As in Case 1, we randomly select three test inputs and compute predictions from both models. For Phase II, we draw 400 samples and report their sample mean. We compute the reference sample mean from 400 numerical-solver realizations for each input.
Figure~\ref{AdvDiff_Case2_Compare_ErrorMap} presents heatmaps of the prediction error between the reference sample mean and the predictions from the deterministic DeepONet (left) and the SON predicted sample mean (right). The results clearly demonstrate the superior predictive accuracy of the SON compared to the deterministic DeepONet.
To better demonstrate both the improved predictive accuracy over the deterministic DeepONet and the ability of SON to quantify model uncertainty, Figure~\ref{AdvDiff_Case2_Compare_CrossSections} presents cross-section predictions for three testing inputs, along with their corresponding uncertainty bands, evaluated at the final column of the solution domain. The first row shows the predictive performance of the DeepONet, while the second row displays the SON predictions together with their associated uncertainty bands. From this figure, we can see that SON not only achieves higher predictive accuracy than the deterministic DeepONet but also provides meaningful uncertainty quantification.
To further assess the uncertainty quantification performance, we compute the std from SON predicted samples and visualize it over the 2D domain in Figure~\ref{AdvDiff_VeloNoise_SingleStd}. By comparing the reference std with the SON-predicted sample std, we observe that SON provides accurate uncertainty estimates across the domain.
%Cross-sections along the last column in Figure~\ref{AdvDiff_Case2_Compare_CrossSections} further illustrate the improvement: compared to Phase I, the Phase-II mean better captures the solution profile and smooths out localized spikes. We did not observe a clear improvement in Case 1 because the Phase-I prediction is already sufficiently accurate; consequently, the Phase-II model primarily fine-tunes the diffusion term.  In addition, the SON model produces reasonable prediction bands for this case.

%To illustrate the dependence of the uncertainty on the training input, we choose another input forcing term and repeat the same procedure to generate a new collection of predicted solution samples and compute the corresponding standard deviation. This predicted standard deviation, together with the corresponding reference one, is also displayed in Figure~\ref{AdvDiff_VeloNoise_SingleStd}. We observe that different input forcing terms lead to different standard deviation maps. Nevertheless, our proposed method is still able to capture the overall shape of the heatmaps and provide reliable uncertainty quantification. More precisely, the maximum absolute errors between the reference and predicted standard deviation maps are around $0.02$-$0.025$, and these are mostly concentrated near the bottom boundary. The remaining errors are around $0.0001$-$0.015$.
\begin{figure}[h!]
\vspace{-0.3cm}
\begin{minipage}{0.333\textwidth}
\includegraphics[scale = 0.227]{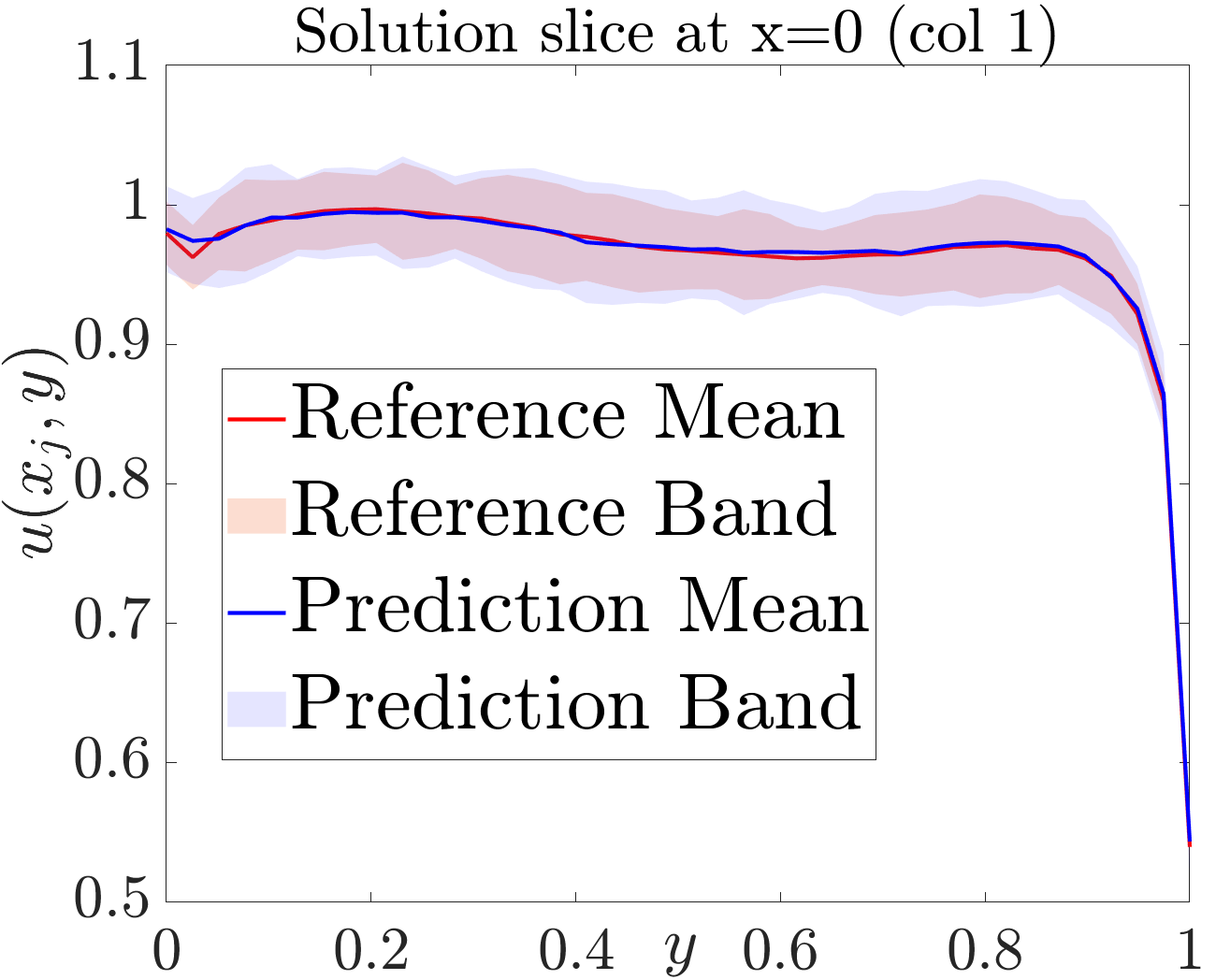}
\end{minipage}%
\begin{minipage}{0.333\textwidth}
\includegraphics[scale = 0.227]{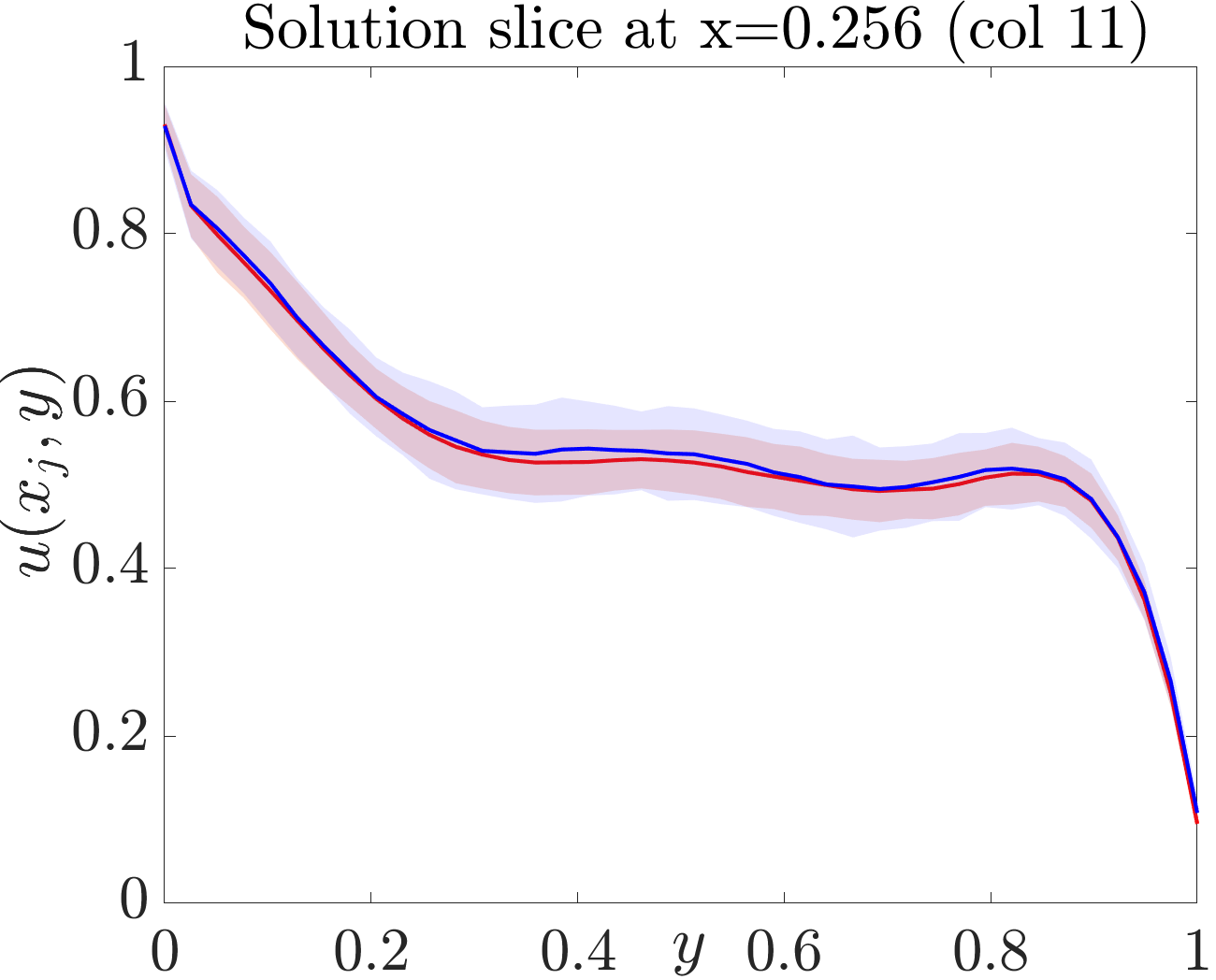}
\end{minipage}%
\begin{minipage}{0.333\textwidth}
\includegraphics[scale = 0.227]{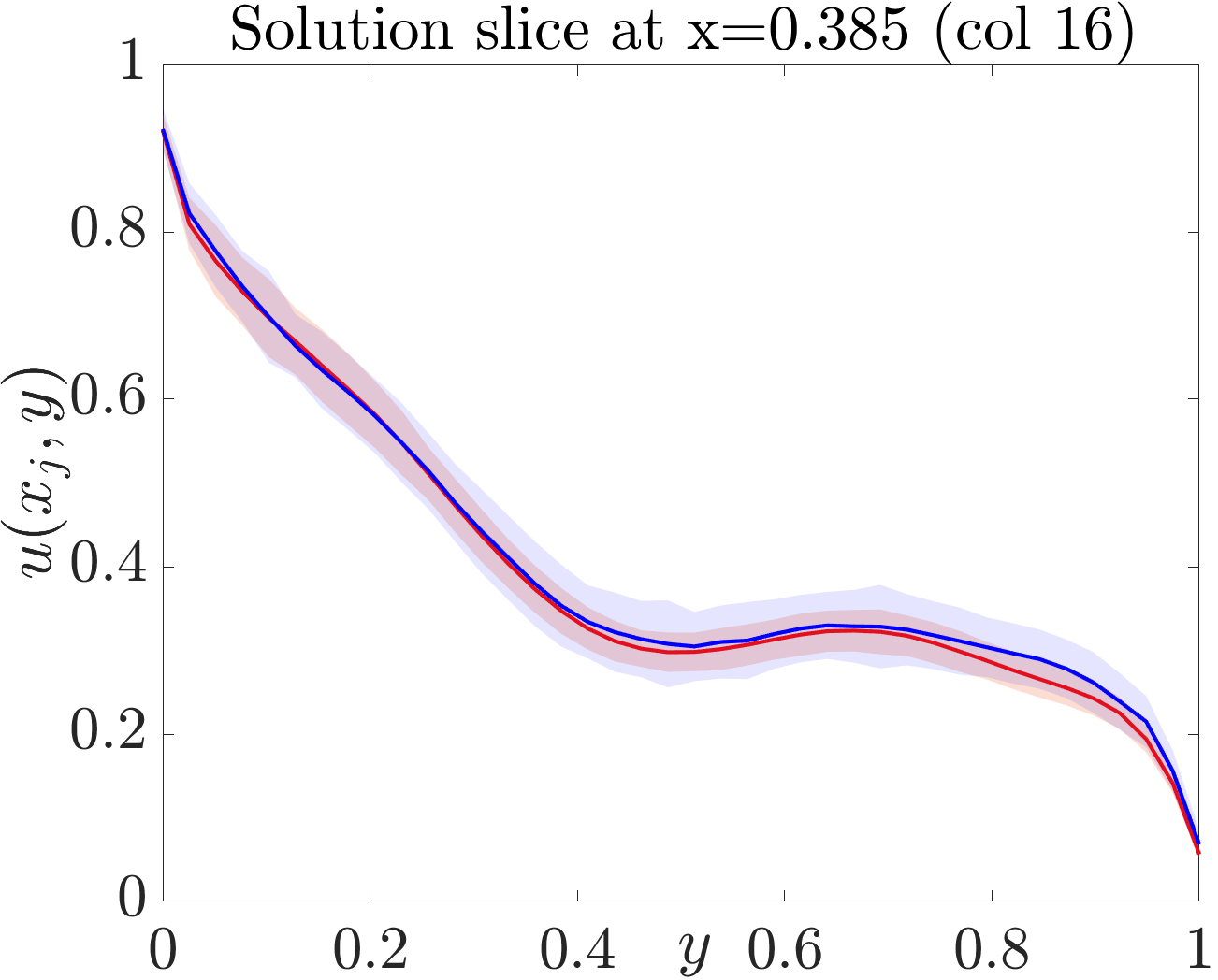}
\end{minipage}

\begin{minipage}{0.333\textwidth}
\includegraphics[scale = 0.227]{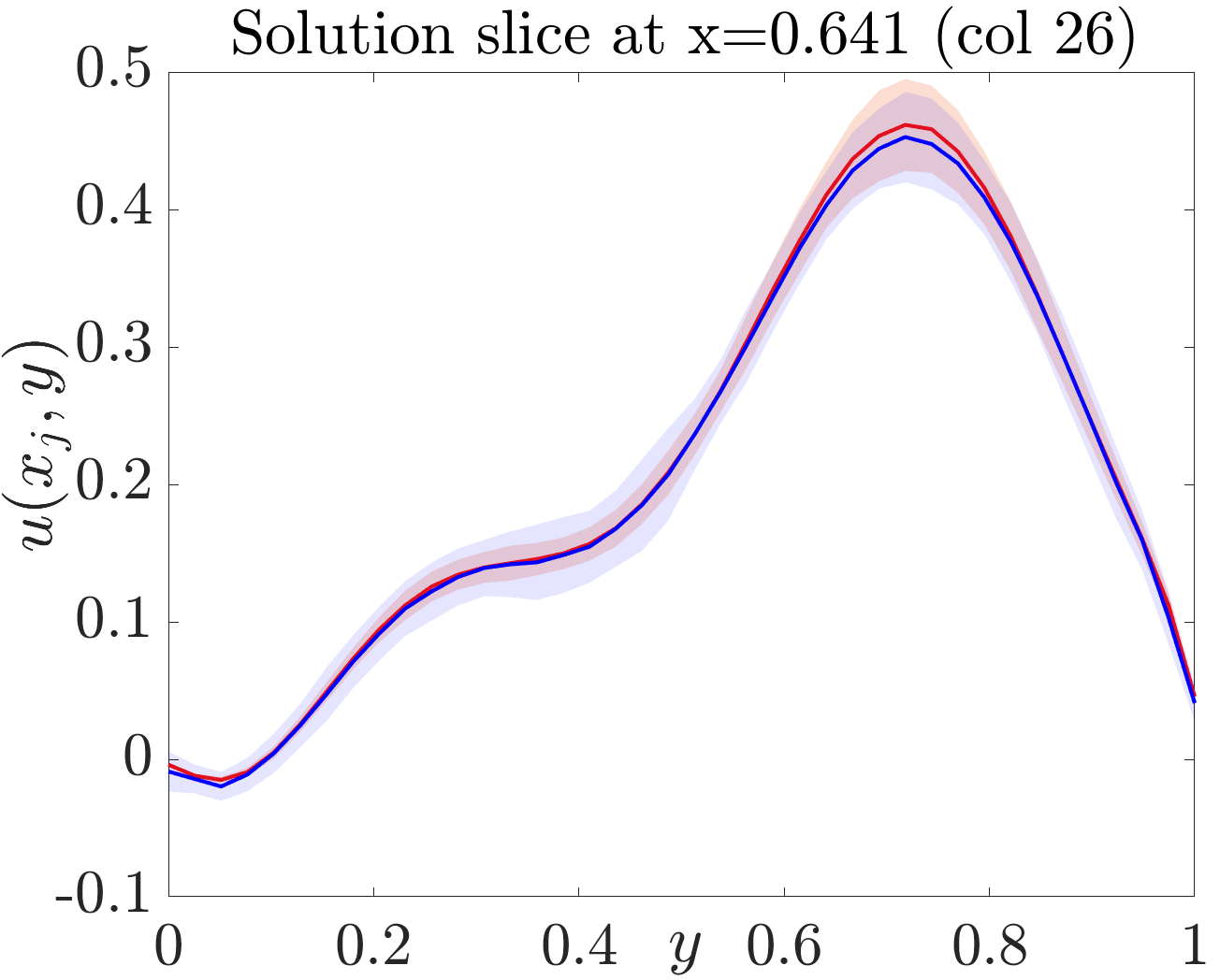}
\end{minipage}%
\begin{minipage}{0.333\textwidth}
\includegraphics[scale = 0.227]{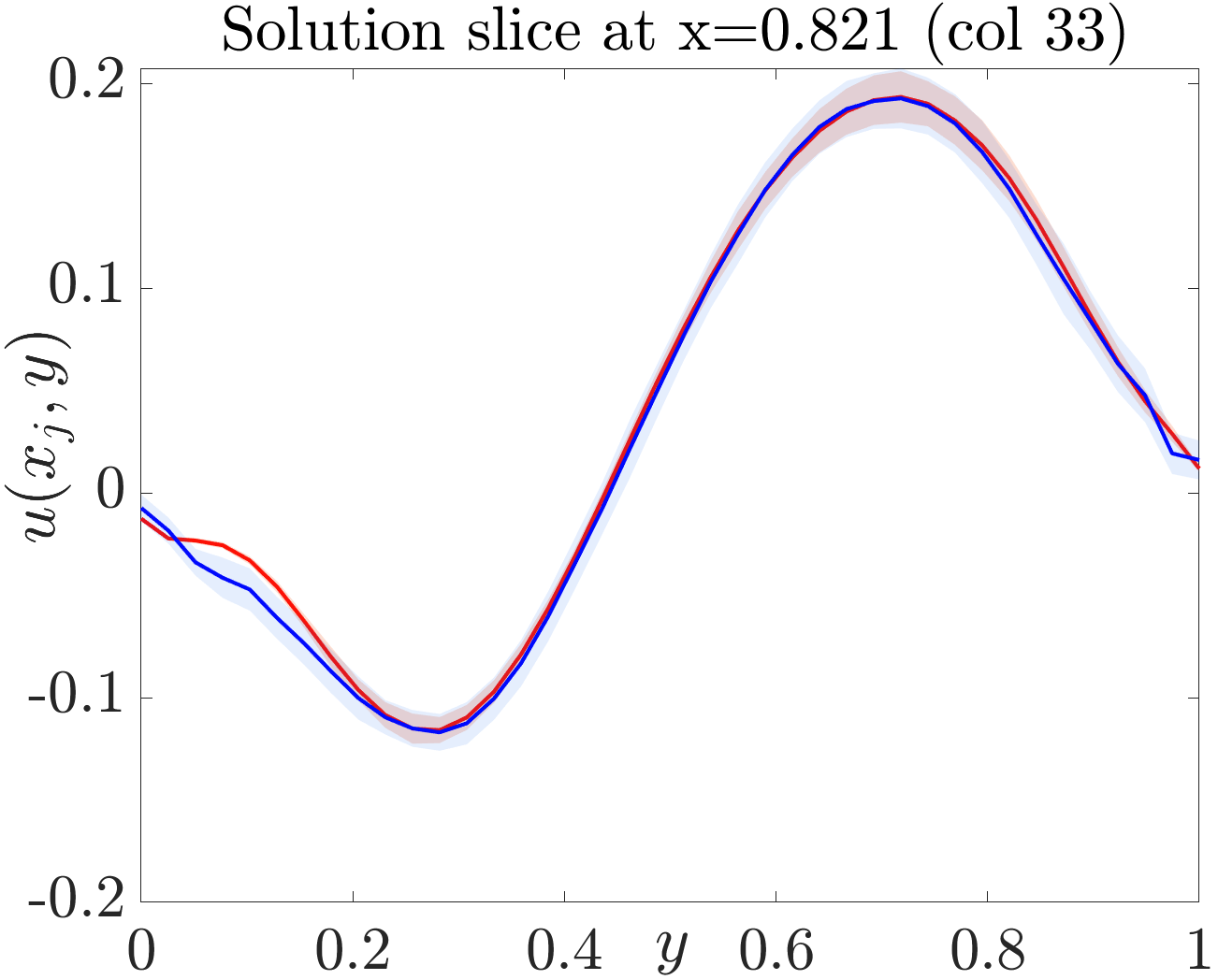}
\end{minipage}%
\begin{minipage}{0.333\textwidth}
\includegraphics[scale = 0.227]{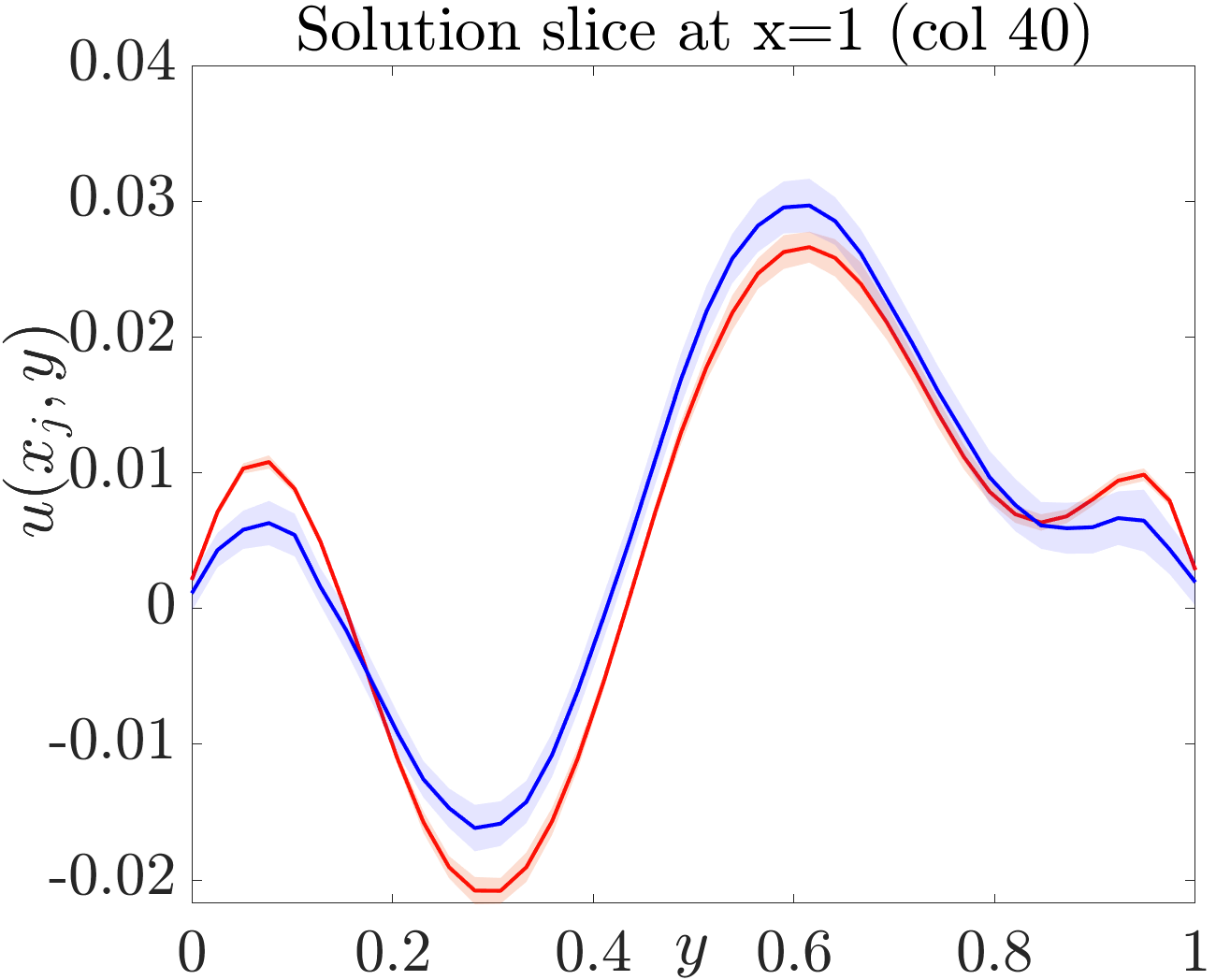}
\end{minipage}
\caption{\footnotesize [Advection-Diffusion Case 2] Cross-sections at columns 1, 11, 16, 26, 33, and 40 of the reference solution and approximation.}
\label{AdvDiff_VeloNoise_CrossSection}
\vspace{-0.3cm}
\end{figure}

To better illustrate SON’s capability in uncertainty quantification, for a representative testing input we plot cross-sections of the reference and predicted means at several selected spatial columns in the solution domain, along with their corresponding confidence bands, in Figure~\ref{AdvDiff_VeloNoise_CrossSection}. From this figure, we can see that the mean of the reference samples is accurately predicted. We also want to point out that the confidence bands exhibit non-uniform behavior, being narrower in some regions than in others, which makes Case 2 more challenging than Case 1. Although the predicted bands do not always perfectly match the reference uncertainty at every spatial location, they preserve a similar overall shape and remain reasonably close in magnitude.

\begin{figure}[h!]
\centering
\begin{minipage}{0.5\textwidth}
\hspace{2cm}\includegraphics[scale = 0.15]{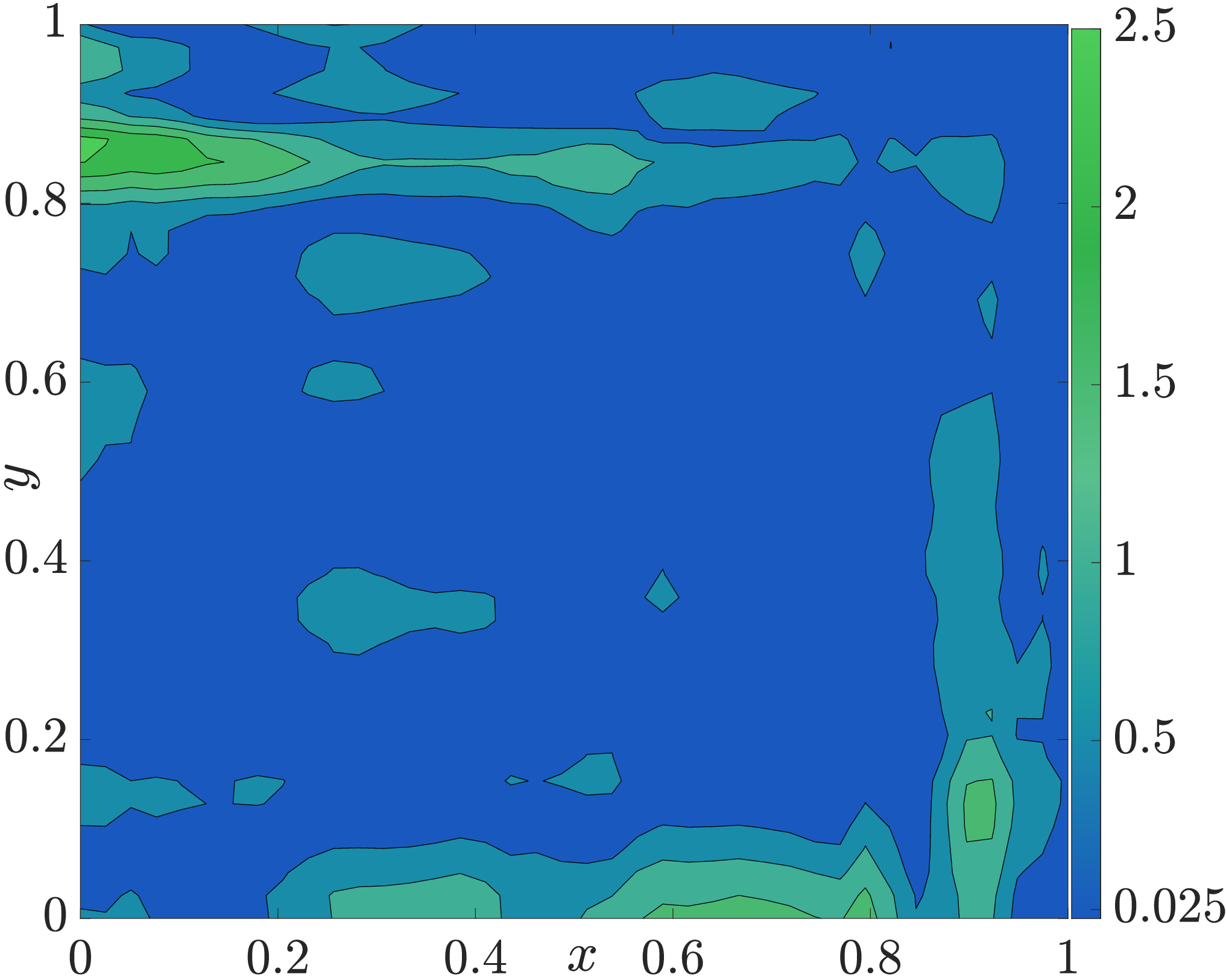}
\end{minipage}%
\begin{minipage}{0.5\textwidth}
\hspace{0.2cm}\includegraphics[scale = 0.15]{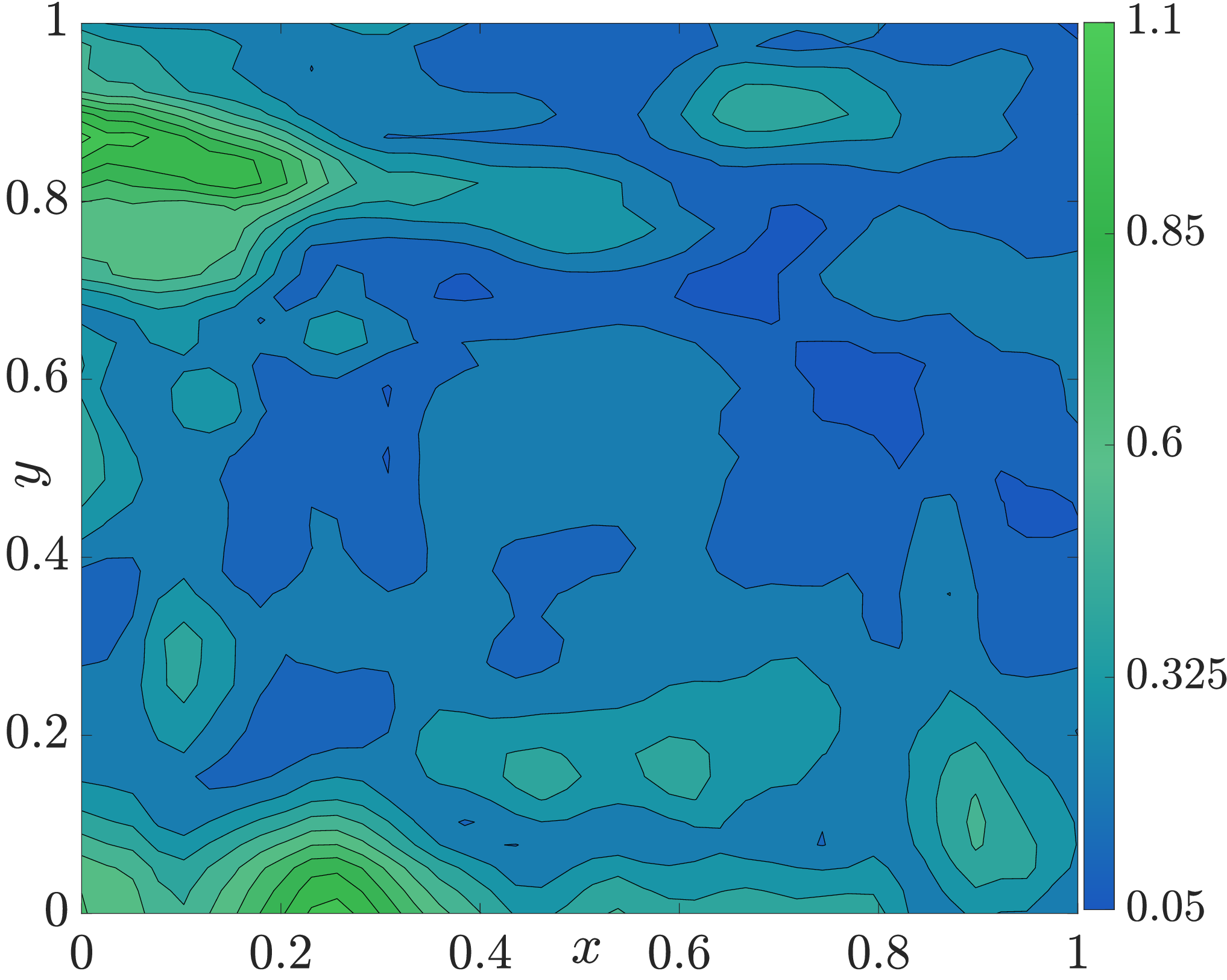}
\end{minipage}
\caption{\footnotesize [Advection-Diffusion Case 2] Errors in the sample mean and std of the B-DeepONet predictions are significantly larger than those of SON under the same input data and training epochs, as shown in Figure~\ref{AdvDiff_Case2_Compare_ErrorMap} (right) and Figure~\ref{AdvDiff_VeloNoise_SingleStd} (right).}
\label{AdvDiff_VeloNoise_Mean_and_Std_BDeepONet}
\end{figure}

\begin{figure}[h!]
\vspace{-0.3cm}
\begin{minipage}{0.333\textwidth}
\includegraphics[scale = 0.227]{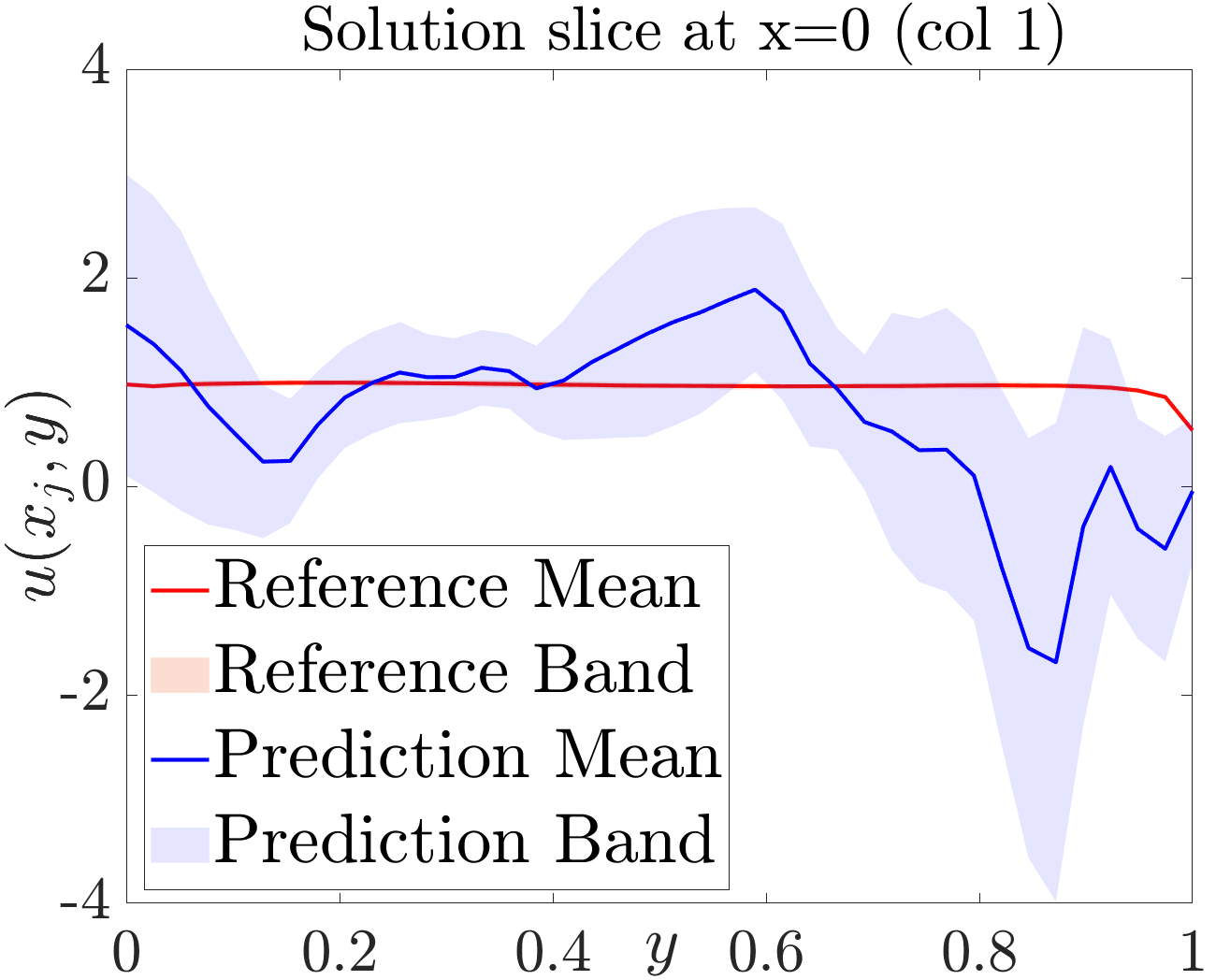}
\end{minipage}%
\begin{minipage}{0.333\textwidth}
\includegraphics[scale = 0.227]{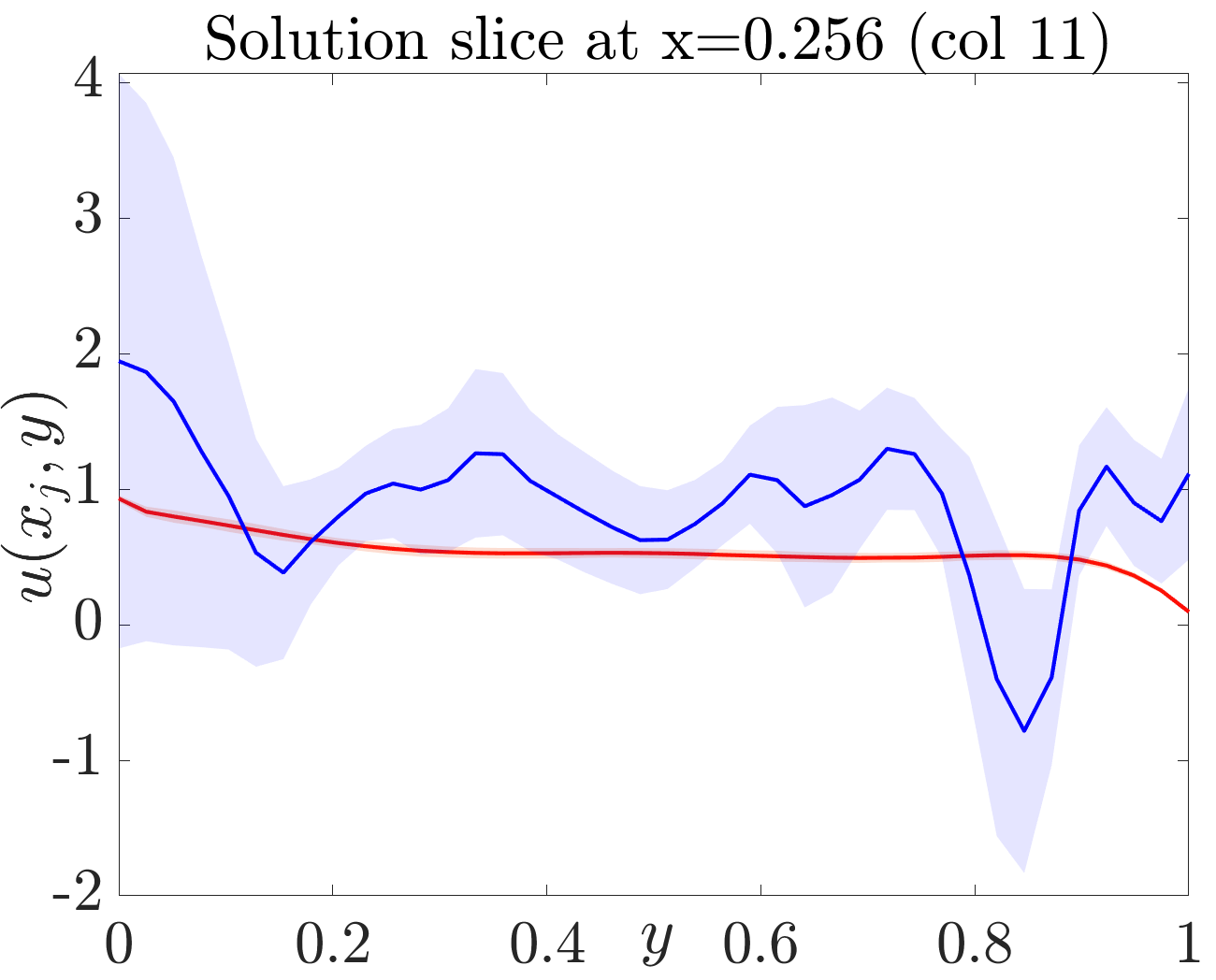}
\end{minipage}%
\begin{minipage}{0.333\textwidth}
\includegraphics[scale = 0.227]{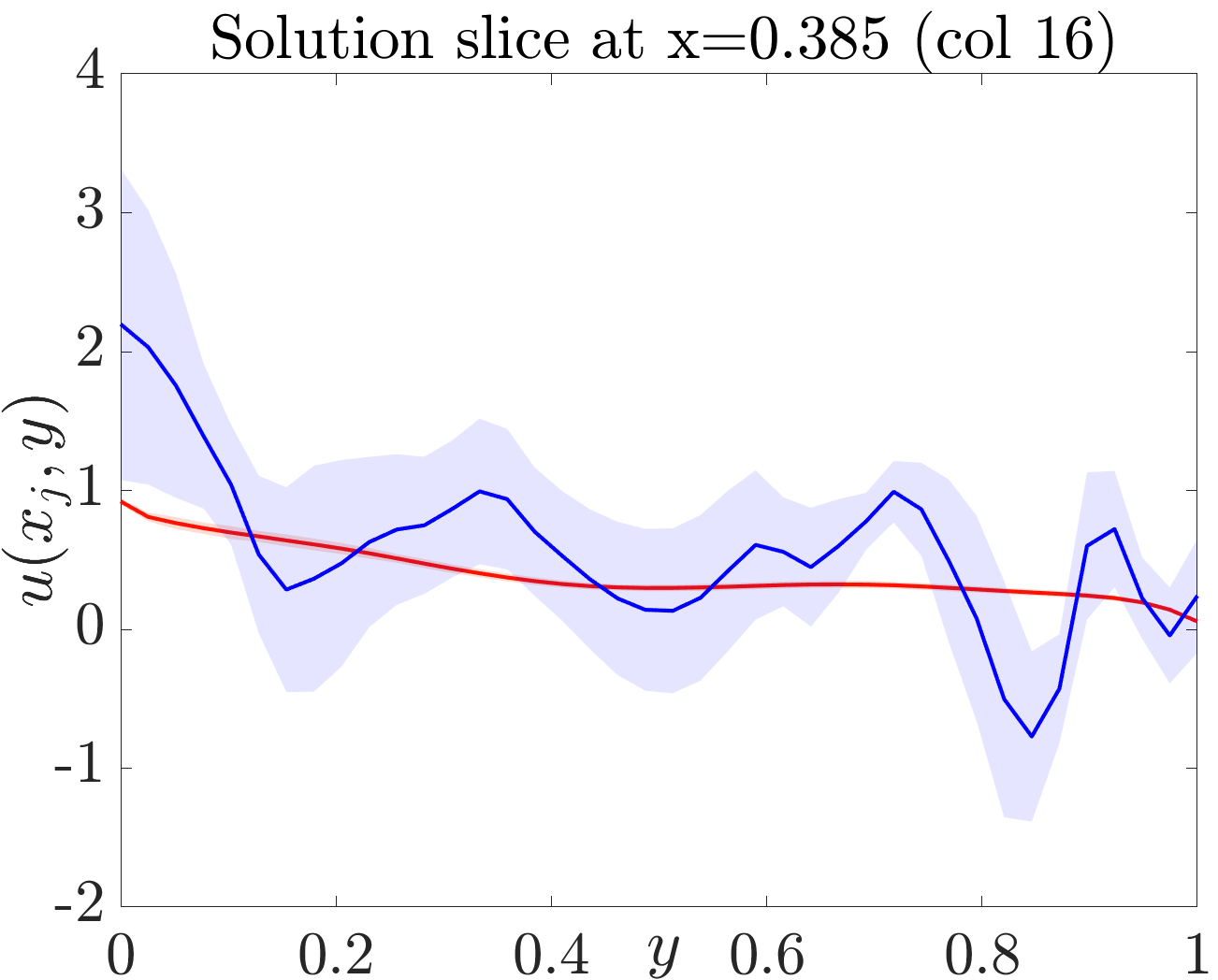}
\end{minipage}

\begin{minipage}{0.333\textwidth}
\includegraphics[scale = 0.227]{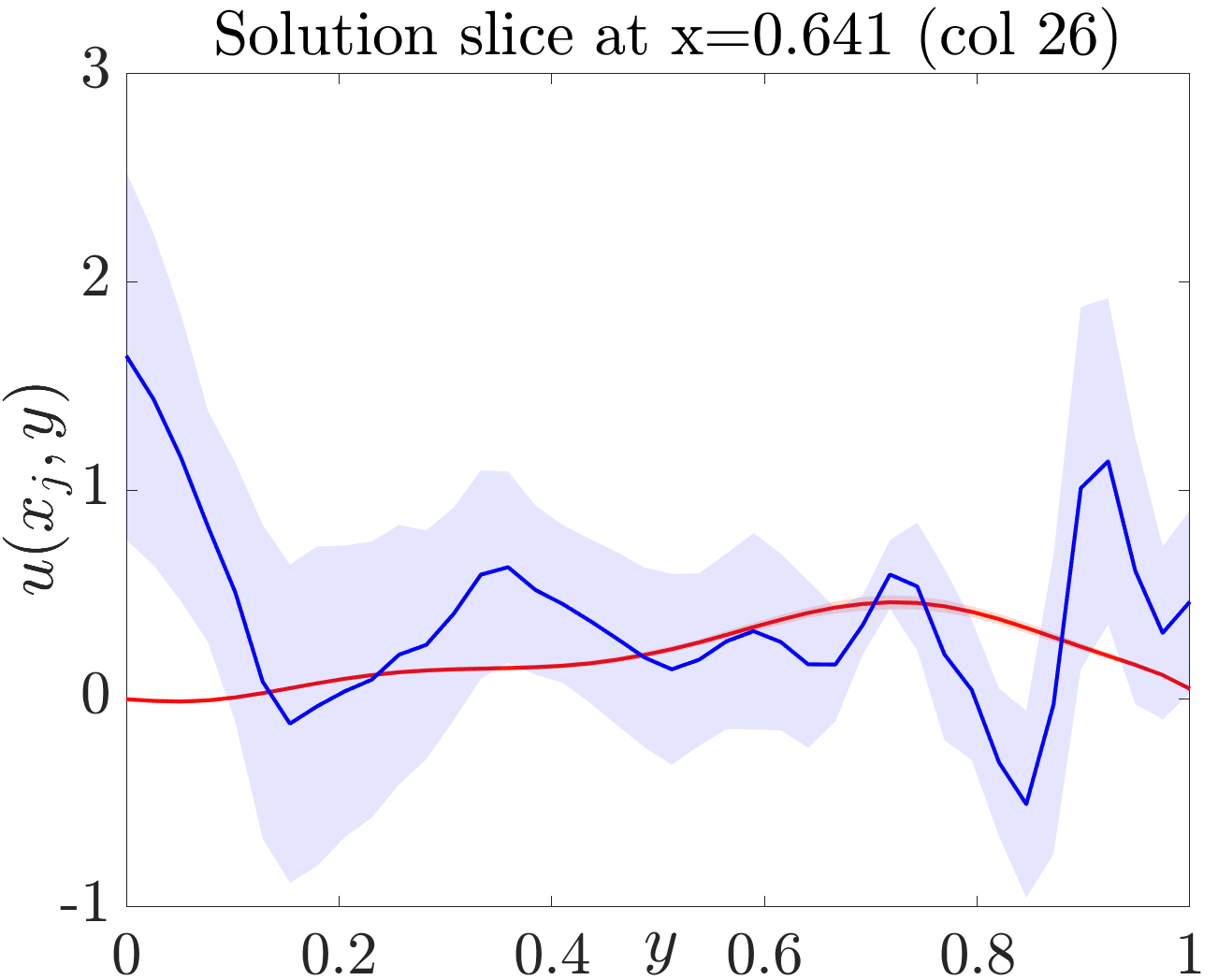}
\end{minipage}%
\begin{minipage}{0.333\textwidth}
\includegraphics[scale = 0.227]{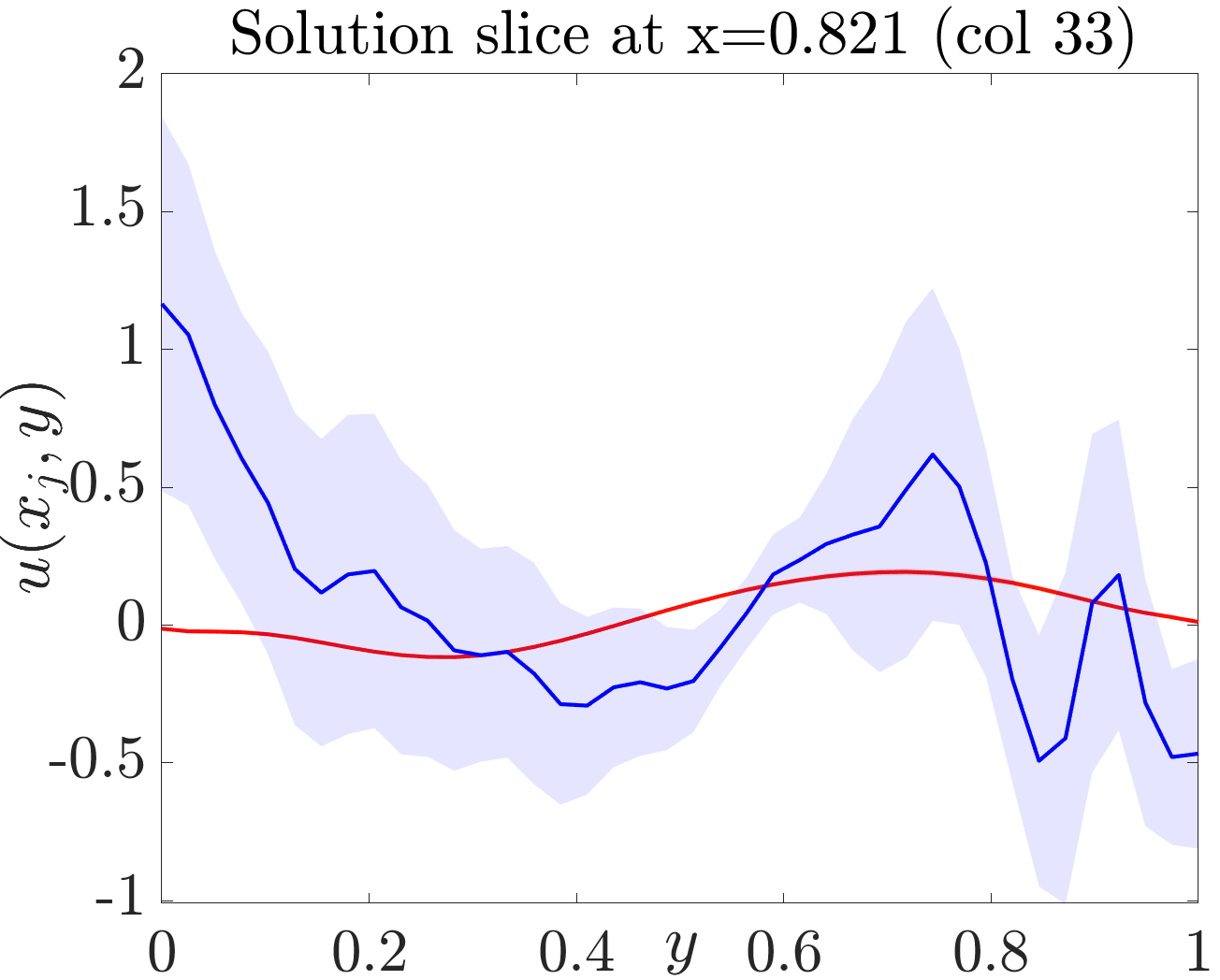}
\end{minipage}%
\begin{minipage}{0.333\textwidth}
\includegraphics[scale = 0.227]{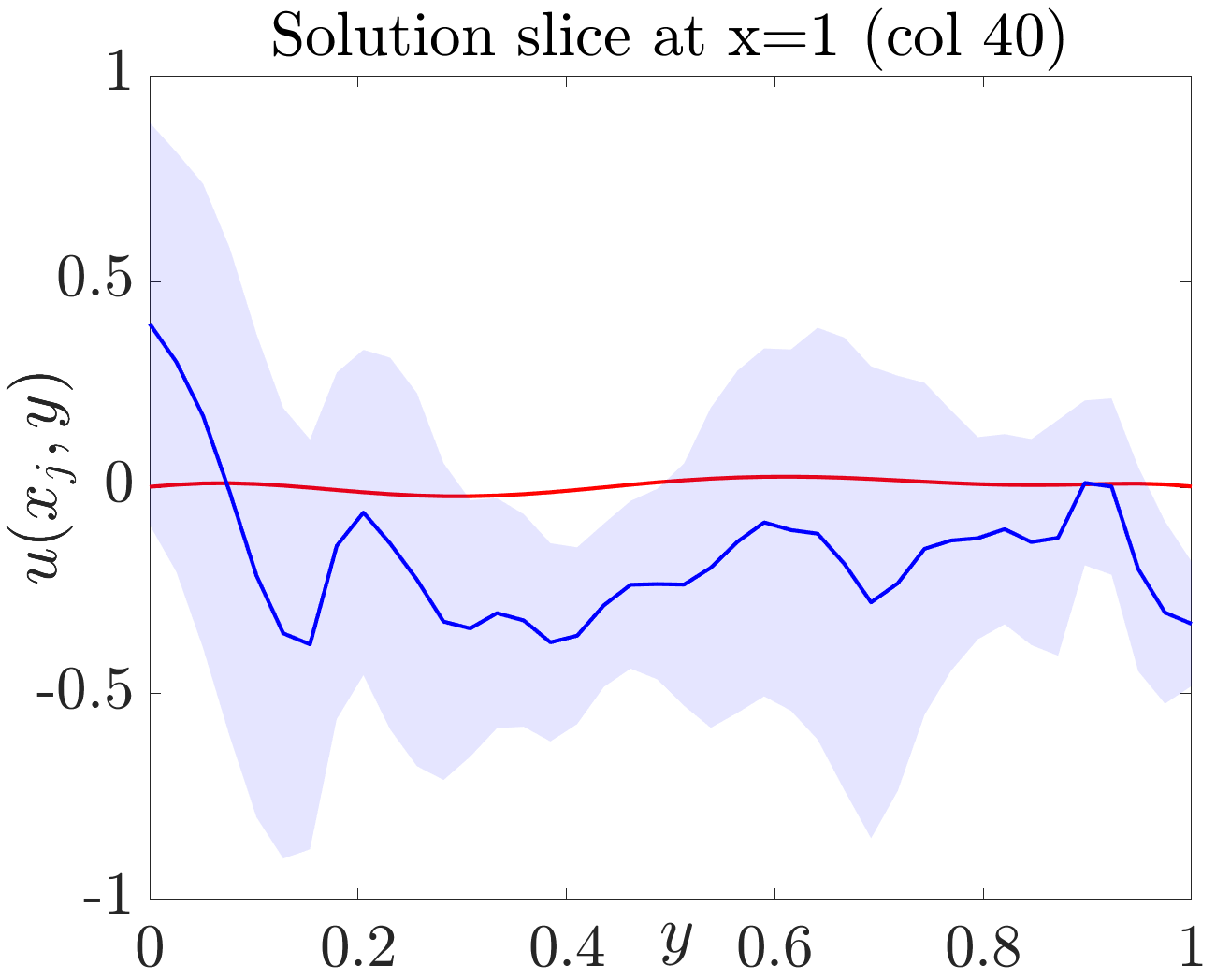}
\end{minipage}
\caption{\footnotesize [Advection-Diffusion Case 2] Cross-sections at columns 1, 11, 16, 26, 33, and 40 of the B-DeepONet prediction, using the same input as in Figure~\ref{AdvDiff_VeloNoise_CrossSection}. Under the same number of training epochs, B-DeepONet produces less accurate predictions compared to both DeepONet and SON.}
\label{AdvDiff_VeloNoise_CrossSection_BDeepONet}
\vspace{-0.3cm}
\end{figure}

In this example, we further demonstrate the superior performance of the SON approach for uncertainty quantification in ML–based predictions by comparing it with the Bayesian DeepONet (B-DeepONet) (introduced in \cite{Lin2023}), in which the network parameters are modeled as random variables and inferred within a Bayesian framework. The B-DeepONet model is trained on the same training dataset and for the same total number of epochs as SON, namely $700$. Using the same input forcing term as in the above experiments, we generate $400$ B-DeepONet predicted samples and compute their sample means and stds. The resulting heatmaps of the mean and std estimation errors are shown in Figure~\ref{AdvDiff_VeloNoise_Mean_and_Std_BDeepONet}, while Figure~\ref{AdvDiff_VeloNoise_CrossSection_BDeepONet} presents the corresponding cross-sections of the sample means and confidence bands along the same selected columns as in Figure~\ref{AdvDiff_VeloNoise_CrossSection}. Compared with the performance of SON in the preceding experiments, our method clearly outperforms B-DeepONet under the same training cost. This is further illustrated by the cross-sections in Figure~\ref{AdvDiff_VeloNoise_CrossSection_BDeepONet}, where the B-DeepONet predictions fail to capture the overall shape of the reference solution, and the associated confidence bands are significantly wider and noisier than those produced by SON. This observation is consistent with the analysis in~\cite{Lin2023} (see Appendix C therein), which indicates that B-DeepONet generally requires a large number of training epochs to achieve satisfactory performance. Overall, this comparison demonstrates that SON is more efficient than B-DeepONet while maintaining high predictive accuracy.

%Since B-DeepONet constructs the predictive ensemble from the last $M$ parameter states generated during training, at least $M$ training steps are needed to obtain an ensemble of size $M$ (see~\cite[Algorimth 2]{Lin2023}. Hence, $700$ training epochs are likely insufficient for B-DeepONet to provide accurate prediction and reliable uncertainty quantification. 

\subsection{Heat Equation}
In this example, we study the performance of the SON model for solving time-dependent problems and consider the following 2D heat equation:
\begin{equation}
\label{2D_Heat}
\left\{\begin{array}{rll}
u_t - \Delta{u} &= \tilde{f}(x, y), \;& \text{for} \; (x, y, t) \in (0, 1)^2 \times (0, 1), \vspace{0.1cm} \\
u(x, y, 0) &= 0.5x(y^2+1), \; &\text{for} \; (x, y) \in (0, 1)^2,
\end{array}\right.
\end{equation}
subject to the time-independent Dirichlet boundary conditions:
\begin{align*}
\begin{array}{lll}
u(x, 0, t) = 0.5x, \; & u(x, 1, t) = x, \; &\text{for} \; x \in  (0, 1), \; t \in (0, 1), \vspace{0.1cm} \\
u(0, y, t) = 0, \; &u(1, y, t) = 0.5(y^2+1), \;& \text{for} \; y \in (0, 1), \; t \in(0, 1).
\end{array}
\end{align*}

For this heat equation example, our goal is to learn a space-time stochastic solution operator associated with Eq.~\eqref{2D_Heat}. In particular, for a given forcing term $\tilde{f}$, we consider solutions driven by an additive perturbation, and we aim to find the neural operator:
$G(\tilde{f})(x,y,t;\epsilon) \approx u_{\tilde{f}}(x,y,t;\epsilon)$, 
where $u_{\tilde{f}}(\cdot,\cdot,\cdot;\epsilon)$ denotes the solution to Eq.~\eqref{2D_Heat} under a random perturbation of the form $\epsilon(t)$ that is constant in $(x,y)$. At the discrete level, we generate the noisy solution by the following propagation rule: at each time step, we add a spatially uniform and temporally i.i.d.\ perturbation  of the form $\epsilon_m(x,y) =\alpha\,\xi_m$, where $\xi_m\stackrel{\text{i.i.d.}}{\sim}\mathcal{N}(0,1)$ and $\alpha=0.28$, to the current numerical solution before advancing to the next time level.
% \begin{equation}
% \epsilon_m(x,y)\equiv \epsilon_m=\alpha\,\xi_m,
% \qquad \xi_m\stackrel{\text{i.i.d.}}{\sim}\mathcal{N}(0,1), \quad \alpha=0.28.
% \end{equation}
%with $\alpha=0.28$. We solve~\eqref{2D_Heat} numerically using backward Euler in time and central finite differences in space. We emphasize that, even though the injected perturbation is spatially uniform at each time level, after the first step the resulting uncertainty in the numerical solution becomes spatially non-uniform due to the action of the implicit time-stepping operator (i.e., the inverse system solve) in the backward Euler scheme.

To generate the training dataset, we run the numerical solver for Eq.~\eqref{2D_Heat} with uniform mesh size $h=1/H$ and time step size $\Delta t = 1/M$, where $H=M=30$. The forcing term is generated from a two-dimensional Chebyshev polynomial space and is evaluated on the $H\times H$ mesh grid. We sample $1800$ forcing terms and split them into $1500$ samples for the training dataset and $300$ samples for the testing dataset. Each forcing term is used to generate a space--time reference solution of size $M\times (H+1)\times (H+1)$. We emphasize that the boundary values are also included in the noisy reference solution. Both datasets then consist of three tensors: the forcing term, the space--time reference solution, and the space--time grid.

We next describe the architecture of the operator network. The branch network takes the forcing-term tensor
$f \in \mathbb{R}^{B \times H \times H}$, where $B=1500$ denotes the number of training samples. We use an $8$-layer
convolutional block $\beta$, whose first two layers are projection layers that downsample the spatial resolution by a factor
of $2$ per layer.
For the trunk network, rather than feeding the full space--time grid into a single trunk, we factor it into two subnetworks:
a spatial trunk $\tau_{\mathrm{space}}$ that processes the spatial grid and a temporal trunk $\tau_{\mathrm{time}}$ that processes
the temporal grid. The spatial trunk $\tau_{\mathrm{space}}$ is a multi-layer two-dimensional convolutional network with two initial
projection layers, reducing the spatial resolution to $\left\lfloor \frac{H}{4} \right\rfloor\times \left\lfloor \frac{H}{4} \right\rfloor$. The temporal trunk $\tau_{\mathrm{time}}$ is implemented
as a multi-layer one-dimensional convolutional network.
To combine these components, the decoder first forms the element-wise product of the branch features $\beta(f)$ and the spatial trunk
features $\tau_{\mathrm{space}}(x,y)$. The result is then fused with $\tau_{\mathrm{time}}(t)$ via an outer product to produce a tensor
in $\mathbb{R}^{B \times M \times \left\lfloor \frac{H}{4} \right\rfloor \times\left\lfloor \frac{H}{4} \right\rfloor}$. We treat the time dimension $M$ as the channel dimension and pass this tensor through an upsampling layer to recover the original spatial resolution, followed by a ResNet-type~\cite{He2016ResNet, He2016Identity} convolutional refinement block.
Regarding the stochastic operator in Phase II, we adopt the architecture introduced in Eq.~\eqref{drift_phase2} for the drift term of the SON. The refinement network $\mathrm{NN}_i$ in Eq.~\eqref{diffusion_phase2} in the case will focus on refining the columns near the boundaries across all time steps. For the diffusion term, we set the latent dimension of the Gaussian random variable to $r=1$ and construct $M$ time-indexed components: $\sigma_{m}\!\left(\theta_{g,m}\right)
= \sum_{l=1}^{L_g} b_{l,m}\,\mu\!\left(c_{l,m}\right)$, $m=1,\dots,M,$
% \begin{equation}
% \label{Time_Diffusion_Single}
% \sigma_{m}\!\left(\theta_{g,m}\right)
% = \sum_{l=1}^{L_g} b_{l,m}\,\mu\!\left(c_{l,m}\right), \qquad m=1,\dots,M,
% \end{equation}
where $\theta_{g,m}$ collects the parameters $\{b_{l,m}\}_{l=1}^{L_g}$ and $\{c_{l,m}\}_{l=1}^{L_g}$.
We then define the time-dependent diffusion vector $\boldsymbol{\sigma}\!\left(\boldsymbol{\theta}_g\right)
:= \big[\sigma_{1}(\theta_{g,1}), \dots, \sigma_{M}(\theta_{g,M})\big]^{\mathsf T}$,
% \begin{equation}
% \label{Time_Diffusion_Combine}
% \boldsymbol{\sigma}\!\left(\boldsymbol{\theta}_g\right)
% := \big[\sigma_{1}(\theta_{g,1}), \dots, \sigma_{M}(\theta_{g,M})\big]^{\mathsf T},
% \end{equation}
where $\boldsymbol{\theta}_g$ denotes the collection of $\{\theta_{g,m}\}_{m=1}^{M}$.
This vector-valued diffusion is shared across all SNN layers.
Note that the drift output is a tensor in
$\mathbb{R}^{B \times M \times  \left\lfloor \frac{H}{4} \right\rfloor  \times \left\lfloor \frac{H}{4} \right\rfloor }$,
whereas the product $\boldsymbol{\sigma}(\boldsymbol{\theta}_g)\,\Delta W_n$ lies in $\mathbb{R}^{M}$.
Therefore, we broadcast $\boldsymbol{\sigma}(\boldsymbol{\theta}_g)\,\Delta W_n$ along the batch and spatial dimensions to match the drift tensor shape.
In this example we use the same optimizer and a mini-batch size of $10$ in both Phase~I and Phase~II. Phase~I is trained for up to $2000$ epochs, while Phase~II is trained for up to $200$ epochs. %In practice, we apply early stopping once the loss values stabilize. 
%To demonstrate the training behavior of SON, we present the loss curves in Figure~\ref{Losses_PhaseI_PhaseII_2DHeat}. In Phase~I, the MSE decreases rapidly and stabilizes after approximately $10{,}000$ iterations. In Phase~II, due to the additional stochastic diffusion term, the MSE starts at $\mathcal{O}(10^{-1})$ and decreases to $\mathcal{O}(10^{-2})$, comparable to Phase~I. The Hamiltonian loss also decreases and becomes stable after about $15{,}000$ iterations. Overall, these trends indicate that both training phases reach a stable regime.
% \begin{figure}[h!]
% \begin{minipage}{0.5\textwidth}
% \includegraphics[scale = 0.18]{Figures/2DHeat/Update/Retraining/LossCurve_PhaseI.eps}
% \end{minipage}%
% \begin{minipage}{0.5\textwidth}
% \includegraphics[scale = 0.18]{Figures/2DHeat/Update/Retraining/LossCurve_PhaseII.eps}
% \end{minipage}
% \caption{\footnotesize [Heat Equation] (Left) MSE loss from Phase 1. (Right) MSE loss and Hamiltonian loss from Phase 2.}
% \label{Losses_PhaseI_PhaseII_2DHeat}
% \vspace{-0.3cm}
% \end{figure}

\begin{figure}[h!]
% \begin{minipage}{0.25\textwidth}
% \includegraphics[scale = 0.11]{Figures/2DHeat/Update/Retraining/PhaseI/Input1/Col1_Time1_Input1_DeepONet.eps}
% \end{minipage}%
% \begin{minipage}{0.25\textwidth}
% \includegraphics[scale = 0.11]{Figures/2DHeat/Update/Retraining/PhaseI/Input1/Col1_Time10_Input1_DeepONet.eps}
% \end{minipage}%
% \begin{minipage}{0.25\textwidth}
% \includegraphics[scale = 0.11]{Figures/2DHeat/Update/Retraining/PhaseI/Input1/Col1_Time20_Input1_DeepONet.eps}
% \end{minipage}%
% \begin{minipage}{0.25\textwidth}
% \includegraphics[scale = 0.11]{Figures/2DHeat/Update/Retraining/PhaseI/Input1/Col1_Time30_Input1_DeepONet.eps}
% \end{minipage}

\begin{minipage}{0.25\textwidth}
\includegraphics[scale = 0.11]{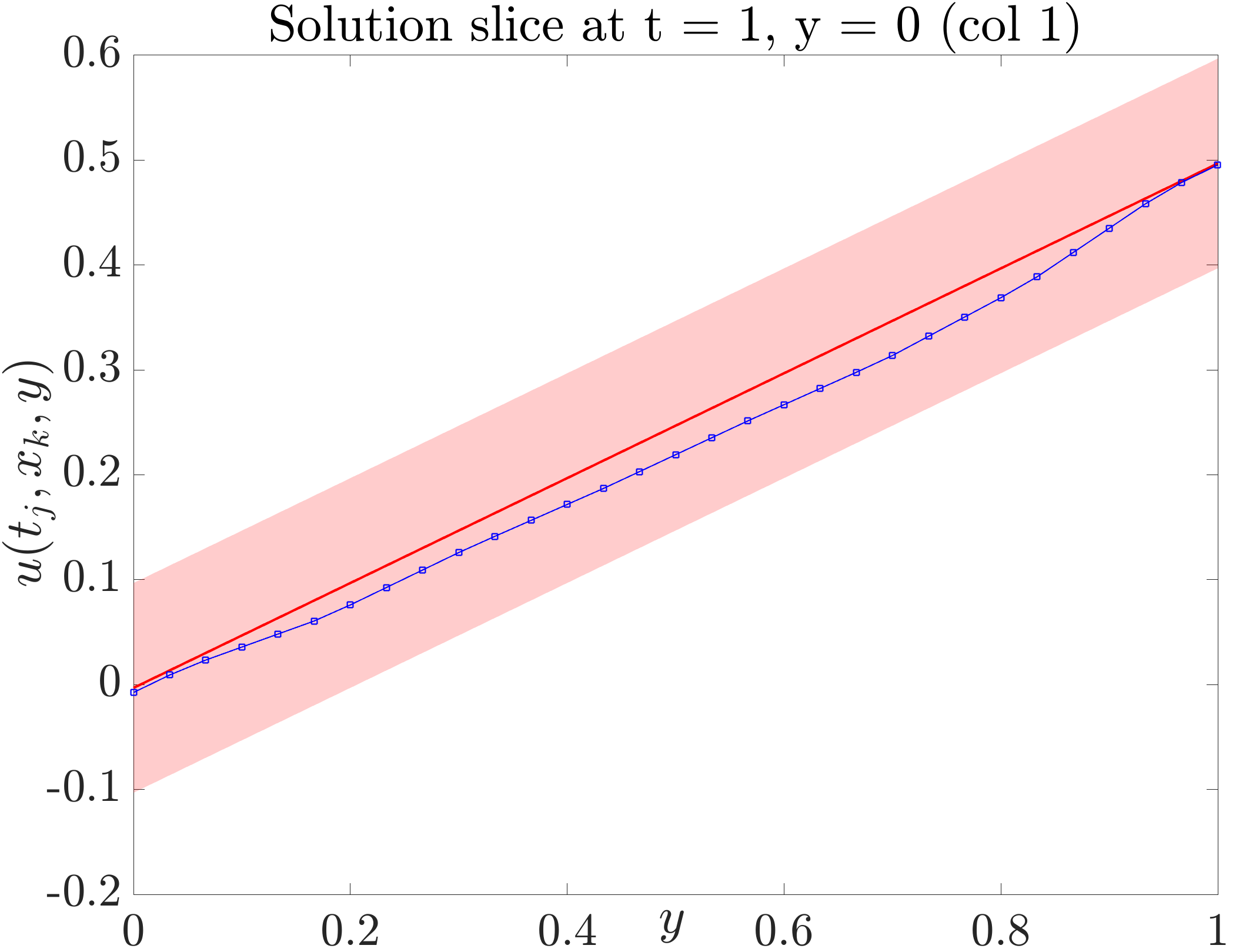}
\end{minipage}%
\begin{minipage}{0.25\textwidth}
\includegraphics[scale = 0.11]{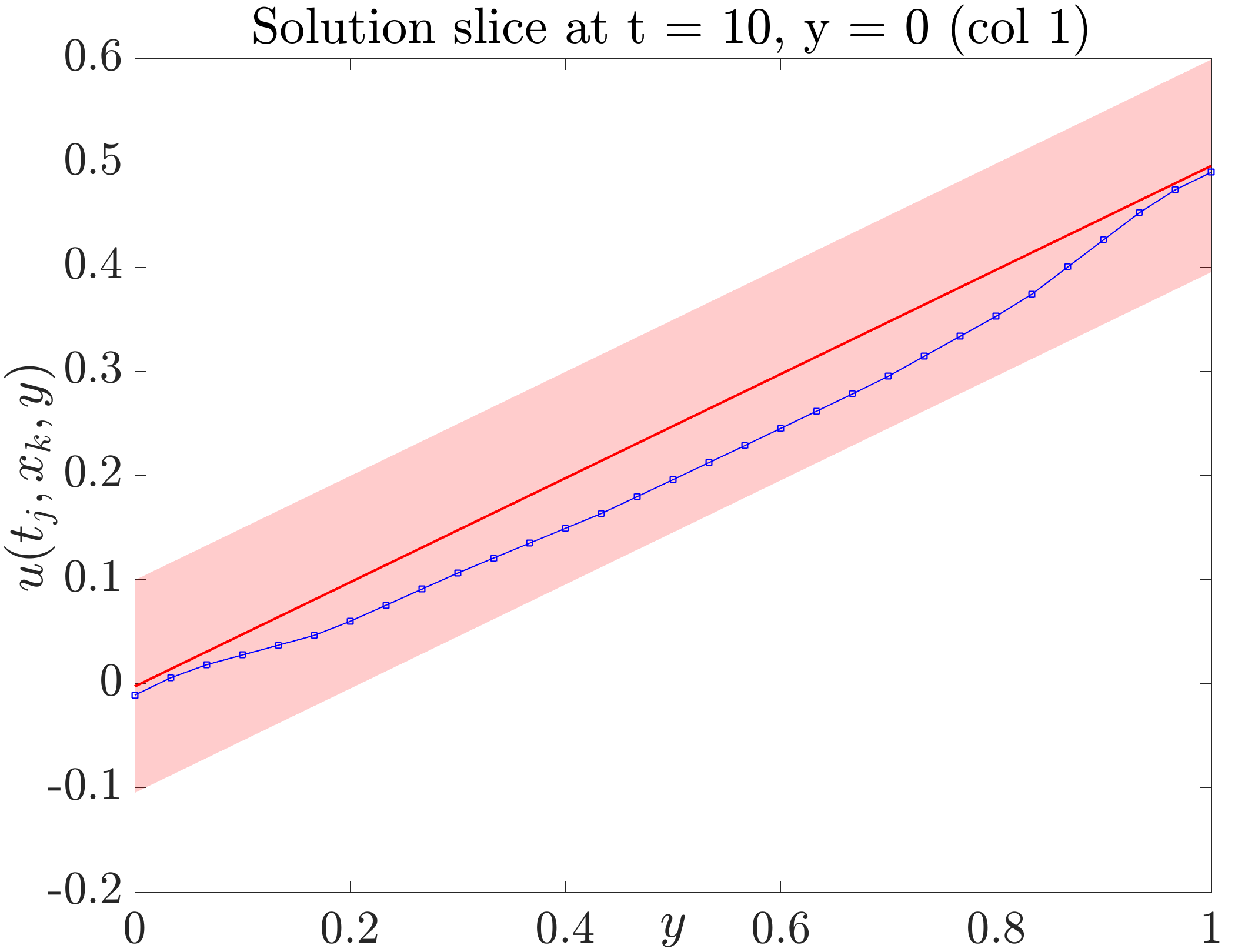}
\end{minipage}%
\begin{minipage}{0.25\textwidth}
\includegraphics[scale = 0.11]{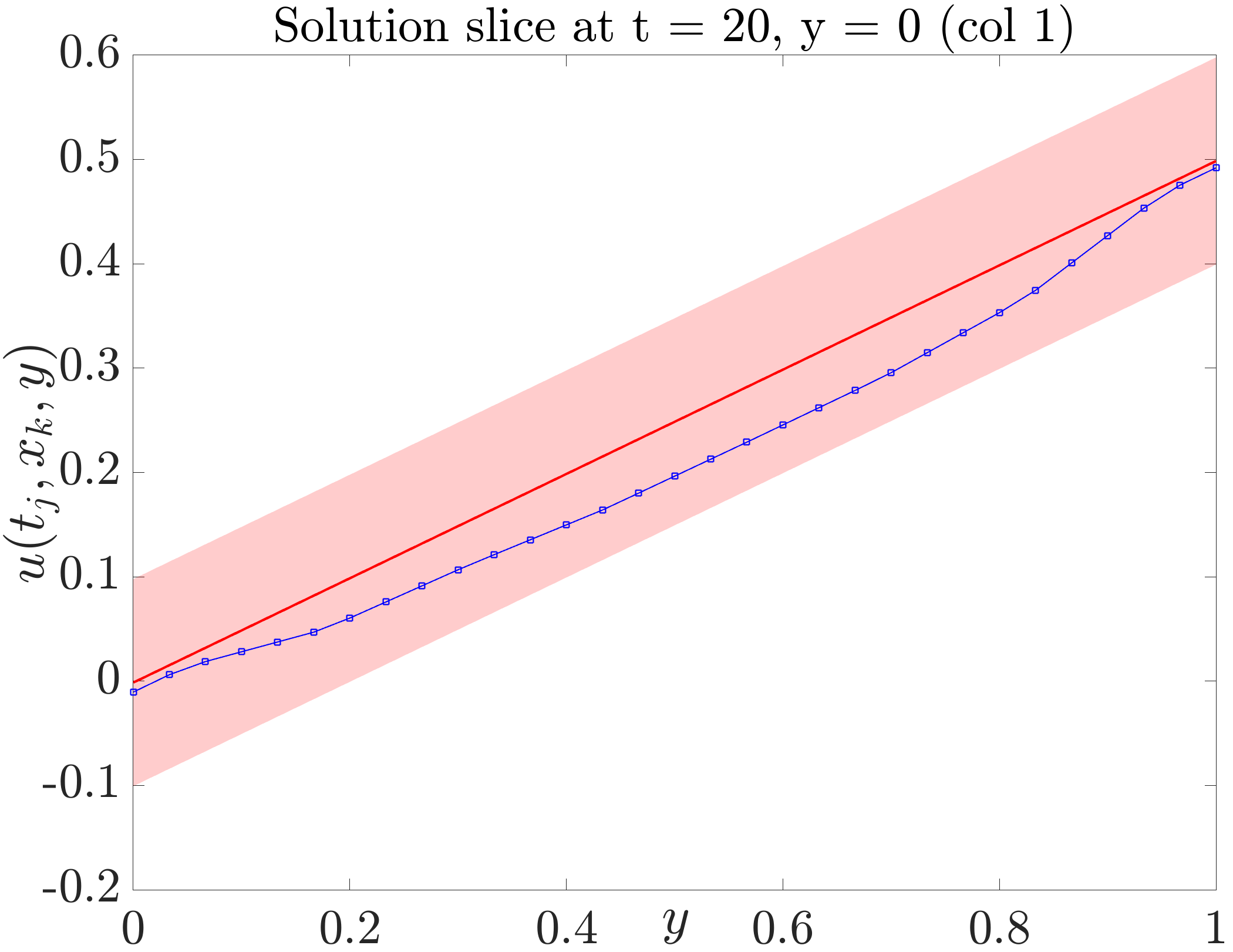}
\end{minipage}%
\begin{minipage}{0.25\textwidth}
\includegraphics[scale = 0.11]{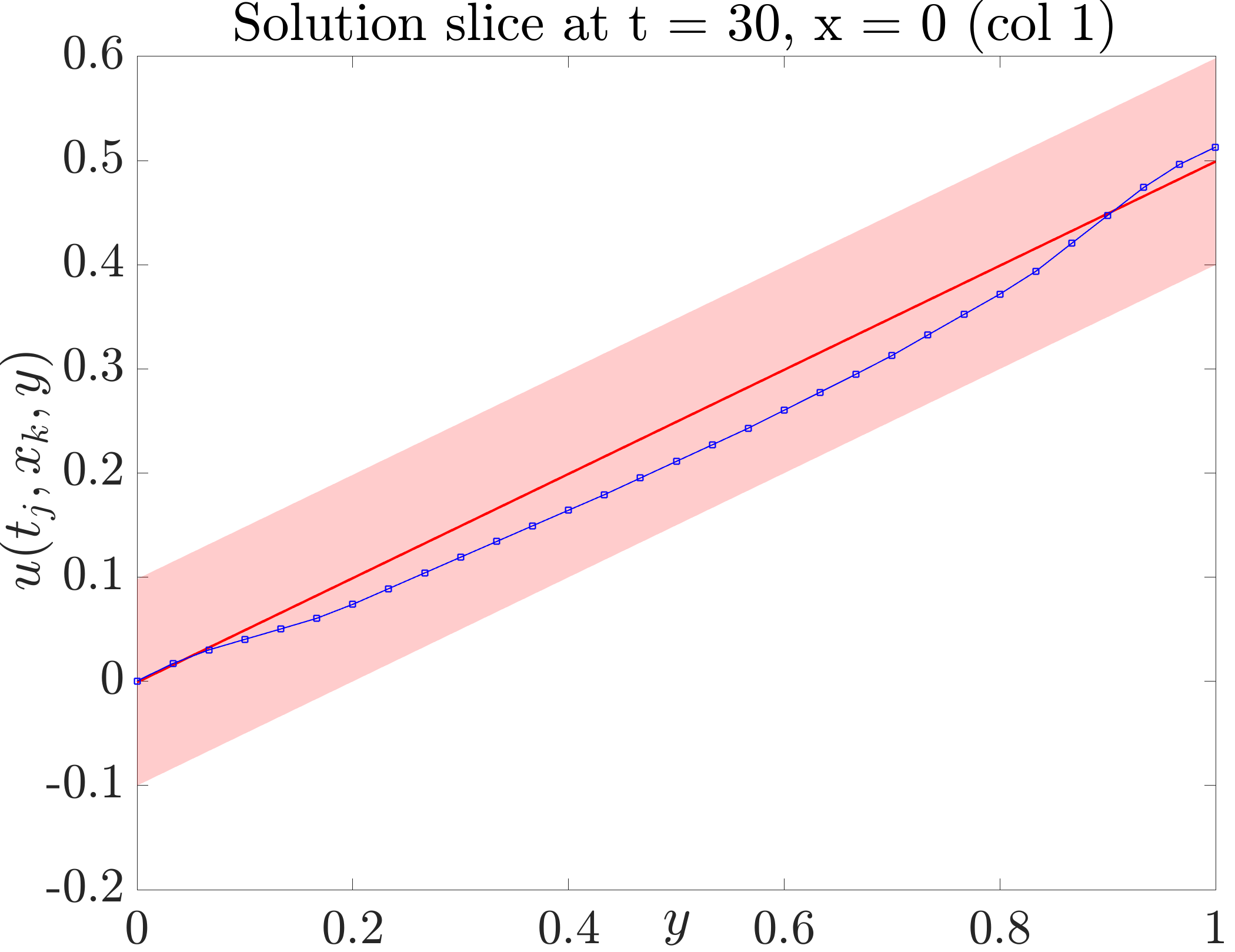}
\end{minipage}

% \begin{minipage}{0.25\textwidth}
% \includegraphics[scale = 0.11]{Figures/2DHeat/Update/Retraining/Input1/Col1_Time1_Input1.eps}
% \end{minipage}%
% \begin{minipage}{0.25\textwidth}
% \includegraphics[scale = 0.11]{Figures/2DHeat/Update/Retraining/Input1/Col1_Time10_Input1.eps}
% \end{minipage}%
% \begin{minipage}{0.25\textwidth}
% \includegraphics[scale = 0.11]{Figures/2DHeat/Update/Retraining/Input1/Col1_Time20_Input1.eps}
% \end{minipage}%
% \begin{minipage}{0.25\textwidth}
% \includegraphics[scale = 0.11]{Figures/2DHeat/Update/Retraining/Input1/Col1_Time30_Input1.eps}
% \end{minipage}%

\begin{minipage}{0.25\textwidth}
\includegraphics[scale = 0.11]{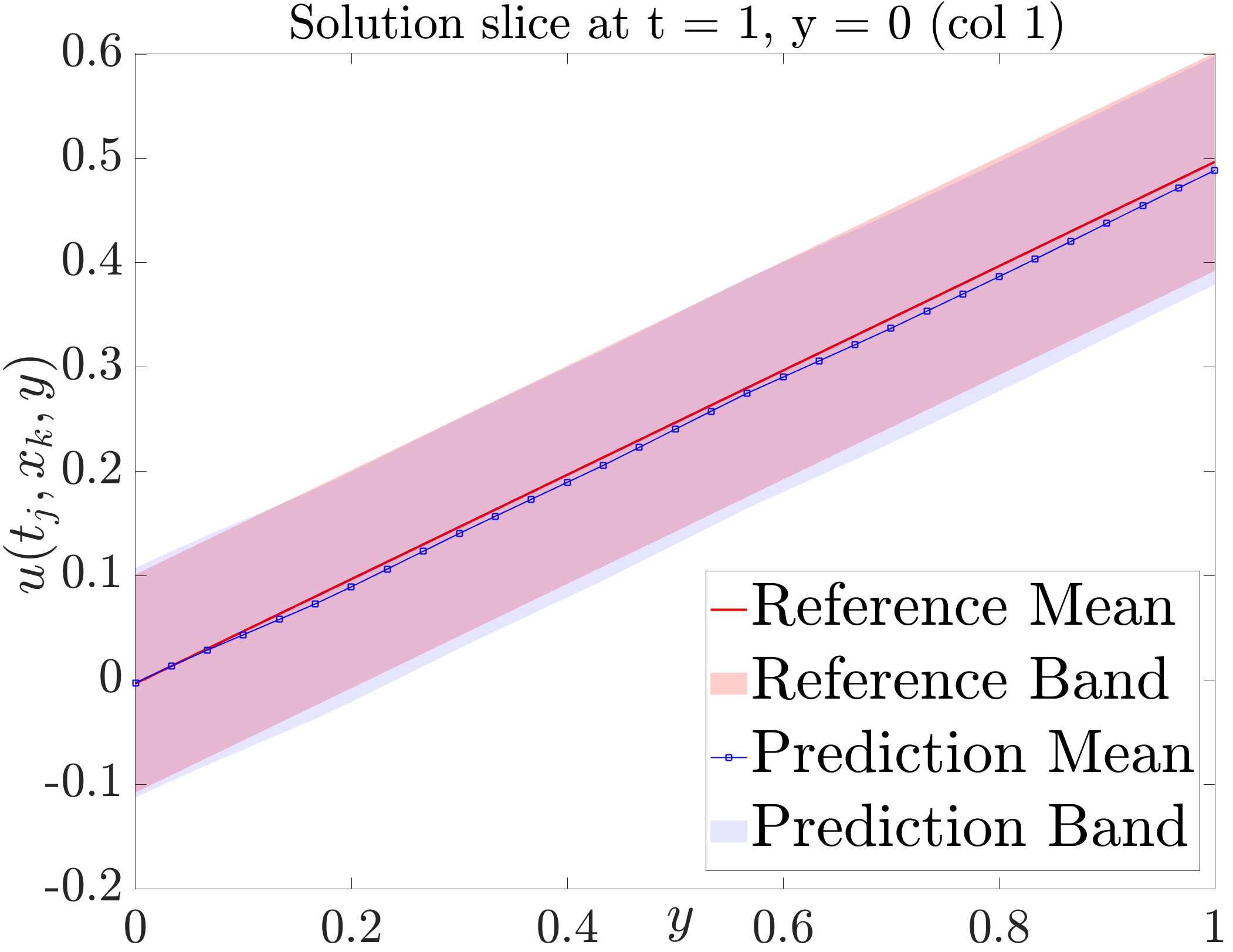}
\end{minipage}%
\begin{minipage}{0.25\textwidth}
\includegraphics[scale = 0.11]{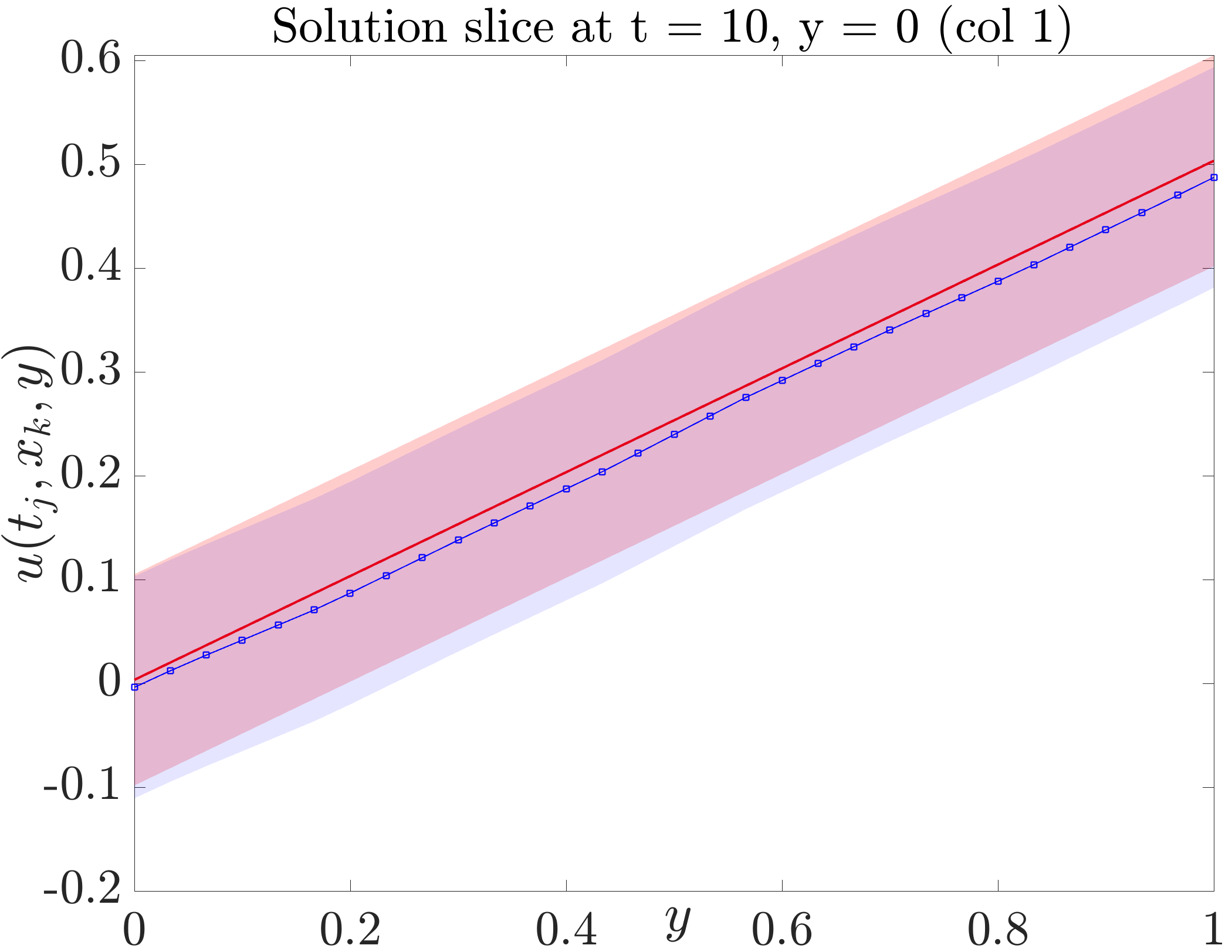}
\end{minipage}%
\begin{minipage}{0.25\textwidth}
\includegraphics[scale = 0.11]{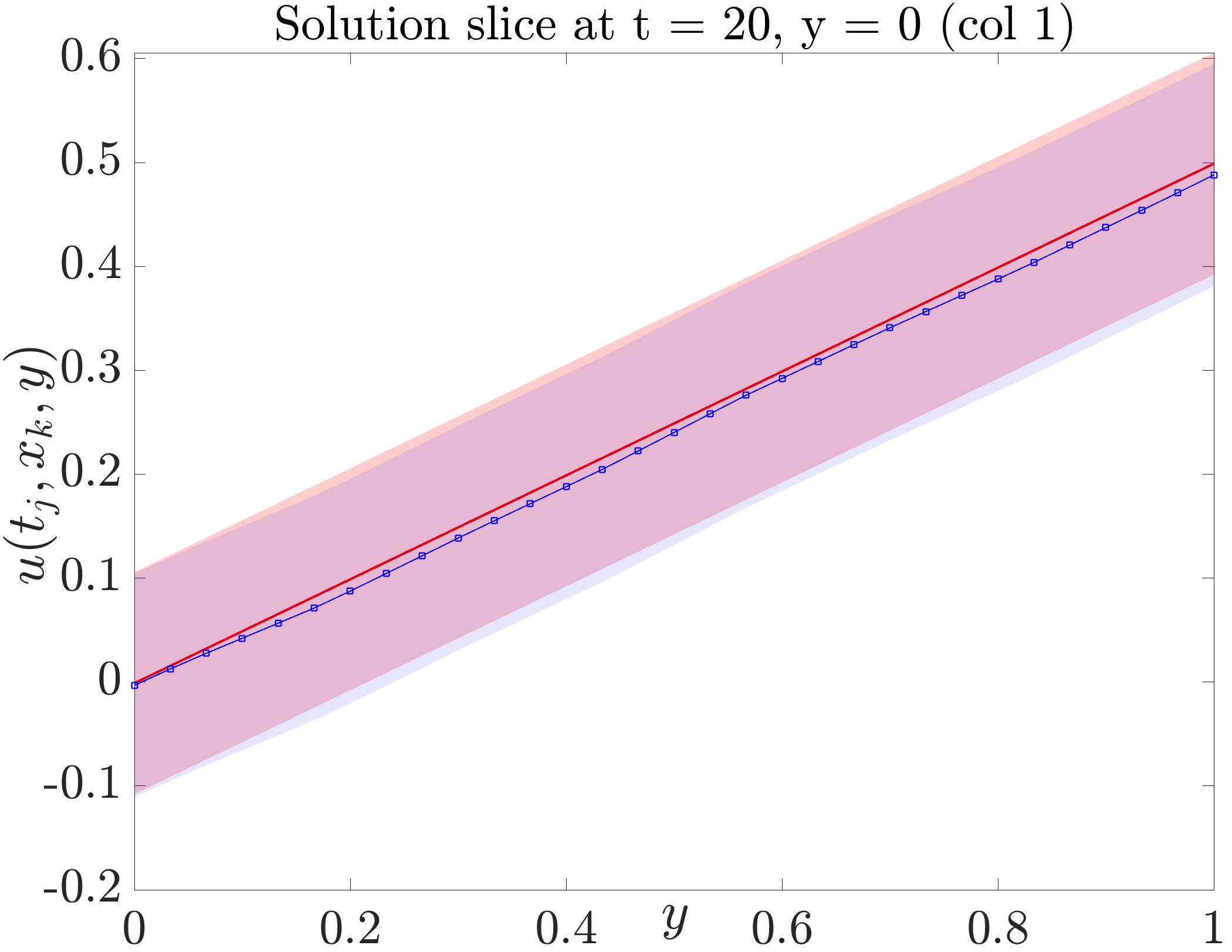}
\end{minipage}%
\begin{minipage}{0.25\textwidth}
\includegraphics[scale = 0.11]{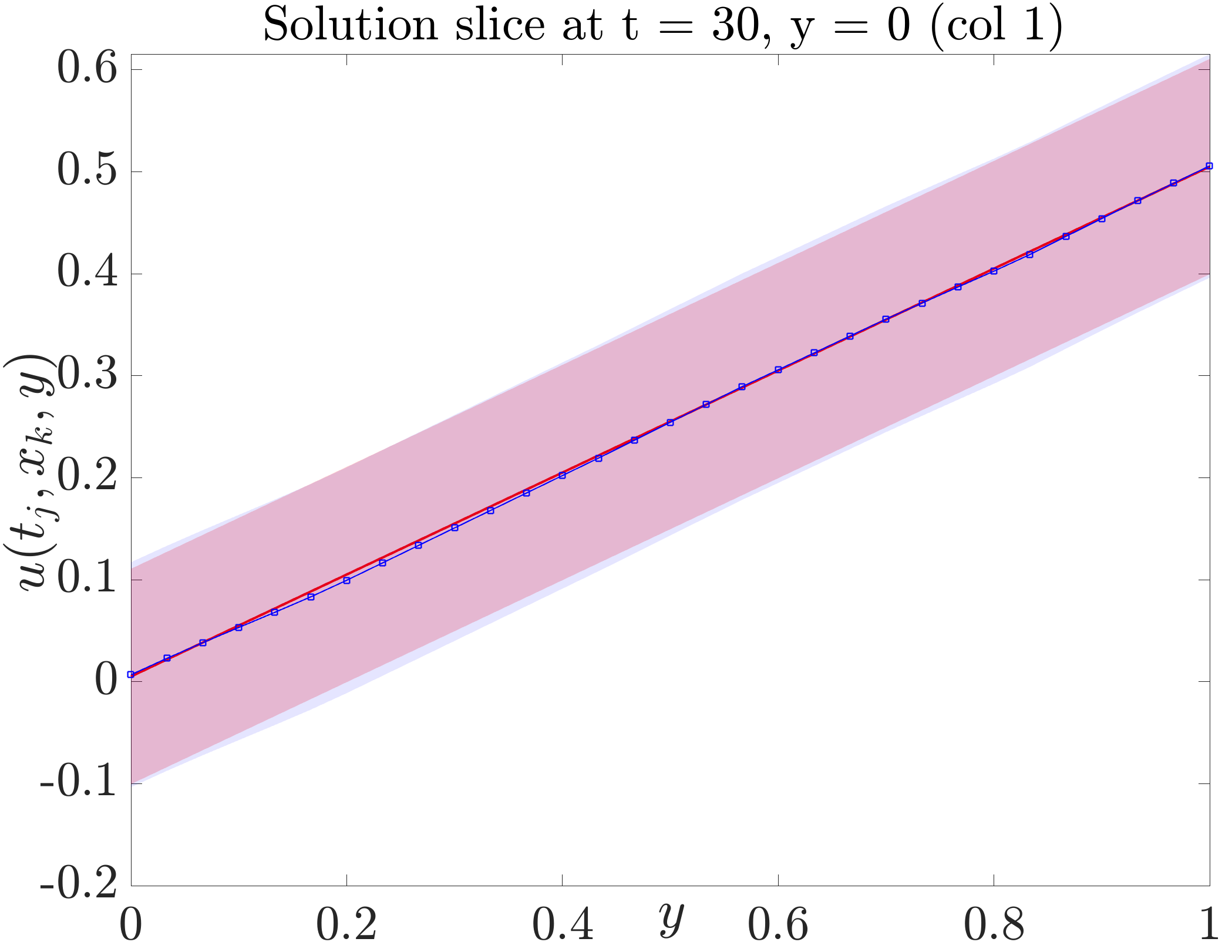}
\end{minipage}
\caption{\footnotesize [Heat Equation] DeepONet vs. SON. Cross-sections of the predicted solutions at the boundary (column 1) and time steps $t_m$ for $m=1, 10, 20, 30$: (First row) DeepONet; (Second row) SON. SON not only provides accurate solution predictions and reliable uncertainty estimates, but also achieves higher predictive accuracy than the deterministic DeepONet.}
\label{2DHeat_Compare}
\vspace{-0.3cm}
\end{figure}

We now examine the performance of SON in solving the heat equation. We randomly select one forcing-term input from the testing dataset and generate predictions using both the deterministic DeepONet and SON. For the SON, we generate 400 predicted solution samples and compute the sample mean along with the corresponding prediction bands. For this testing input, we also generate 400 reference solution samples directly using the numerical solver and compute the corresponding solution sample mean and prediction bands. Figure~\ref{2DHeat_Compare} shows cross-sections along the boundary $x=0$ at time steps $m=1, 10, 20$, and $30$. From this figure, we observe that SON provides accurate solution predictions as well as reliable uncertainty estimates, and it outperforms the deterministic DeepONet in accuracy.

%Compared with the previous section, the SON model yields more clearly improved predictions near the boundary. Moreover, the spread of the predicted distributions agrees well with that of the reference solutions. 

\begin{figure}[h!]
\begin{minipage}{0.333\textwidth} 
\includegraphics[scale= 0.14]{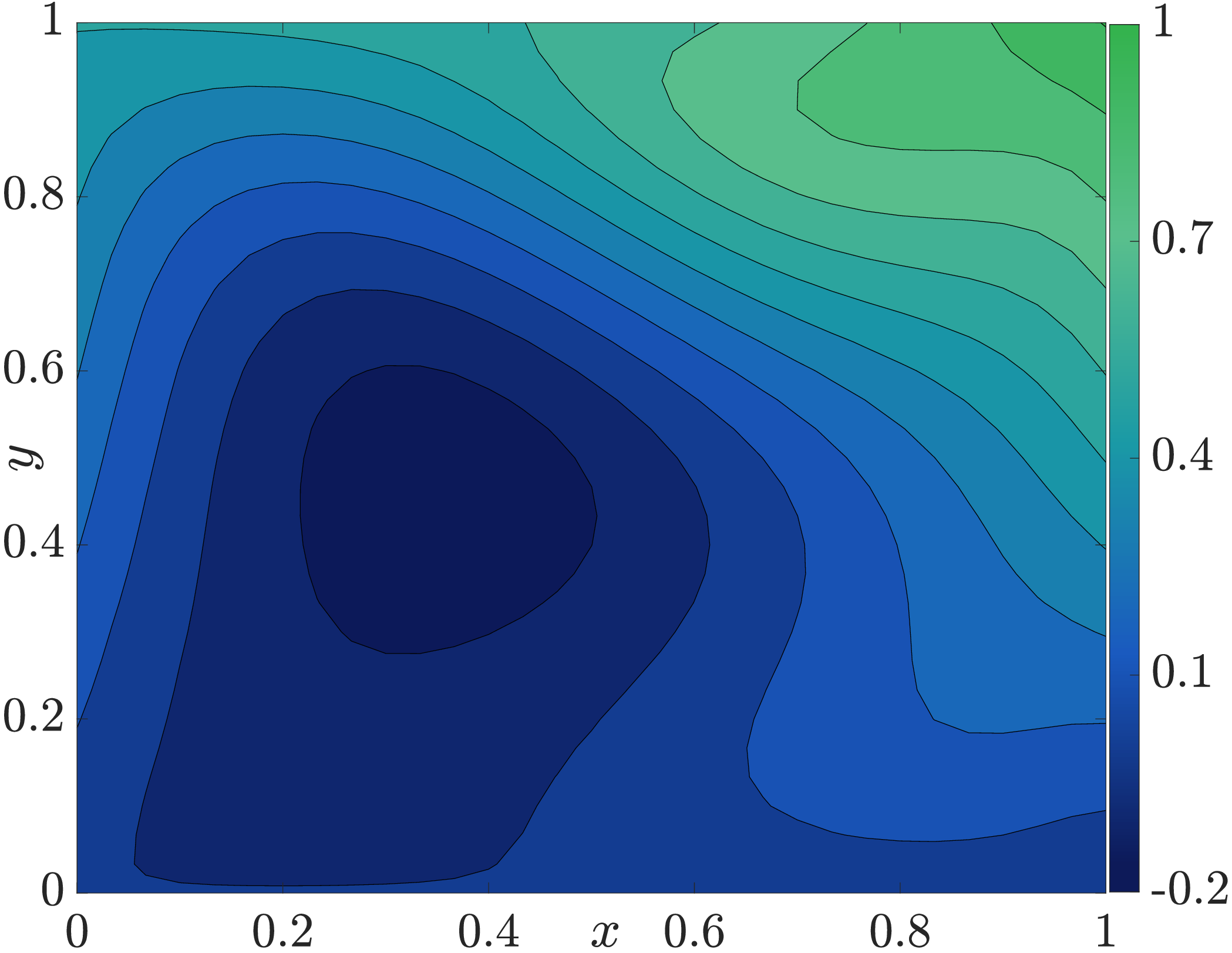}
\end{minipage}%
\begin{minipage}{0.333\textwidth} 
\includegraphics[scale= 0.14]{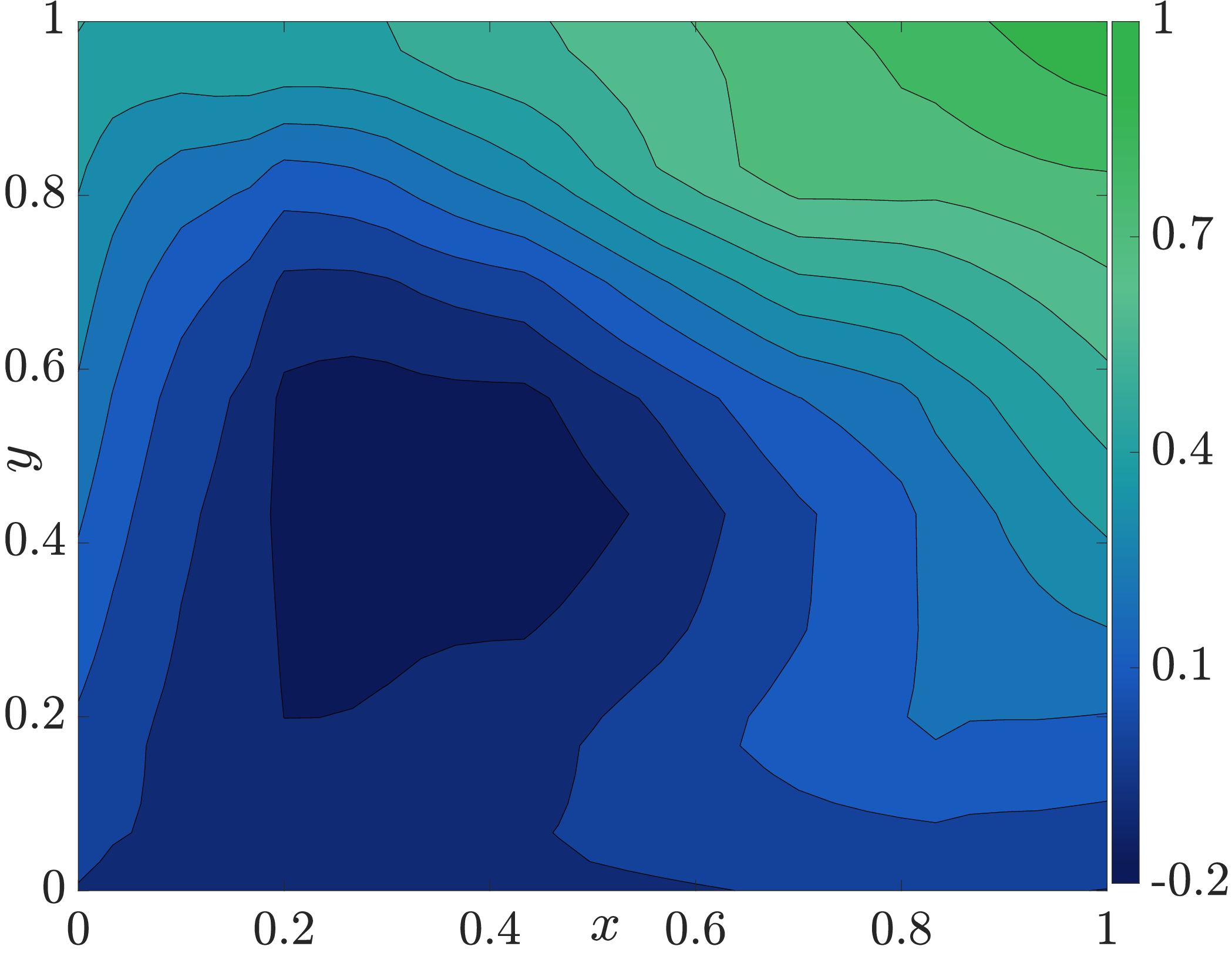}
\end{minipage}
\begin{minipage}{0.333\textwidth}  
\includegraphics[scale= 0.14]{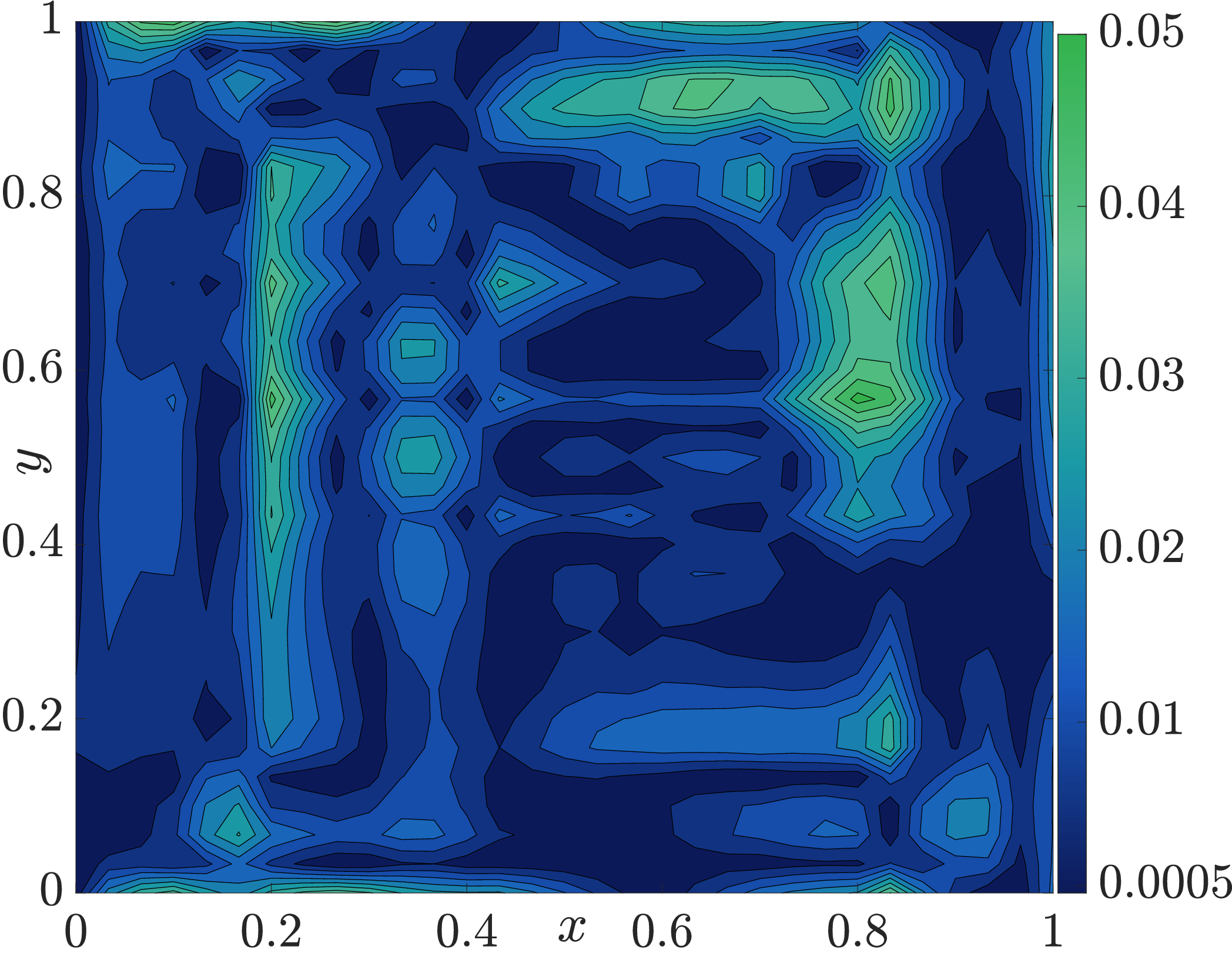}
\end{minipage}
\caption{\footnotesize [Heat Equation] SON performance: (Left) Reference solution at the final time; (Middle) SON predicted sample mean at the final time.(Right) SON prediction errors.}
\label{Single_2DHeat_Mean}
\end{figure}

\begin{figure}[h!]
\begin{minipage}{0.333\textwidth} 
\includegraphics[scale= 0.14]{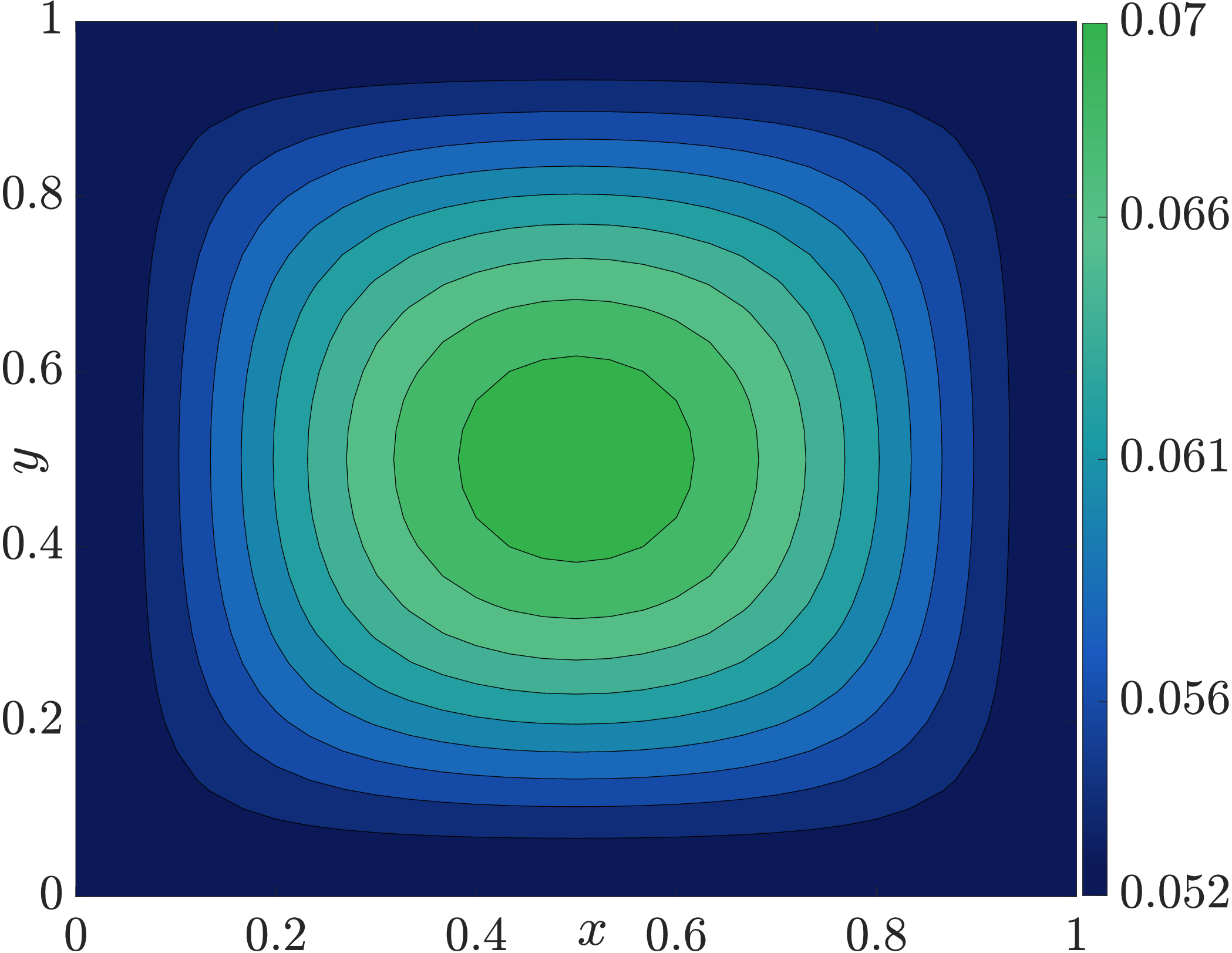}
\end{minipage}%
\begin{minipage}{0.333\textwidth} 
\includegraphics[scale= 0.14]{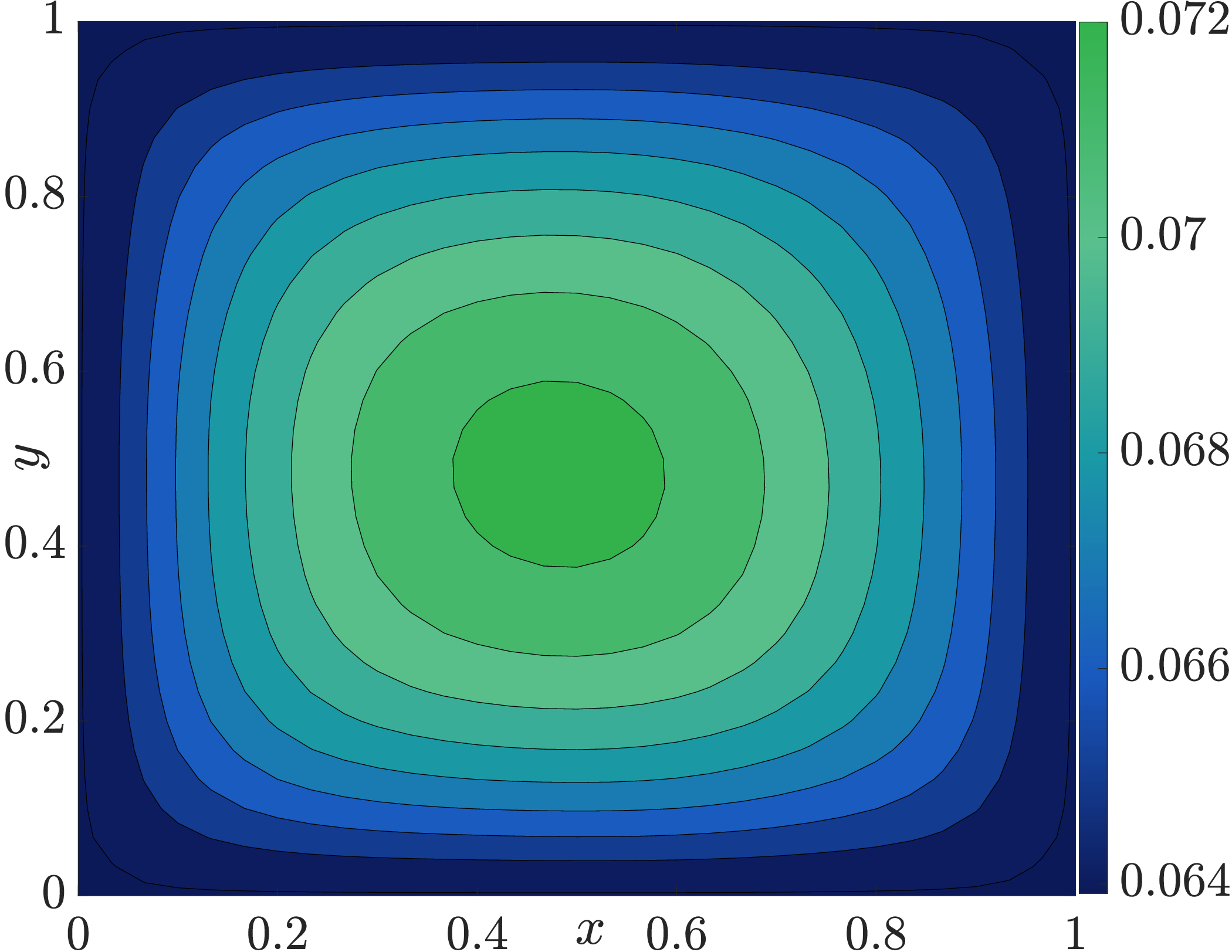}
\end{minipage}
\begin{minipage}{0.333\textwidth}  
\includegraphics[scale= 0.14]{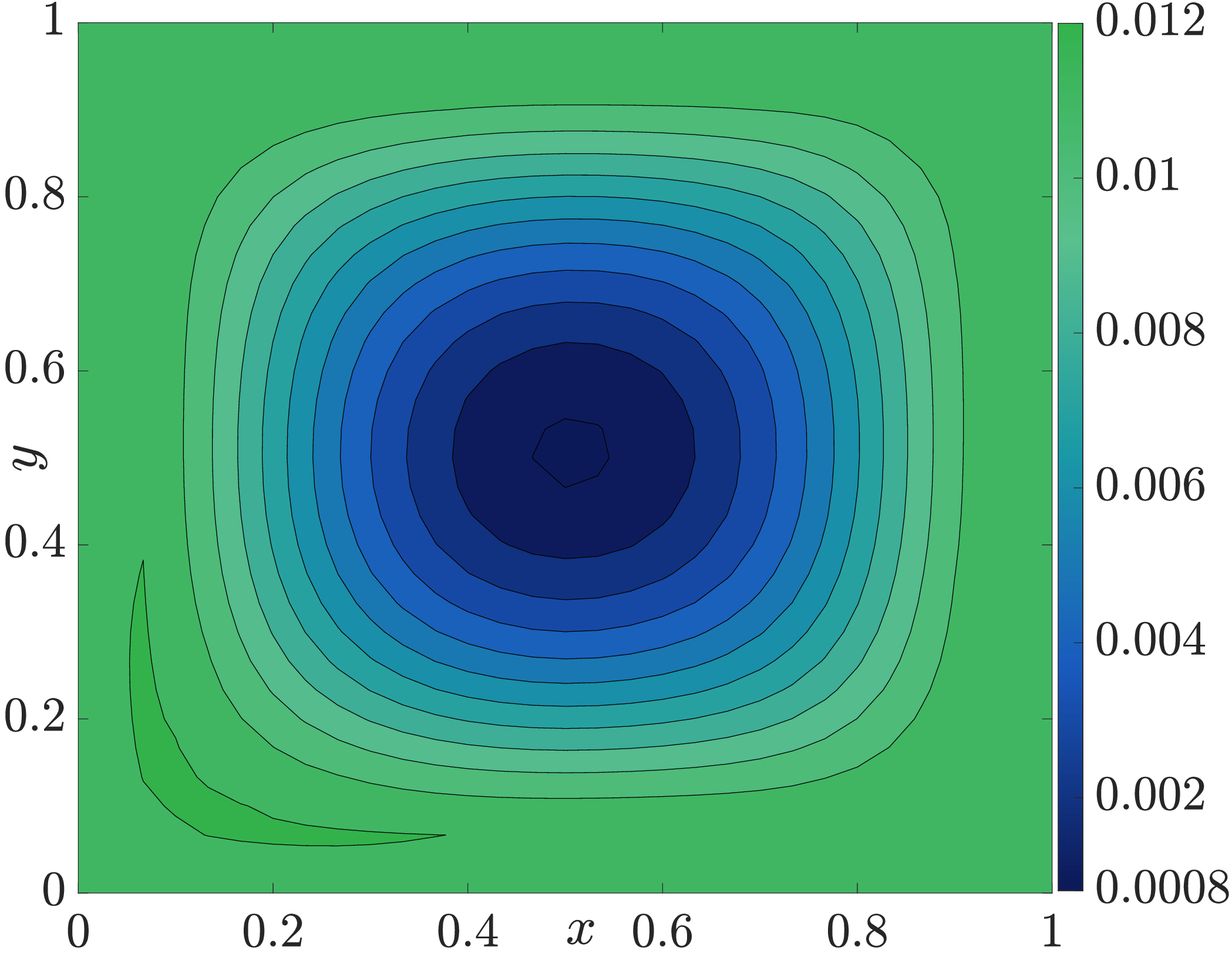}
\end{minipage}
\caption{\footnotesize [Heat Equation] SON performance:  (Left) Reference std at the final time. (Middle) SON output sample std at the final time.(Right) SON std prediction error.}
\label{Single_2DHeat_Std}
\end{figure}

We then focus on the predictive performance of the SON by selecting another forcing term from the testing dataset as a representative input. Specifically, we plot the 2D heatmaps of the reference sample mean and the corresponding SON-predicted sample mean at the final time $T$, along with the estimation error heatmap in Figure~\ref{Single_2DHeat_Mean}. To further assess SON’s capability in uncertainty quantification, we also present the heatmaps of the reference and predicted std, as well as their estimation error, in Figure~\ref{Single_2DHeat_Std}. From these figures, we observe that SON provides accurate predictions of the solution mean as well as reliable uncertainty estimates.

%From the first panel of Figure~\ref{Single_2DHeat}, the two mean solutions agree closely. This is further supported by the error heatmap in the second panel of Figure~\ref{Single_2DHeat}, whose maximum value is approximately $0.08$. Finally, from the last panel of Figure~\ref{Single_2DHeat}, we observe that the predicted standard deviation remains close to the reference standard deviation at the final time.

\begin{figure}[h!]
\vspace{-0.3cm}
\begin{minipage}{0.248\textwidth}
\includegraphics[scale = 0.11]{Figures/2DHeat/Update/Retraining/Input1/Col1_Time1_Input1.eps}
\end{minipage}%
\begin{minipage}{0.248\textwidth}
\includegraphics[scale = 0.11]{Figures/2DHeat/Update/Retraining/Input1/Col1_Time10_Input1.eps}
\end{minipage}%
\begin{minipage}{0.248\textwidth}
\includegraphics[scale = 0.11]{Figures/2DHeat/Update/Retraining/Input1/Col1_Time20_Input1.eps}
\end{minipage}%
\begin{minipage}{0.248\textwidth}
\includegraphics[scale = 0.11]{Figures/2DHeat/Update/Retraining/Input1/Col1_Time30_Input1.eps}
\end{minipage}

\begin{minipage}{0.248\textwidth}
\includegraphics[scale = 0.11]{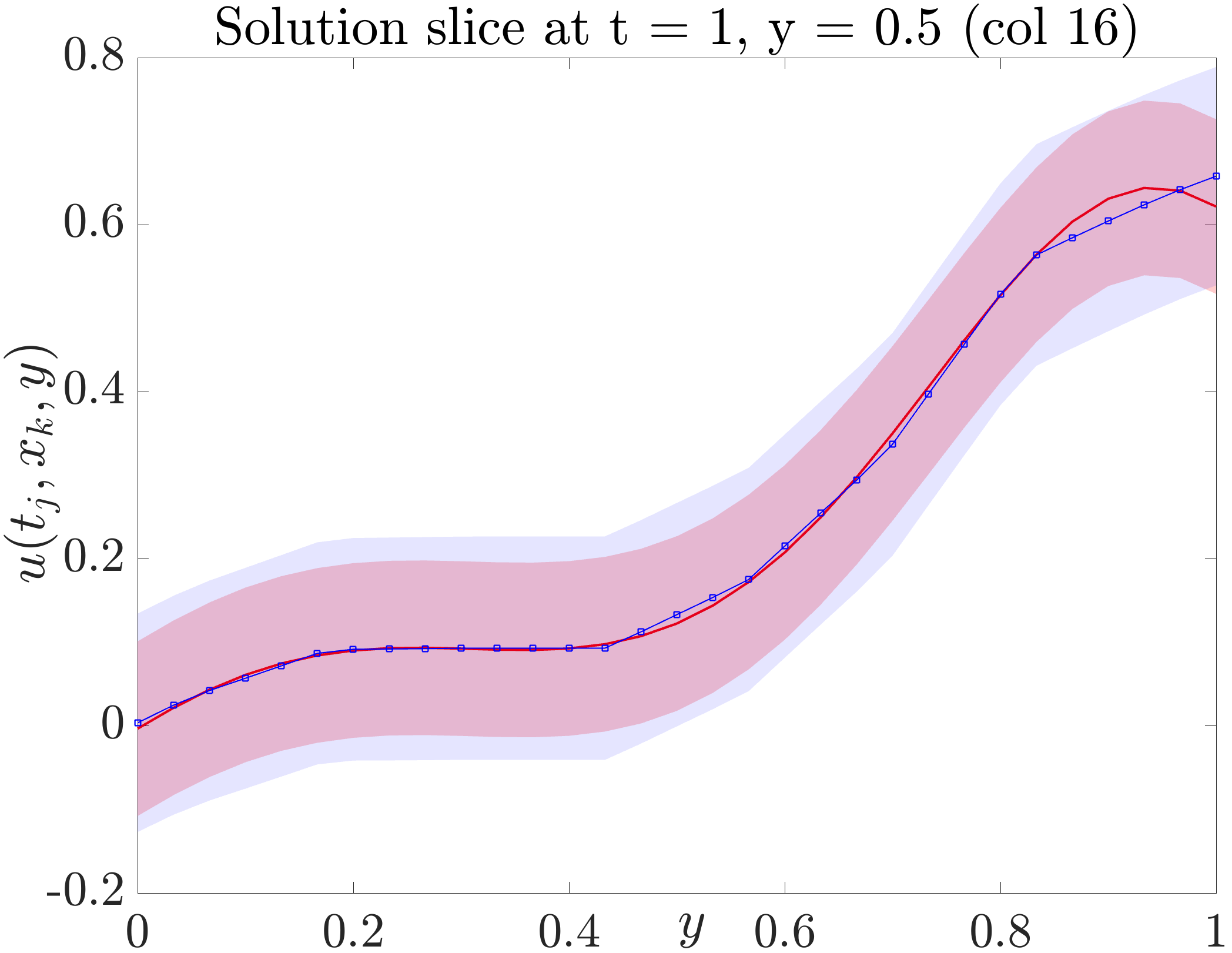}
\end{minipage}%
\begin{minipage}{0.248\textwidth}
\includegraphics[scale = 0.11]{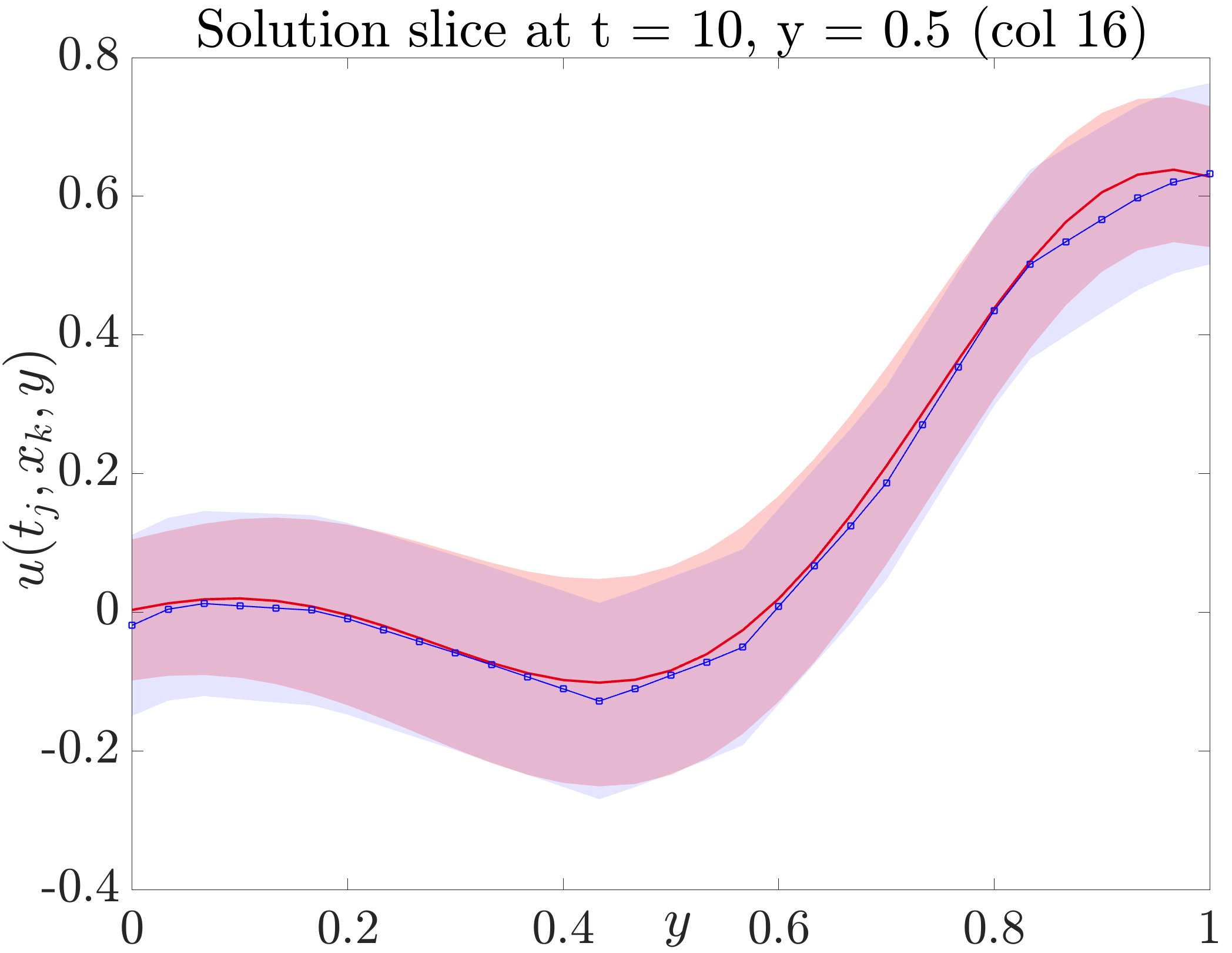}
\end{minipage}%
\begin{minipage}{0.248\textwidth}
\includegraphics[scale = 0.11]{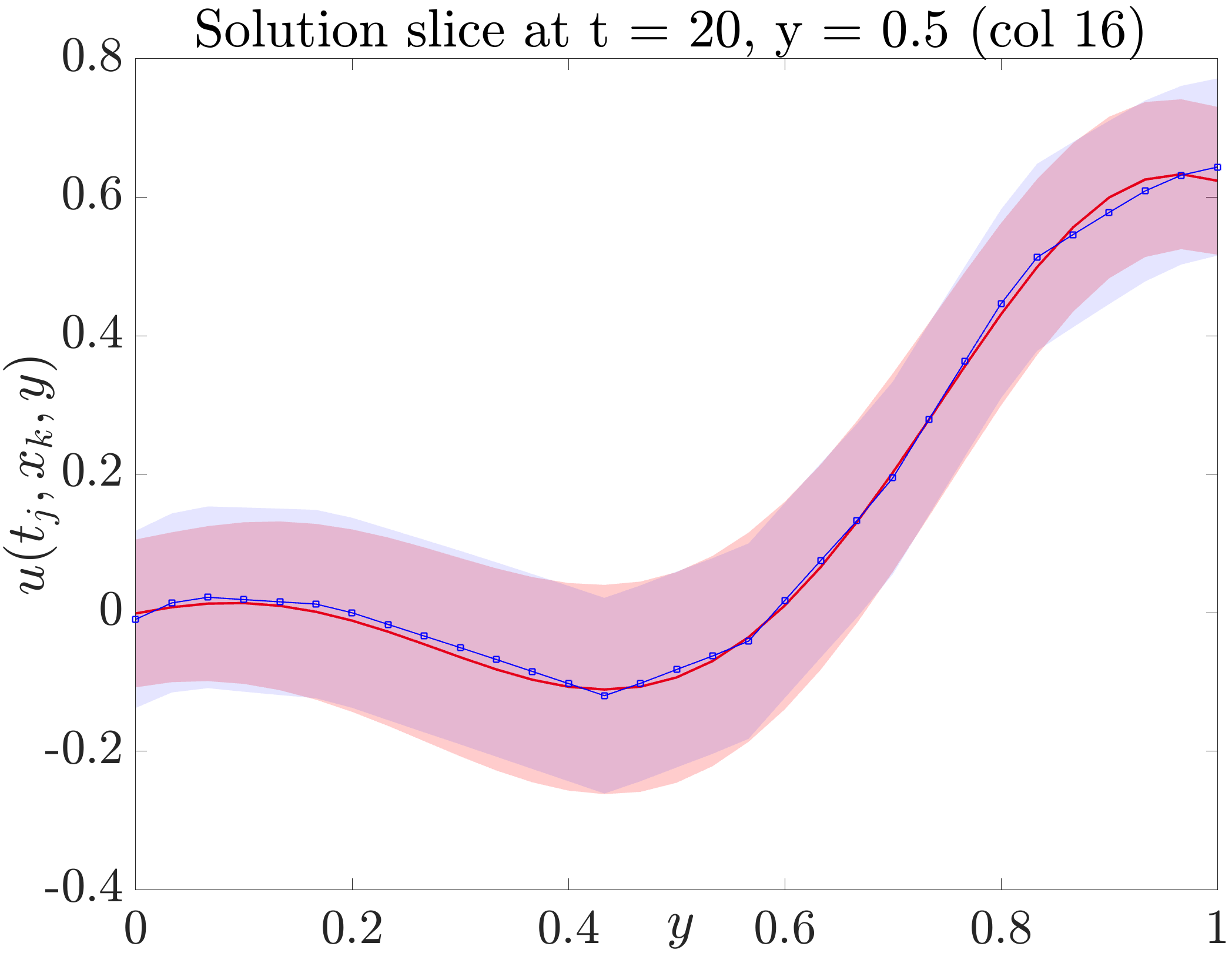}
\end{minipage}%
\begin{minipage}{0.248\textwidth}
\includegraphics[scale = 0.11]{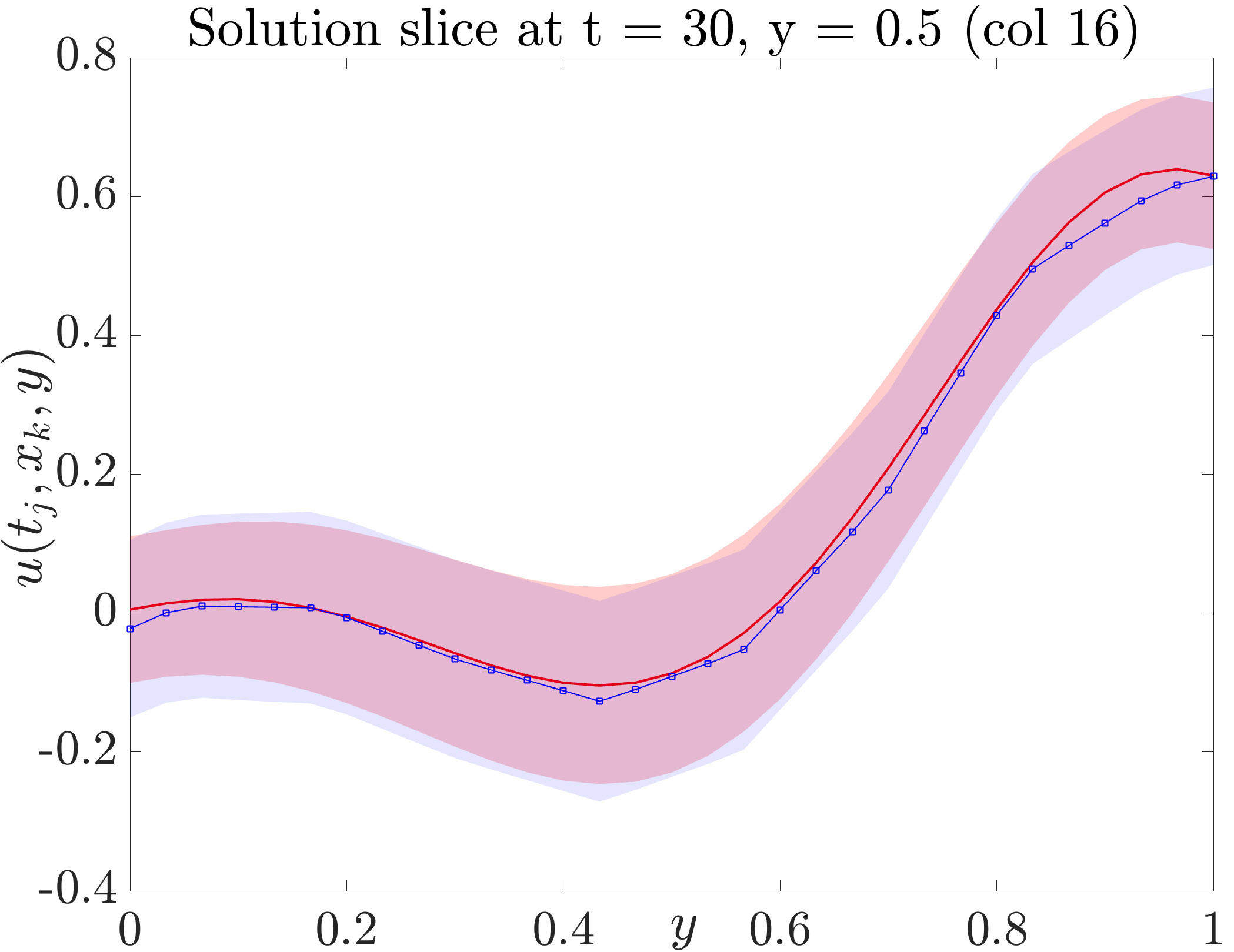}
\end{minipage}

\begin{minipage}{0.248\textwidth}
\includegraphics[scale = 0.11]{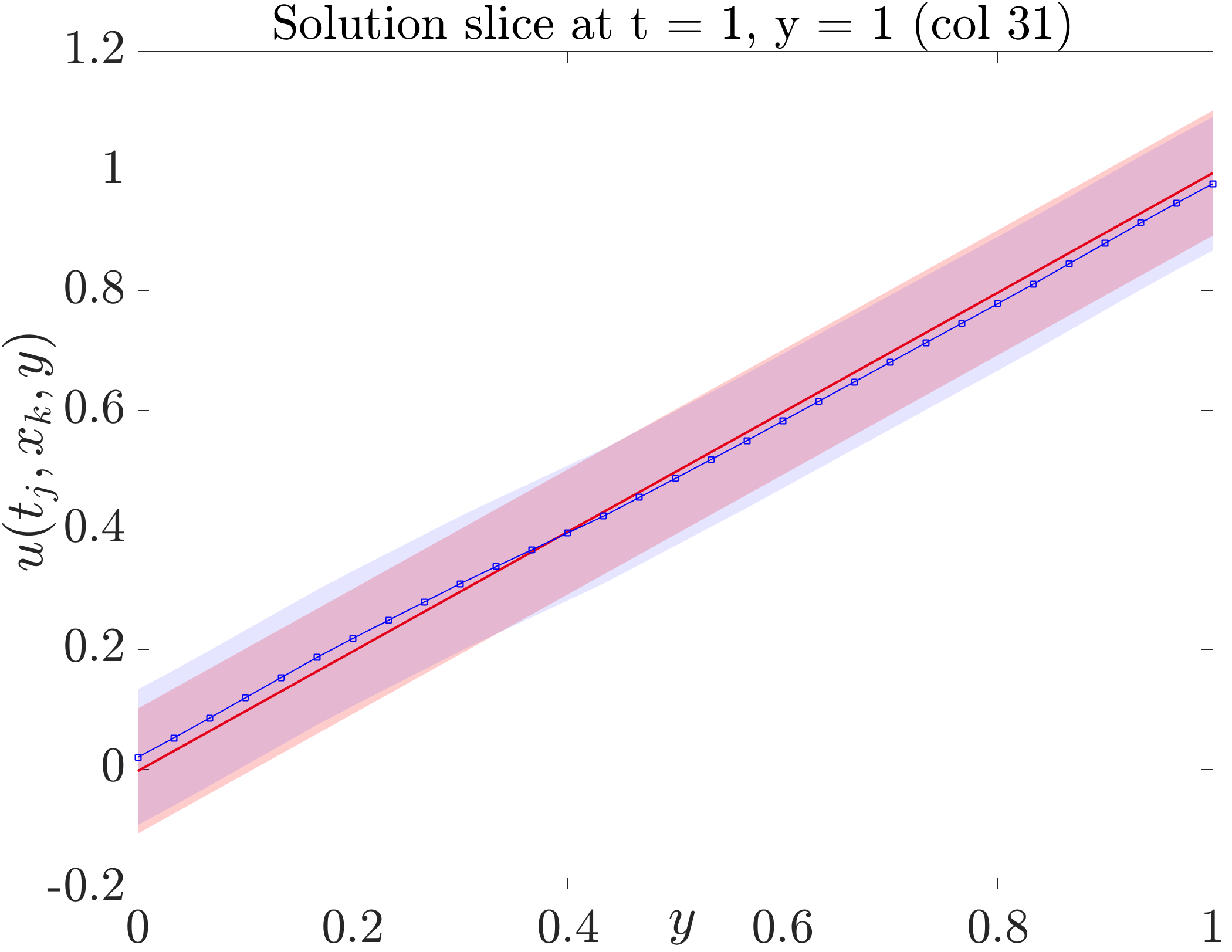}
\end{minipage}%
\begin{minipage}{0.248\textwidth}
\includegraphics[scale = 0.11]{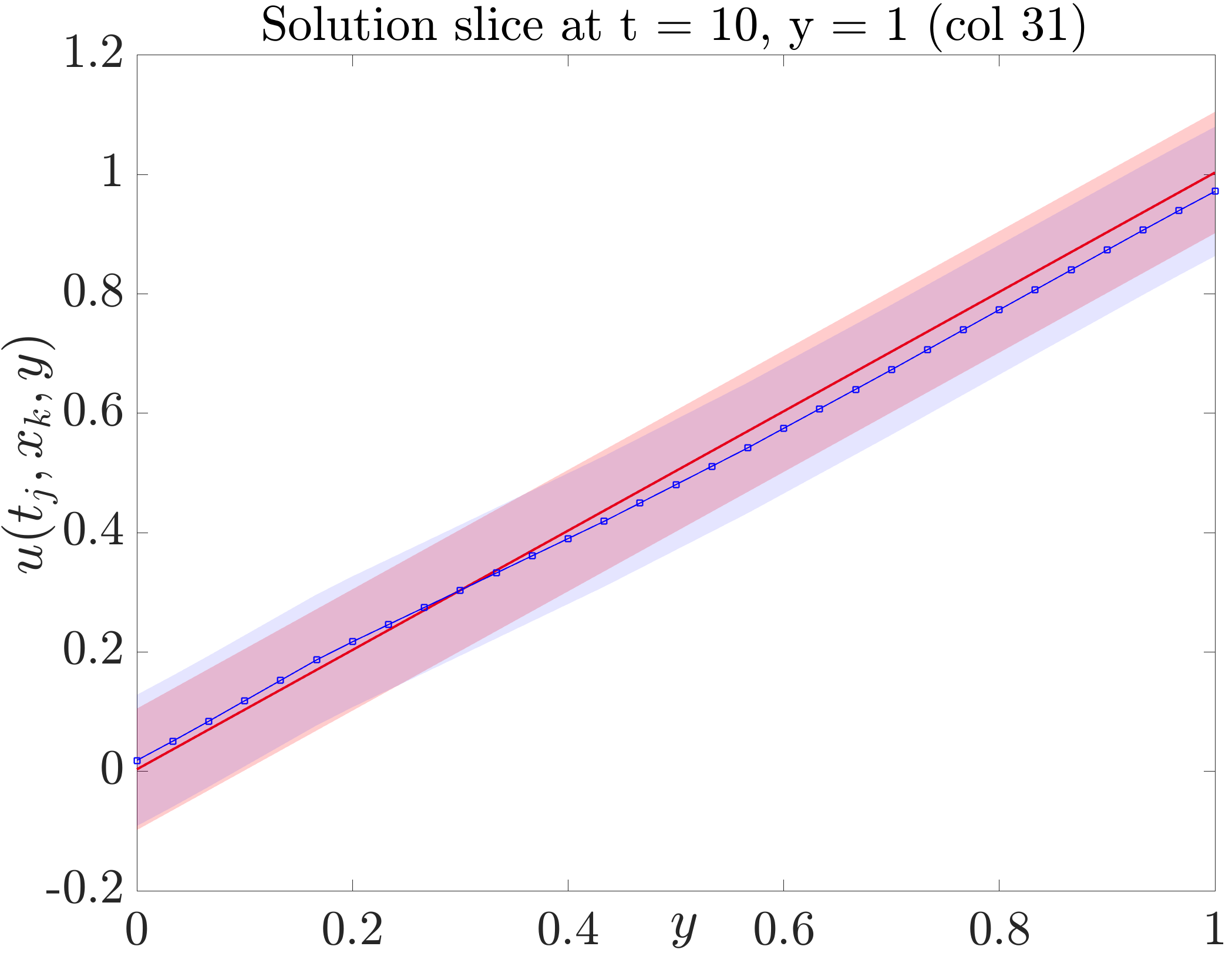}
\end{minipage}%
\begin{minipage}{0.248\textwidth}
\includegraphics[scale = 0.11]{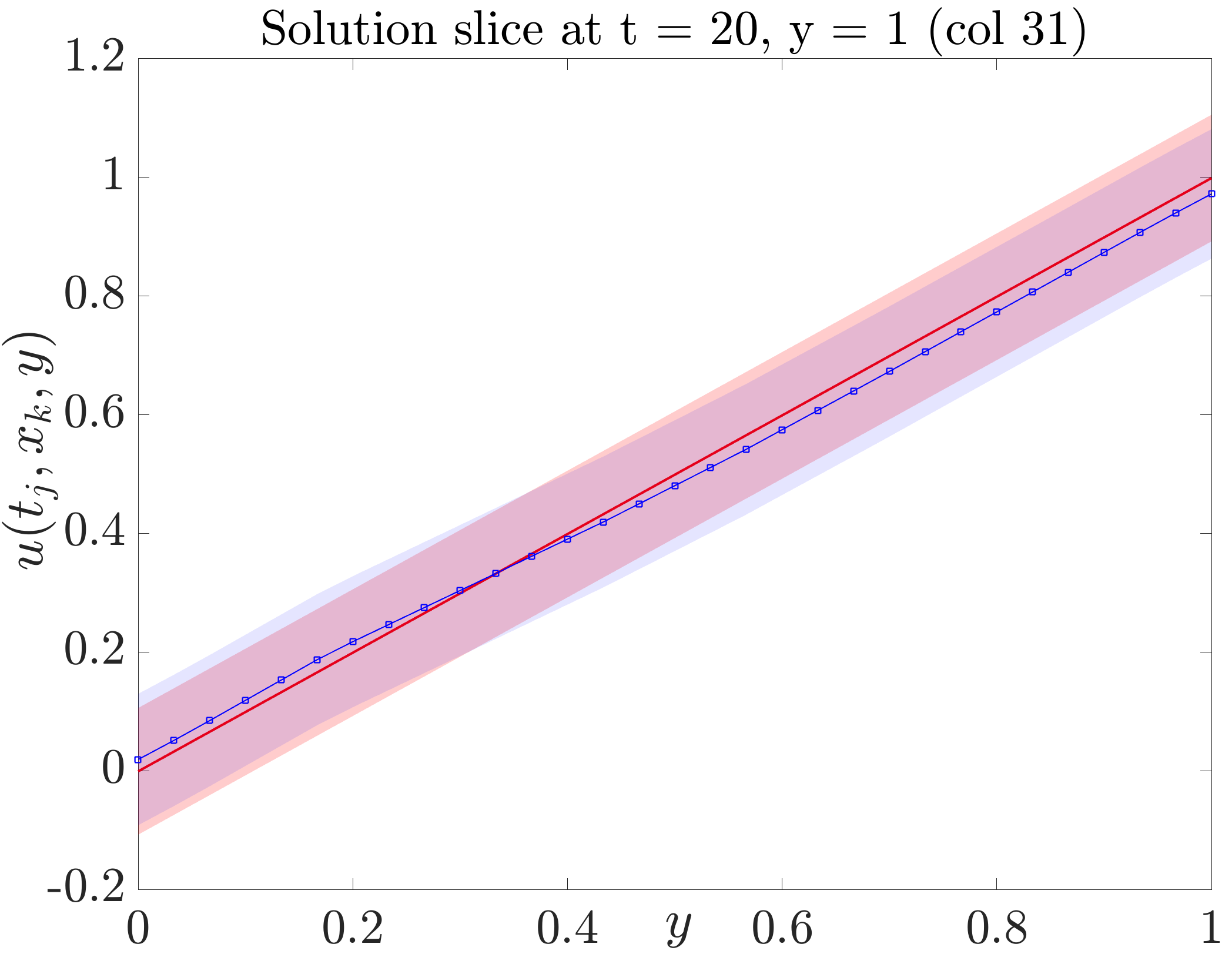}
\end{minipage}%
\begin{minipage}{0.248\textwidth}
\includegraphics[scale = 0.11]{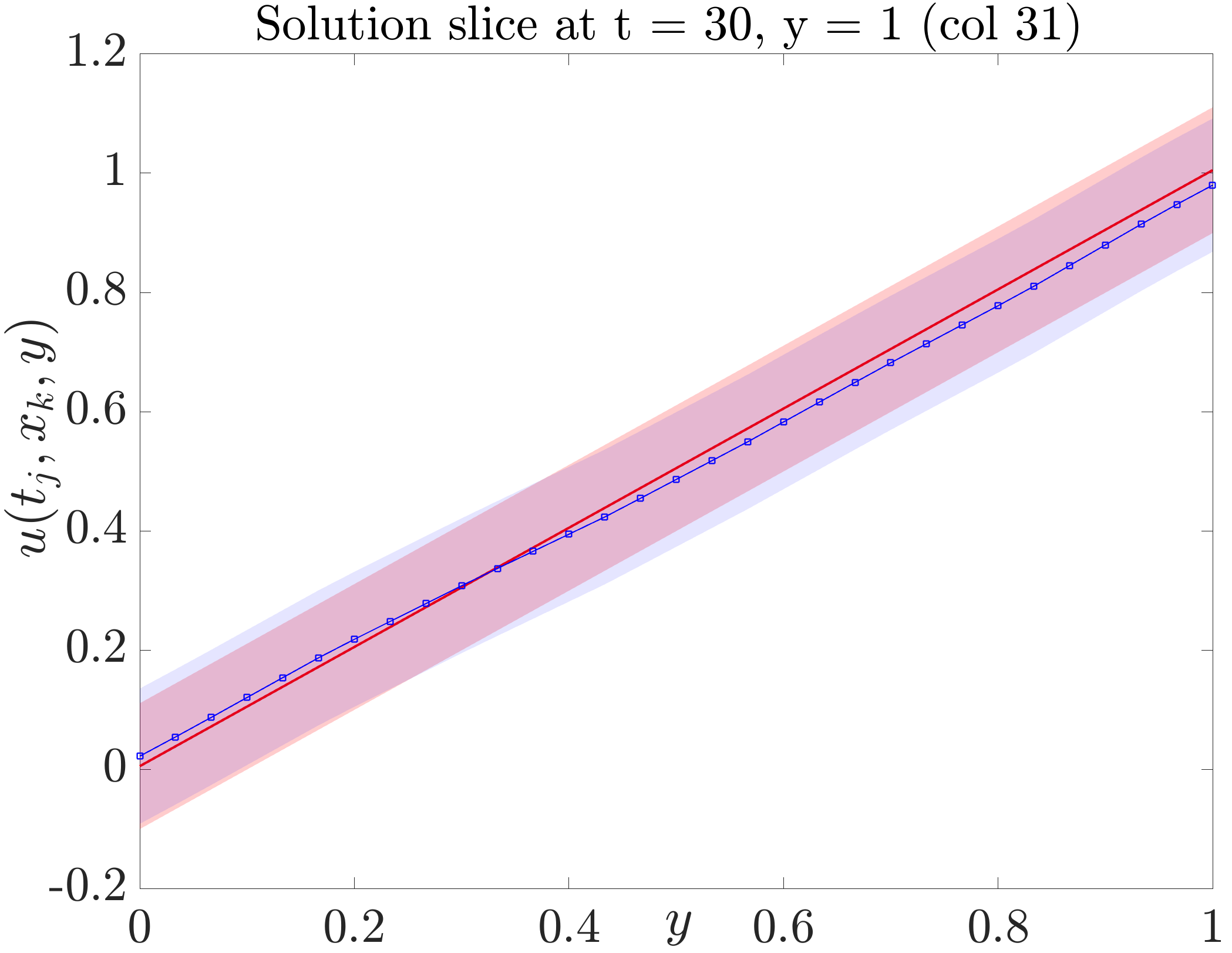}
\end{minipage}
\caption{\footnotesize [Heat Equation] Cross-sections of the SON predicted solutionand its associated uncertainty at time steps $t_m$ for $m=1,10,20,30$: (Top row) Column 1; (Middle row) Column 16; (Bottom row) Column 31.}
\label{2DHeat_CrossSections}
\end{figure}

To assess the temporal evolution of SON predictions, we plot cross-sections of the reference and SON-predicted solutions at time steps $m=1,10,20,30$ along several spatial columns. These results, shown in Figure~\ref{2DHeat_CrossSections}, demonstrate that the SON accurately captures both the overall solution profile and its evolution over time.

%Finally, we evaluate the robustness of the method across different testing input forcings. We repeat the same procedure over 8 distinct testing inputs and compute the averages of the corresponding error metrics. The averaged mean-error heatmaps are shown in Figure~\ref{Average_MeanErr_2DHeat}, while the averaged standard-deviation error heatmaps are presented in Figure~\ref{Average_StdErr_2DHeat}. Overall, these results demonstrate consistent predictive accuracy and stable uncertainty estimates across multiple inputs.

% \begin{figure}[h!]
% \begin{minipage}{0.333\textwidth} 
% \includegraphics[scale= 0.14]{Figures/2DHeat/Update/Retraining/Input1/Mean_FinalTime_Input1.eps}
% \end{minipage}%
% \begin{minipage}{0.333\textwidth} 
% \includegraphics[scale= 0.14]{Figures/2DHeat/Update/Retraining/Input1/ErrorMean_Time30_Input1.eps}
% \end{minipage}
% %
% \begin{minipage}{0.333\textwidth}  
% \includegraphics[scale= 0.14]{Figures/2DHeat/Update/Retraining/Input1/ErrorStd_Time30_Input1_v2.eps}
% \end{minipage}
% \caption{\footnotesize [2D Heat Equation] Single-Input Comparison: (First) Mean at Final Time (Ref. vs Pred.). (Second) Absolute Mean Error at Final Time.(Third) Absolute Std Error at Final Time.}
% \label{Single_2DHeat}
% \end{figure}

\subsection{Burgers' equation}
In this numerical example, we consider a more challenging problem: the 2D Burgers' equation,
\begin{equation}
\label{2DBurgers}
\begin{array}{l}
u_t + \left(\dfrac{u^2}{2}\right)_{x} + \left(\dfrac{u^2}{2}\right)_{y} = 0, \; (x, y, t) \in (-1, 1)^2 \times (0, T), \vspace{0.1cm} \\
u(x, y, 0) = u_0(x, y), \; x \in (-1, 1)^2,
\end{array}
\end{equation}
subject to the periodic boundary conditions, where $T =0.08$ and $u_0$ is the initial condition which is sampled from a 2D Gaussian process with a periodic radial basis function kernel. We consider two scenarios for incorporating uncertainty: Case 1, where time-dependent but space-independent noise is added to the solution at each time step, and Case 2, where the noise is space–time dependent. We emphasize that the nonlinear flux in the Burgers equation couples the perturbations with the state, so that even additive noise leads to solution-dependent stochastic effect. As a result, the stochastic solution may develop spatial correlations in both cases.

To generate the reference solution samples, we solve Eq.~\eqref{2DBurgers} by using a discontinuous Galerkin (DG) method~\cite{Cockburn1991a} combined with a second order strong-stability-preserving Runge–Kutta (SSP–RK2) schemes~\cite{Shu1998a} on a uniform grid over $[-1,1]^2$ with $H=40$ intervals in each spatial direction (so $h=2/H=1/20$) and time step $\Delta t = T/M$ with $M=40$. At each time level $t_m$, we perturb the numerical solution by adding a random perturbation $\xi_m$ before advancing to $t_{m+1}$.
As no forcing term is present in Eq.~\eqref{2DBurgers}, we treat the initial condition $u_0$ as the input to the solution operator. We generate initial conditions by sampling a two-dimensional Gaussian random field with a periodic kernel. For both cases, we use $1500$ training samples and $300$ testing samples. Since the stochastic solution may exhibit spatial correlations, we set the dimension of the driving Gaussian random variable to $r=12$ in Case~1 and $r=16$ in Case~2. Consequently, the diffusion term in Eq.~\eqref{SDE_A} is a tensor-valued function in $\mathbb{R}^{M\times H\times H\times r}$ given by $\pmb{\sigma}\!\left(\mathrm{A}, \pmb{\theta}_{g}\right)
:= \bigl[\pmb{\sigma}_1\!\left(\mathrm{A}, \pmb{\theta}_{g,1}\right)\ \cdots\ \pmb{\sigma}_r\!\left(\mathrm{A}, \pmb{\theta}_{g,r}\right)\bigr],$
% \begin{equation}
% \label{Time_Diffusion_Tensor_Combined}
% \pmb{\sigma}\!\left(\mathrm{A}, \pmb{\theta}_{g}\right)
% := \bigl[\pmb{\sigma}_1\!\left(\mathrm{A}, \pmb{\theta}_{g,1}\right)\ \cdots\ \pmb{\sigma}_r\!\left(\mathrm{A}, \pmb{\theta}_{g,r}\right)\bigr],
% \end{equation}
where, for each $k=1,\dots,r$, the $k$-th channel $\pmb{\sigma}_k\!\left(\mathrm{A}, \pmb{\theta}_{g,k}\right)$ consists of the entries $\sigma_{k,m}\!\left(\mathrm{A}, \theta_{g,k,m}\right)
= \sum_{l=1}^{L_g} b_{l,k,m}\,\mu\!\left(c_{l,k,m}\mathrm{A}\right)$,
$m=1,\dots,M,$
% \begin{equation}
% \label{Time_Diffusion_Tensor_Single}
% \sigma_{k,m}\!\left(\mathrm{A}, \theta_{g,k,m}\right)
% = \sum_{l=1}^{L_g} b_{l,k,m}\,\mu\!\left(c_{l,k,m}\mathrm{A}\right),
% \qquad m=1,\dots,M,
% \end{equation}
where $\theta_{g,k,m}$ denotes the collection of coefficients at time level $t_m$ in the $k$-th channel, i.e.,$\theta_{g,k,m} := \Bigl(\{b_{l,k,m}\}_{l=1}^{L_g},\ \{c_{l,k,m}\}_{l=1}^{L_g}\Bigr)$, $m=1,\dots,M$.
Accordingly, we define
$\pmb{\theta}_{g,k}:=\{\theta_{g,k,m}\}_{m=1}^{M}$,
and
$\pmb{\theta}_{g}:=\{\pmb{\theta}_{g,k}\}_{k=1}^{r}$.

Unlike the Heat Equation example where the injected noise is effectively damped at each time step by diffusion, in Eq.~\eqref{2DBurgers} the uncertainty propagates through the nonlinear dynamics and persists in subsequent time steps. To reflect the resulting growth in uncertainty, we enforce the sequences $\{b_{l,k,m}\}_{m=1}^{M}$ and $\{c_{l,k,m}\}_{m=1}^{M}$ to be increasing in $m$ for all $l=1,\dots,L_g$ and $k=1,\dots,r$. Finally, to avoid overly large or overly small diffusion magnitudes, we generate $r$ intervals $\{[\alpha_k,\beta_k]\}_{k=1}^{r}$, where $\alpha_k\sim\mathcal{U}([0.03,0.05])$ and $\beta_k\sim\mathcal{U}([0.1,0.2])$.

\vspace{0.5em}
%As in the previous section, we observe that the Phase-II stochastic model refines the Phase-I prediction near the boundaries and yields acceptable prediction bands. We therefore focus on the predictive performance of the Phase-II model and omit the Phase-I results for brevity.
Since we have demonstrated the superior performance of SON over both the deterministic DeepONet and the stochastic B-DeepONet in previous examples, we focus here on evaluating SON on this more challenging stochastic Burgers' Equation problem.

\subsubsection{Case 1: Time-dependent, space-independent noise}

In Case~1, we let the uncertainty be a time-dependent random field. 
Let $u_{h,m}\in\mathbb{R}^{H\times H}$ denote the numerical solution on the spatial grid at time $t_m$. 
The initial condition for the next step is the perturbed solution
$ \tilde u^{\epsilon}_{h,m}=u_{h,m}+\gamma_m \epsilon_m\,\mathbf 1_{H\times H}, \epsilon_m\sim\mathcal N(0,1)
$, where $\gamma_m \sim \mathcal{U}([0.1,0.2])$. SON's Phase I is trained over $2000$ epochs, while Phase II is trained with $100$ epochs. %We use the AdamW optimizer for both phases with the same scheduler. 

\begin{figure}[h!]
\begin{minipage}{0.3333\textwidth} 
\includegraphics[scale= 0.16]{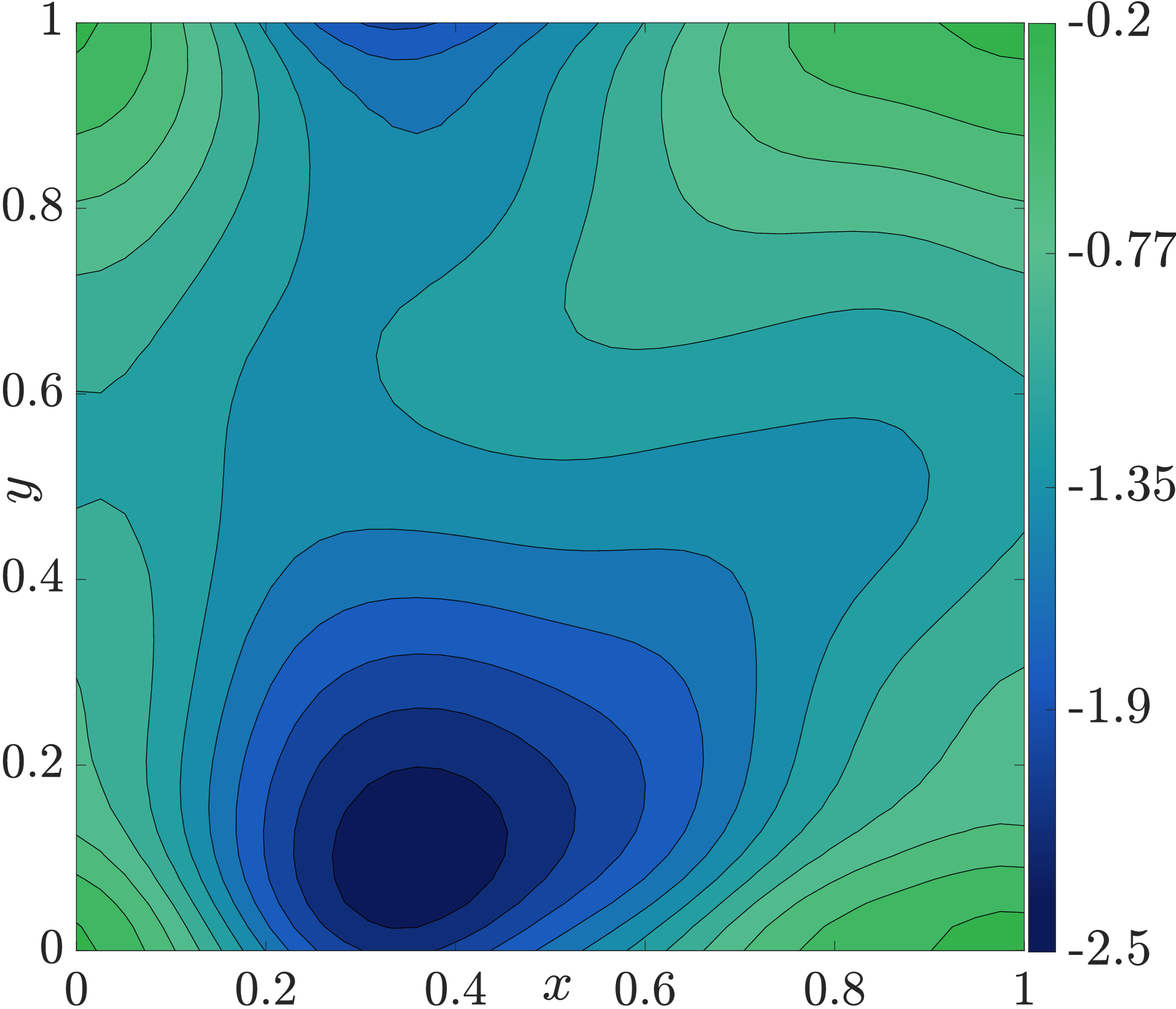}
\end{minipage}%
\begin{minipage}{0.3333\textwidth} 
\includegraphics[scale= 0.16]{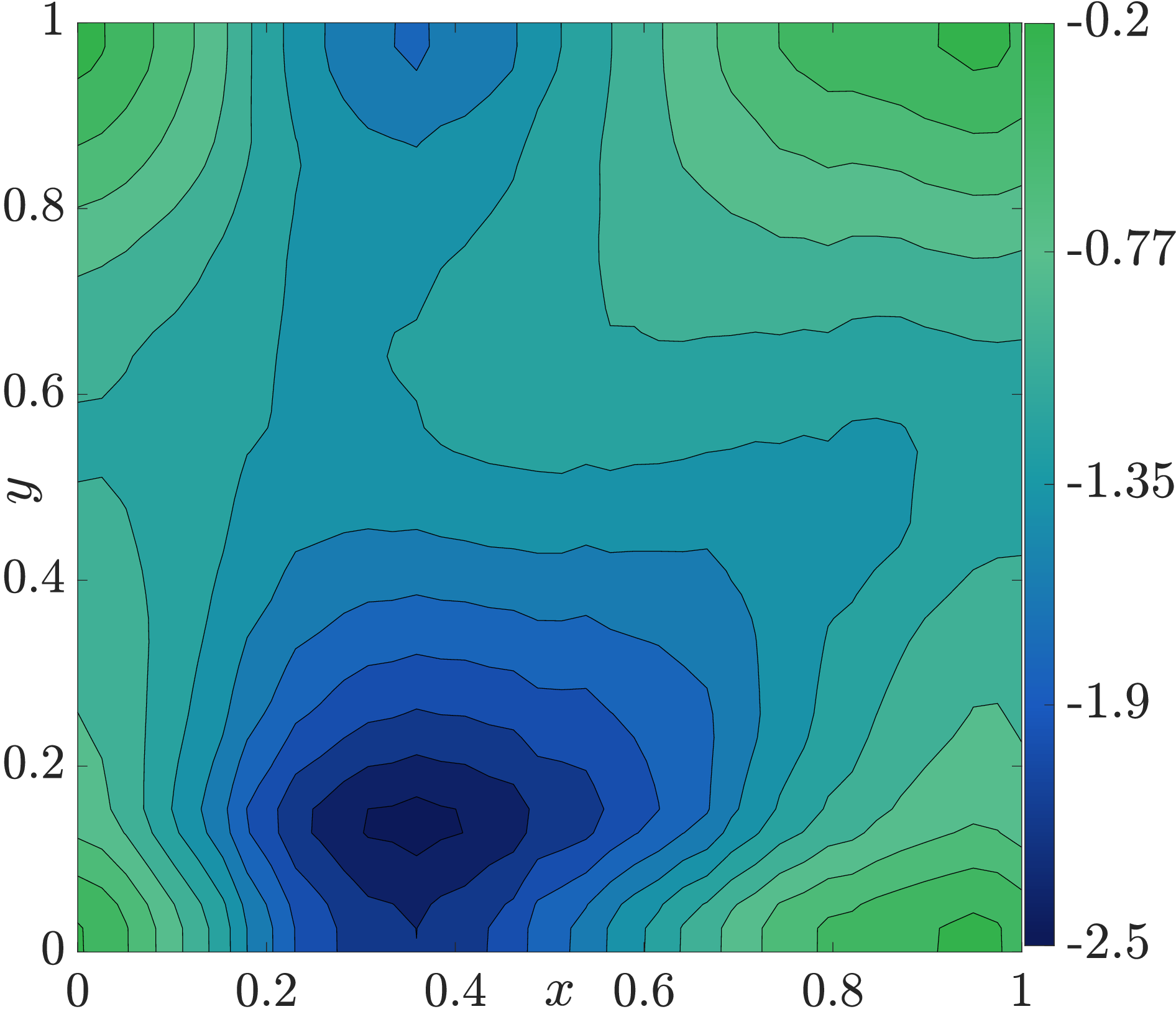}
\end{minipage}%
\begin{minipage}{0.3333\textwidth} 
\includegraphics[scale= 0.16]{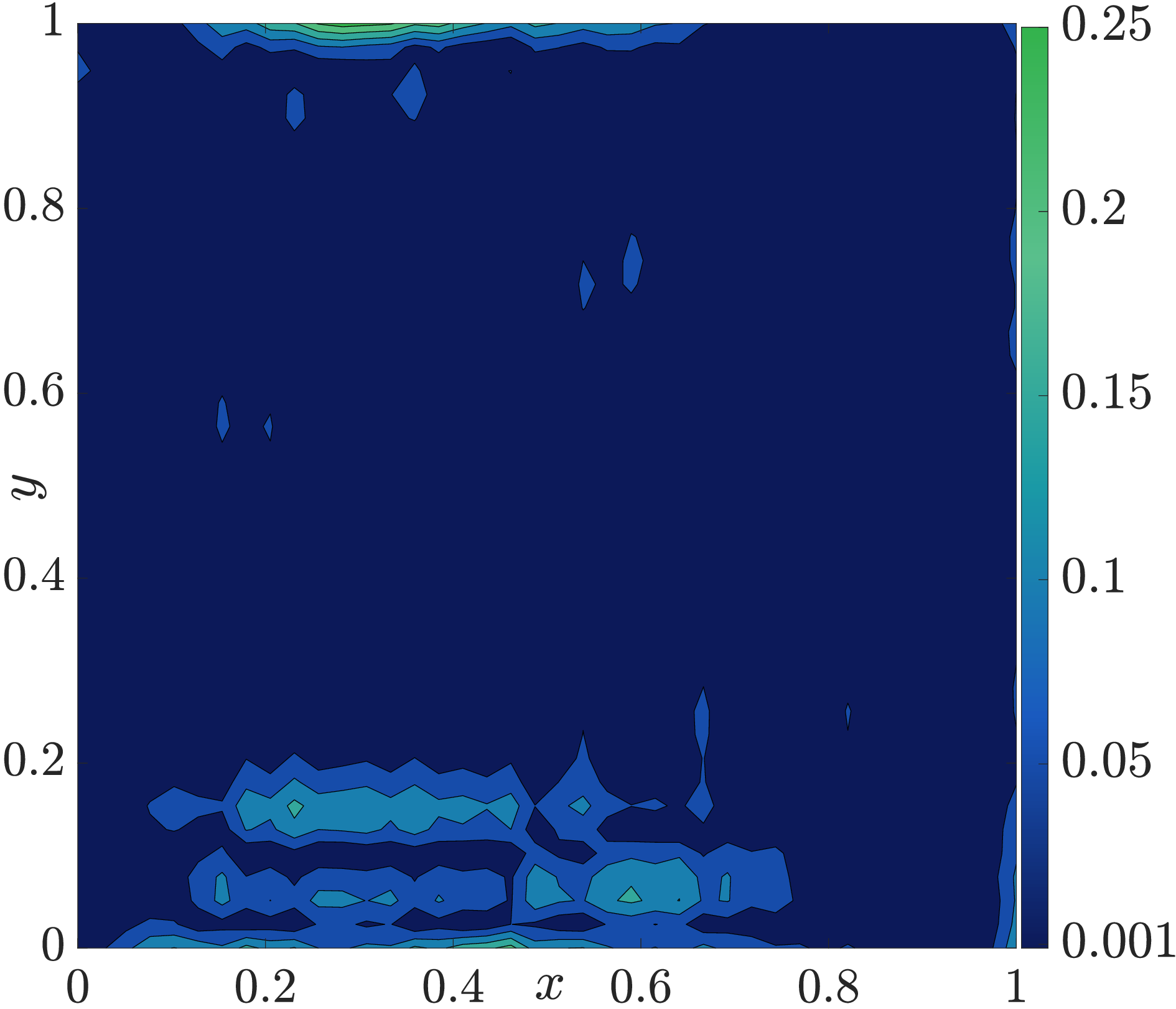}
\end{minipage}
\caption{\footnotesize [Burger Equation Case 1] Prediction mean at final time: (Left) Reference mean; (Middle) Prediction mean; (Right) Prediction error.}
\label{Single_2DBurger_Mean_Case1}
\end{figure}

\begin{figure}[h!]
\begin{minipage}{0.3333\textwidth} 
\includegraphics[scale= 0.16]{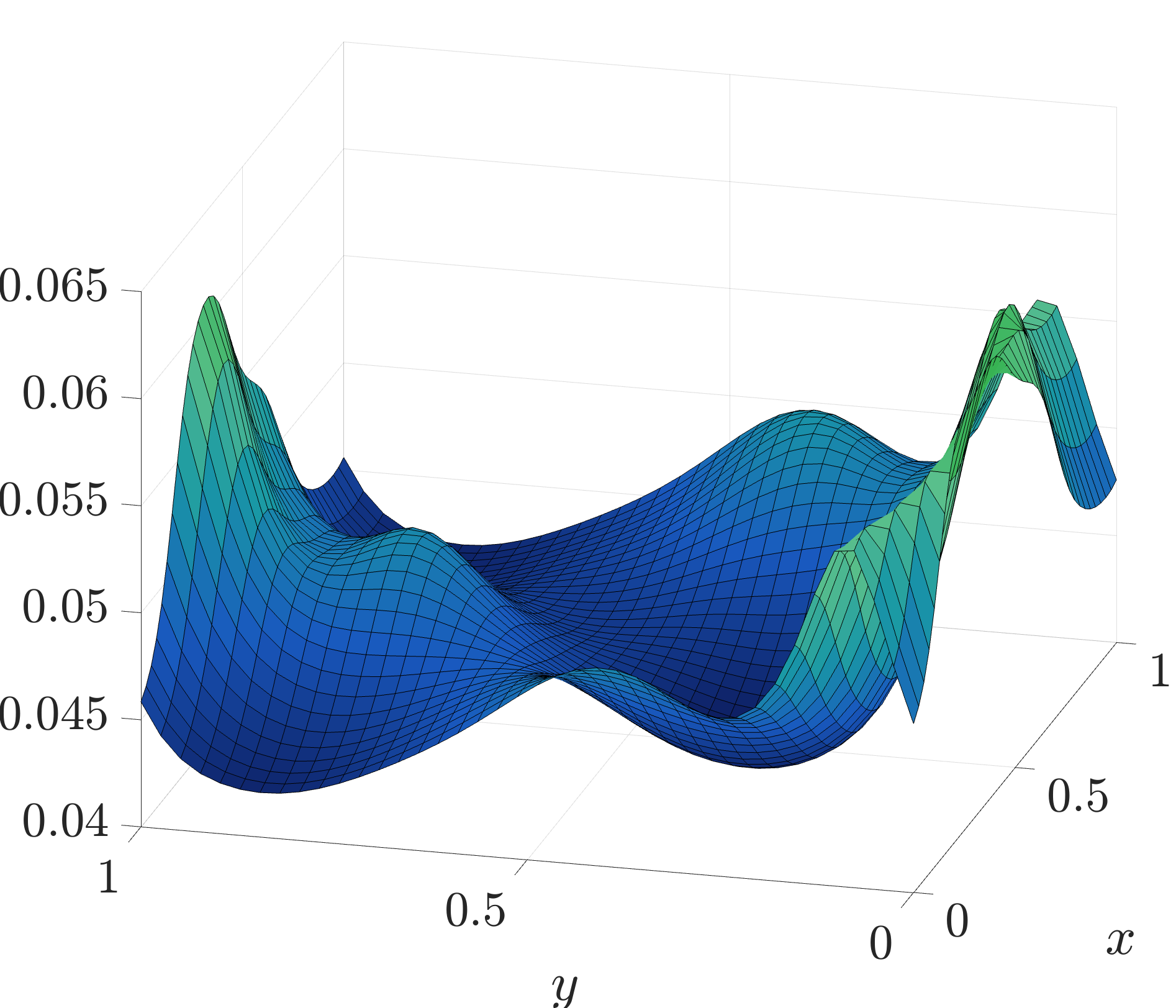}
\end{minipage}%
\begin{minipage}{0.3333\textwidth} 
\includegraphics[scale= 0.16]{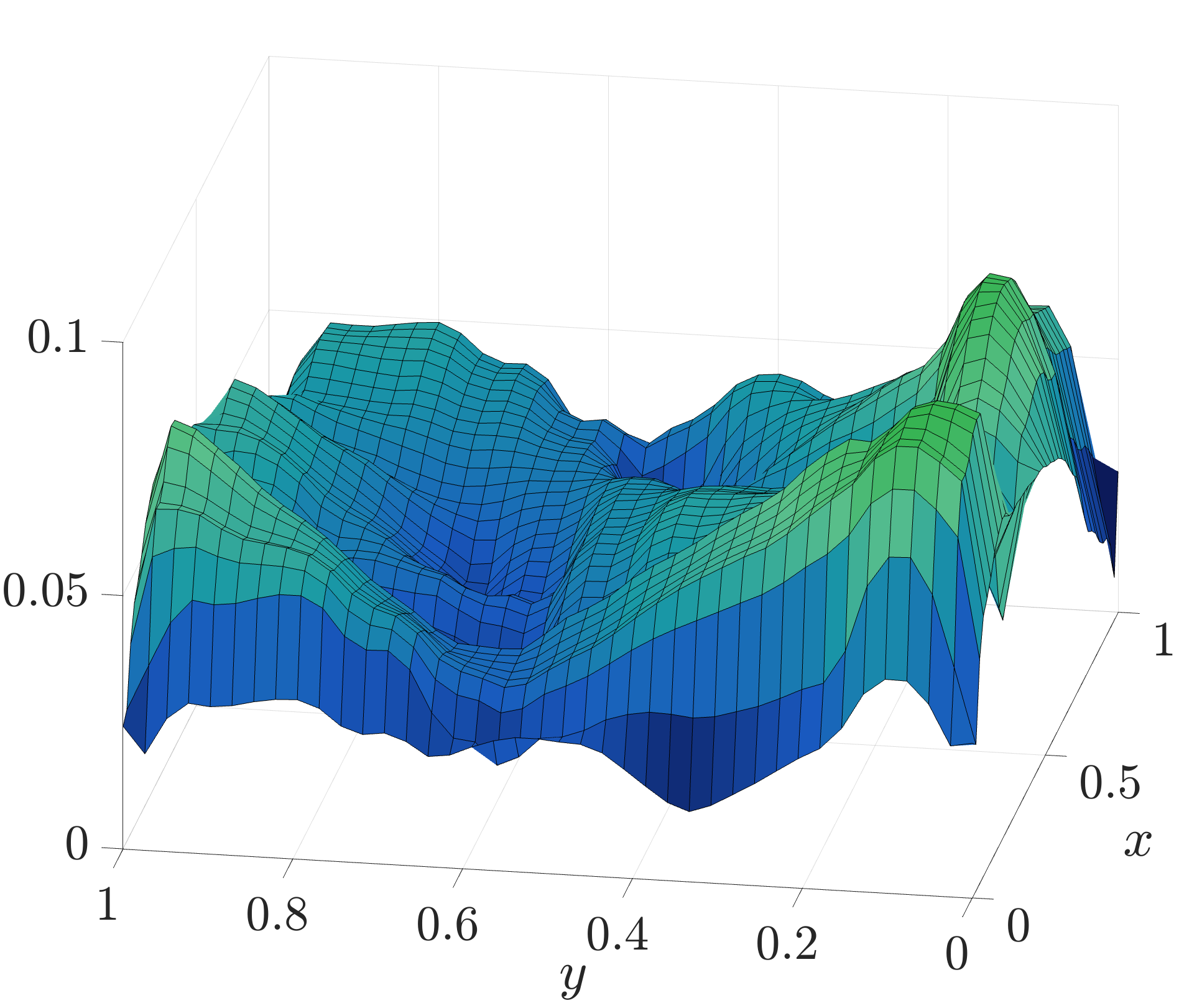}
\end{minipage}%
\begin{minipage}{0.3333\textwidth} 
\includegraphics[scale= 0.16]{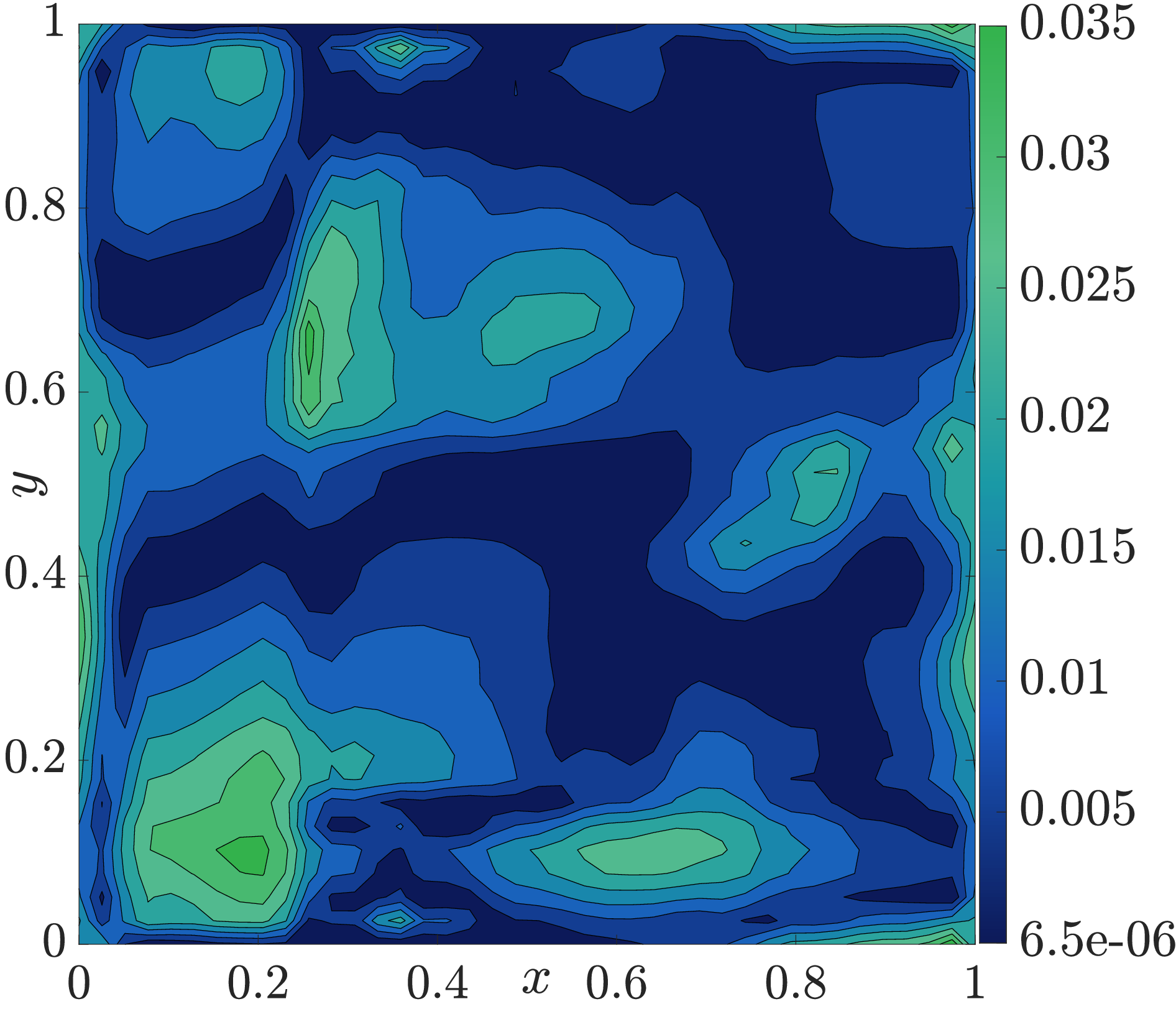}
\end{minipage}
% \begin{minipage}{0.3333\textwidth} 
% \includegraphics[scale= 0.16]{Figures/2DBurgers/TimeOnly/Update/Redraw/Prediction2/RefStd_3D_Nt40_Input2_v1.eps}
% \end{minipage}%
% \begin{minipage}{0.3333\textwidth} 
% \includegraphics[scale= 0.16]{Figures/2DBurgers/TimeOnly/Update/Redraw/Prediction2/PredStd_3D_Nt40_Input2_v2.eps}
% \end{minipage}%
% \begin{minipage}{0.3333\textwidth} 
% \includegraphics[scale= 0.16]{Figures/2DBurgers/TimeOnly/Update/Redraw/Prediction2/ErrorStd_Nt40_Input2_v2.eps}
% \end{minipage}
\caption{\footnotesize [Burger Equation Case 1] Prediction std at final time: (Left) Reference std. (Middle) Prediction std. (Third) Std estimation errors.}
\label{Single_2DBurger_Std_Case1}
\end{figure}

We first present results for a representative input. We generate 400 solution samples using SON and 400 reference samples using the direct numerical solver, and compute the corresponding sample means and std. Figure~\ref{Single_2DBurger_Mean_Case1} shows the sample means at the final time $T$, along with heatmaps of the mean prediction errors, while the std prediction performance is presented in Figure~\ref{Single_2DBurger_Std_Case1}. From Figure~\ref{Single_2DBurger_Mean_Case1}, we can see that the predicted and reference means closely agree, and most mean errors fall within the range 
$0.001-0.05$. Figure~\ref{Single_2DBurger_Std_Case1} further demonstrates that SON effectively quantifies uncertainty, with most std error below $0.015$.

% \begin{figure}[h!]
% \begin{minipage}{0.3333\textwidth} 
% \includegraphics[scale= 0.16]{Figures/2DBurgers/TimeOnly/Update/Redraw/Prediction1/Mean_FinalTime_Input1.eps}
% \end{minipage}%
% \begin{minipage}{0.3333\textwidth} 
% \includegraphics[scale= 0.16]{Figures/2DBurgers/TimeOnly/Update/Redraw/Prediction1/ErrorMap_Nt40_Input1.eps}
% \end{minipage}%
% \begin{minipage}{0.3333\textwidth} 
% \includegraphics[scale= 0.16]{Figures/2DBurgers/TimeOnly/Update/Redraw/Prediction1/StdMap_Nt40_Input1.eps}
% \end{minipage}
% \caption{\footnotesize [2D Burger Case 1] Single-Input Comparison. (First) Mean at final time (Ref. vs Pred.). (Second) Absolute Mean Error Heat Map. (Third) Absolute Std Error Heat Map.}
% \label{Single_2DBurger_Case1}
% \end{figure}

\begin{figure}[h!]
\vspace{-0.3cm}
\begin{minipage}{0.245\textwidth}
\includegraphics[scale = 0.12]{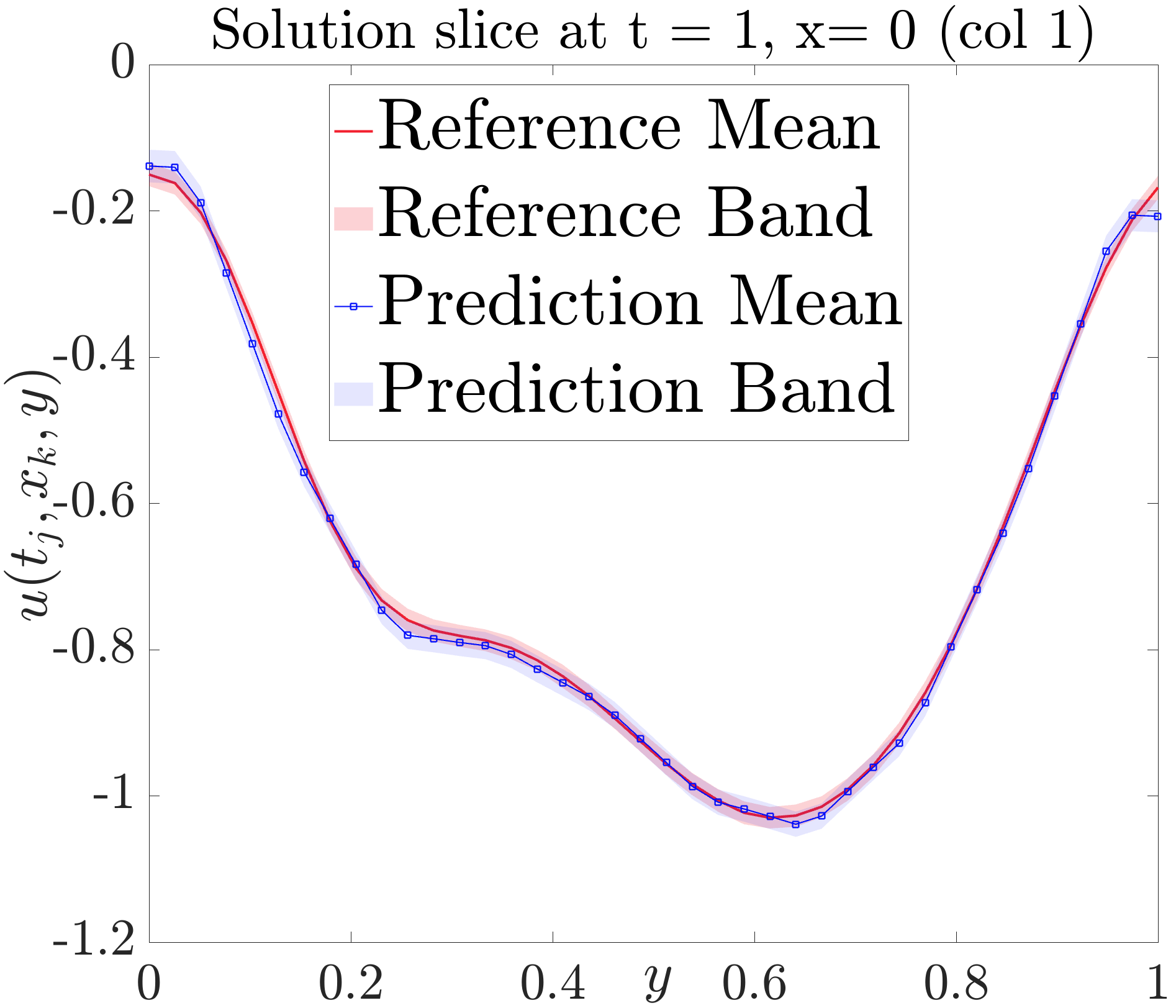}
\end{minipage}%
\begin{minipage}{0.245\textwidth}
\includegraphics[scale = 0.12]{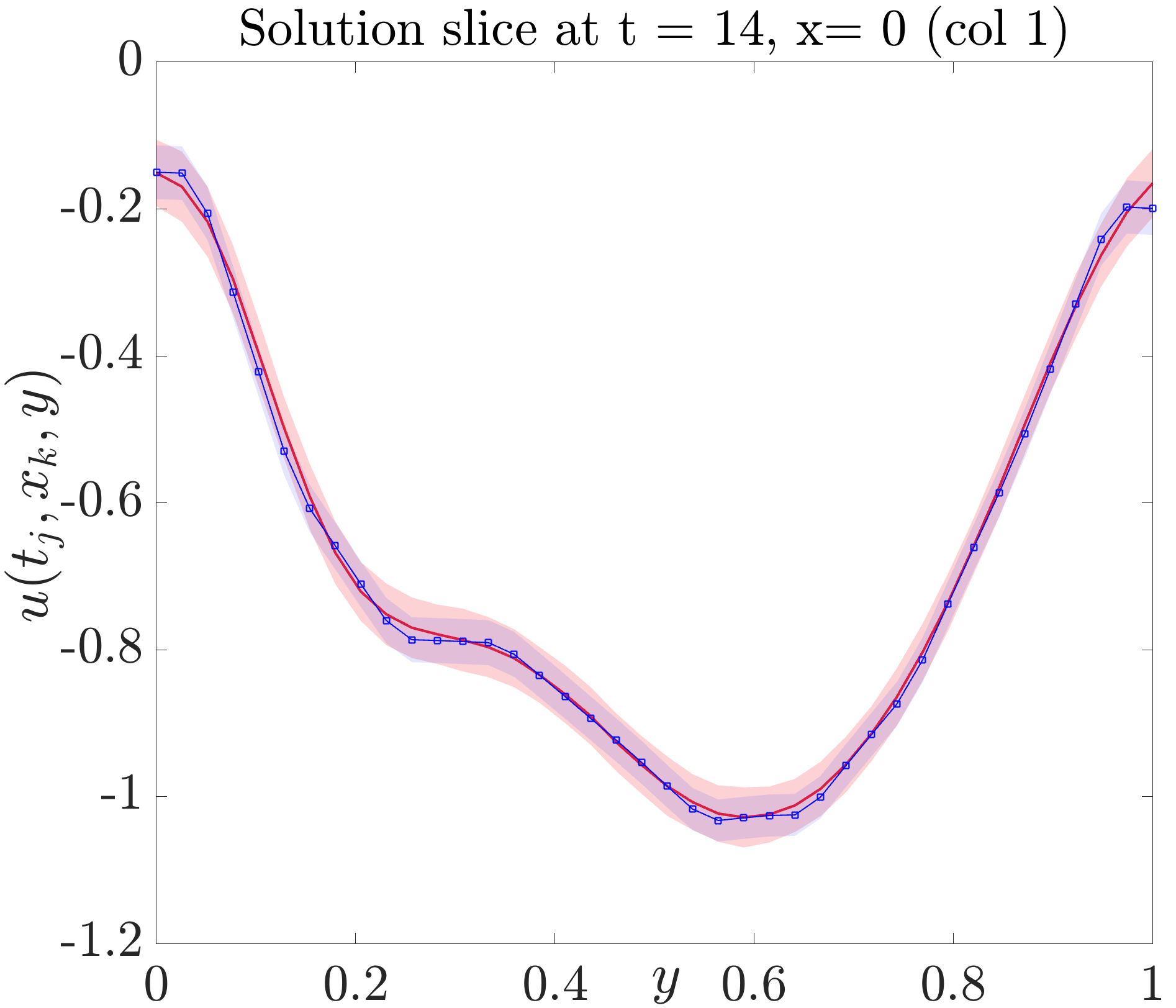}
\end{minipage}%
\begin{minipage}{0.245\textwidth}
\includegraphics[scale = 0.12]{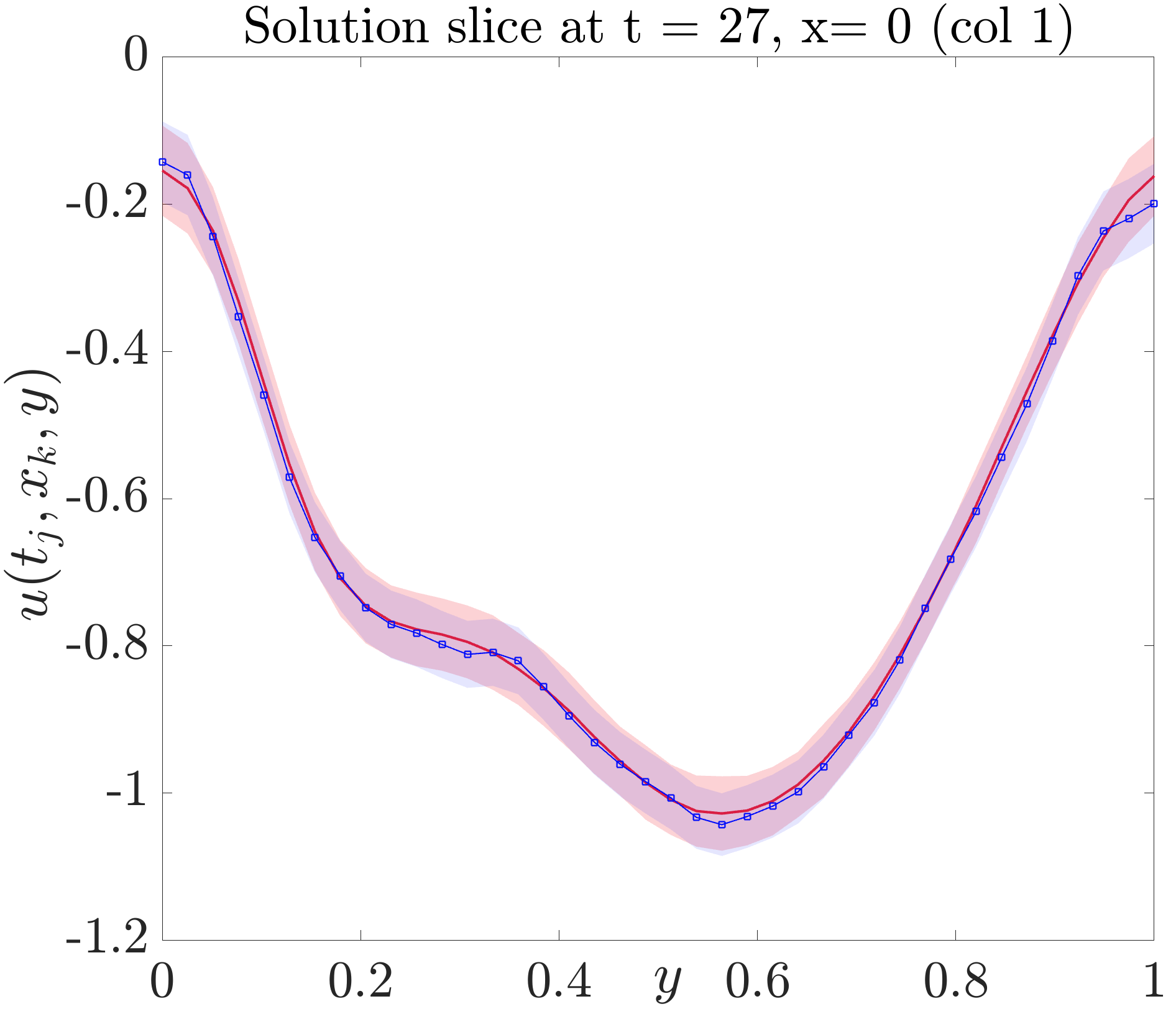}
\end{minipage}%
\begin{minipage}{0.245\textwidth}
\includegraphics[scale = 0.12]{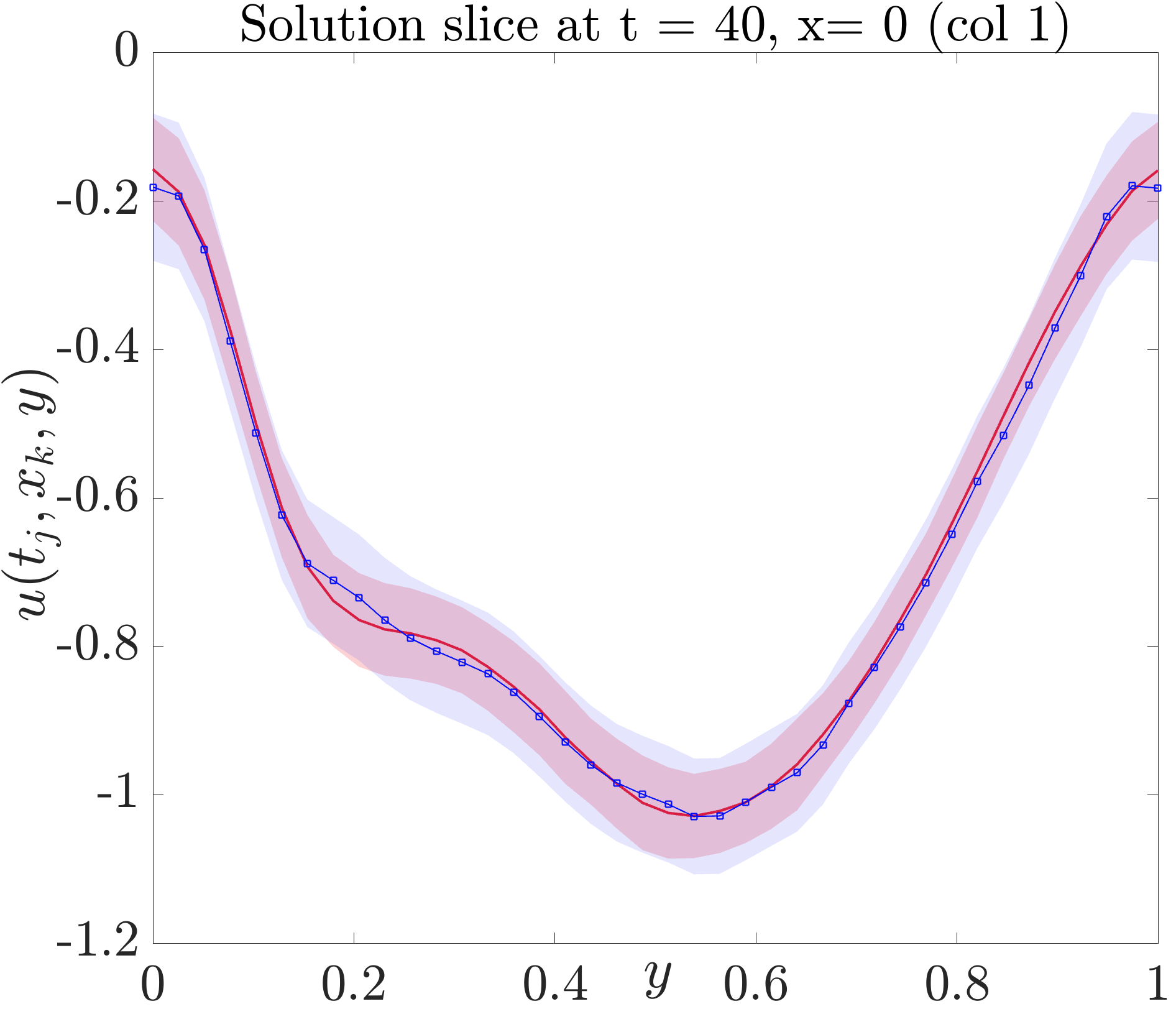}
\end{minipage}

\begin{minipage}{0.245\textwidth}
\includegraphics[scale = 0.12]{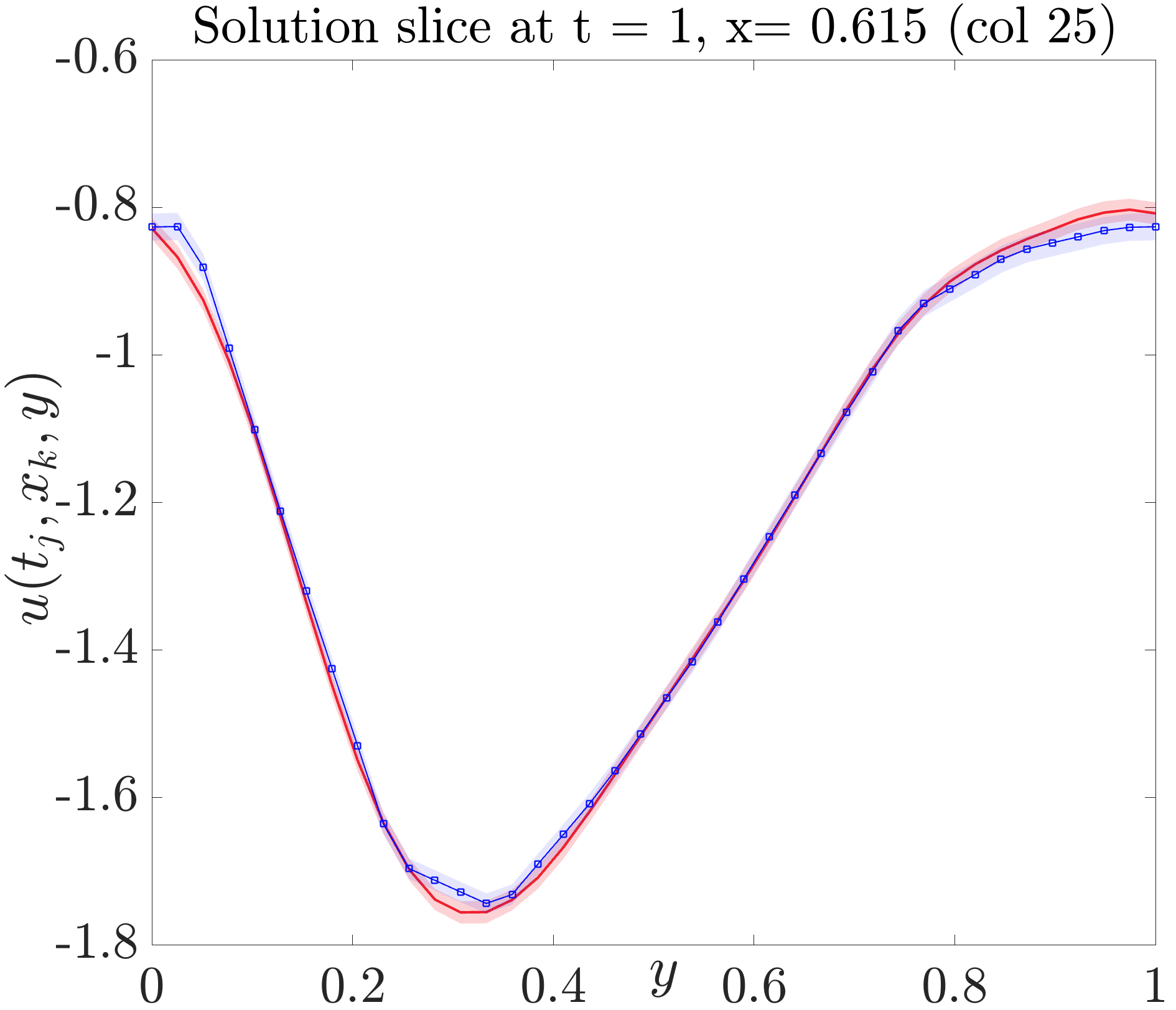}
\end{minipage}%
\begin{minipage}{0.245\textwidth}
\includegraphics[scale = 0.12]{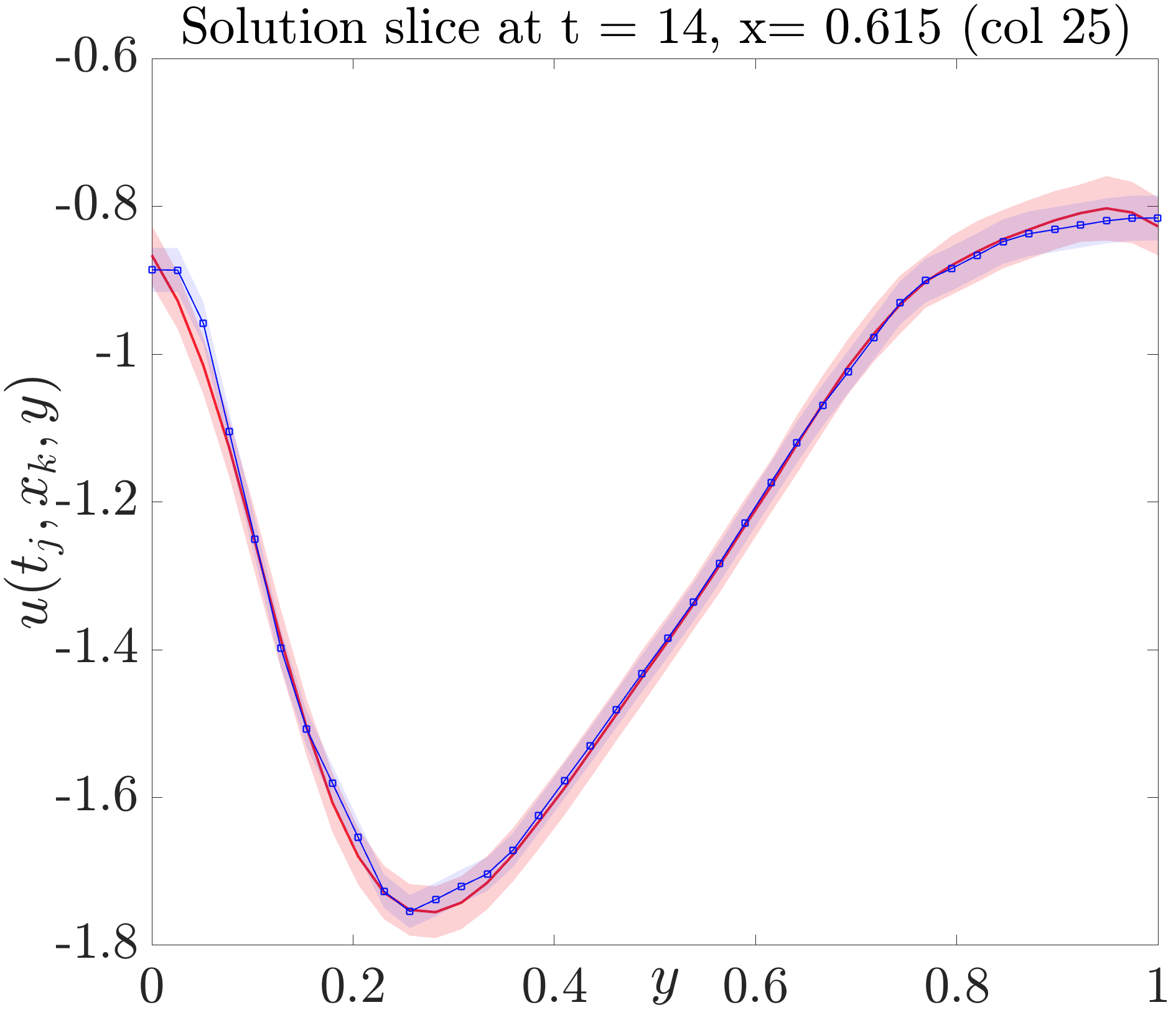}
\end{minipage}%
\begin{minipage}{0.245\textwidth}
\includegraphics[scale = 0.12]{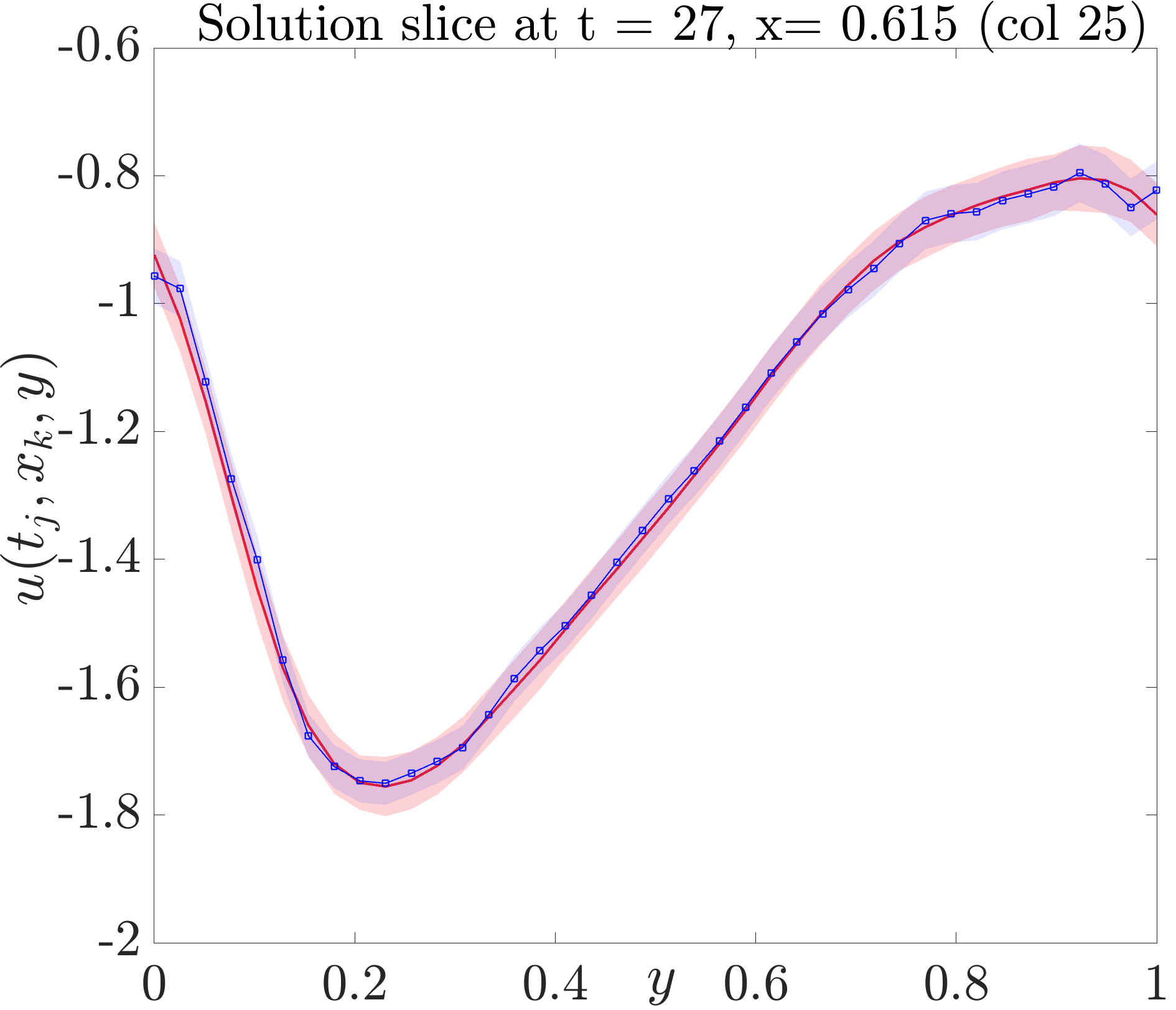}
\end{minipage}%
\begin{minipage}{0.245\textwidth}
\includegraphics[scale = 0.12]{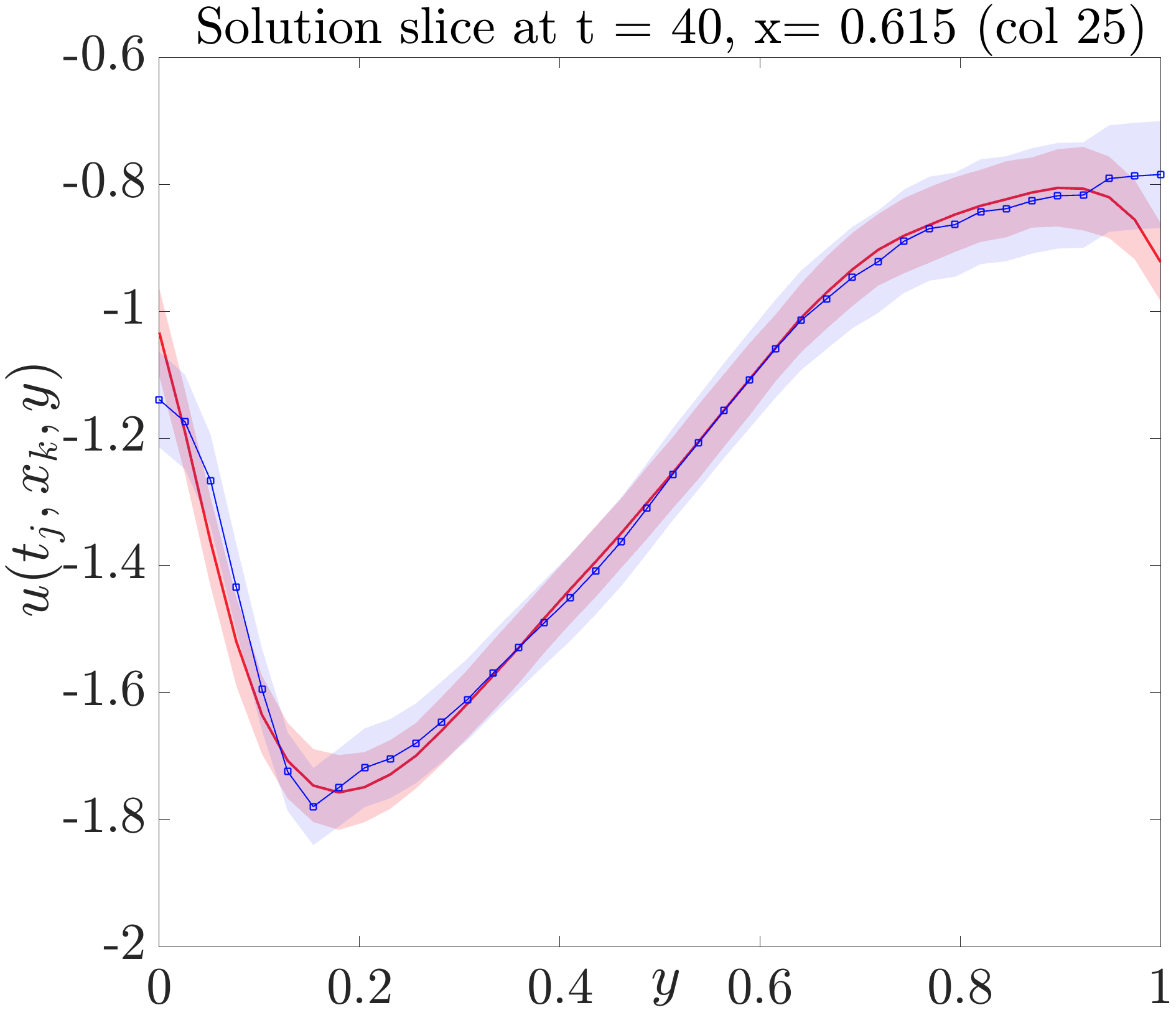}
\end{minipage}

\begin{minipage}{0.245\textwidth}
\includegraphics[scale = 0.12]{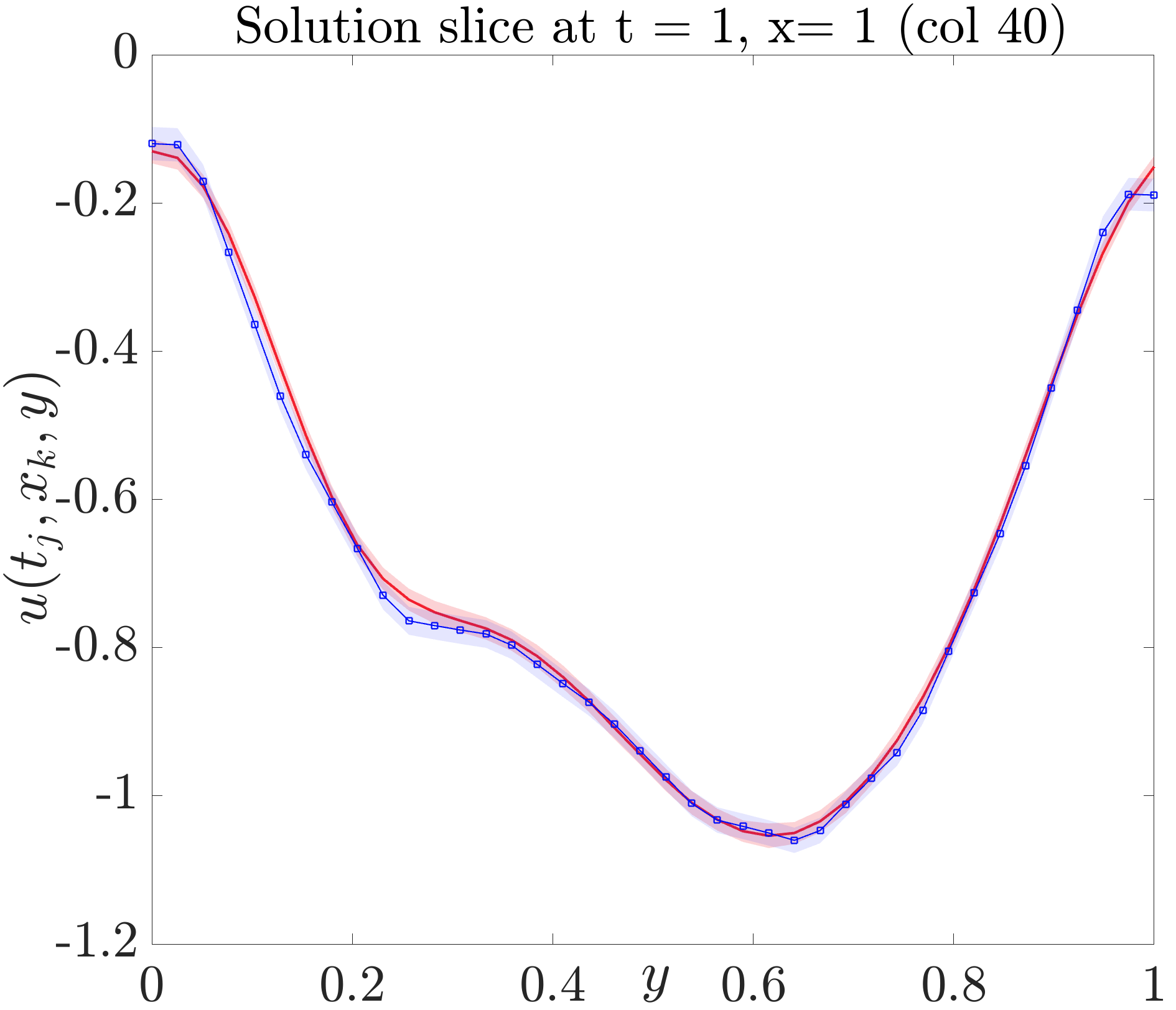}
\end{minipage}%
\begin{minipage}{0.245\textwidth}
\includegraphics[scale = 0.12]{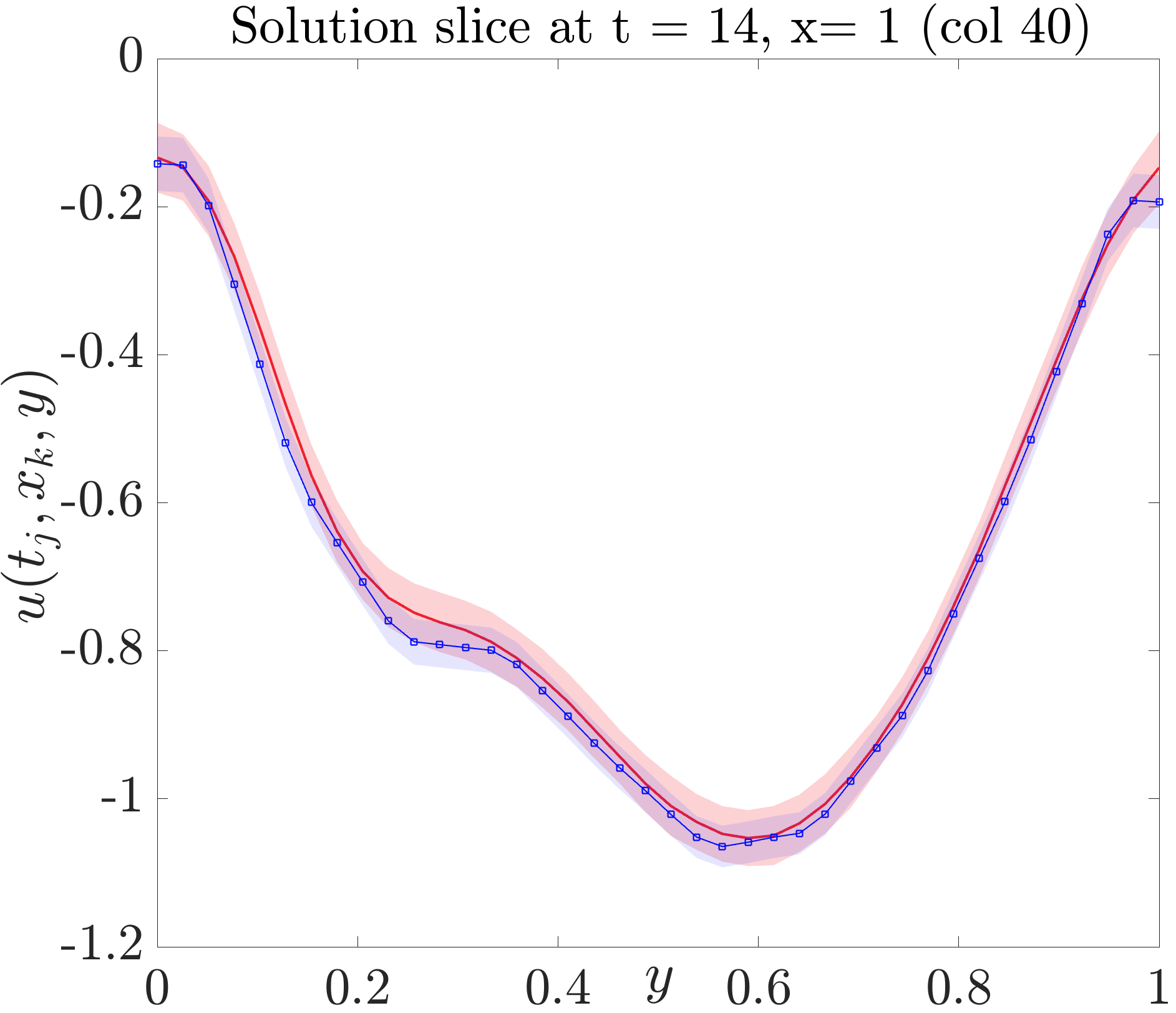}
\end{minipage}%
\begin{minipage}{0.245\textwidth}
\includegraphics[scale = 0.12]{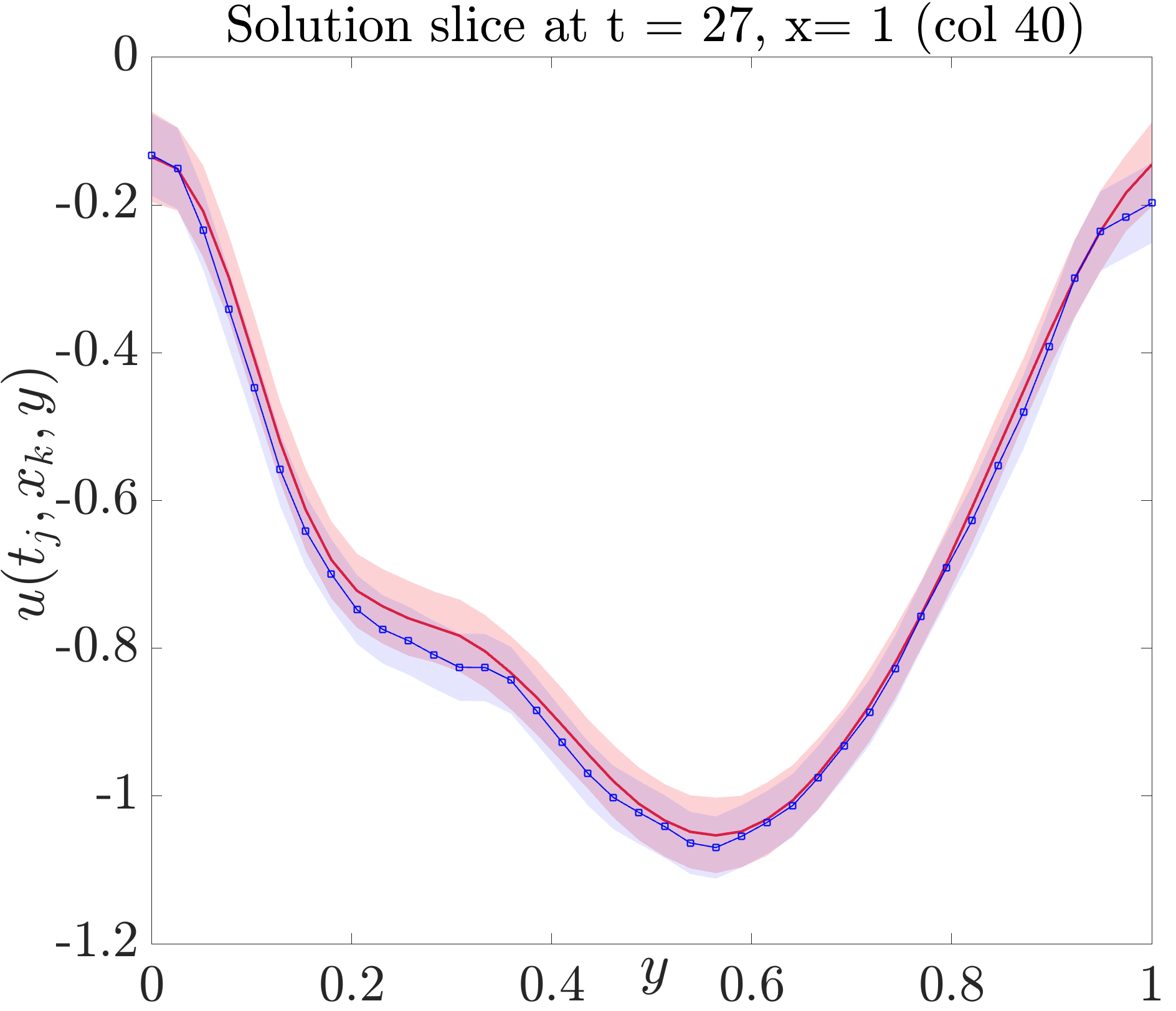}
\end{minipage}%
\begin{minipage}{0.245\textwidth}
\includegraphics[scale = 0.12]{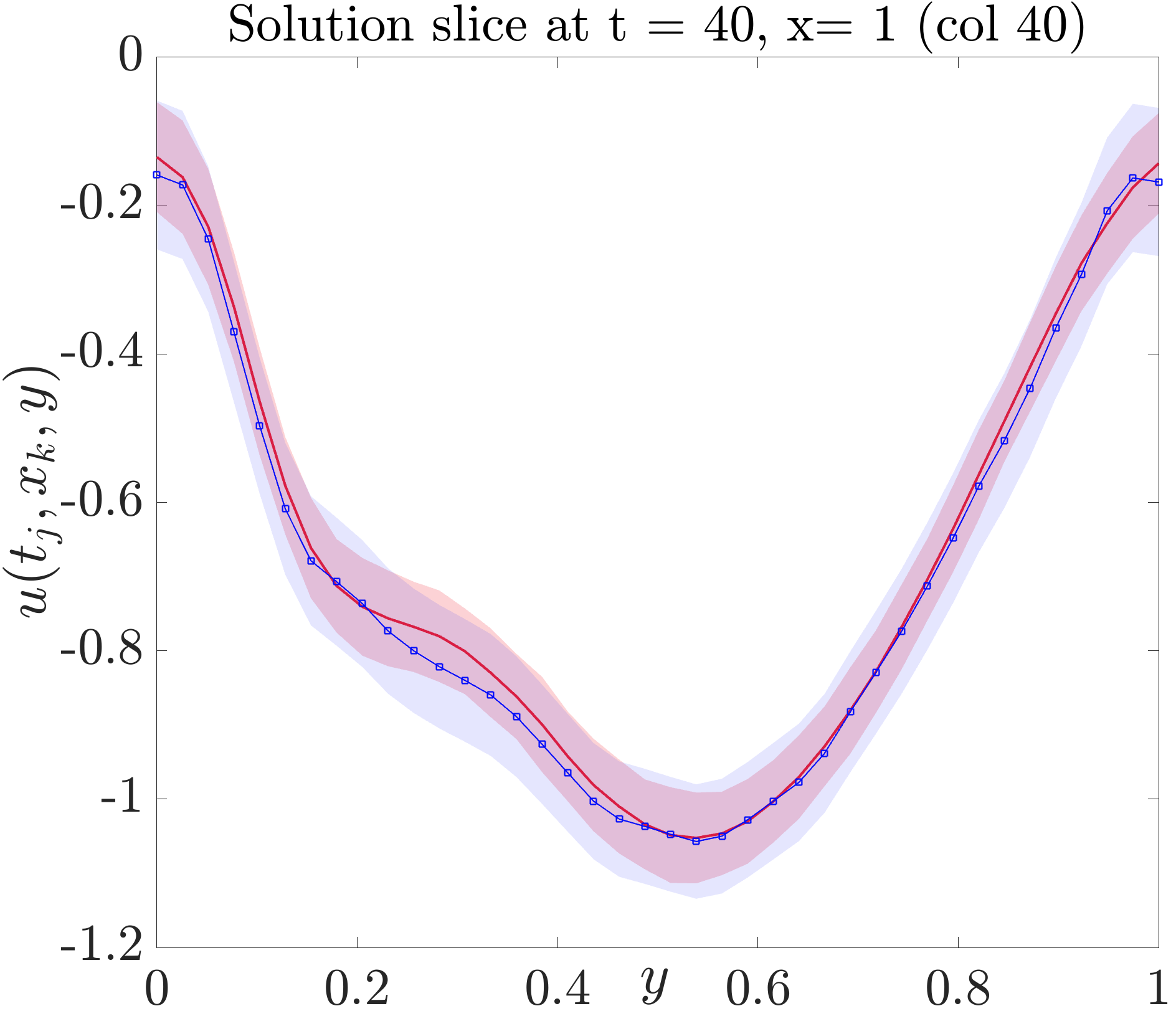}
\end{minipage}
\caption{\footnotesize [Burger Equation Case 1] Cross-sections of the predicted solution at time steps $t_m$ for $m=1, 14, 27, 40$. (First row) Column 1. (Second row) Column 25. (Third row) Column 40.}
\label{2DBurger_Case1_CrossSection}
\vspace{-0.1cm}
\end{figure}

We then examine SON's predictive performance more closely by plotting cross-sections of predicted sample means and their corresponding confidence bands at $x=0$, $x=25$ and $x=40$ for time steps $m=1, 14, 27$ and $40$ in Figure~\ref{2DBurger_Case1_CrossSection}. The figure shows that the predicted and reference means are in close agreement. Moreover, the predicted confidence bands accurately capture both the shape and the width of the reference uncertainty bands.

\begin{figure}[h!]
\vspace{-0.3cm}
\begin{minipage}{0.25\textwidth}
\includegraphics[scale = 0.123]{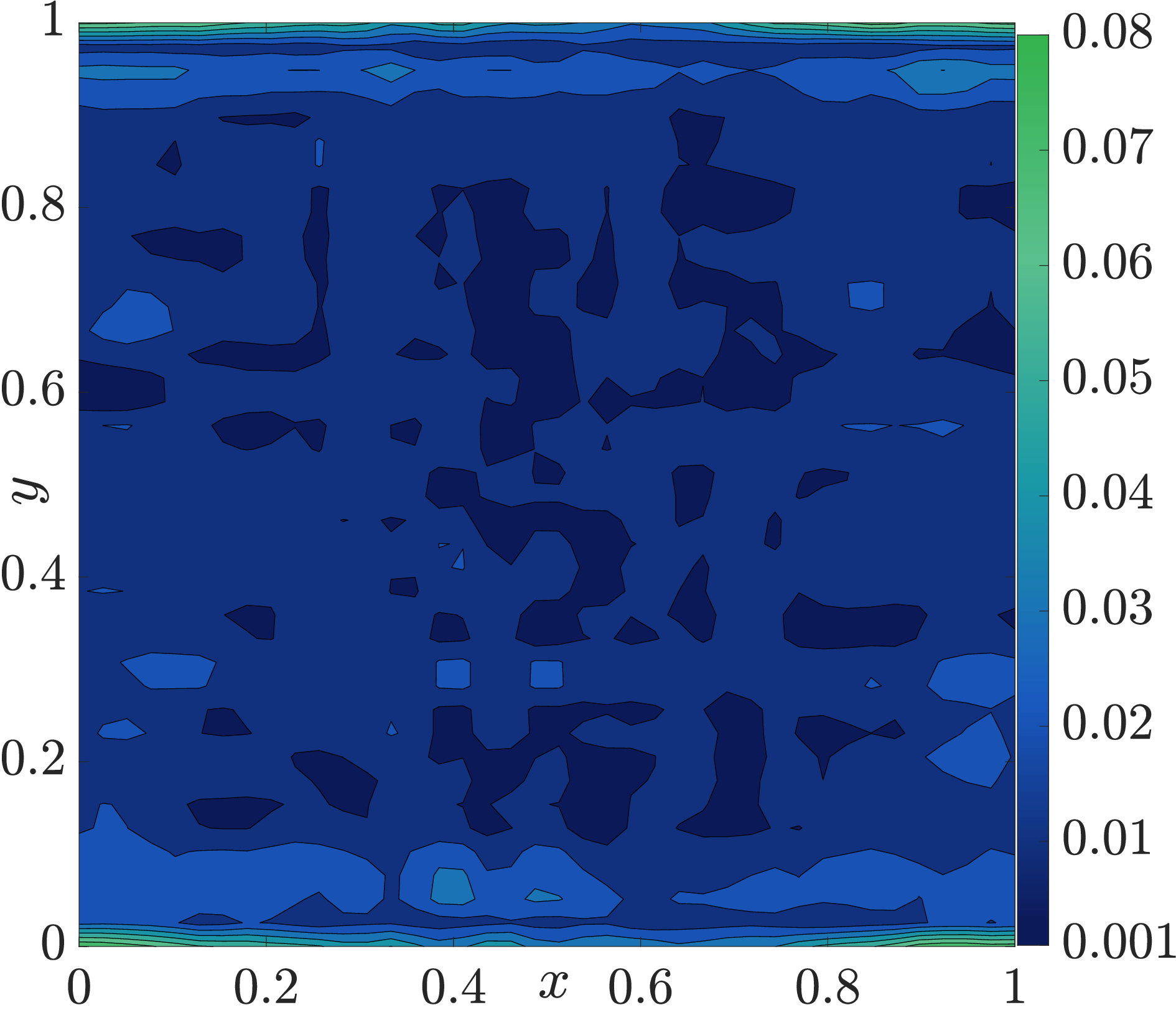}
\end{minipage}%
\begin{minipage}{0.25\textwidth}
\includegraphics[scale = 0.123]{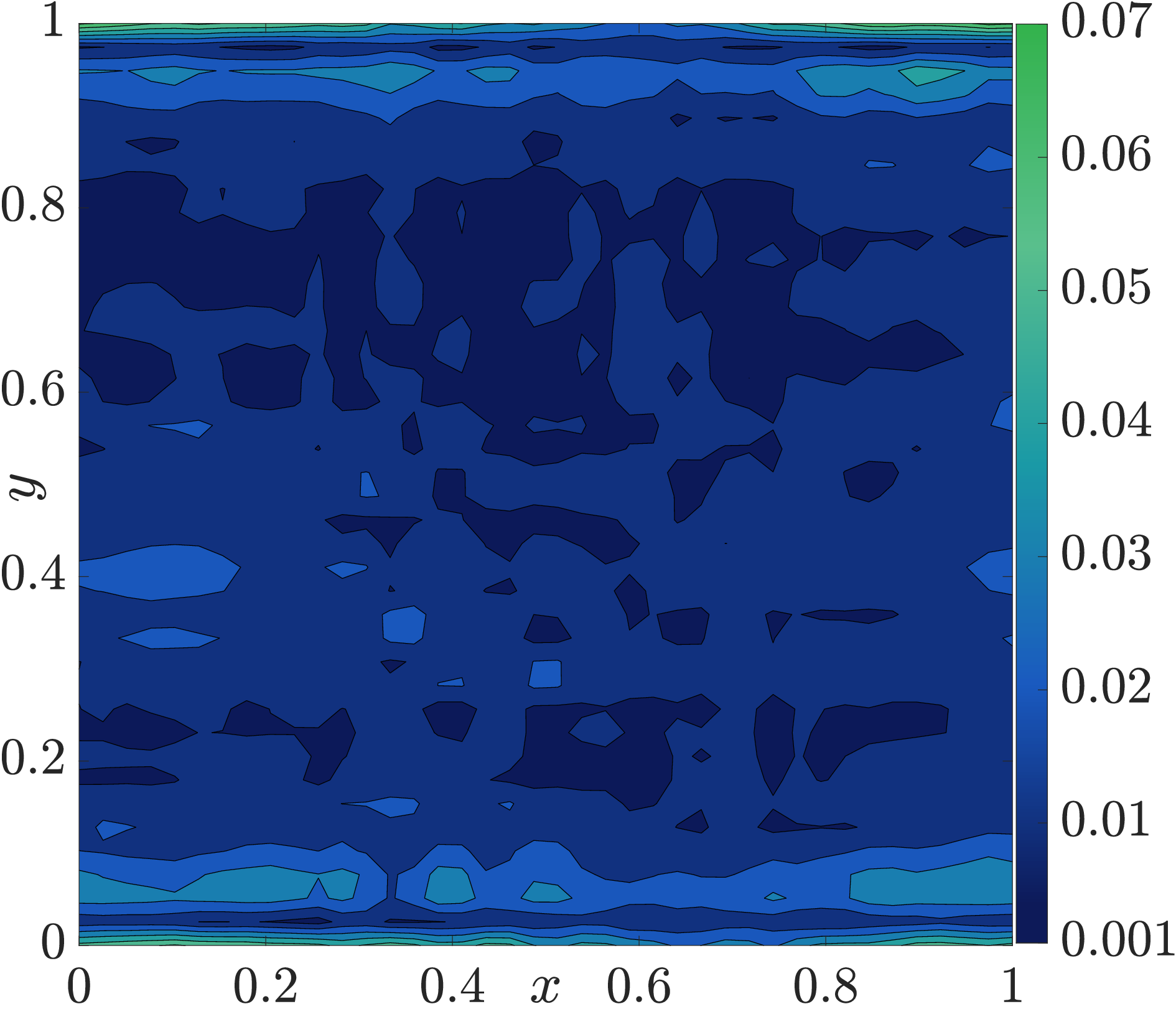}
\end{minipage}%
\begin{minipage}{0.25\textwidth}
\includegraphics[scale = 0.123]{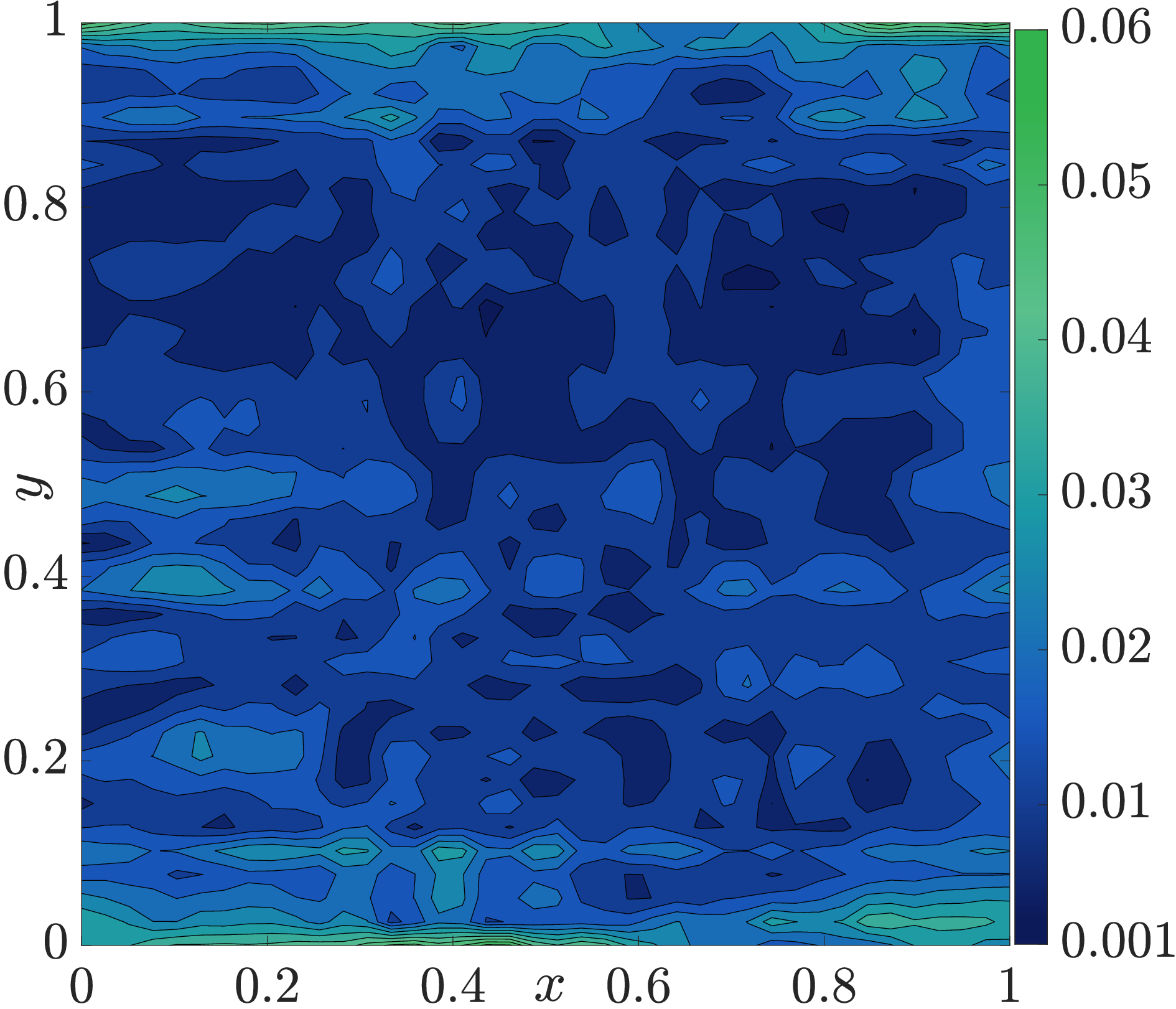}
\end{minipage}%
\begin{minipage}{0.25\textwidth}
\includegraphics[scale = 0.123]{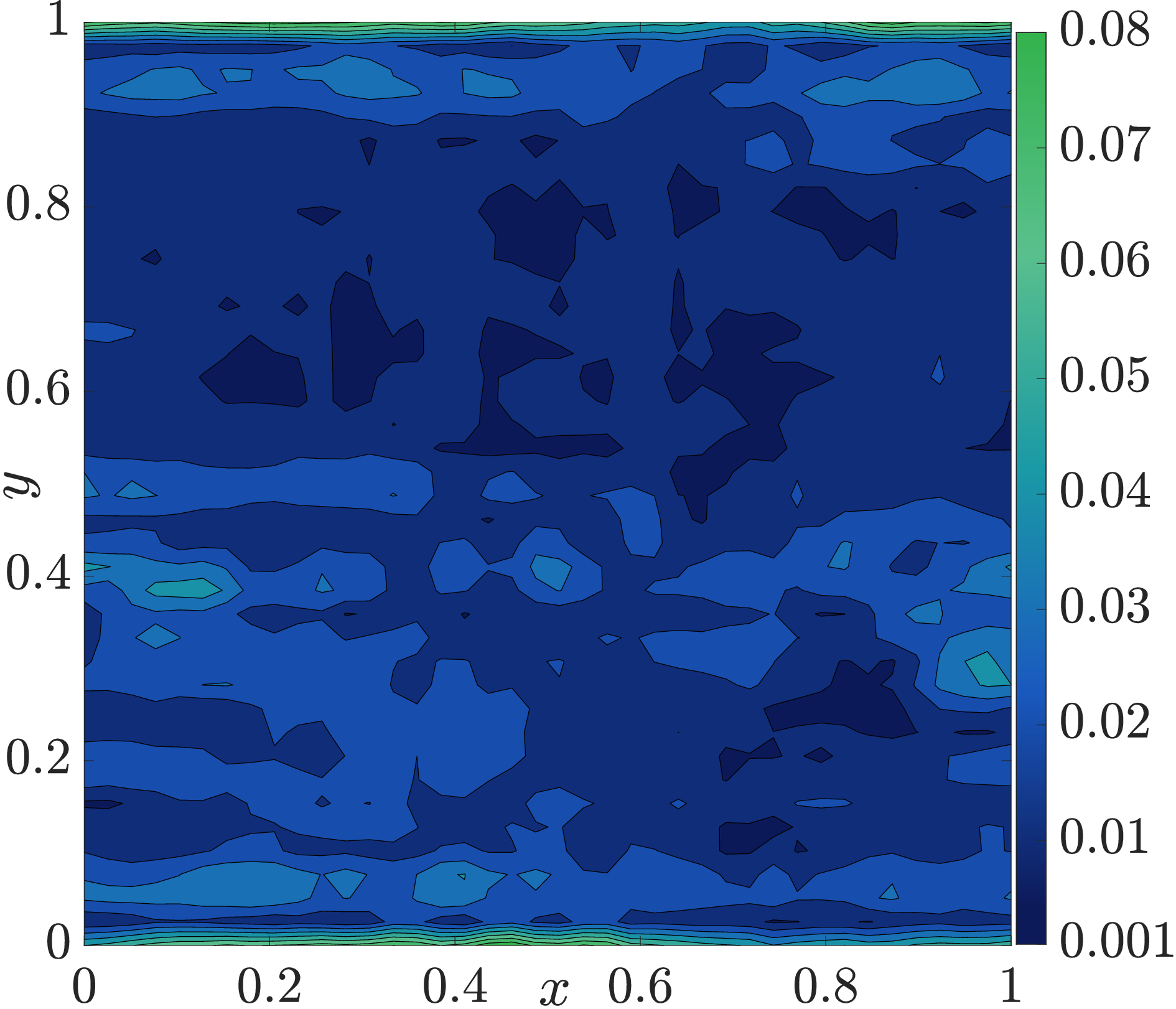}
\end{minipage}
\caption{\footnotesize [Burger Equation Case 1] Average mean error over 8 inputs at time step $t_m$ for $m=1, 14, 27, 40$.}
\label{Average_Mean_2DBurger_Case1}
\end{figure}

\begin{figure}[h!]
\vspace{-0.3cm}
\begin{minipage}{0.25\textwidth}
\includegraphics[scale = 0.123]{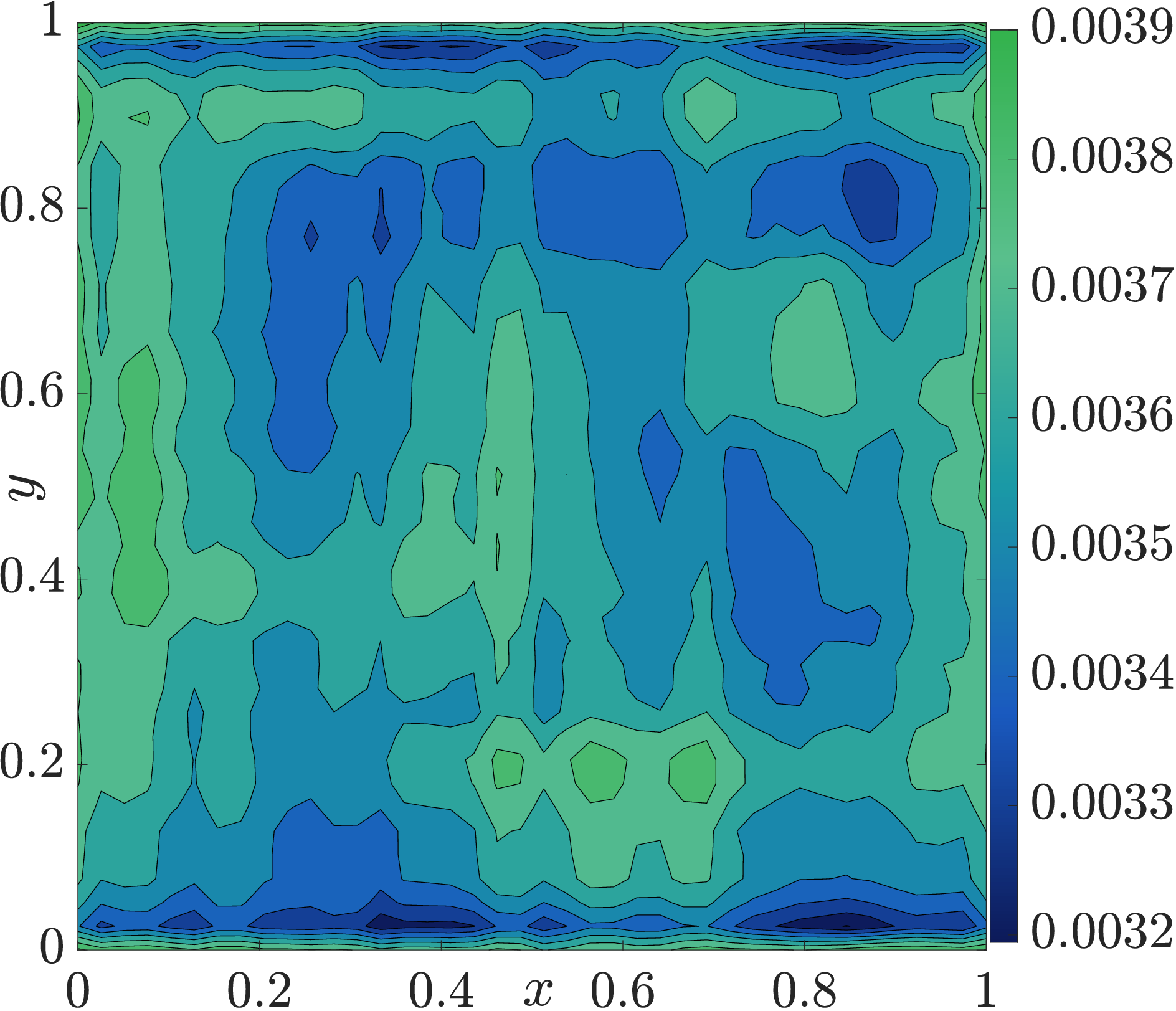}
\end{minipage}%
\begin{minipage}{0.25\textwidth}
\includegraphics[scale = 0.123]{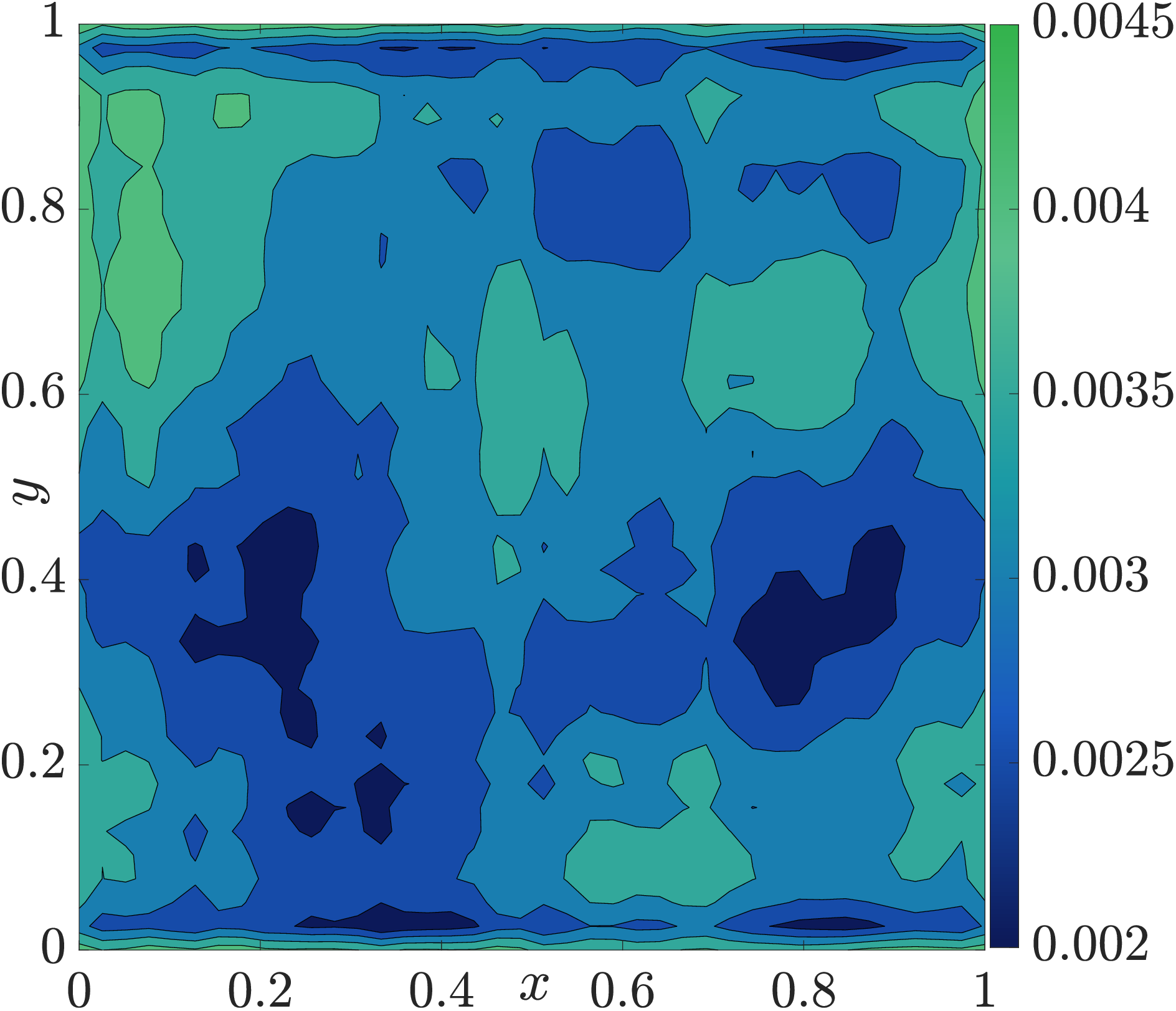}
\end{minipage}%
\begin{minipage}{0.25\textwidth}
\includegraphics[scale = 0.123]{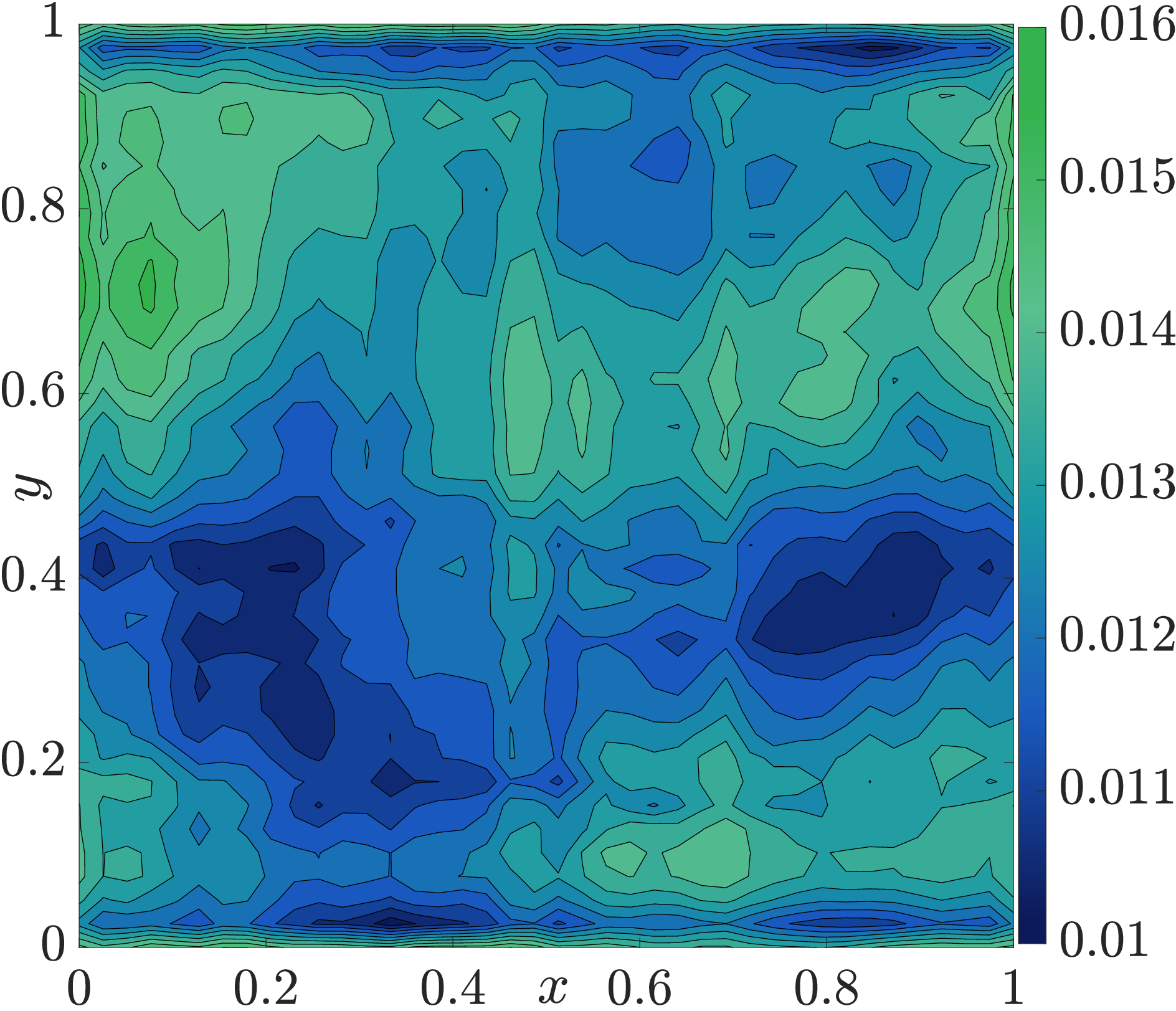}
\end{minipage}%
\begin{minipage}{0.25\textwidth}
\includegraphics[scale = 0.123]{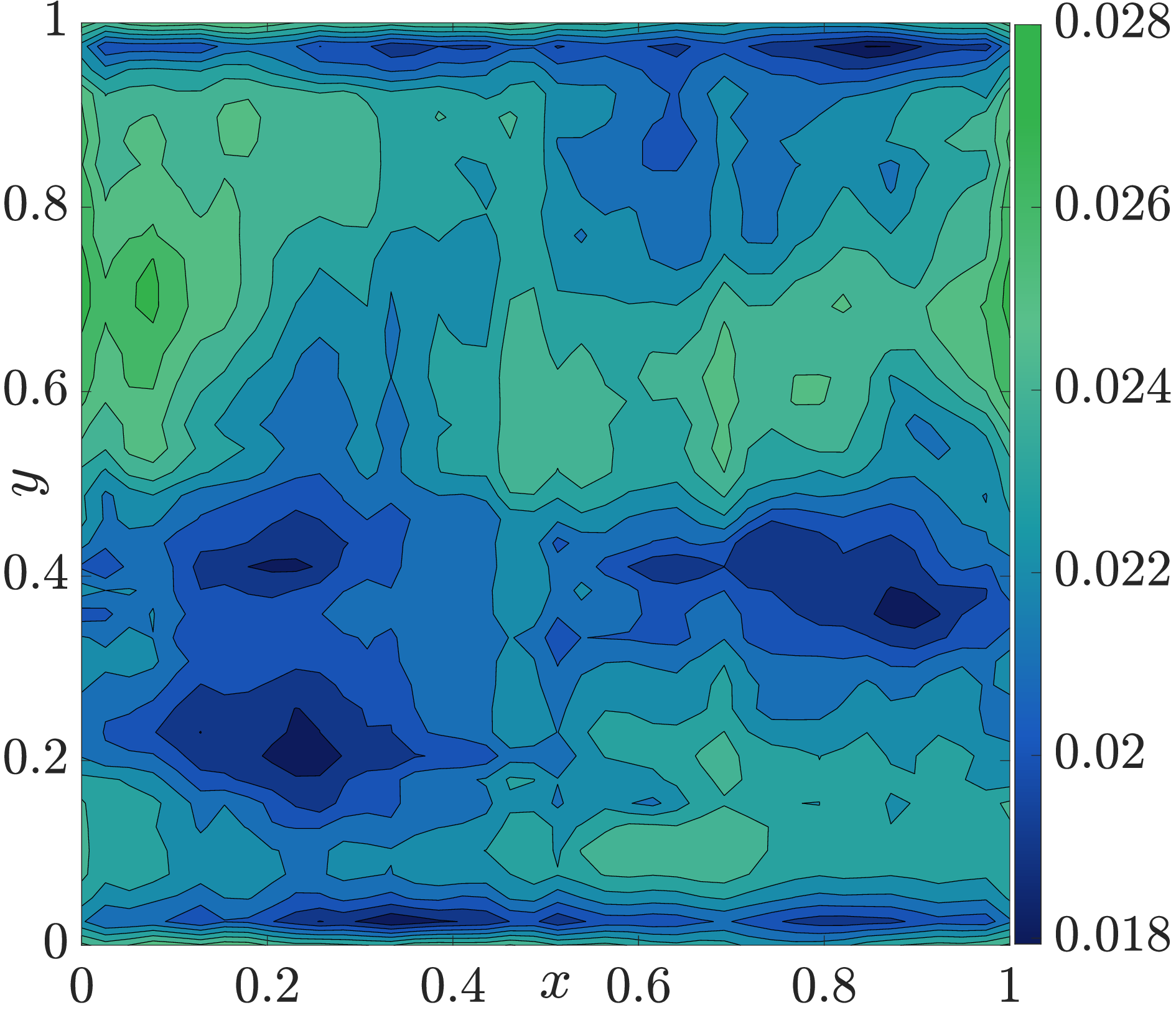}
\end{minipage}
\caption{\footnotesize [Burger Equation Case 1] Average std error over 8 inputs at time step $t_m$ for $m=1, 14, 27, 40$.}
\label{Average_Std_2DBurger_Case1}
\vspace{-0.4cm}
\end{figure}

For this more challenging Burgers' equation problem, we further assess the performance robustness of the SON model. To this end, we repeat the above experiment with $8$ randomly selected testing inputs and average the prediction errors of the sample mean and sample std across these tests at time step $t_m$ for $m=1, 14, 27, 40$. We visualize these averages as heatmaps in Figures~\ref{Average_Mean_2DBurger_Case1} and~\ref{Average_Std_2DBurger_Case1}. These results demonstrate the consistent accuracy and reliability of the SON model in capturing the solution’s spatial structure and in quantifying uncertainty across the domain.

\vspace{0.5em}
\subsubsection{Case 2: Space-time dependent noise}
Let $\mathbf u_{h,m} \in\mathbb R^{N}$ denote the numerical solution at time $t_m$, where $N=H\times H$ is the number of spatial degrees of freedom. In Case~2, we introduce Gaussian noise with spatially and temporally varying amplitude:
\[
\tilde{\mathbf u}_{h,m}^{(b)}
=
\mathbf u_{h,m}^{(b)}
+
\sqrt{\Delta t}\,\big(\boldsymbol{\sigma}_m \odot \boldsymbol{\xi}_{m}^{(b)}\big),
\qquad
\boldsymbol{\xi}_{m}^{(b)}\sim\mathcal N(\mathbf 0, I_{N}),
\]
where $\boldsymbol{\sigma}_m=[\sigma_{m,1},\dots,\sigma_{m,H}] \in \mathbb{R}^N$ is a matrix and $\odot$ denotes element-wise multiplication. For all time steps $m$, the entries of $\pmb{\sigma}_m$ are sampled from $\mathcal{U}(0.15, 0.25)$.

\begin{figure}[h!]
\begin{minipage}{0.3333\textwidth} 
\includegraphics[scale= 0.16]{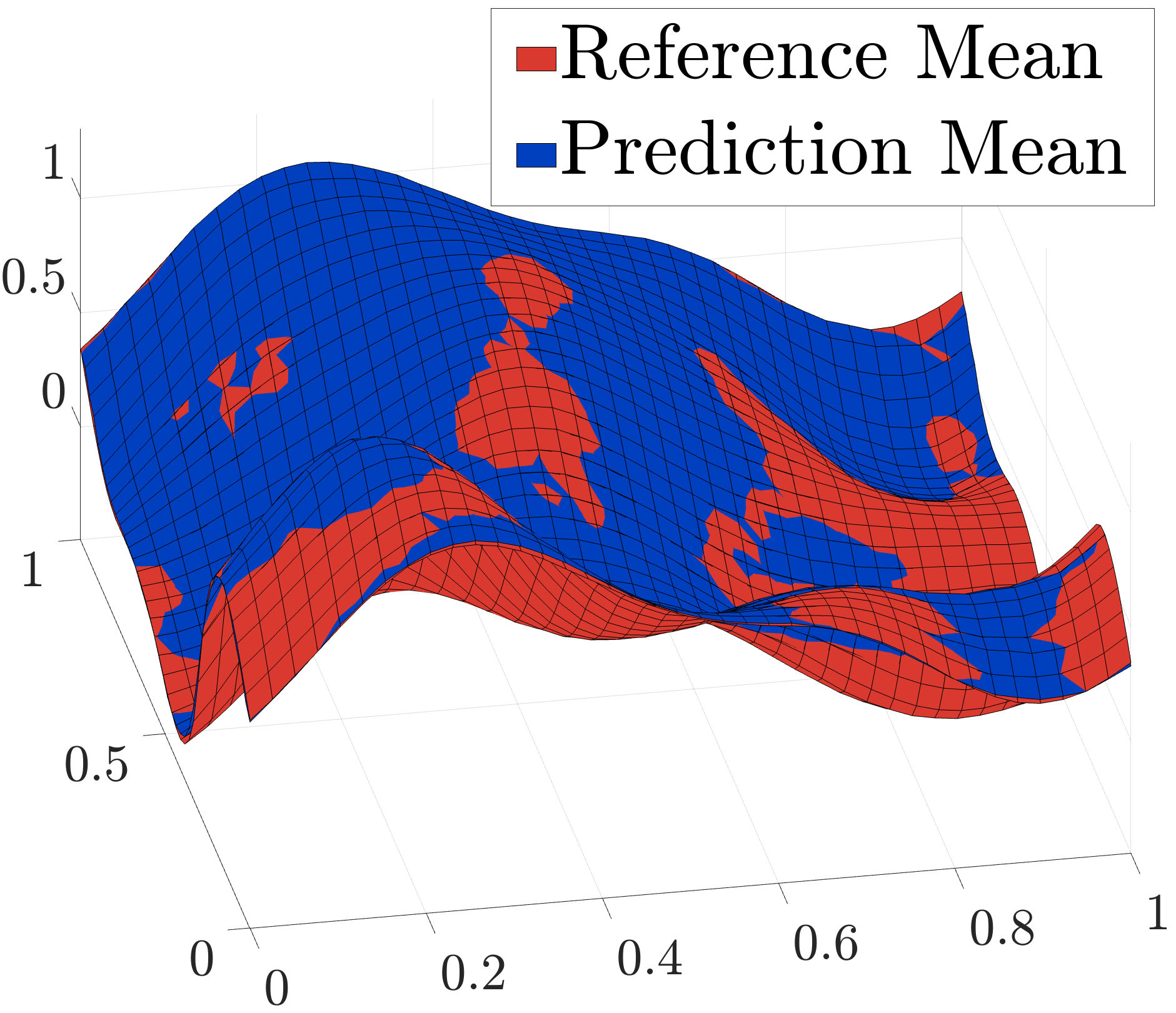}
\end{minipage}%
\begin{minipage}{0.3333\textwidth} 
\includegraphics[scale= 0.16]{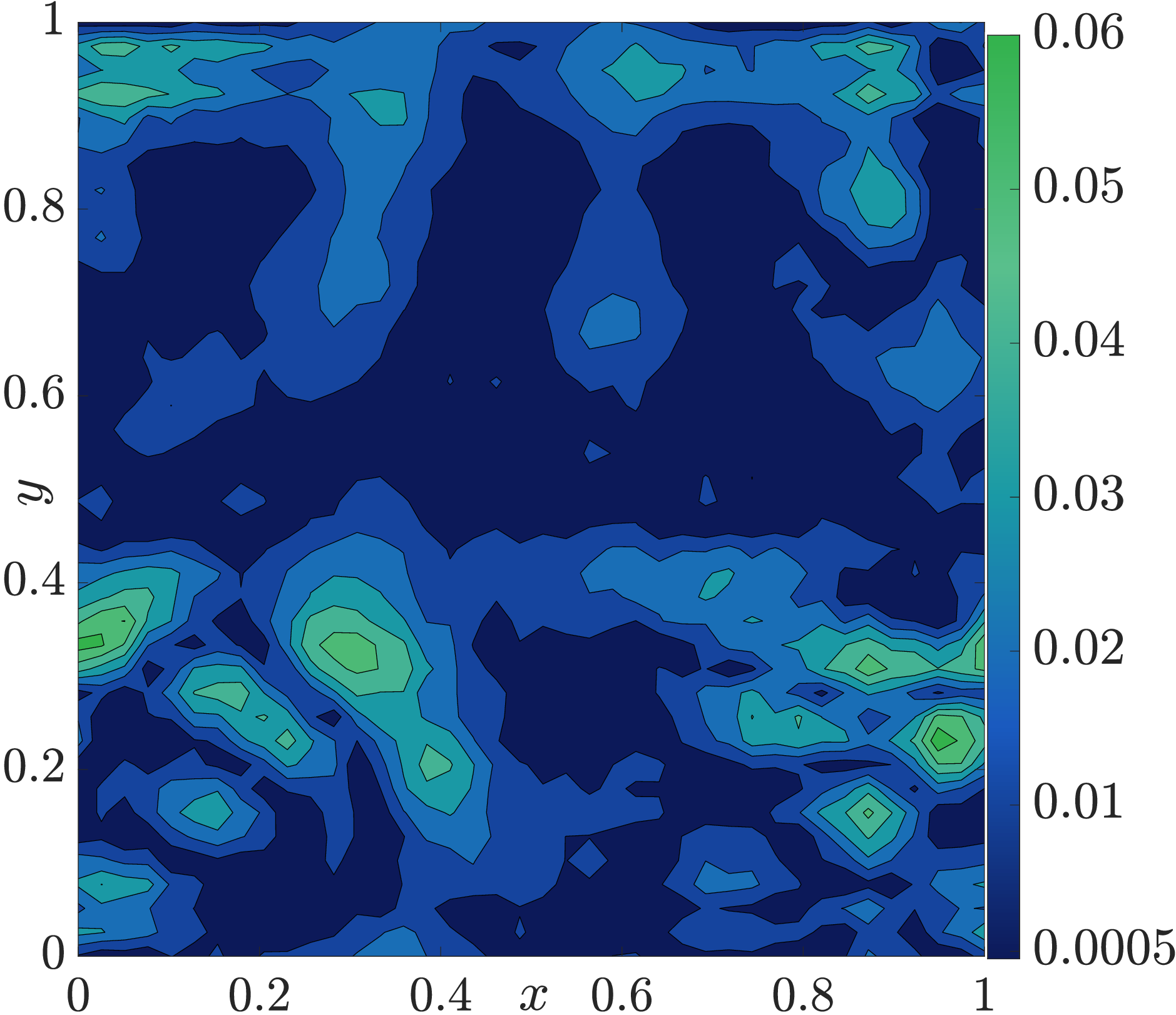}
\end{minipage}%
\begin{minipage}{0.3333\textwidth} 
\includegraphics[scale= 0.16]{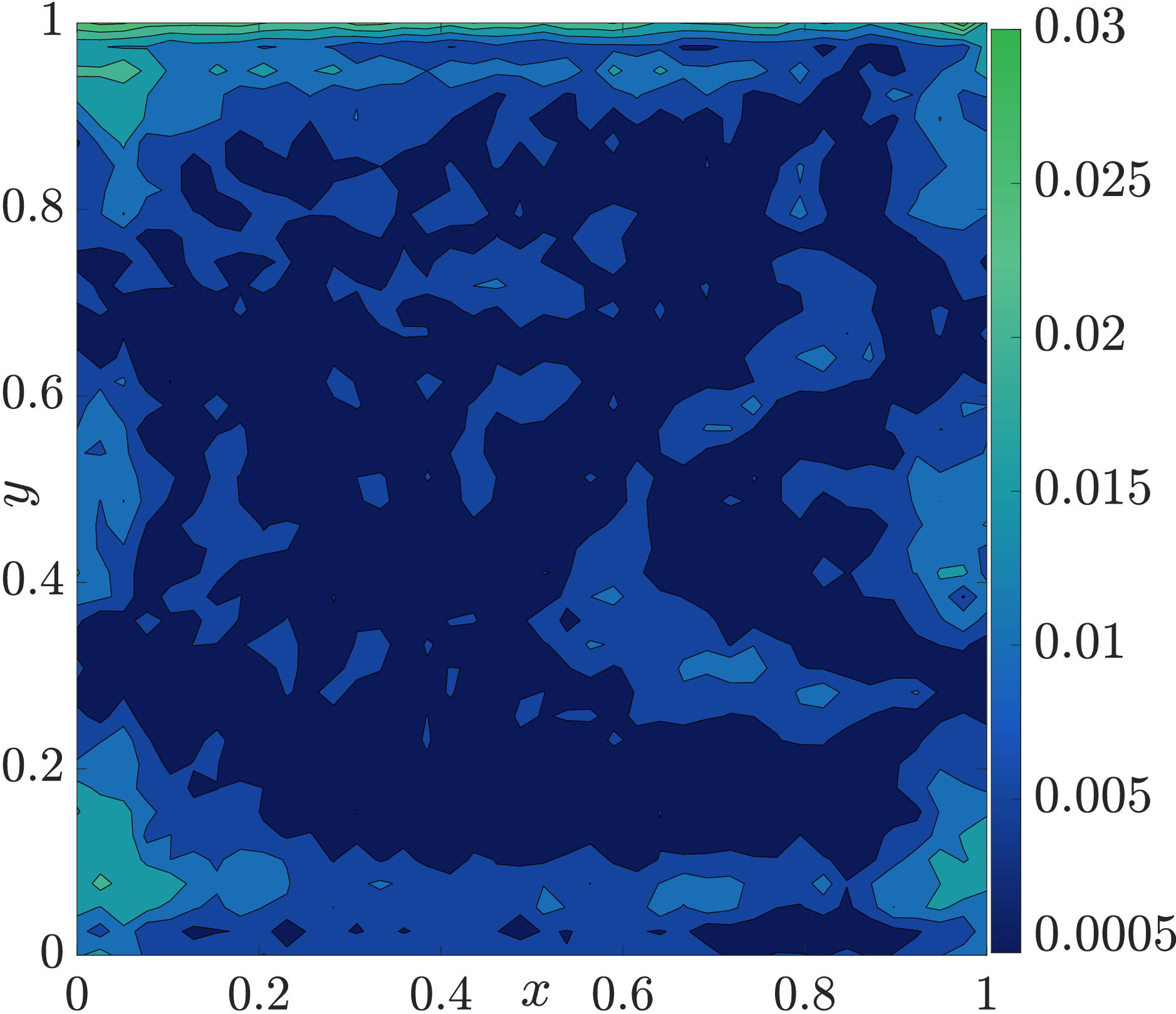}
\end{minipage}
\caption{\footnotesize [Burger Equation Case 2] Prediction performance. (Left) Prediction mean at final time. (Middle) Prediction mean error. (Right) Prediction std error.}
\label{Single_2DBurger_Case2}
\end{figure}
\begin{figure}[h!]
\vspace{-0.3cm}
\begin{minipage}{0.245\textwidth}
\includegraphics[scale = 0.15]{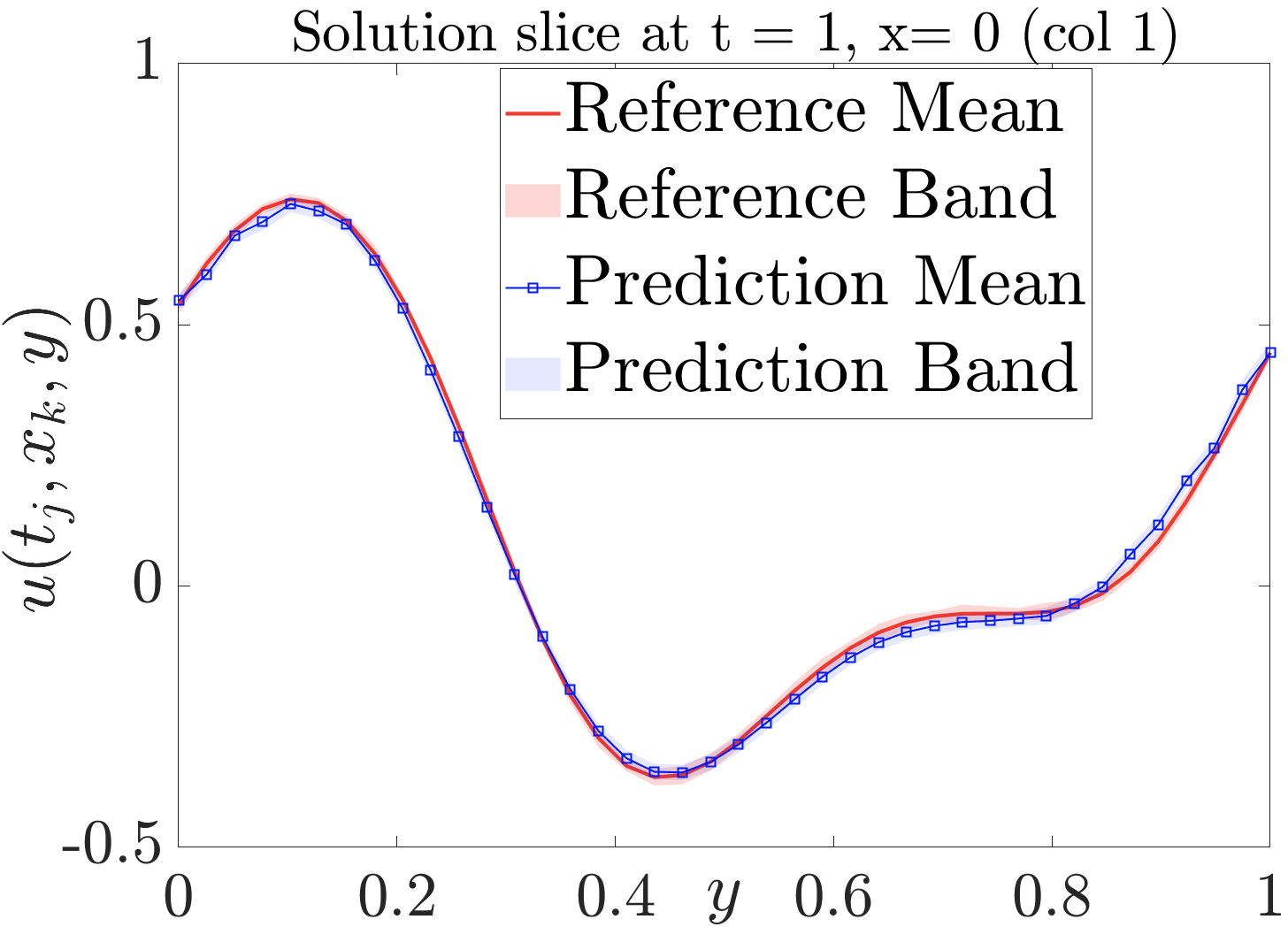}
\end{minipage}%
\begin{minipage}{0.245\textwidth}
\includegraphics[scale = 0.15]{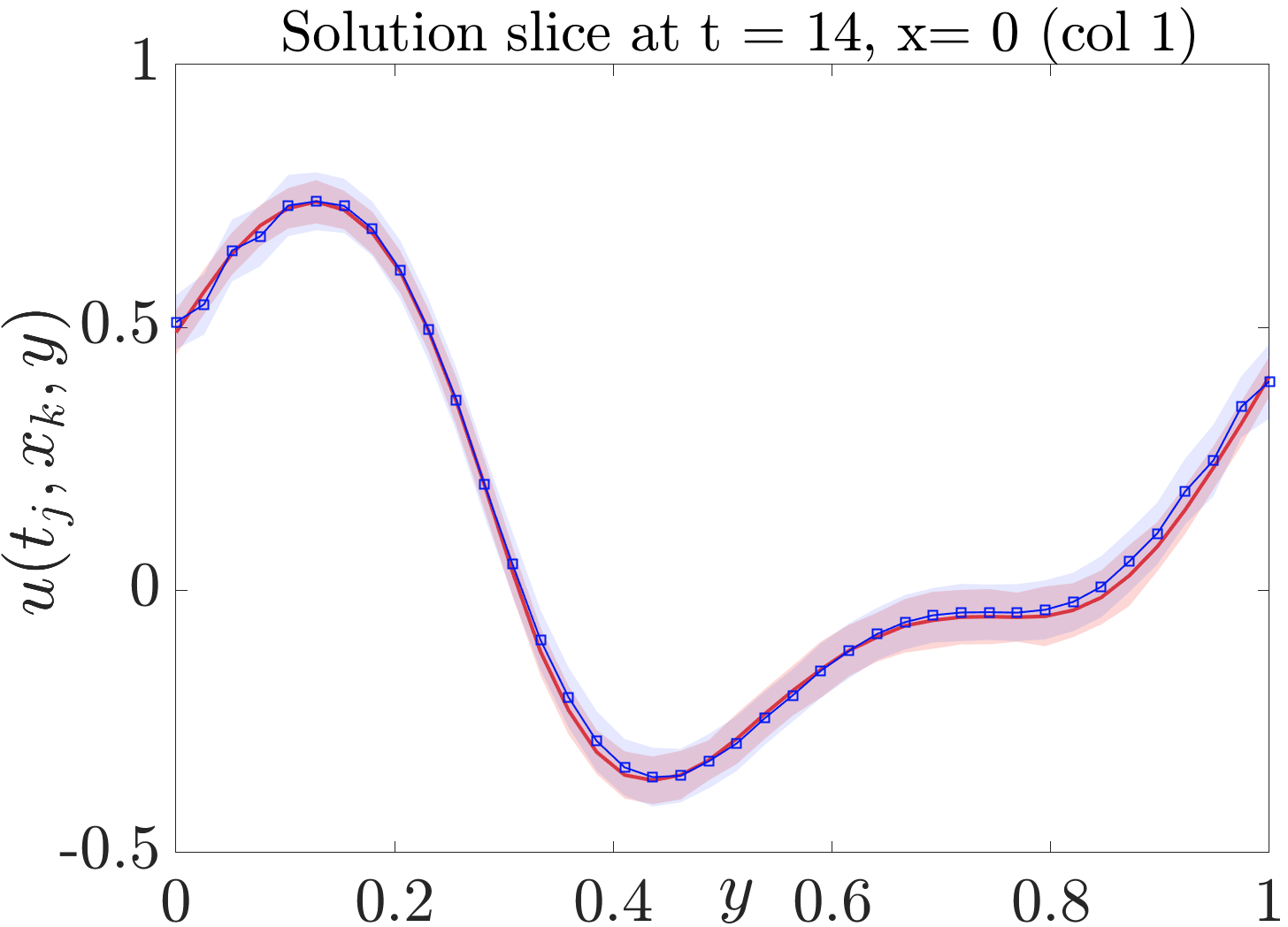}
\end{minipage}%
\begin{minipage}{0.245\textwidth}
\includegraphics[scale = 0.15]{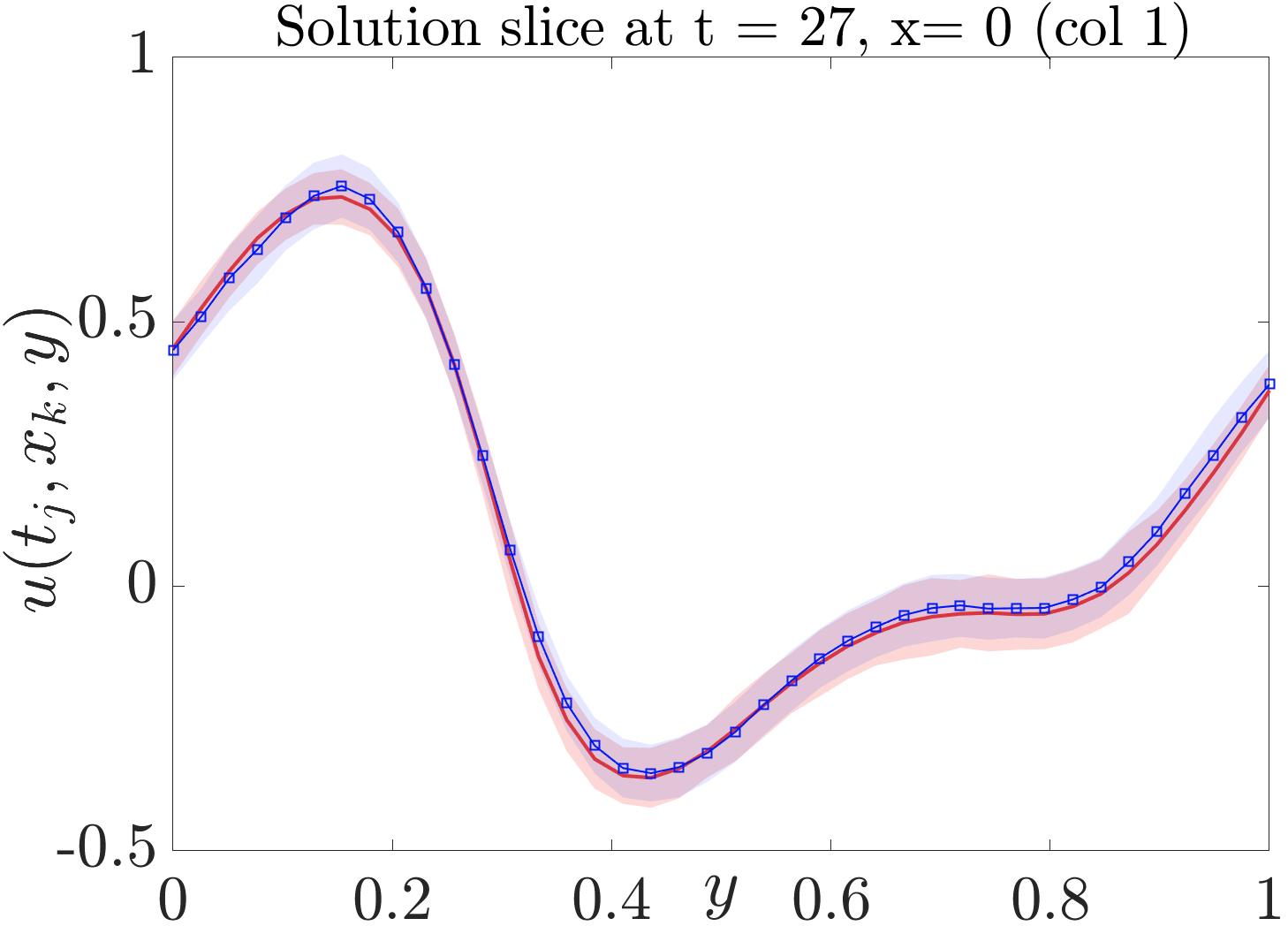}
\end{minipage}%
\begin{minipage}{0.245\textwidth}
\includegraphics[scale = 0.15]{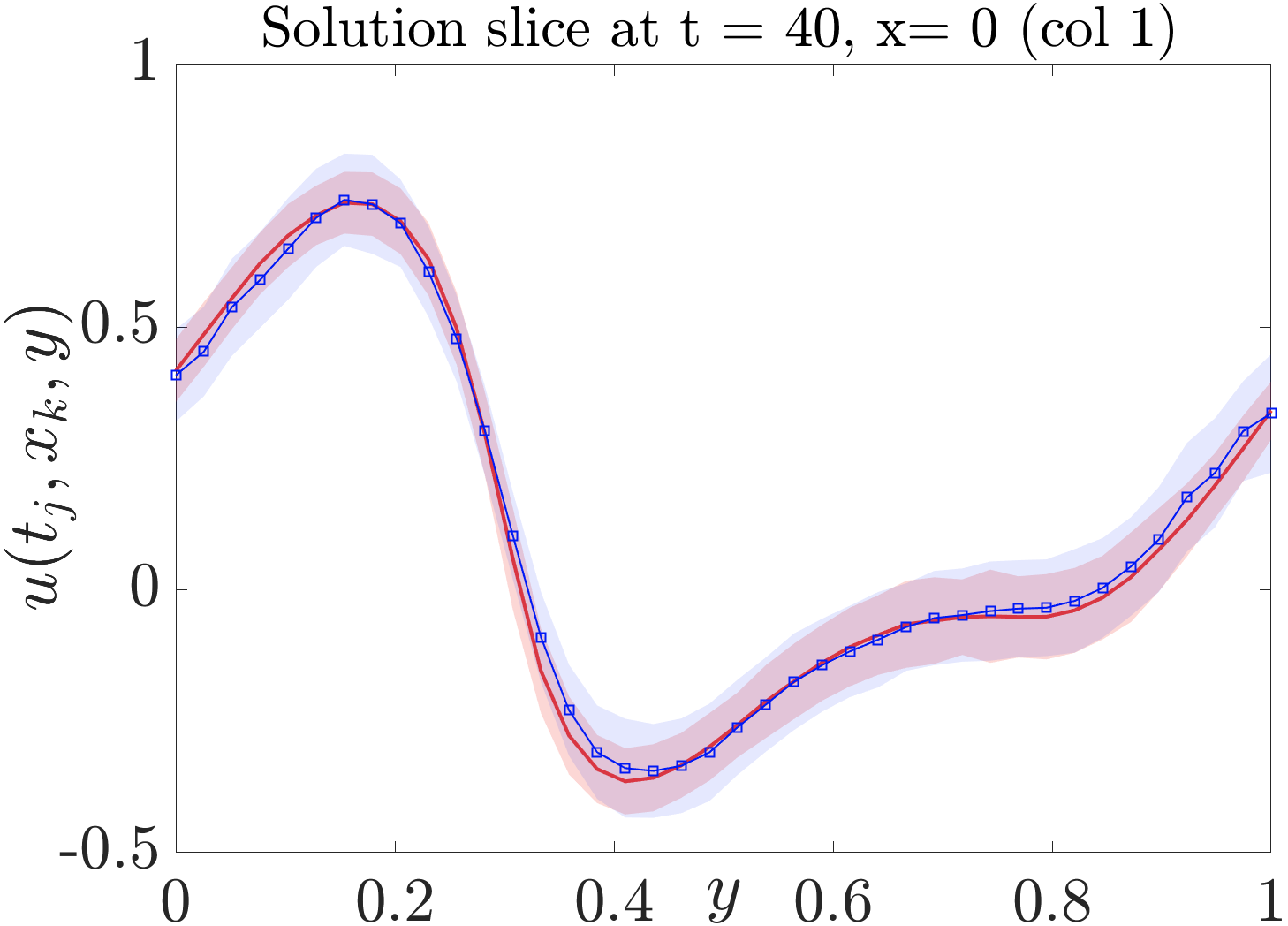}
\end{minipage}

\begin{minipage}{0.245\textwidth}
\includegraphics[scale = 0.15]{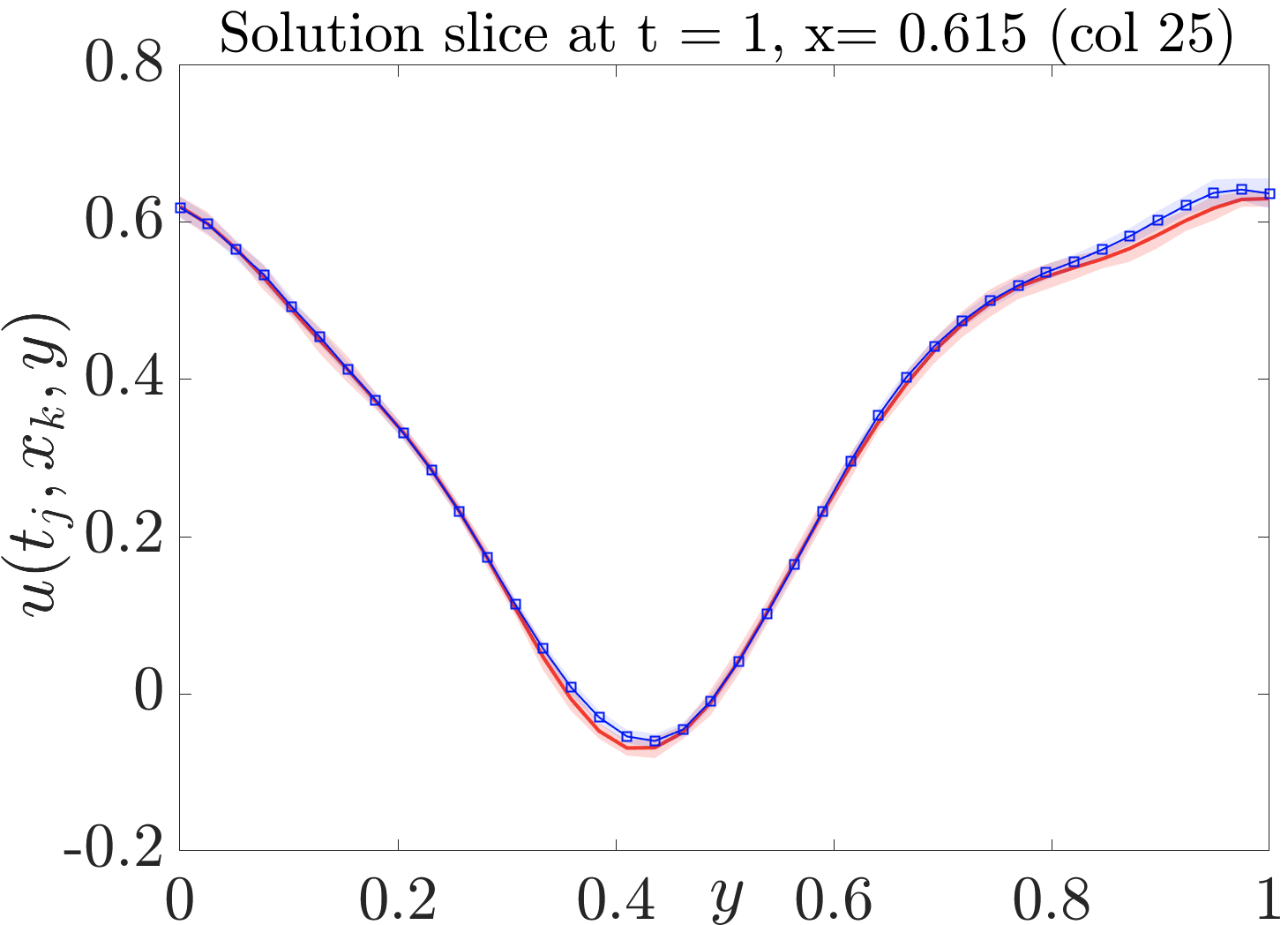}
\end{minipage}%
\begin{minipage}{0.245\textwidth}
\includegraphics[scale = 0.15]{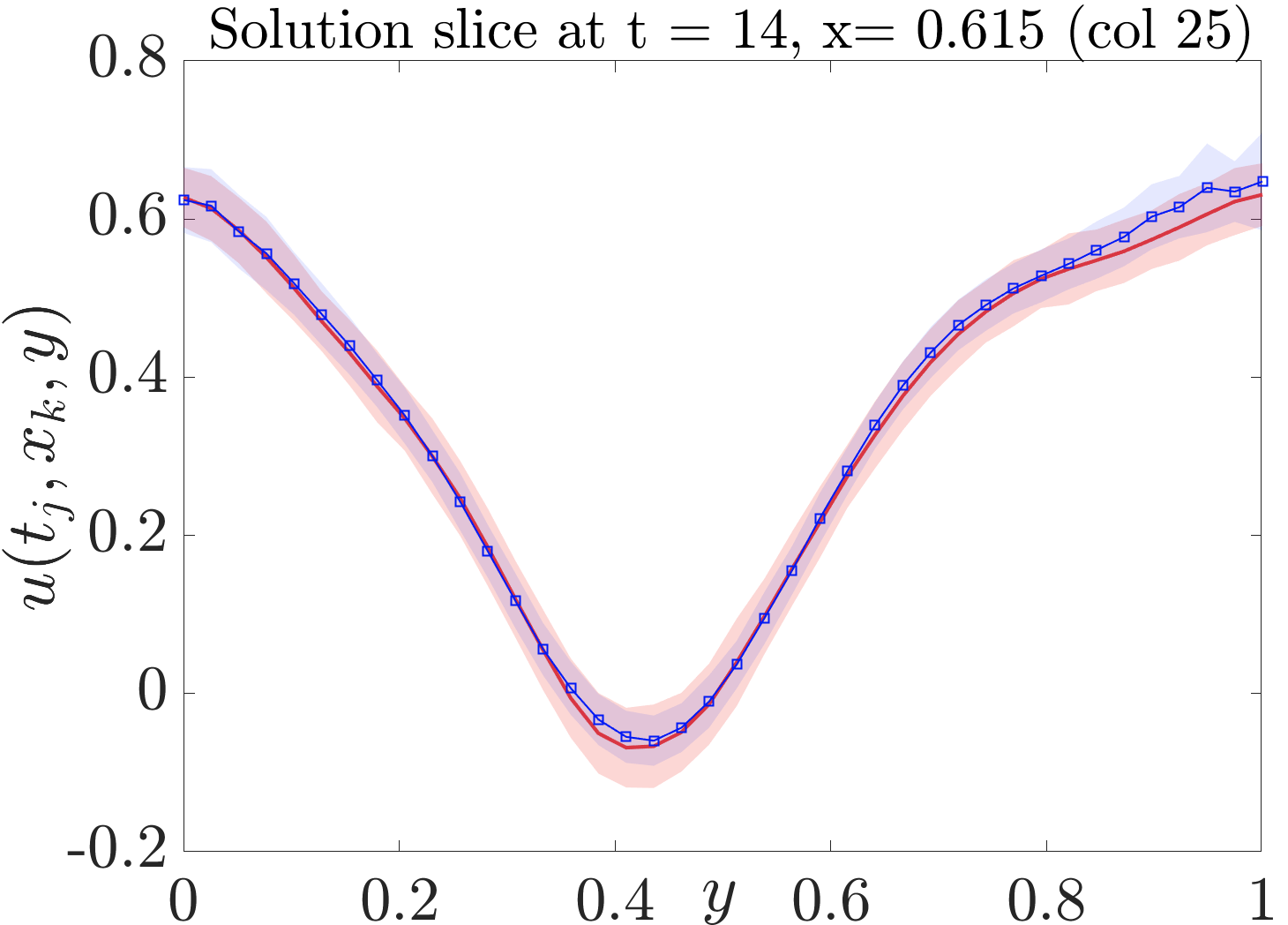}
\end{minipage}%
\begin{minipage}{0.245\textwidth}
\includegraphics[scale = 0.15]{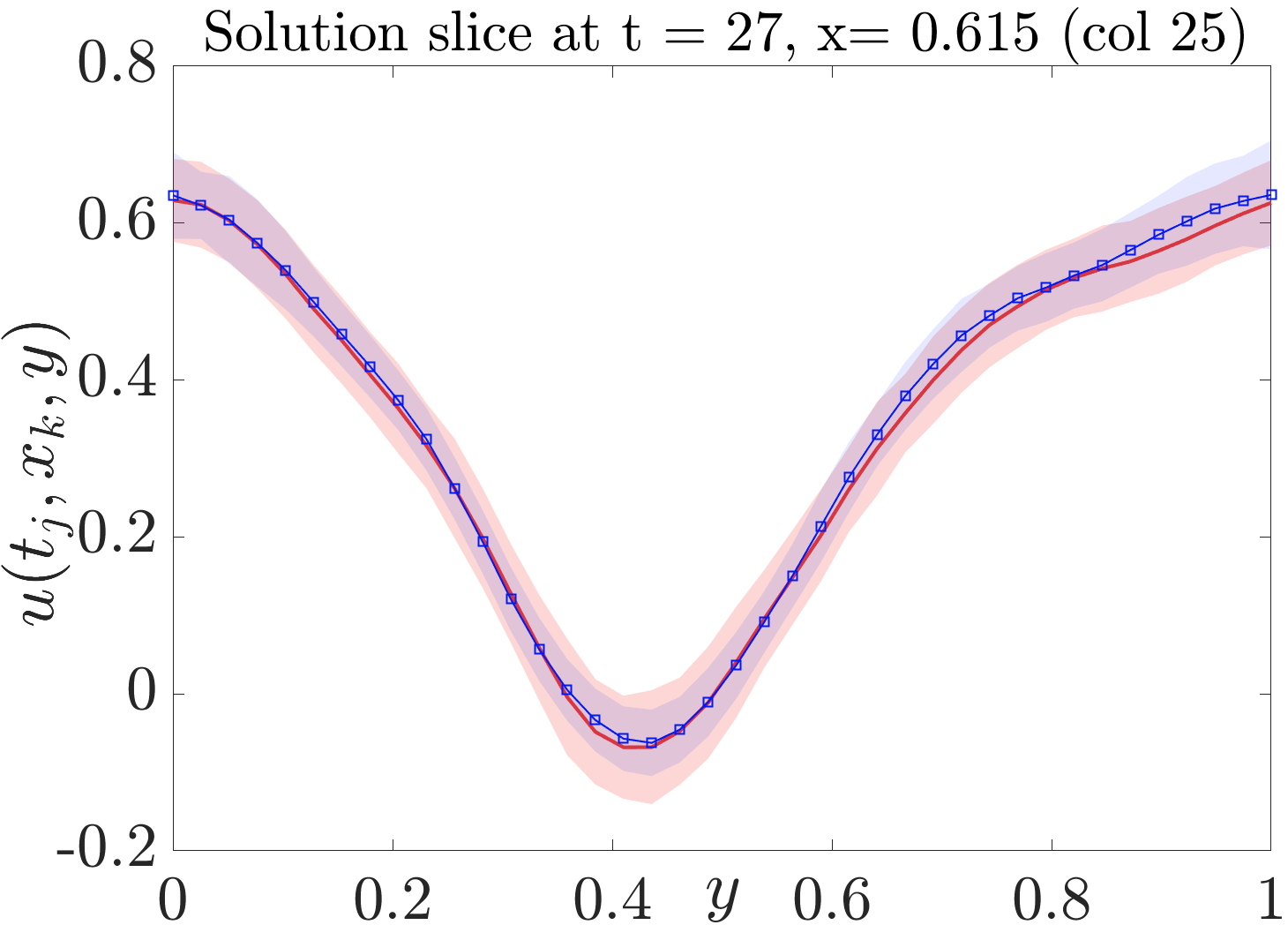}
\end{minipage}%
\begin{minipage}{0.245\textwidth}
\includegraphics[scale = 0.15]{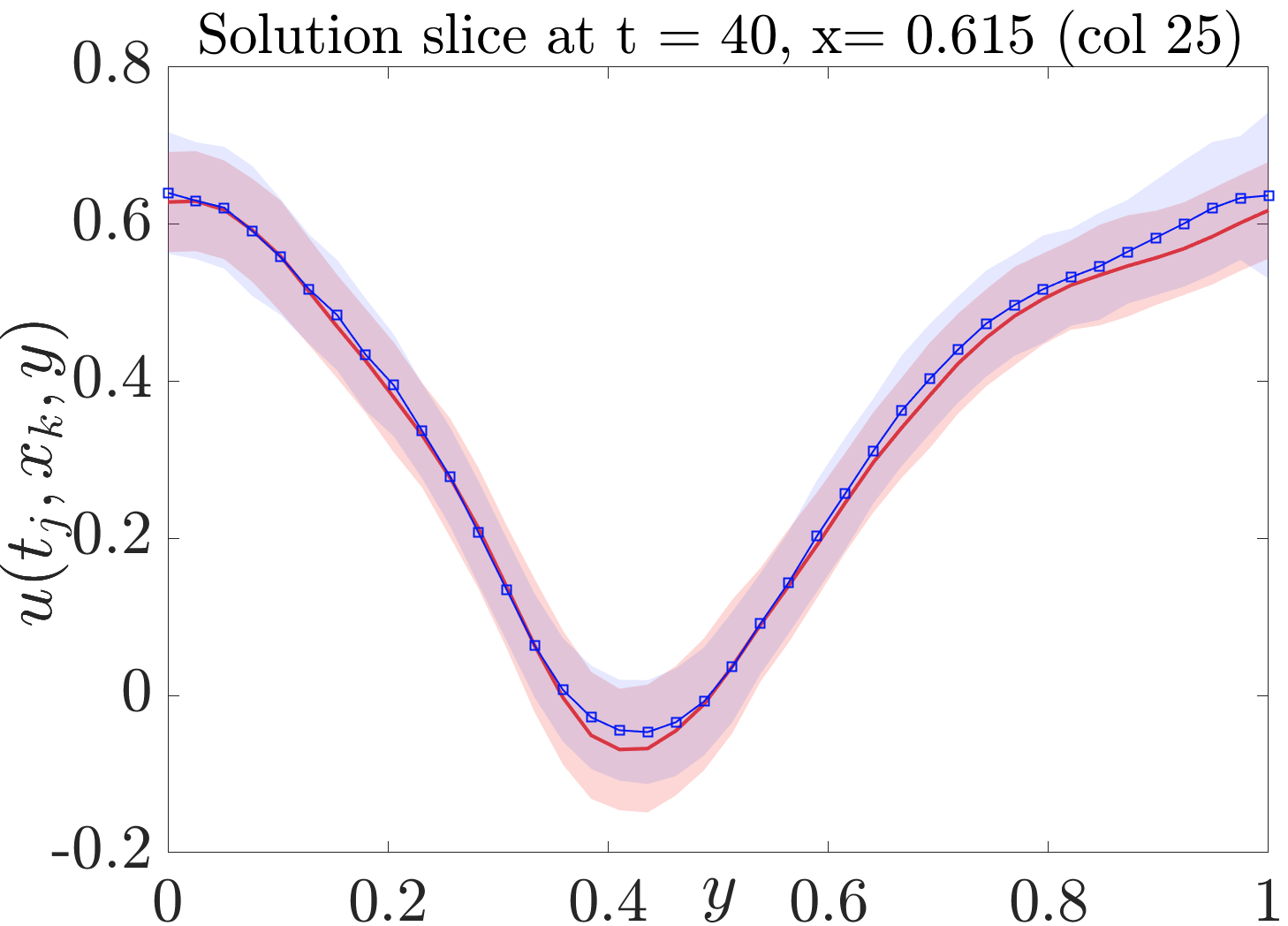}
\end{minipage}

\begin{minipage}{0.245\textwidth}
\includegraphics[scale = 0.15]{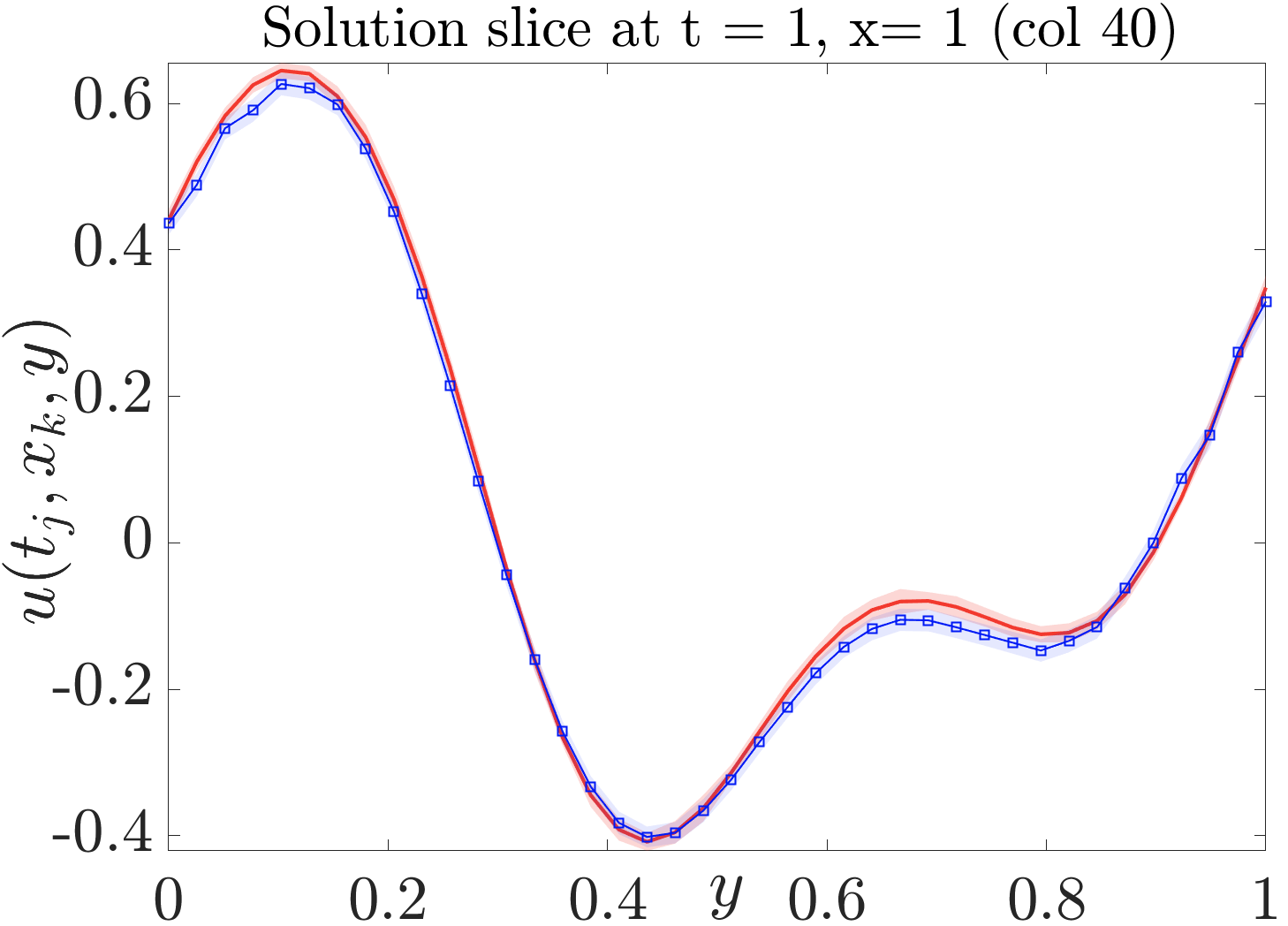}
\end{minipage}%
\begin{minipage}{0.245\textwidth}
\includegraphics[scale = 0.15]{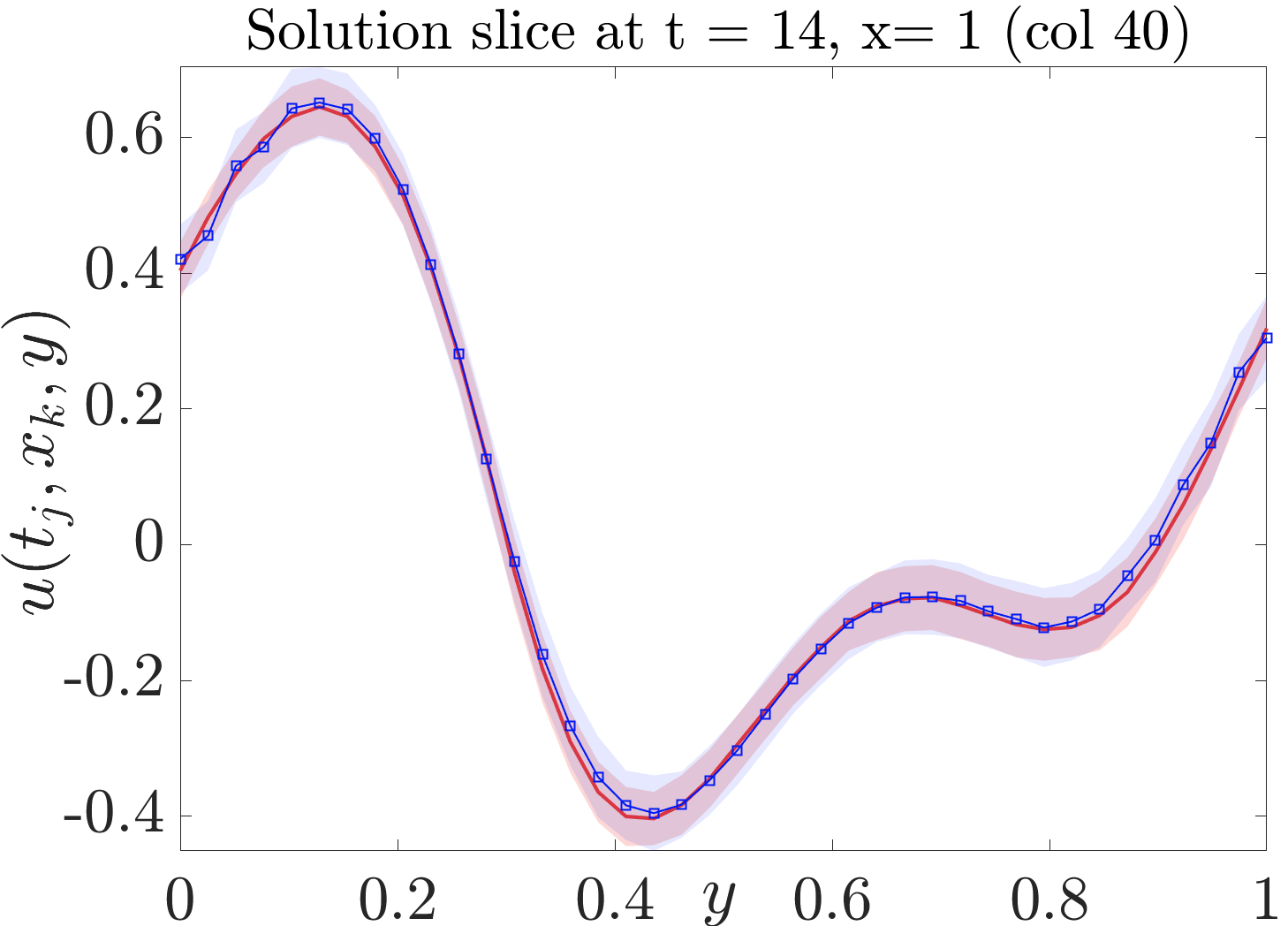}
\end{minipage}%
\begin{minipage}{0.245\textwidth}
\includegraphics[scale = 0.15]{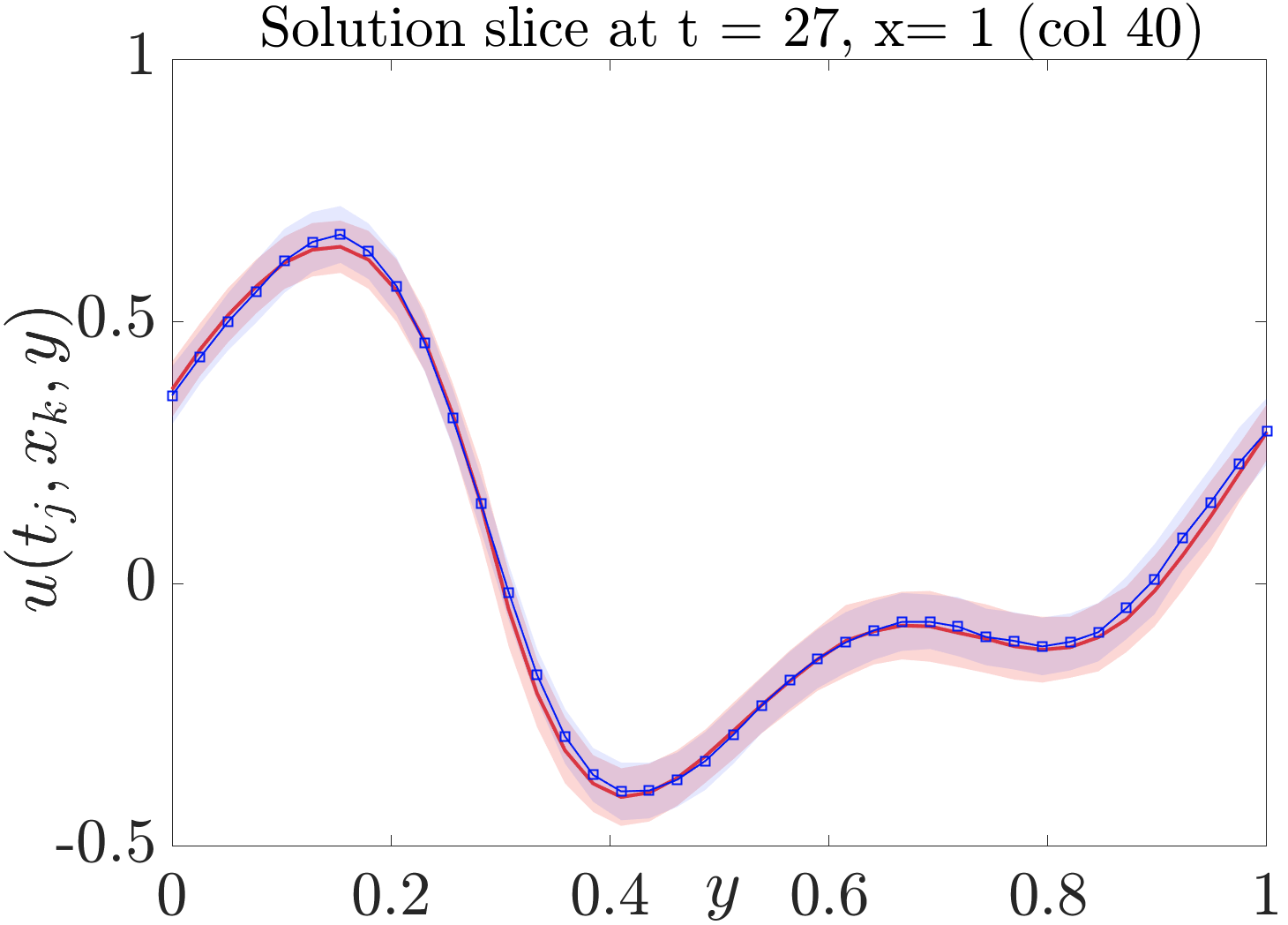}
\end{minipage}%
\begin{minipage}{0.245\textwidth}
\includegraphics[scale = 0.15]{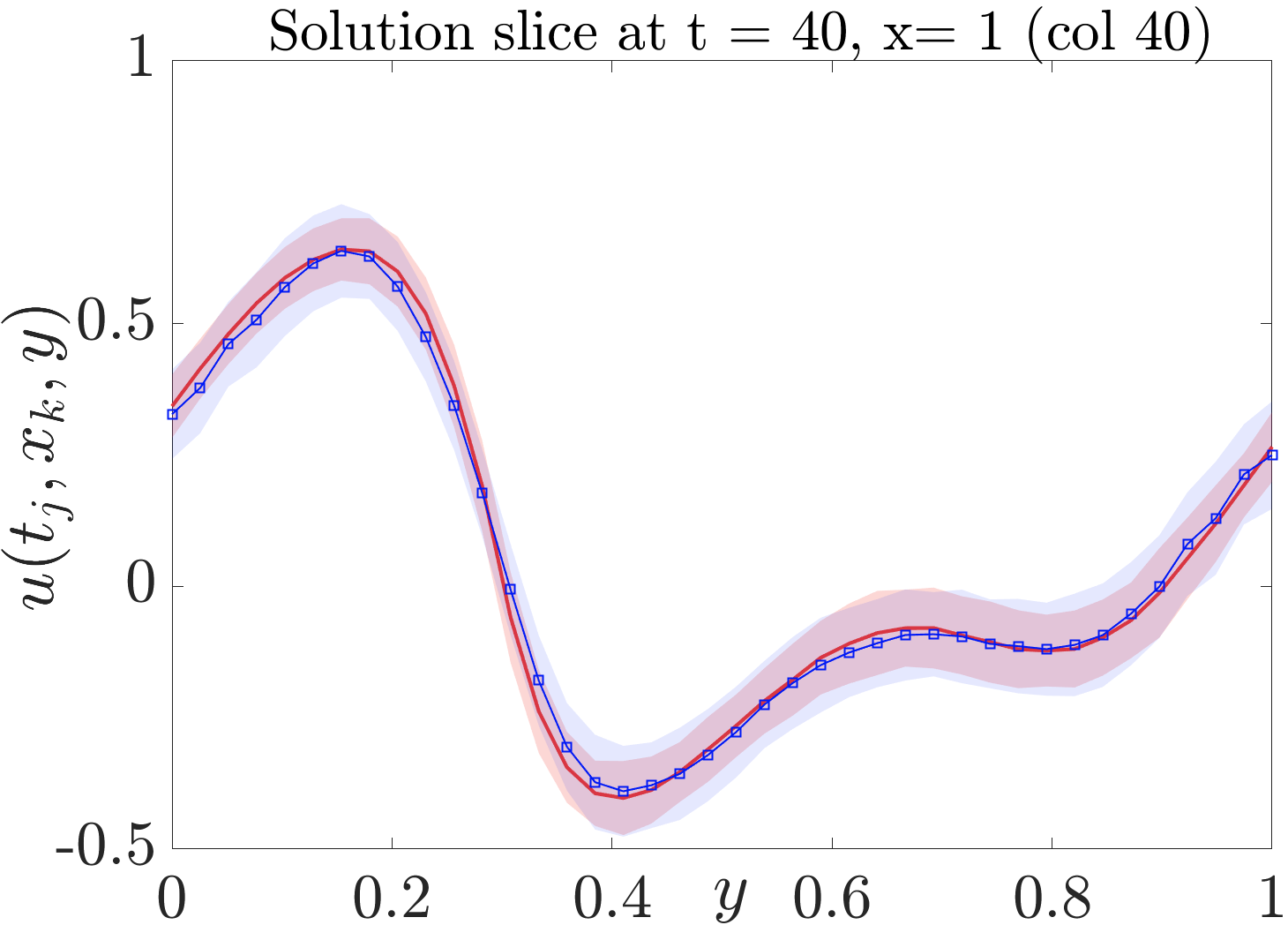}
\end{minipage}
\caption{\footnotesize [Burger Equation Case 2] Cross-sections of the predicted solution in Figure~\eqref{Single_2DBurger_Case2} at time steps $t_m$ for $m=1, 14, 27, 40$. (First row) Column 1. (Second row) Column 25. (Third row) Column 40.}
\label{2DBurger_Case2_CrossSection}
\end{figure}

We first present the performance of SON for a randomly selected representative input. For this input, we generate $400$ reference and predicted solution samples and compute the corresponding sample means and stds. Figure~\ref{Single_2DBurger_Case2} displays the 3D views of the mean fields together with heatmaps of the prediction errors in the means and stds. Figure~\ref{2DBurger_Case2_CrossSection} further compares cross-sections of the predicted means and the corresponding uncertainty bands at selected time steps along the three vertical slices $x=0$, $x=0.615$, and $x=1$. Overall, these results reaffirm that the SON model can accurately capture the mean behavior and provides reliable uncertainty quantification for this representative input.

\begin{figure}[h!]
\vspace{-0.3cm}
\begin{minipage}{0.25\textwidth}
\includegraphics[scale = 0.123]{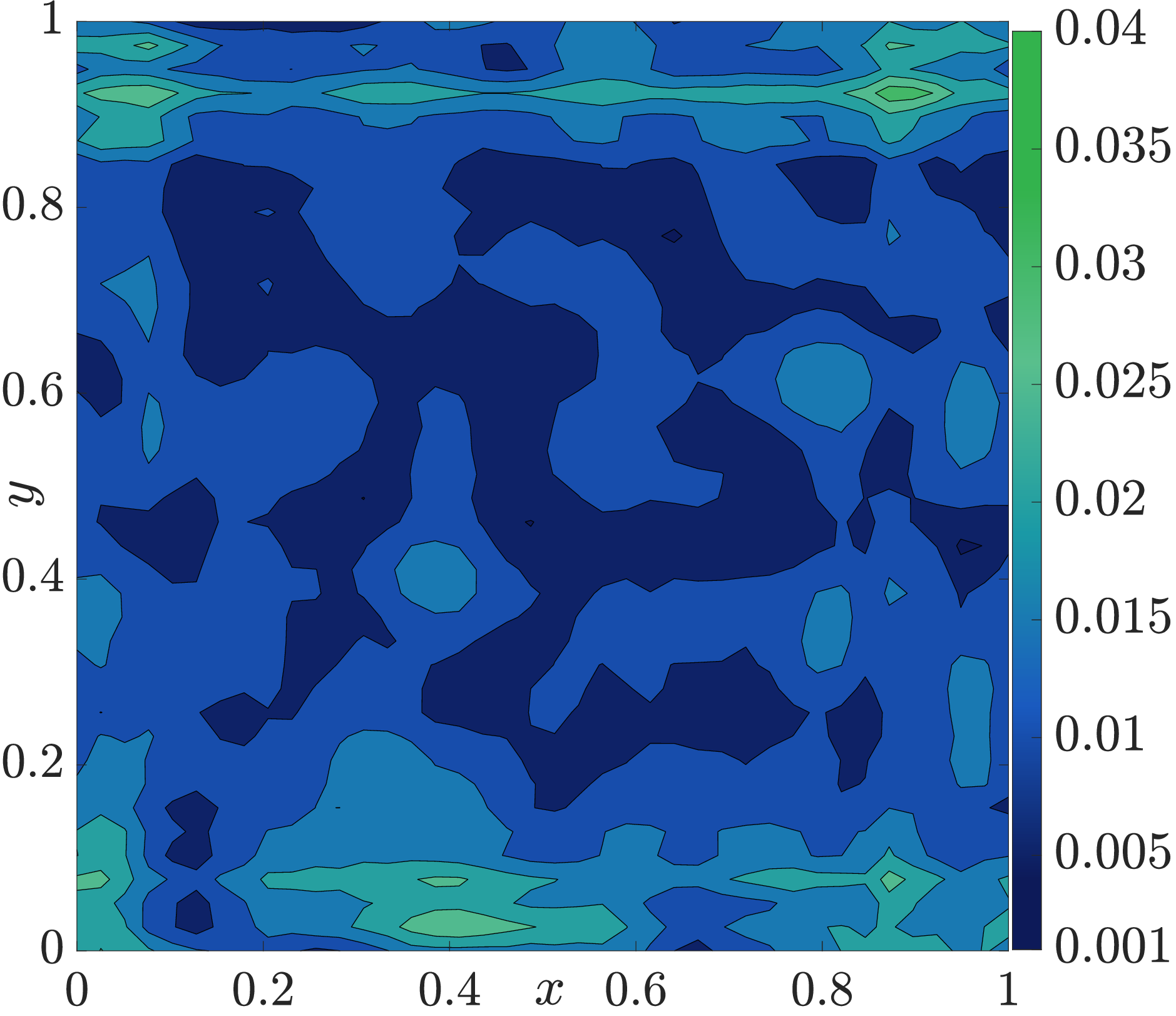}
\end{minipage}%
\begin{minipage}{0.25\textwidth}
\includegraphics[scale = 0.123]{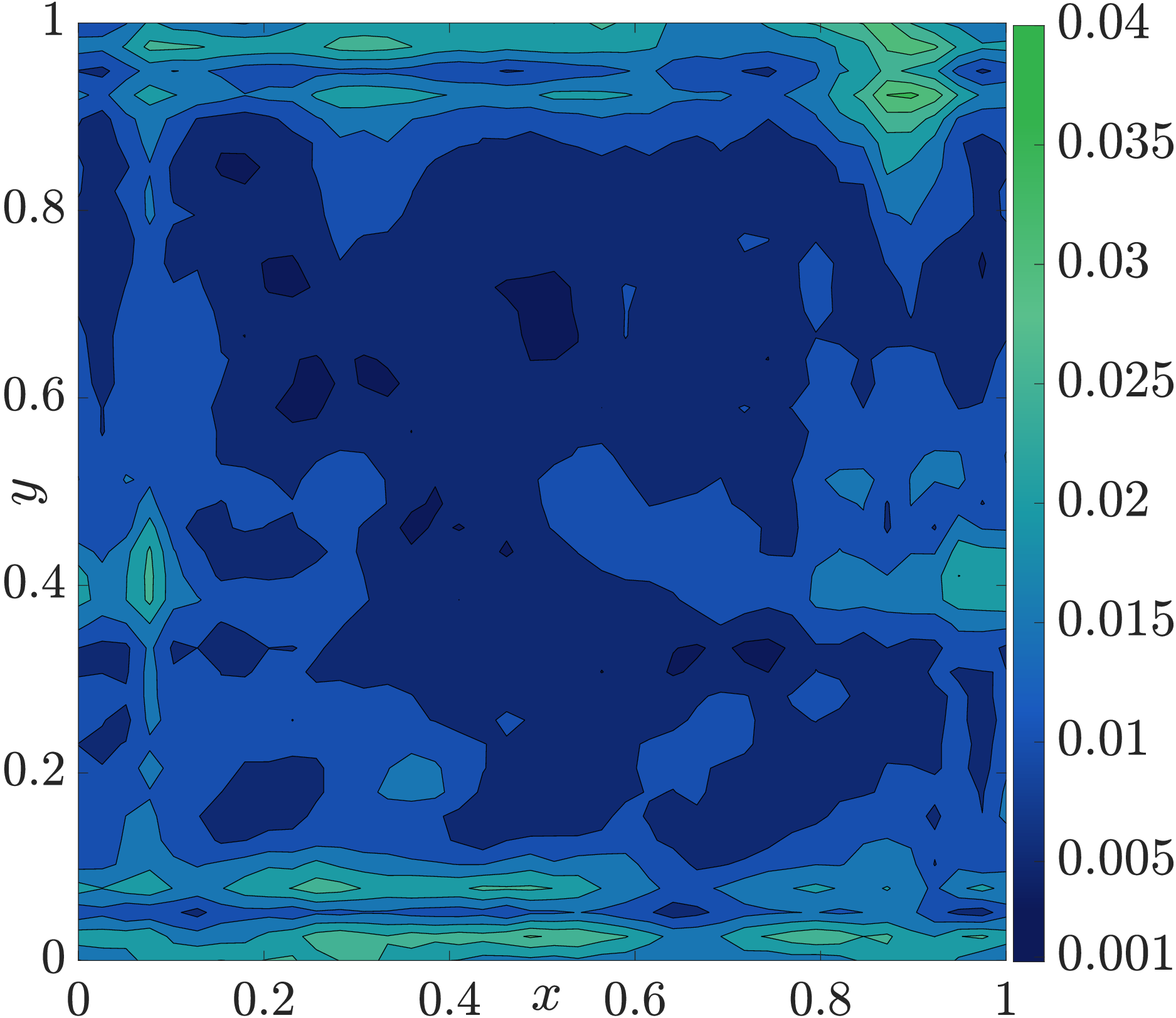}
\end{minipage}%
\begin{minipage}{0.25\textwidth}
\includegraphics[scale = 0.123]{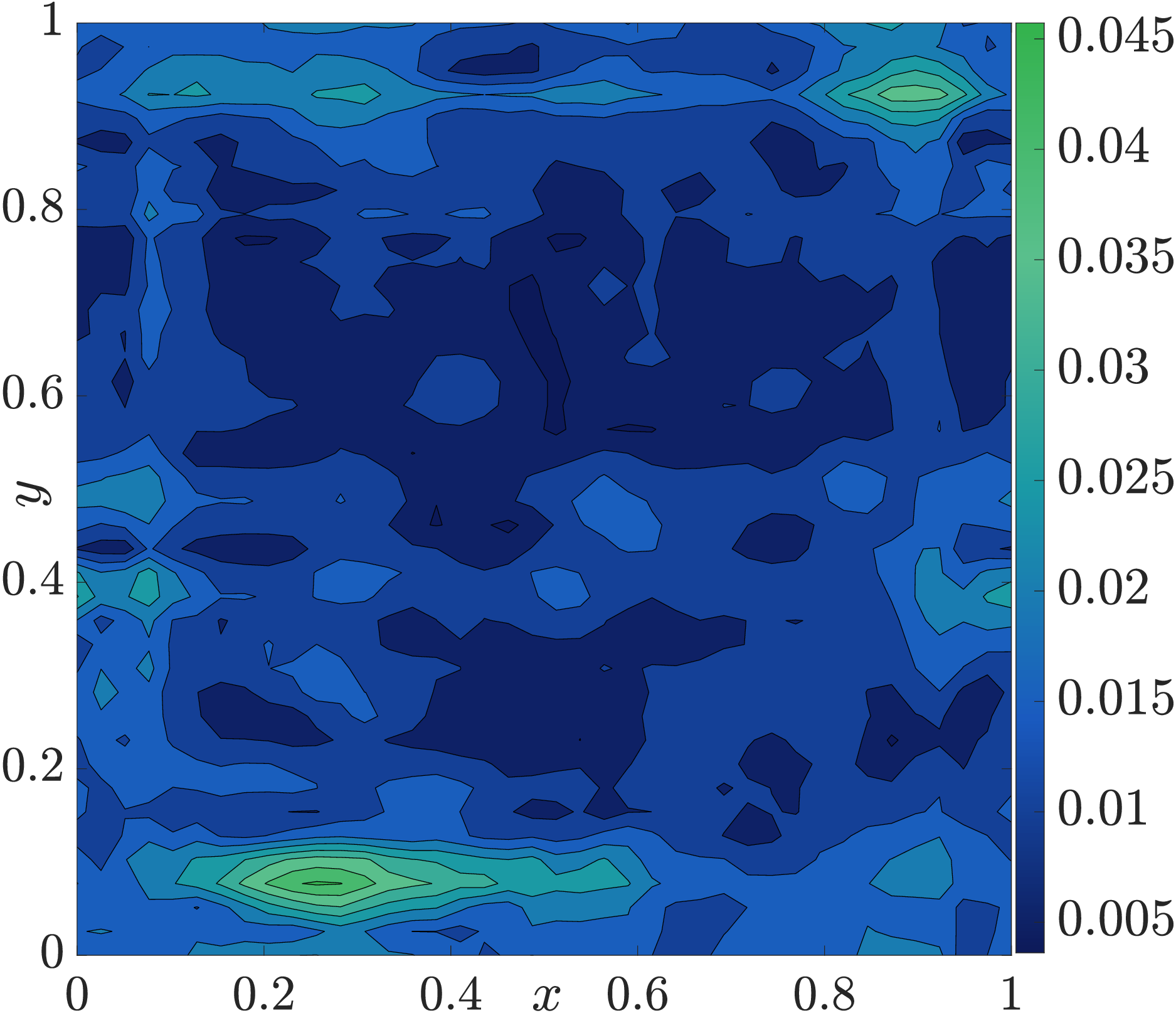}
\end{minipage}%
\begin{minipage}{0.25\textwidth}
\includegraphics[scale = 0.123]{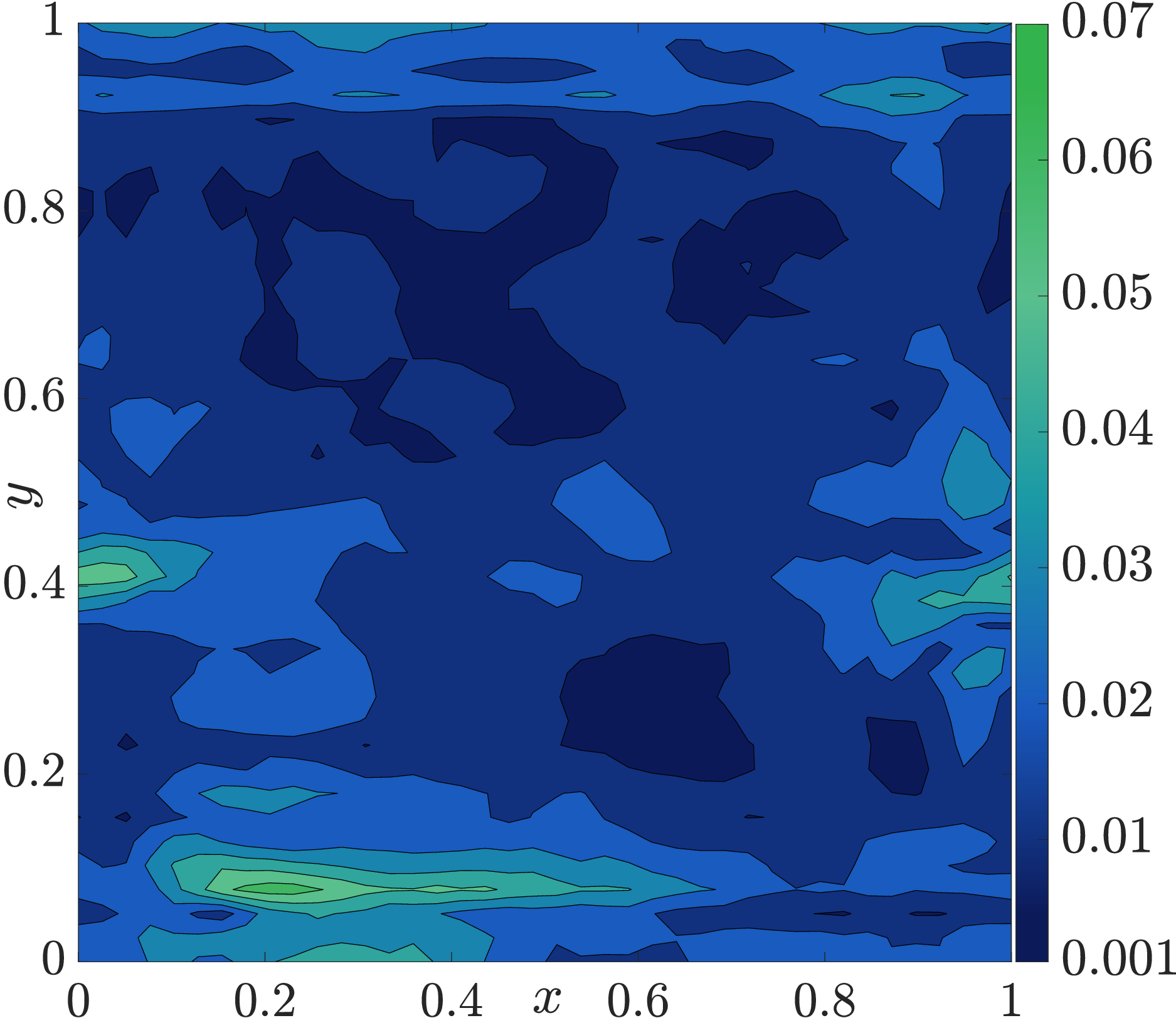}
\end{minipage}
\caption{\footnotesize [Burger Equation Case 2] Average mean error over 8 inputs at time step $t_m$ for $m=1, 14, 27, 40$.}
\label{Average_Mean_2DBurger_Case2}
\end{figure}
\begin{figure}[h!]
\vspace{-0.3cm}
\begin{minipage}{0.25\textwidth}
\includegraphics[scale = 0.123]{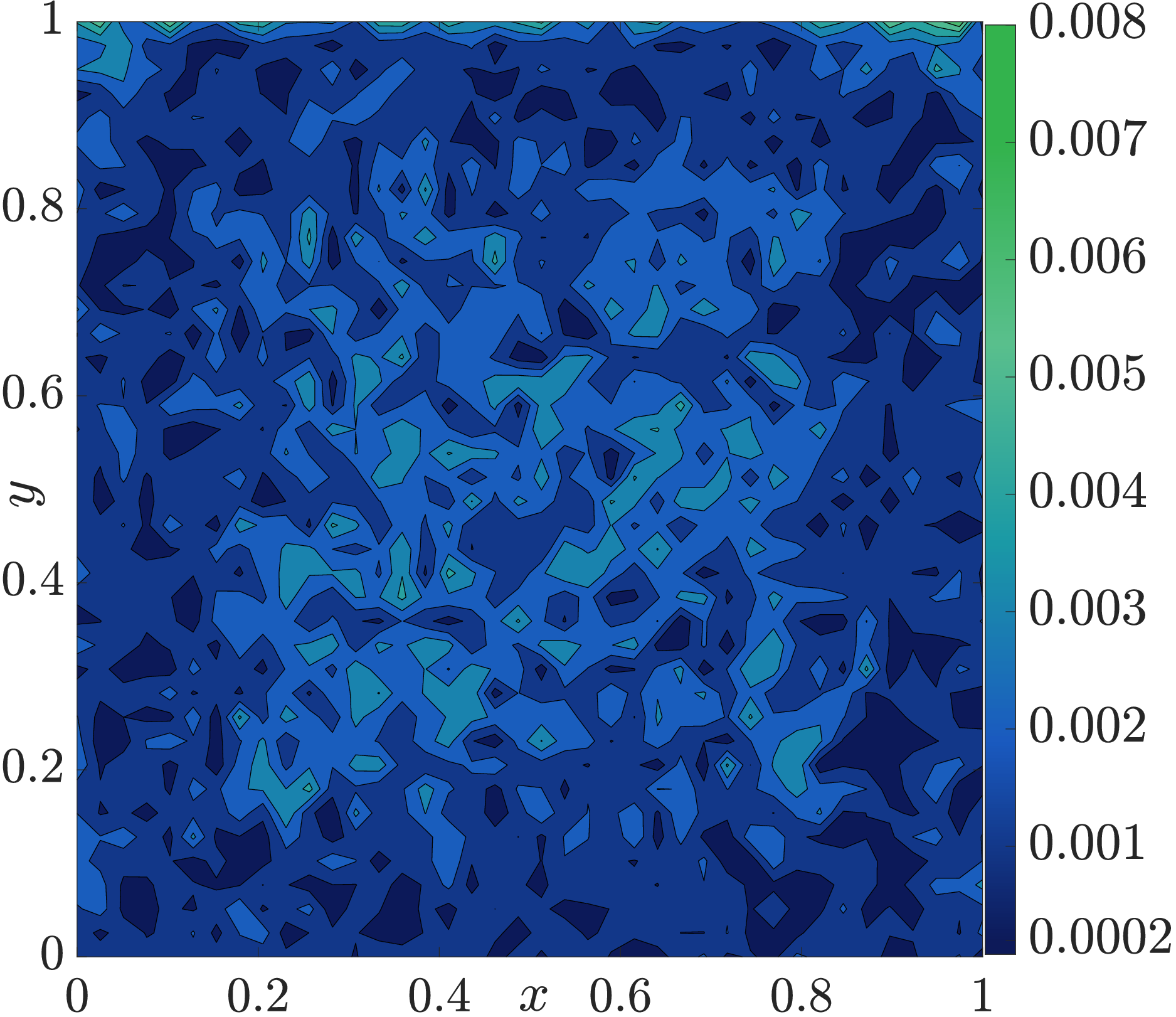}
\end{minipage}%
\begin{minipage}{0.25\textwidth}
\includegraphics[scale = 0.123]{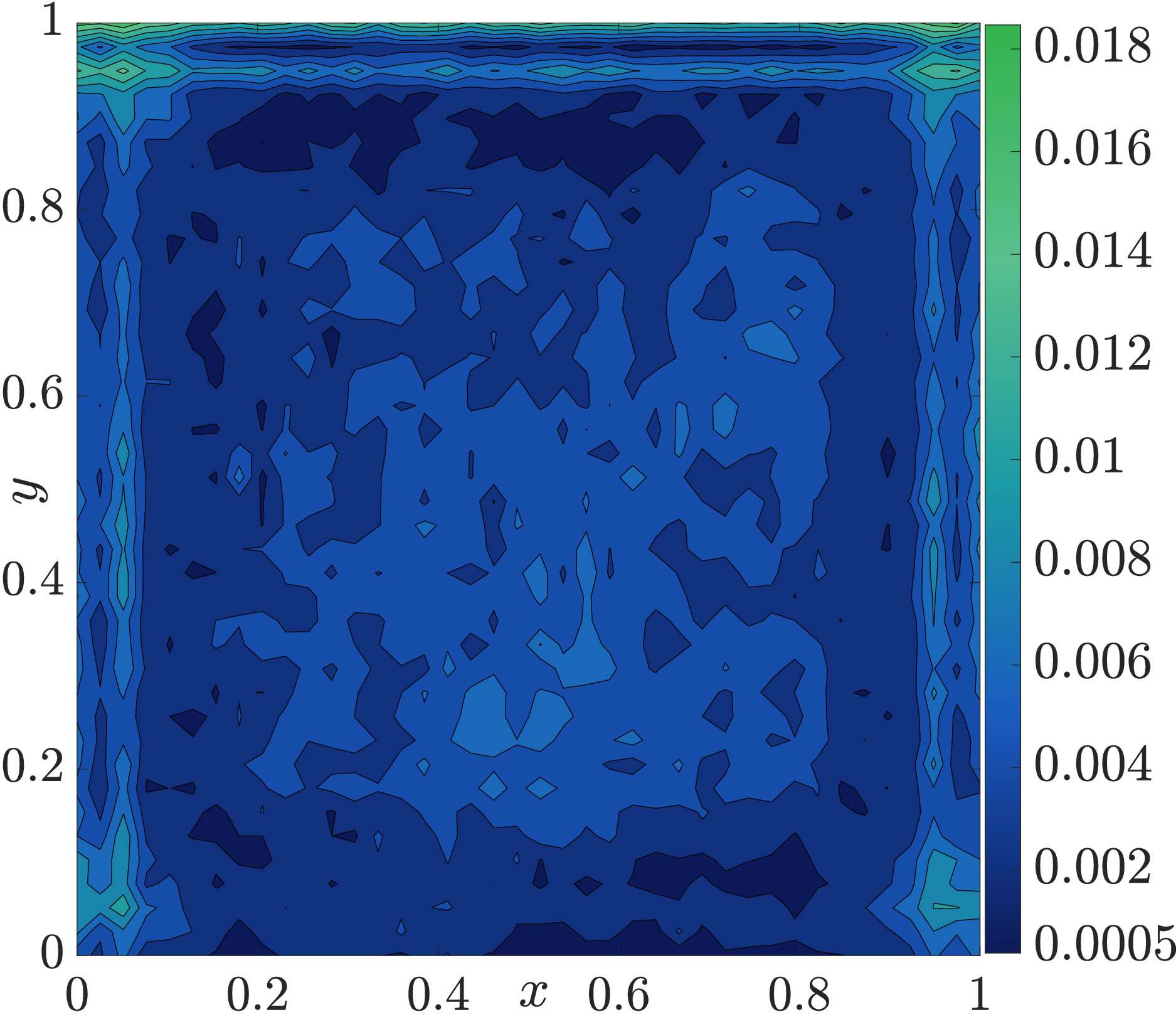}
\end{minipage}%
\begin{minipage}{0.25\textwidth}
\includegraphics[scale = 0.123]{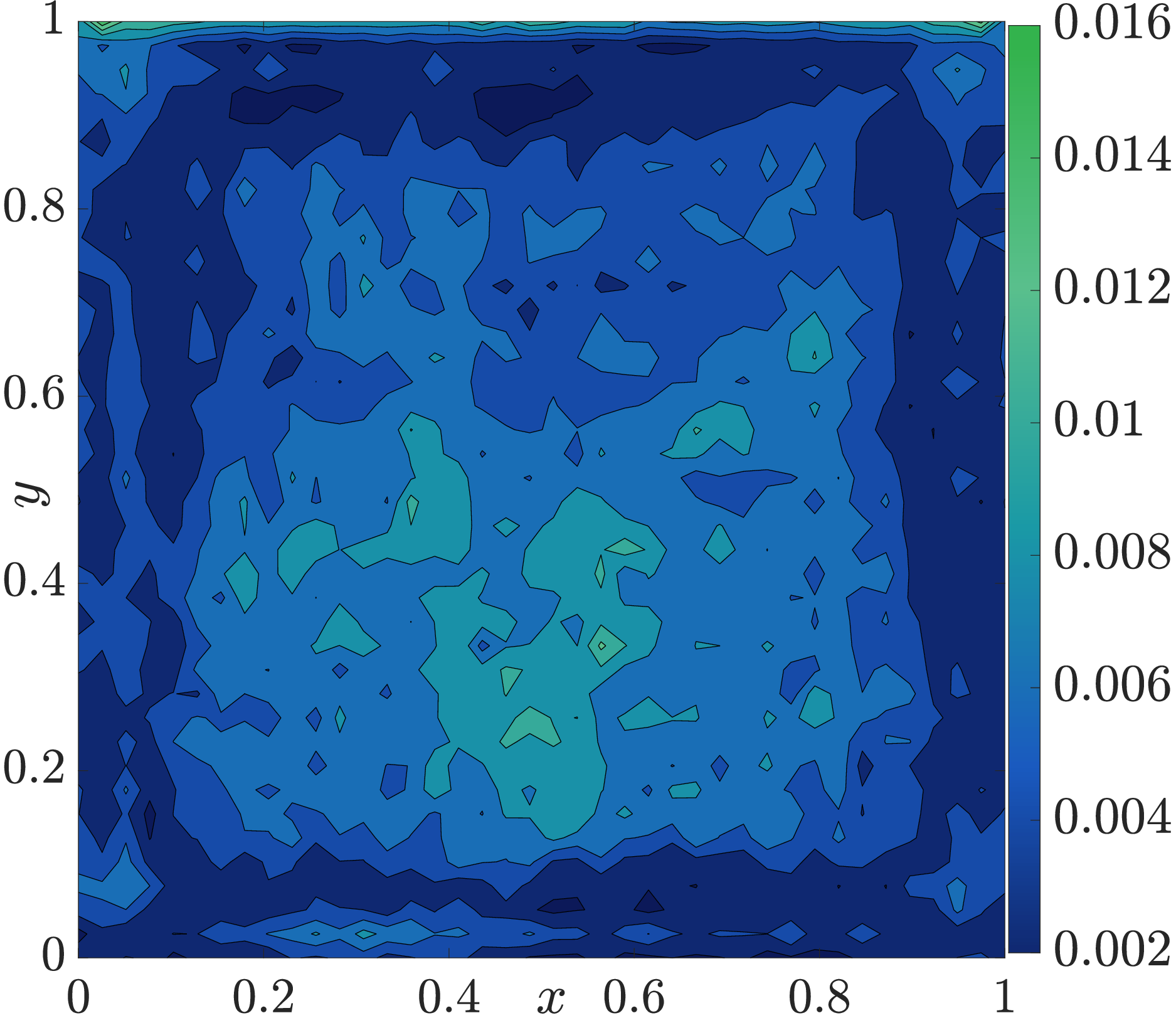}
\end{minipage}%
\begin{minipage}{0.25\textwidth}
\includegraphics[scale = 0.123]{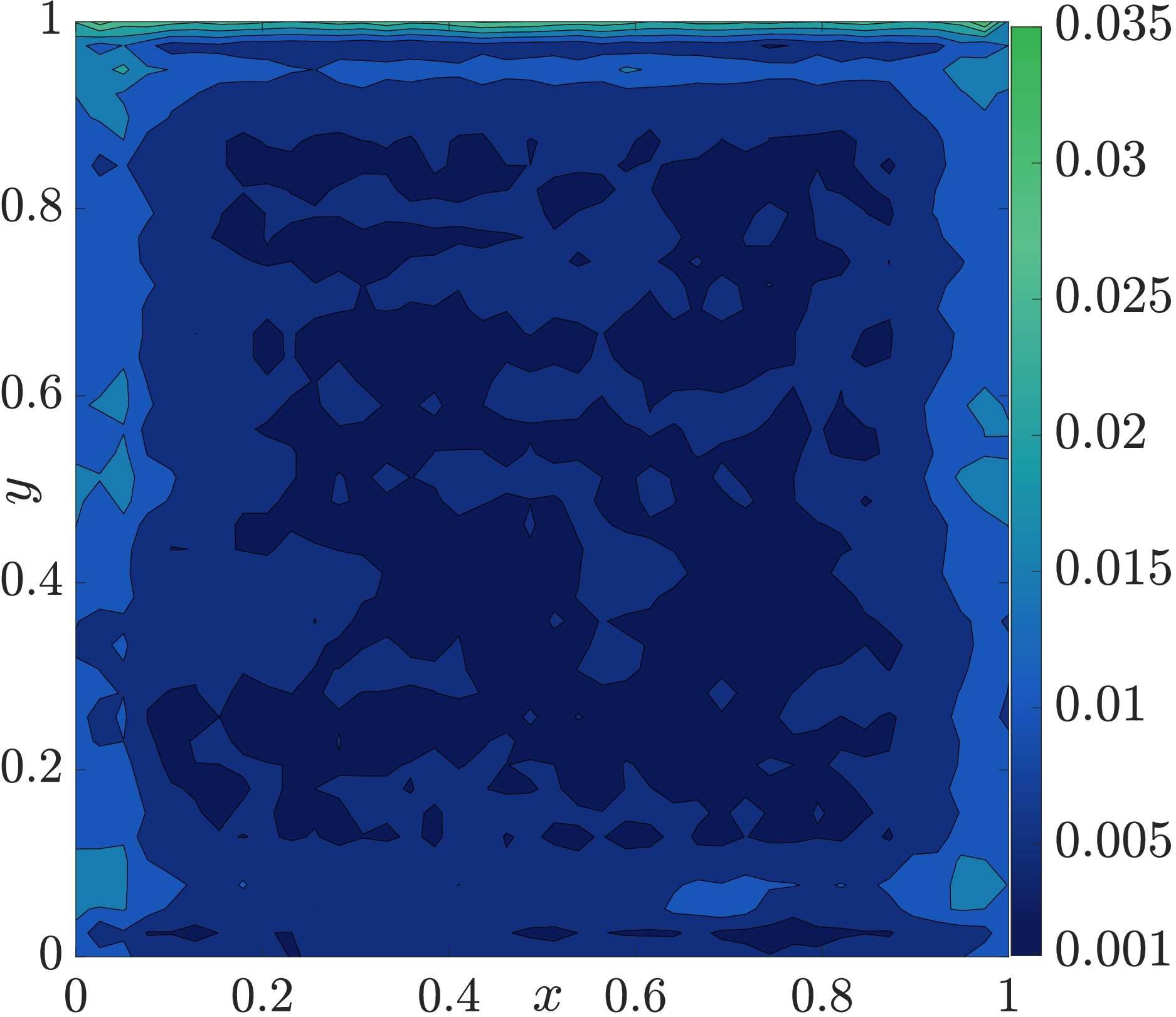}
\end{minipage}
\caption{\footnotesize [Burger Equation Case 2] Average std error over 8 inputs at time step $t_m$ for $m=1, 14, 27, 40$.}
\label{Average_Std_2DBurger_Case2}
\end{figure}

Finally, we repeat the above experiment using $8$ randomly selected input functions from the testing dataset. For each input, we generate ensembles of $400$ reference and predicted samples and compute the corresponding sample means and stds. We then evaluate the prediction errors in the means and stds for each input and present their averages over the $8$ test inputs. The resulting heatmaps for average prediction mean errors and prediction std errors (at time indices $m=1,14,27,$ and $40$) are shown in Figure~\ref{Average_Mean_2DBurger_Case2} and Figure~\ref{Average_Std_2DBurger_Case2}, respectively. Overall, these results again demonstrate the consistency of our SON model in predicting the mean behavior and quantifying uncertainty across multiple inputs. 

\section{Conclusion}\label{Conclusion}
In this work, we introduced the Stochastic Operator Network (SON) framework for learning solution operators associated with various SPDEs and for quantifying intrinsic uncertainty in model predictions. SON is derived from the DeepONet architecture by replacing the branch network with a Stochastic Neural Network (SNN) equipped with an additive diffusion term, thereby enabling uncertainty quantification. The training mechanism developed for SNNs was adapted to optimize the SON parameters, and the corresponding target loss function was extended to incorporate the parameters of the trunk network. To further improve computational efficiency while maintaining predictive accuracy, we proposed a two-phase training strategy for SON that reduces the complexity and computational cost.

We validated the effectiveness and robustness of the proposed approach through comprehensive numerical experiments on several random PDEs and SPDEs, including two stationary problems, namely the reaction-diffusion and advection-diffusion equations, and two time-dependent problems, namely heat and Burgers' equations. For the advection-diffusion and Burgers' equations, we also considered multiple ways of incorporating uncertainty into the numerical solver to generate randomness in the model solutions. These tests demonstrate the capability of SON to predict solution trajectories and quantify model uncertainty.

\subsection*{Acknowledgments}
This material is based upon work supported by the U.S. National Science Foundation under Grant DMS-2142672 and by the U.S. Department of Energy, Office of Science, Office of Advanced Scientific Computing Research, Applied Mathematics Program under Grant DE-SC0025412 and through the SciDAC FASTMath Institute.

Notice: This manuscript has been authored by UT-Battelle, LLC, under contract DE-AC05-00OR22725 with the US Department of Energy (DOE). The US government retains and the publisher, by accepting the article for publication, acknowledges that the US government retains a nonexclusive, paid-up, irrevocable, worldwide license to publish or reproduce the published form of this manuscript, or allow others to do so, for US government purposes. DOE will provide public access to these results of federally sponsored research in accordance with the DOE Public Access Plan (\url{https://www.energy.gov/doe-public-access-plan}).

\bibliographystyle{plain}
\bibliography{References}
\end{document}